\documentclass{article}

\usepackage[preprint]{neurips_2024}

\usepackage[frozencache,cachedir=.]{minted}

\usepackage[utf8]{inputenc} %
\usepackage[T1]{fontenc}    %
\usepackage{hyperref}
\usepackage{url}            %
\usepackage{booktabs}       %
\usepackage{amsfonts}       %
\usepackage{nicefrac}       %
\usepackage{microtype}      %
\usepackage{xcolor}         %

\usepackage{graphicx}
\usepackage[position=top]{subfig}
\usepackage{multirow}
\usepackage{multicol}
\usepackage[normalem]{ulem}
\usepackage{wrapfig}

\usepackage[toc,page,header]{appendix}
\usepackage{minitoc}

\definecolor{LightGray}{gray}{0.95}
\definecolor{mydarkblue}{rgb}{0,0.08,0.45}
\definecolor{mydarkgreen}{rgb}{0,0.45,0.08}
  \hypersetup{
    colorlinks=true,
    linkcolor=mydarkblue,
    citecolor=mydarkblue,
    filecolor=mydarkblue,
    urlcolor=mydarkblue
}
\usepackage{enumitem}

\makeatletter
\def\@fnsymbol#1{\ensuremath{\ifcase#1\or * \or w \or p \or mcp \or \dagger\or \mathsection\or
   \ddager\or \mathparagraph\or \|\or **\or \dagger\dagger
   \or \ddagger\ddagger \else\@ctrerr\fi}}
\makeatletter

\title{Self-Improving Transformers Overcome Easy-to-Hard and Length Generalization Challenges}

\author{%
  Nayoung Lee\thanks{Authors contributed equally to this paper.} \\
  University of Wisconsin-Madison\\
  \texttt{nayoung.lee@wisc.edu} \\
  \And
  Ziyang Cai\footnotemark[1] \\
  University of Wisconsin-Madison\\
  \texttt{zcai75@wisc.edu} \\
  \And
  Avi Schwarzschild \\
  Carnegie Mellon University\\
  \texttt{schwarzschild@cmu.edu} \\
  \And
  Kangwook Lee \\
  University of Wisconsin-Madison\\
  \texttt{kangwook.lee@wisc.edu} \\
  \And
  Dimitris Papailiopoulos \\
   Microsoft Research, AI Frontiers\\
   University of Wisconsin-Madison\\
  \texttt{dimitris@papail.io} \\
}

\usepackage{amsmath,amsfonts,bm}

\def\eqref#1{equation~\ref{#1}}

\def\1{\bm{1}}

\DeclareMathAlphabet{\mathsfit}{\encodingdefault}{\sfdefault}{m}{sl}
\SetMathAlphabet{\mathsfit}{bold}{\encodingdefault}{\sfdefault}{bx}{n}

\usepackage{amsmath}
\usepackage{amssymb}
\usepackage{mathtools}
\usepackage{amsthm}
\usepackage{thm-restate}

\usepackage{cuted}
\usepackage{flushend}

\usepackage{soul}

\usepackage[linesnumbered,ruled,vlined]{algorithm2e}
\usepackage{algorithmicx}
\usepackage{algpseudocode}

\usepackage[capitalize,noabbrev]{cleveref}

\usepackage[most]{tcolorbox}

\usepackage{tabularx}
    \newcolumntype{L}{>{\raggedright\arraybackslash}X}

\definecolor{commentcolour}{rgb}{0.3,0.7,0.2}
\definecolor{blue}{RGB}{33, 144, 141}
\definecolor{lightblue}{RGB}{240, 248, 248} 

\newtcolorbox{finding}{
    enhanced,
    breakable,
    colback=lightblue,         %
    colframe=blue,        %
    boxrule=1.5pt,
    arc=0.25em,
    left=1em,
    right=1em,
    top=1em,
    bottom=0.75em,
    before=\vspace{1em},
    overlay unbroken and first={
        \node[
            fill=blue,
            text=white,
            font=\bfseries,
            anchor=west,
            inner xsep=0.75em,
            inner ysep=0.5em,
            rounded corners=0.25em
        ] 
        at ([xshift=0.75em]frame.north west) {Findings};
    }
}

\usepackage{soul}
\usepackage{adjustbox}
\usepackage{minted}

\usepackage{dsfont}

\theoremstyle{definition}

\renewcommand{\arraystretch}{1.1} %

\ifdefined\usebigfont

\usepackage{times}
\usepackage[fontsize=13pt]{scrextend}
\makeatletter
\@ifpackageloaded{geometry}{\AtBeginDocument{\newgeometry{letterpaper,left=1.56in,right=1.56in,top=1.71in,bottom=1.77in}}}{\usepackage[letterpaper,left=1.56in,right=1.56in,top=1.71in,bottom=1.77in]{geometry}}
\makeatother
\else
\fi

\usepackage{thm-restate}

\appto\appendix{\addtocontents{toc}{\protect\setcounter{tocdepth}{0}}}

\begin{document}
\doparttoc

\maketitle

\begin{abstract}

Large language models often struggle with length generalization and solving complex problem instances beyond their training distribution. We present a self-improvement approach where models iteratively generate and learn from their own solutions, progressively tackling harder problems while maintaining a standard transformer architecture. Across diverse tasks including arithmetic, string manipulation, and maze solving, self-improving enables models to solve problems far beyond their initial training distribution—for instance, generalizing from 10-digit to 100-digit addition without apparent saturation. We observe that in some cases filtering for correct self-generated examples leads to exponential improvements in out-of-distribution performance across training rounds. Additionally, starting from pretrained models significantly accelerates this self-improvement process for several tasks. Our results demonstrate how controlled weak-to-strong curricula can systematically teach a model logical extrapolation without any changes to the positional embeddings, or the model architecture.

\end{abstract}

\section{Introduction}\label{sec:intro}

\begin{figure*}
    \centering
    \includegraphics[width=1\linewidth]{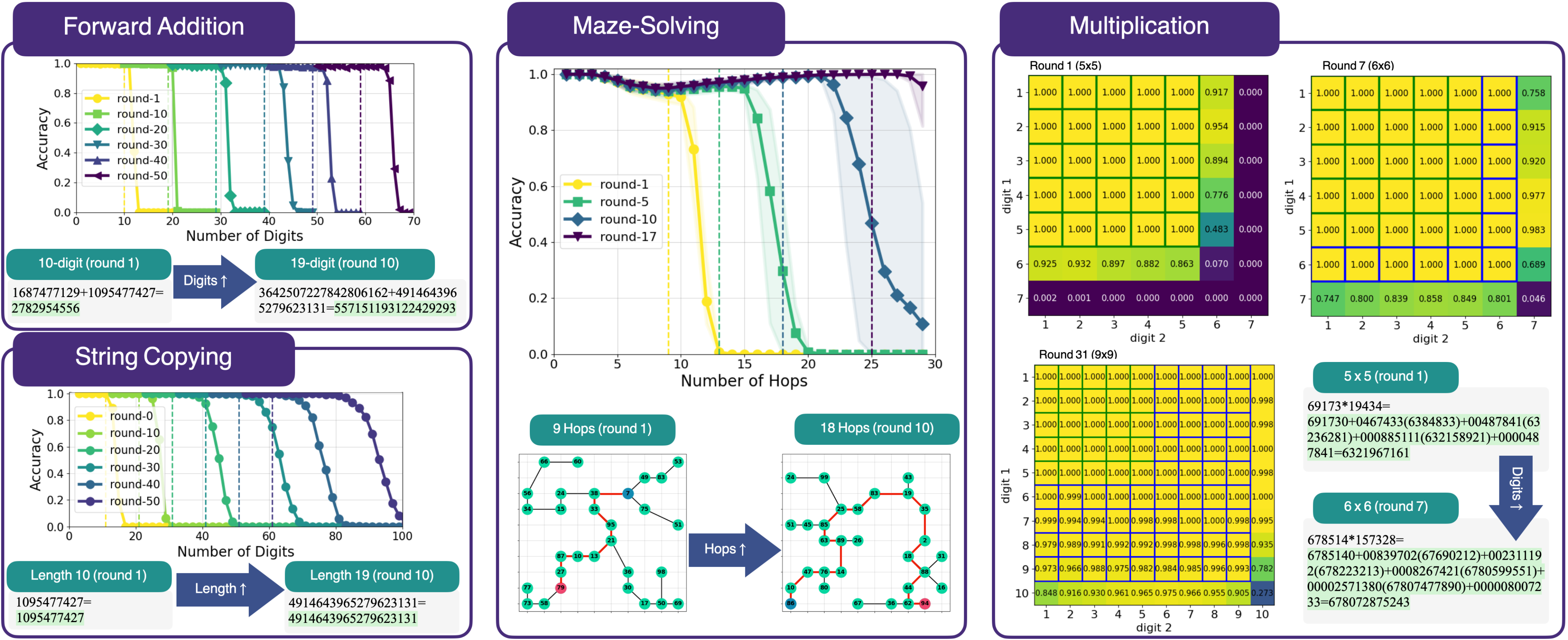}
    \caption{Overview of self-improvement results. Models trained with self-improvement can tackle increasingly complex tasks that extend far beyond their initial training distributions, achieving significant generalization \textbf{without any additional supervision}.}
    \label{fig:result_overview}
    \vspace{-6mm}
\end{figure*}

Despite the remarkable success of transformer-based language models~\citep{vaswani2017attention} across a wide range of tasks, these models exhibit significant limitations in \textit{length generalization}—the ability to extrapolate to longer sequences than those seen during training. Even in simple algorithmic tasks such as arithmetic, standard transformer models trained with autoregressive objectives struggle to generalize to longer problem instances~\citep{dubois2019location,hupkes2020compositionality,newman2020eos,anil2022exploring}. 

To address this, prior work has explored various approaches, including changes to positional embeddings~\citep{ruoss2023randomized,li2023functional,mcleish2024transformers,kazemnejad2024impact,sabbaghi2024explicitly,Cho2024PositionCI,zhou2024transformers}, architectural modifications~\citep{fan2024looped,duan2023interpolation}, and data format changes such as index hinting~\citep{zhou2023algorithms,zhou2024transformers}. While effective in controlled setups, these approaches are often incompatible with how large language models (LLMs) are trained in practice, as they introduce task-specific modifications that are unclear how and to what extent they would transfer to the general purpose settings.

In this work, we attempt to overcome length generalization challenges in the standard transformer setting, by building around an interesting phenomenon that transformers exhibit, i.e., ``transcendence''~\citep{zhang2024transcendence}. Transcendence is the ability of a student model to  generalize slightly beyond the difficulty of the data provided by a teacher during training. Specifically, models trained on simple instances of a task, say $n$ digit arithmetic, can sometimes generate correct outputs for slightly harder instances, e.g., $n+1$ digit arithmetic, with some accuracy. We leverage this property by applying a \textbf{self-improvement} framework, drawing significant inspiration by STaR~\citep{zelikman2022star} and ReST~\citep{gulcehre2023reinforced}, where we alternate between collecting output predictions and finetuning using the self-generated dataset. 

Self-improvement has been widely studied in various contexts~\citep{singh2023beyond,gulcehre2023reinforced,liang2024sheep}, typically in settings where external verifiers, weak supervision, or filtering mechanisms are used to ensure data quality. We demonstrate that extreme length generalization is indeed possible under this framework, \textit{without any  modification to the base transformer architecture}. For tasks like reverse addition and string copying, self-improvement succeeds with no explicit data filtering. However, for harder problems such as multiplication and shortest-path finding in mazes, self-improvement without data filtering fails due to error accumulation. We show that simple filtering techniques—such as length filtering and majority voting—suffice to maintain data quality and enable self-improvement to extend far beyond the initial training distribution.

Our findings suggest that self-improvement is not limited to length generalization but also enables \textit{easy-to-hard generalization}, where a model trained on simpler tasks successfully learns harder tasks without additional supervision. Notably, our approach does not introduce a new self-improvement framework but instead demonstrates its effectiveness across diverse algorithmic tasks. 

Furthermore, we investigate the dynamics of self-improvement and show that: (1) controlling the weak-to-strong curriculum is crucial, as models require a structured difficulty schedule to avoid catastrophic failure, (2) self-improvement accelerates over time, as models increasingly benefit from harder examples, leading in some cases to exponential extrapolation, and (3) starting with a pretrained models singificantly accelerates self-improvement, allowing to generalize further and faster than models trained from scratch. 

Our findings provide evidence that learn self-improvement is a general purpose and scalable solution for length and easy-to-hard generalization. Our contributions can be summarized as:

\begin{enumerate}[left=5pt] %
    \setlength{\parskip}{0pt}
    \setlength{\itemsep}{0pt plus 1pt}
    \item We apply an iterative self-training framework to train transformers on the arithmetic, maze and string manipulation tasks, and successfully tackle \textbf{easy-to-hard generalization} to extreme out-of-distribution test data. 
    \item We motivate the importance of a carefully crafted self-improvement schedule and \textbf{label filtering} based on length and majority voting, which are central to consistent self-improvement.
    \item We show that the rate of self-improvement can be exponential and pretrained models can achieve faster acceleration in easy-to-hard generalization.
    \item We investigate some key failure modes of self-correction due to label noise leading to an \textbf{error avalanche}, and discuss how they can be overcome through weak verification.
\end{enumerate}

\section{Related Works}\label{sec:related_work}

\paragraph{Length and Easy-to-hard Generalization.} 

Length generalization is concerned with extrapolating to longer sequence lengths than those seen during training~\citep{dubois2019location,hupkes2020compositionality,newman2020eos,anil2022exploring}. Previous approaches to improve length generalization includes architectural modifications, including specialized positional embeddings~\citep{press2021train,li2023functional,ruoss2023randomized,kazemnejad2024impact,sabbaghi2024explicitly,Cho2024PositionCI,zhou2024transformers}, looping~\cite{fan2024looped}, novel attention mechanisms~\citep{duan2023interpolation}, and input format augmentation~\citep{zhou2023algorithms,zhou2024transformers}. Beyond arithmetic, ~\citet{yehudai2021local} studies length generalization in graph tasks. In contrast, our approach adheres to the standard transformer architecture without introducing significant modifications to architecture, positional encoding, or input structure. While prior approaches typically rely on fixed-length training datasets without further updates to model weights, we iteratively update model weights on self-generated datasets, enabling the model to progressively improve and extend its generalization capabilities. 

More generally, easy-to-hard-generalization is the paradigm where human annotation is provided for easier tasks, but aiming to enable generalization to harder tasks with no additional supervision~\citep{schwarzschild2021can,bansal2022end,burns2023weak,hase2024unreasonable,sun2024easy}. 
For instance, ~\citet{zhang2024transcendence} study this \textit{transcendence} phenomenon in chess, showing that chess transformers can sometimes outperform all players in the training dataset. Similarly, ~\citet{sun2024easy} finds that a reward model trained on easier mathematical problems can be effectively transferred to harder problems, facilitating generalization through reinforcement learning. ~\citet{shin2024weak} identifies overlap data points—instances containing both easy and hard patterns—as a key mechanism for weak-to-strong generalization, allowing weak models to pseudolabel easier patterns while stronger models use these labels to learn harder patterns. Our work shows that a similar mechanism emerges naturally within self-improvement, where progressively increasing difficulty enables models to generate useful supervision signals for harder tasks without explicit human intervention.

\paragraph{Self-Improvement.}

When high-quality training labels are unavailable or costly to obtain, training on self-generated labels provides an efficient way to broaden the capabilities of a model. Typically, this involves generating candidate labels, filtering or verifying them to prune errors, and retraining on the refined self-generated data~\citep{zelikman2022star,wang2022selfinstruct,huang2022large,singh2023beyond,chen2023teaching,gulcehre2023reinforced,madaan2024selfrefine,yuan2024self,liang2024sheep,pang2024iterativereasoningpreferenceoptimization}. This approach has been successfully applied across various domains, including reasoning~\citep{zelikman2022star,huang2022large,singh2023beyond,pang2024iterativereasoningpreferenceoptimization}, mathematics~\citep{zhang2023chain,charton2024patternboost,alfarano2024global,liang2024sheep}, coding~\citep{chen2023teaching}, and general instruction tuning~\citep{wang2022selfinstruct,yuan2024self}.
Recent studies further analyze and refine the self-improvement process. \citet{song2024mind} identify the generation-verification gap as a key limiting factor, while \citet{huang2024selfimprovementlanguagemodelssharpening} introduce a "sharpening mechanism" that improves reliability by training on best-of-N model outputs.
Our work builds on STaR~\citep{zelikman2022star} and ReST~\citep{gulcehre2023reinforced}, leveraging iterative prediction, filtering, and fine-tuning to enhance model capabilities.

\paragraph{Model Collapse.}
A key challenge in self-improvement is model collapse, where iterative training on self-generated outputs leads to performance degradation~\citep{,Hataya_2023_ICCV,Arcaute2023CombiningGA,Alemohammad2023SelfConsumingGM}. While some work suggests this degradation is inevitable~\citep{shumailov2024ai,shumailov2023curse,zhang2023chain}, several mitigation strategies have emerged, including maintaining original training data~\citep{gerstgrasser2024model}, careful data mixing~\citep{gerstgrasser2024model,Dohmatob2024ATO,Briesch2023LargeLM}, and reliable verification mechanisms~\citet{gillman2024self,feng2024beyond}. Our approach incorporates these insights through unsupervised filtering techniques and controlled data generation, effectively preventing collapse while enabling sustained improvement.
\textit{We provide additional discussion of related works in Appendix~\ref{sec:related_work_extended}. 
}%

\section{Preliminaries and Experimental Setup}\label{sec:prelim}

\begin{table*}%
\centering
\caption{Examples of Tasks Considered %
}
\label{tab:task_examples}
  \vspace{-1mm}
  \centering
  \small
    \setlength{\tabcolsep}{4pt} %
    \renewcommand{\arraystretch}{1.0}
		 {
\begin{tabular}{c|l|c}
\toprule
\textbf{ Task Type} & \multicolumn{1}{c|}{\textbf{Input} (\textbf{Q}: Prompt, \textbf{A}: label) } & \textbf{Task Difficulty} \\ \hline
Reverse Addition & Q: 31558+91786= \, A: 232451  & \multirow{3}{*}{ \shortstack[c]{\vspace{1mm} \\ Max digit length of \\the two operands}}  \\ %
Forward Addition & Q: 85513+68719= \, A: 154232 &  \\ %
\shortstack[l]{Multiplication \\ \quad } & \shortstack[l]{Q: 34895*148= \\ \quad} \hspace{4mm} \shortstack[l]{A: 348950+0273932(3653542)\\ \quad +00447874=36972305} & \\ \midrule
Copy & Q: 12345= \hspace{10mm} A:12345 & \multirow{2}{*}{Length of string} \\ %
Reverse & Q: 12345= \hspace{10mm} A: 54321 &  \\ \midrule
Maze Solving &
\begin{tabularx}{0.6\linewidth}{L@{}L@{}}\includegraphics[width=0.2\textwidth]{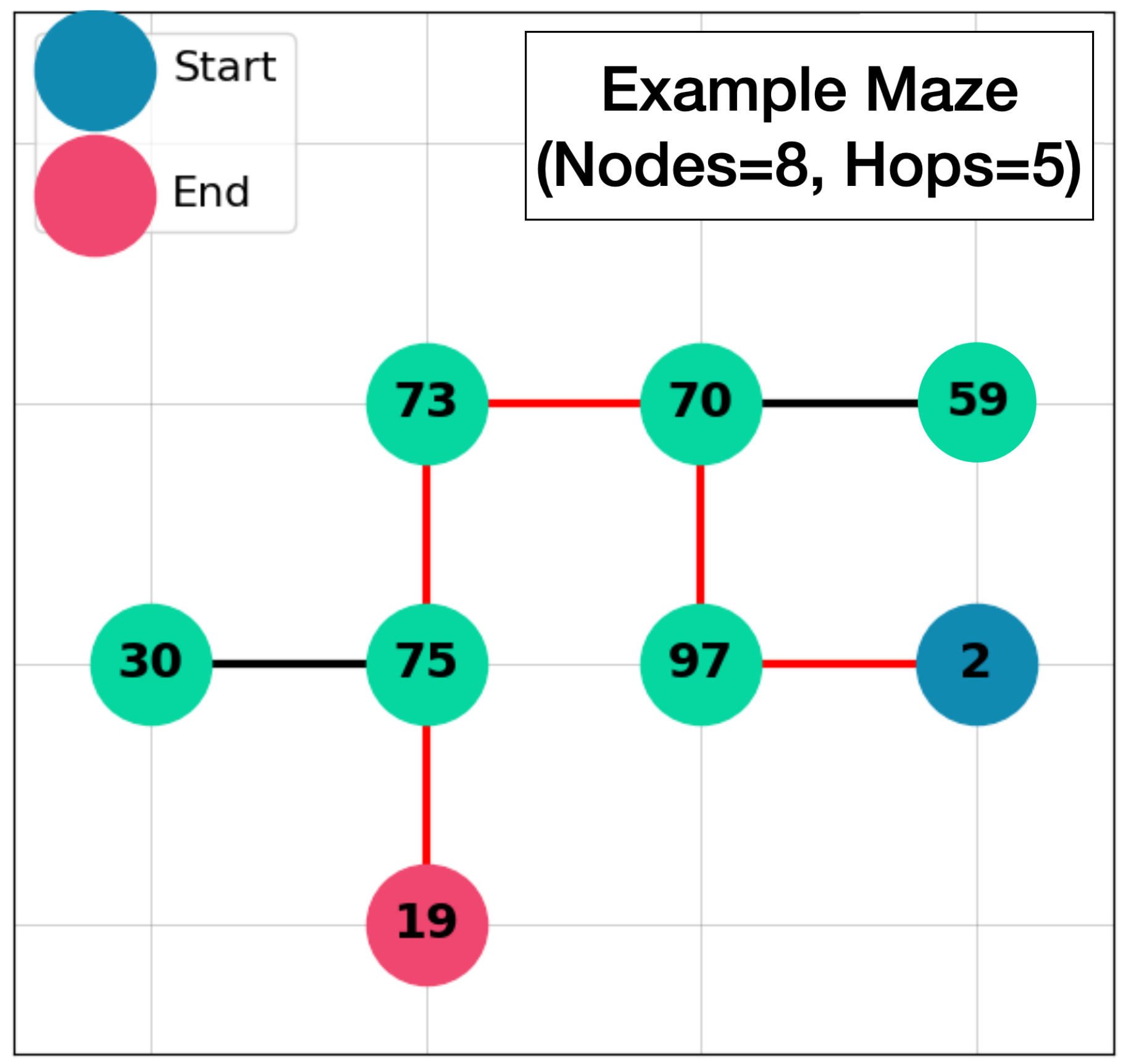} & \hspace{-12mm} \shortstack[l]{Finding shortest path from node {\color{blue}2} to {\color{red}19}\\ {\scriptsize($\leftarrow$ example image for illustration)}\\
\vspace{2mm} \\Q: {\color{blue}2}>{\color{red}19}\#73:70,75-97:2,70-70:73,97,59\\-75:73,30,19-2:97-30:75-59:70-19:75=\\ \vspace{3mm} \\ A: 2>97>70>73>75>19 \vspace{4mm} }\hspace{-5mm} 
\end{tabularx} 
& \shortstack[l]{ (1) Number of hops \\ between start \& end \\ \vspace{1mm} \\ (2) Number of nodes} \\ %

\bottomrule
\end{tabular}
}
\vspace{-3mm}
\end{table*}

In this section, we describe the experimental setup, including the model architecture, tasks, training methodology, evaluation criteria, and the self-improvement framework.

\paragraph{Models. }
We adopt the LLaMA architecture with six layers, six attention heads, and an embedding dimension of 384 and a total of 14M parameters. Positional embeddings are excluded, using the No Positional Encoding (NoPE) method~\citep{kazemnejad2024impact}. To evaluate applicability to large language models (LLMs), we extend our experiments to pretrained models (Llama-1B, Llama-3B) in Section~\ref{sec:pretrained}. Character-level tokenization is used across all tasks, except for the maze-solving task, where numbers (0–99) are tokenized as individual tokens instead of characters.

\paragraph{Tasks. }
We evaluate our approach on a diverse set of tasks, categorized into arithmetic operations, string manipulation, and maze solving. Table~\ref{tab:task_examples} provides examples for each task.

\begin{itemize}[left=10pt]
\item \textbf{Arithmetic operations:} 
\begin{enumerate} 
\item \textit{Addition} (Section~\ref{sec:reverse_addition},~\ref{sec:forward_addition}):  We consider both reverse and forward addition of two numbers of equal length. In reverse addition, both operands and the answers are reversed, so they are written with the least significant digit first. Forward addition, in contrast, follows the standard format, with the most significant digit first.%
\item \textit{Multiplication} (Section~\ref{sec:mult}): Multiplication tasks are presented in a chain-of-thought (CoT) data format~\citep{deng2024explicit}, which includes intermediate steps to guide the computation. %
\end{enumerate}
\item \textbf{String manipulation:} 
\begin{enumerate} 
\item \textit{Copy} (Section~\ref{sec:string_copy}): The task is to replicate the input sequence exactly. %
\item \textit{Reverse} (Section~\ref{sec:string_copy}): The task is to reverse the input sequence.
\end{enumerate} 
\item \textbf{Maze solving} (Section~\ref{sec:maze}): The task is to find the shortest path between a start node and an end node in a tree-structured graph. The shortest path is defined as the path with the fewest number of hops, where each hop represents a transition between two adjacent nodes.
\end{itemize}

Each task presents distinct challenges that test different aspects of model generalization. Reverse addition~\citep{lee2023teaching} has been widely adopted task for length generalization. Forward addition, by contrast, is significantly harder due to its increasing dependency in length, making it more challenging for transformers to extrapolate~\citep{zhou2023algorithms}. Copying and reversing sequences are considered fundamental operations but remain difficult for standard transformers without architectural modifications~\citep{anil2022exploring,zhou2023algorithms}. Multiplication is challenging even in-distribution~\citep{dziri2024faith}, and fine-tuning large language models with CoT reasoning has shown limited success. Finally, maze solving extends our evaluation to structured reasoning tasks, requiring models to internalize search behavior rather than relying on local token-to-token dependencies~\citep{bachmann2024pitfalls}. These tasks collectively provide a controlled yet diverse testbed for studying the effectiveness of self-improvement across different problem domains.

\begin{figure}
    \centering
    \includegraphics[width=0.95\linewidth]{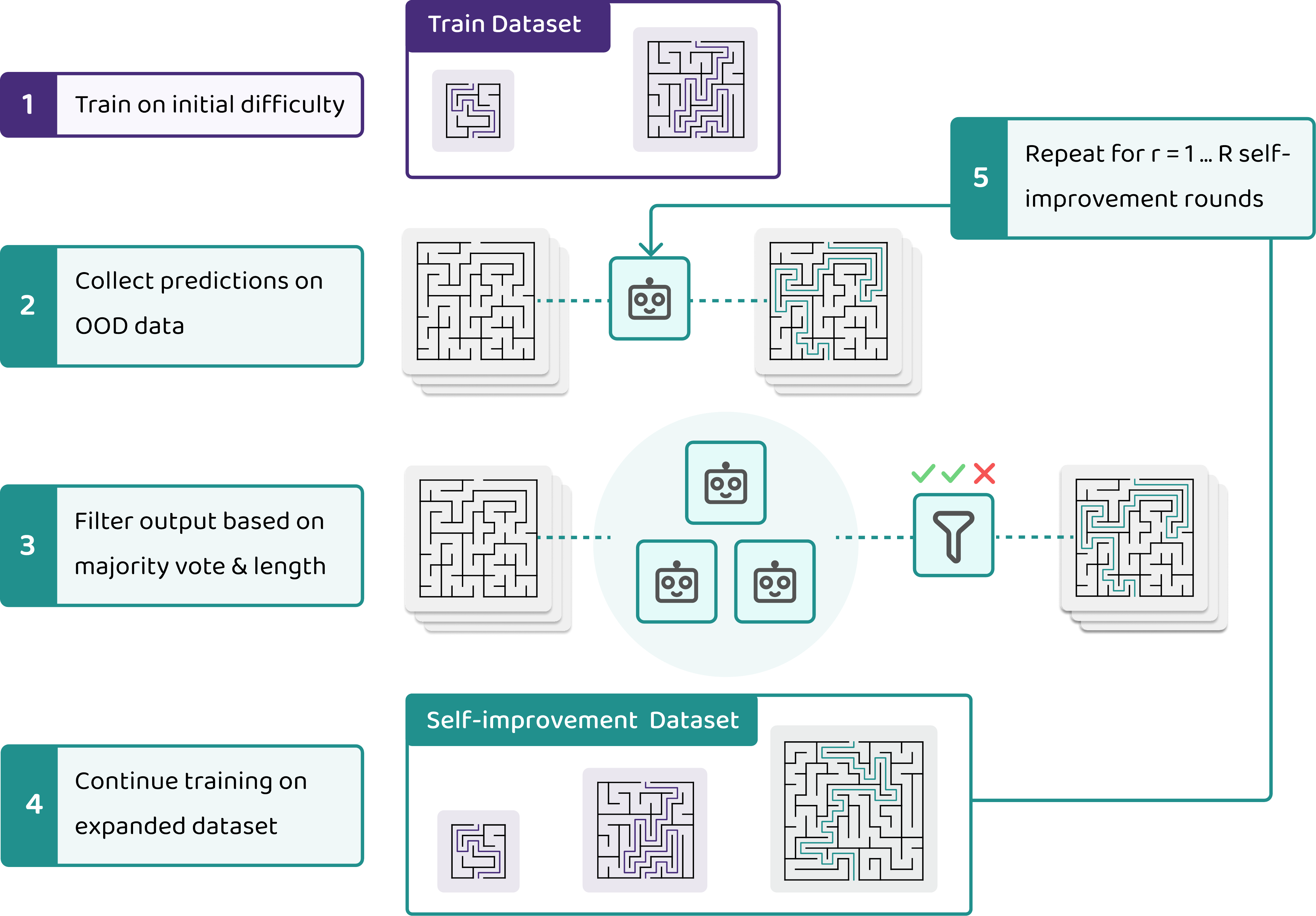}
    \caption{Illustration of our self-improvement procedure. At each round, the training data is updated with the model's predictions on progressively harder problems. }
    \label{fig:SI_illustration}
\end{figure}

\paragraph{Task Difficulty. }
All tasks we consider admit a straightforward notion of difficulty. 

\begin{itemize}[left=10pt]
    \item \textbf{Addition: } The maximum length of the two operands.
    \item \textbf{Multiplication: } The maximum length of the two operands. Intuitively, a 5-by-5 multiplication problem is harder than that of 4-by-4. But even a 6-by-1 multiplication is considered harder than 5-by-5, because the model has never seen training data containing length more than 5.
    \item \textbf{String Copy \& Reverse: } The length of the input string. 
    \item \textbf{Maze Solving: } We define difficulty as 1) the number of hops between the start and end nodes and 2) the total number of nodes in the graph. The number of hops corresponds to the number of transitions required to reach the goal. 
\end{itemize}

We denote the difficulty level of a problem instance $x$ as an integer $\text{Difficulty}(x)$.

\paragraph{Data Generation and Sampling. }
We generate an initial supervised training dataset $\mathcal{D}_0$ of up to a fixed difficulty level $d_0$ by uniformly sampling the difficulty level $d \leq d_0$, followed by independent sampling of the data conditioned on the difficulty. Denoting the input as $x_i$, labels as $y_i$,

\begin{align*}
    \mathcal{D}_0=\{(x_i,y_i)\}_{i=1}^{N_0},\quad\text{where 
 }\text{Difficulty}(x_i)\leq d_0. 
\end{align*}

For arithmetic tasks such as addition or multiplication, each problem instance is represented as a tuple $x_i = (a_i, b_i)$, with $\mathcal{D}_0$ containing problems of up to $d_0$-digit numbers. The digit lengths $(d_{a_i}, d_{b_i})$ are uniformly sampled from $\{1, \dots, d_0\}^2$, and the numbers $a_i$ and $b_i$ are uniformly sampled from the ranges $[10^{d_{a_i}-1}, 10^{d_{a_i}}-1]$ and $[10^{d_{b_i}-1}, 10^{d_{b_i}}-1]$, respectively.

For string manipulation tasks (e.g., copying or reversing), we uniformly sample string lengths up to $d_0$ and generate random sequences. Similarly, for maze-solving tasks, we uniformly sample the number of hops or total nodes in the maze and generate random graphs that satisfy these constraints. %
This strategy ensures balanced coverage across all difficulty levels up to $d_0$. %

\paragraph{Self-Improvement Framework. }

The self-improvement framework begins by training a model using the labeled training dataset $\mathcal{D}_0$, which gives us our base model $M_0$. %

For each subsequent round $r$ ($r = 1, 2, 3, \dots$), we increase the problem difficulty, such as the number of digits or string length for arithmetic and string manipulation tasks, or the number of hops for maze-solving tasks, to $d_r$. Using the previous model $M_{r-1}$, we generate $N_r$ new self-improve data samples $\mathcal{D}_r$ defined as:

\begin{align*}
    \mathcal{D}_r = \{(x_i, {\color{blue}M_{r-1}(x_i)})\}_{i=1}^{N_r},\quad \text{where }d_{r-1}\leq \text{Difficulty}(x_i)\leq d_r
\end{align*}
Instead of the true labels $y_i$, we obtain the predicted labels $M_{r-1}(x_i)$ from the output of the model.

At each self-improvement round \( r \), the model is trained on the combined dataset \( \mathcal{D}_0 \cup \mathcal{D}_1 \cup \dots \cup \mathcal{D}_{r-1} \), which includes the initial labeled dataset and all subsequent self-improvement datasets. To ensure sufficient training on the most recently generated data \( \mathcal{D}_{r-1} \), we up-sample it with a sampling probability of 50\%. The remaining datasets \( \mathcal{D}_0, \dots, \mathcal{D}_{r-2} \) are sampled uniformly at random. %
This iterative process allows the model to gradually tackle harder problems, leveraging its own predictions to expand the training data and improve generalization. 

\paragraph{Data Filtering. }
We employ two unsupervised data-filtering methods to refine our self-improvement dataset: 1) length filtering and 2) majority voting.  For a given self-improved dataset \( \mathcal{D}_r = \{(x_i, M_{r-1}(x_i))\}_{i=1}^{N_r} \) at round \( r \), data is filtered based on specific criteria on the model-generated outputs \( M_{r-1}(x_i) \), producing a smaller, refined dataset \( \tilde{\mathcal{D}}_r = \{(x_i, M_{r-1}(x_i))\}_{i=1}^{\tilde{N}_r} \). We given detailed motivations and description in Section~\ref{sec:data_filter}.

\paragraph{Training and Evaluation. } Except for the experiments on pretrained Llama 3.2 models, all models are trained from scratch using the conventional next-token prediction objective. The loss is computed solely on the completion, meaning that the input prompt is masked, and only the model's predictions are included in the loss computation. Detailed settings, including hyperparameters and training schedules, are provided in the Appendix.

During inference, we use greedy decoding and exact-match accuracy as the primary metric for evaluation. A prediction is deemed correct if all tokens in the output sequence match the ground truth; any discrepancy in the generated tokens is classified as an incorrect prediction.
\begin{figure}
    \centering
    \includegraphics[width=0.8\linewidth]{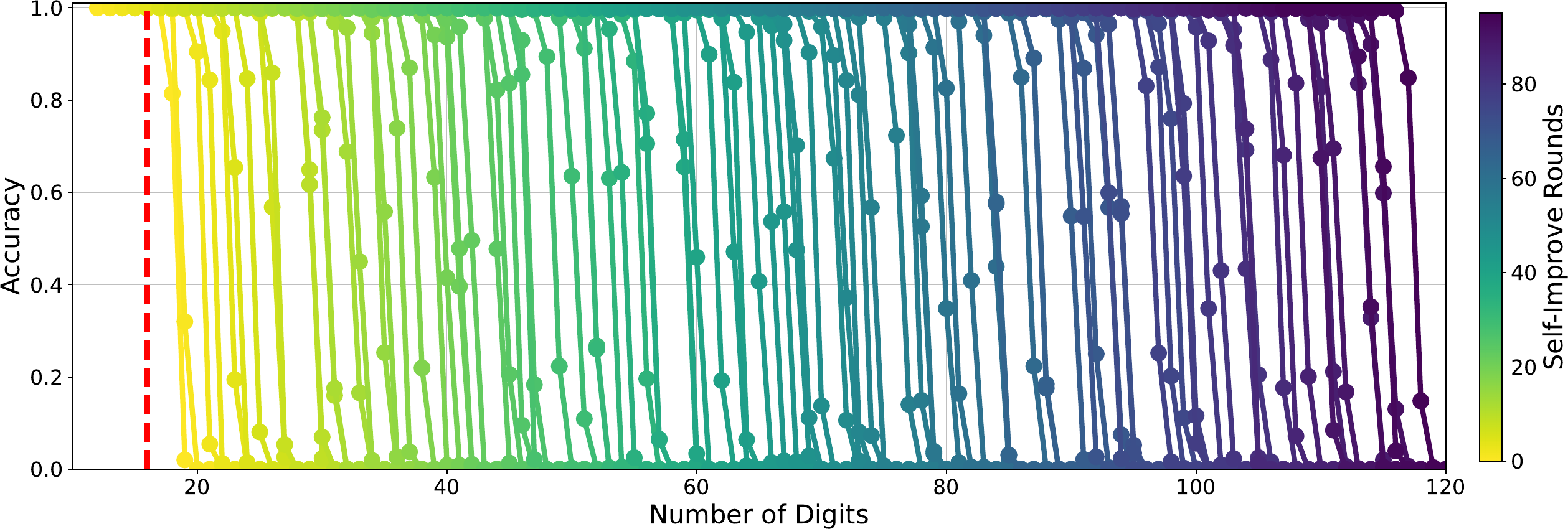}
    \caption{Results on the reverse addition task, where both operands and the output are represented in reverse order, with the least significant digit first. The self-improvement framework enables a model initially trained on 1-16 digit examples to generalize perfectly to over 100-digit addition. }
    \label{fig:result_reverse_add}
\end{figure}

\section{Length Generalization on Reverse Addition and String Copying/Reversing}\label{sec:addition}
\begin{finding}
    With self-improvement training, transformers can achieve state-of-the-art length generalization on reversed addition, string copying and maze solving, without further architectural modifications.
\end{finding}

In this section, we apply our self-improvement framework to relatively simple tasks such as reverse addition and string copying/reversing. These tasks serve as testbeds to demonstrate that our framework can extend model capabilities far beyond the original training distribution through multiple iterations of self-improvement, even in the absence of additional data filtering.

\subsection{Reverse Addition}\label{sec:reverse_addition}

Reversed addition, where the operands and output are written with the least significant digit first, has been shown to enhance sample efficiency and performance~\citep{lee2023teaching}. Reversed addition has become a popular setting for studying length generalization in arithmetic tasks~\citep{lee2023teaching,shen2023positional,zhou2023algorithms,zhou2024transformers,Cho2024PositionCI,mcleish2024transformers}. 
Here, we show that the self-improvement framework achieves substantial length generalization on reversed addition without any modifications to positional encodings, input formats, or the Transformer architecture.

\paragraph{Setting.} 
The initial supervised dataset $\mathcal{D}_0$ contains 2 million examples of reverse addition, with operand lengths ranging from 1 to 16 digits. This dataset is used to train the model for 10,000 steps. In subsequent self-improvement rounds, we sample 50,000 additional training examples at each round, extending the operand length by one digit. Specifically, at self-improvement round \( r \), the self-generated data $\mathcal{D}_r$ consists of length-\( 16+r \) examples produced by the model \( M_r \). The model is fine-tuned on the combined dataset $\mathcal{D}_0 \cup \mathcal{D}_1 \cup \dots \cup \mathcal{D}_r$ for 1,500 steps, resulting in an improved model \( M_{r+1} \).

\paragraph{Results.} Figure~\ref{fig:result_reverse_add} demonstrates that, starting with a model trained on 1 to 16-digit reverse addition, the self-improvement framework enables near-perfect length generalization up to 100 digits without any additional supervision. 

\begin{figure}
    \centering
    \includegraphics[width=0.75\linewidth]{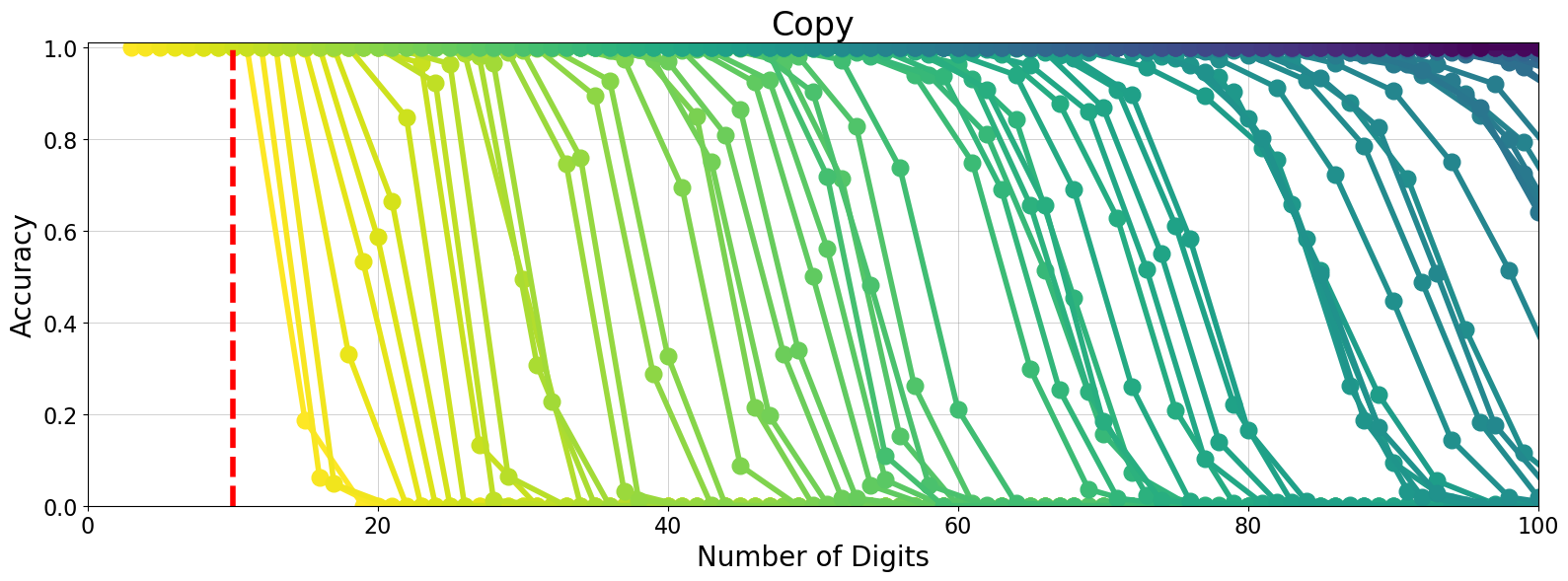}
    \includegraphics[width=0.75\linewidth]{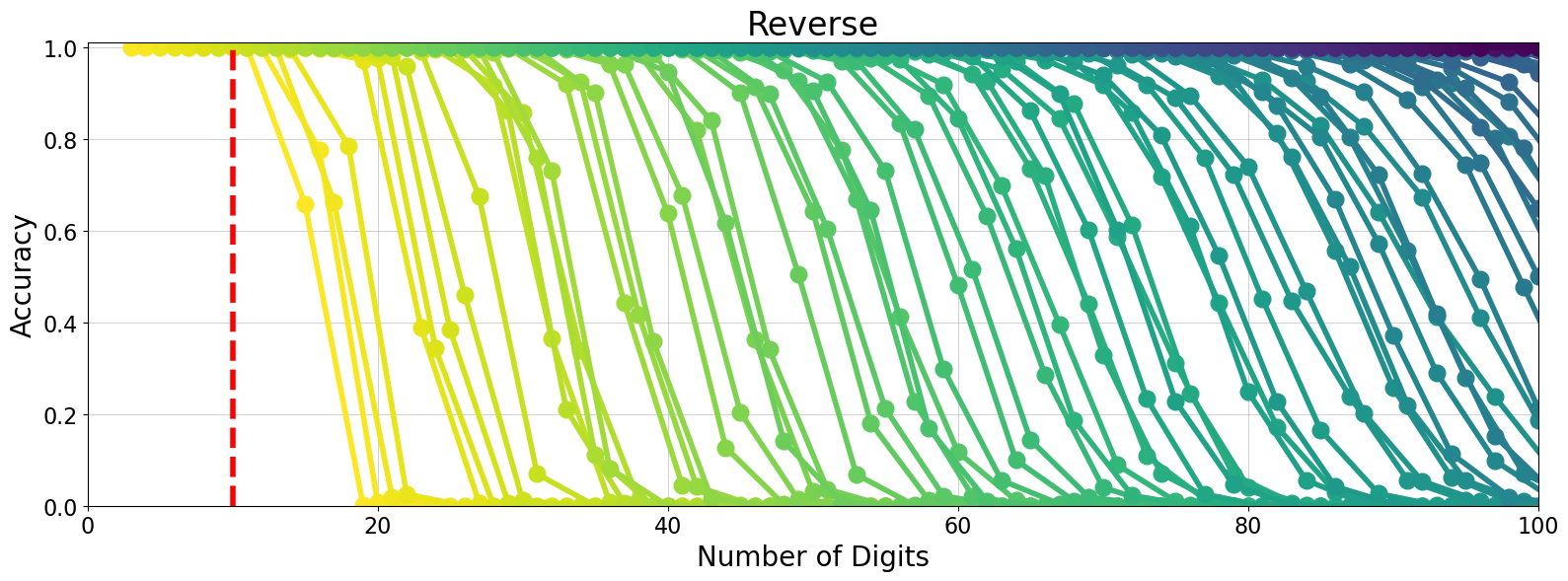}
    \caption{Results on string manipulation tasks. (Top) Copy: the model replicates the input string exactly. (Bottom) Reverse: the model outputs the input string in reverse order. The model initially trained on strings of length 1 to 10 generalizes to sequences of over 120. %
    }
    \label{fig:result_copy_reverse}
\end{figure}

\subsection{String Copy \& String Reverse}\label{sec:string_copy}

Copying and reversing a given input string is another task that is considered hard for vanilla transformers~\citep{anil2022exploring,zhou2023algorithms}. The input string consists of digits from 0 to 9. 

\paragraph{Setting.} 
The initial training set $\mathcal{D_0}$ consists of 2 million examples of strings of length 1 to 10. The vocabulary of the string is the digits $0$ to $9$. For each subsequent round $r$, we sample $D_{r}$ consisting of $50000$ examples of length $10+r$ from the model $M_r$. Then we continue training $M_r$ on the combined dataset $D_1\cup\dots\cup D_r$ for 500 steps to obtain $M_{r+1}$. 

\paragraph{Results.} 
Figure~\ref{fig:result_copy_reverse} demonstrates that starting with strings of length 1 to 10, the self-improvement framework enables the model to perfectly generalize to string lengths of over 120 after approximately 100 self-improvement rounds.

\section{Unsupervised Data Filtering}~\label{sec:data_filter}
\begin{finding}
     Filtering self-improvement data using length-based filtering and majority voting is crucial for sustaining the self-improvement process while keeping the method general-purpose.
\end{finding}

\subsection{Motivation for Data Filtering}

\begin{wrapfigure}{r}{0.45\textwidth}
    \vspace{-11mm}
    \centering
    \includegraphics[width=0.43\textwidth]{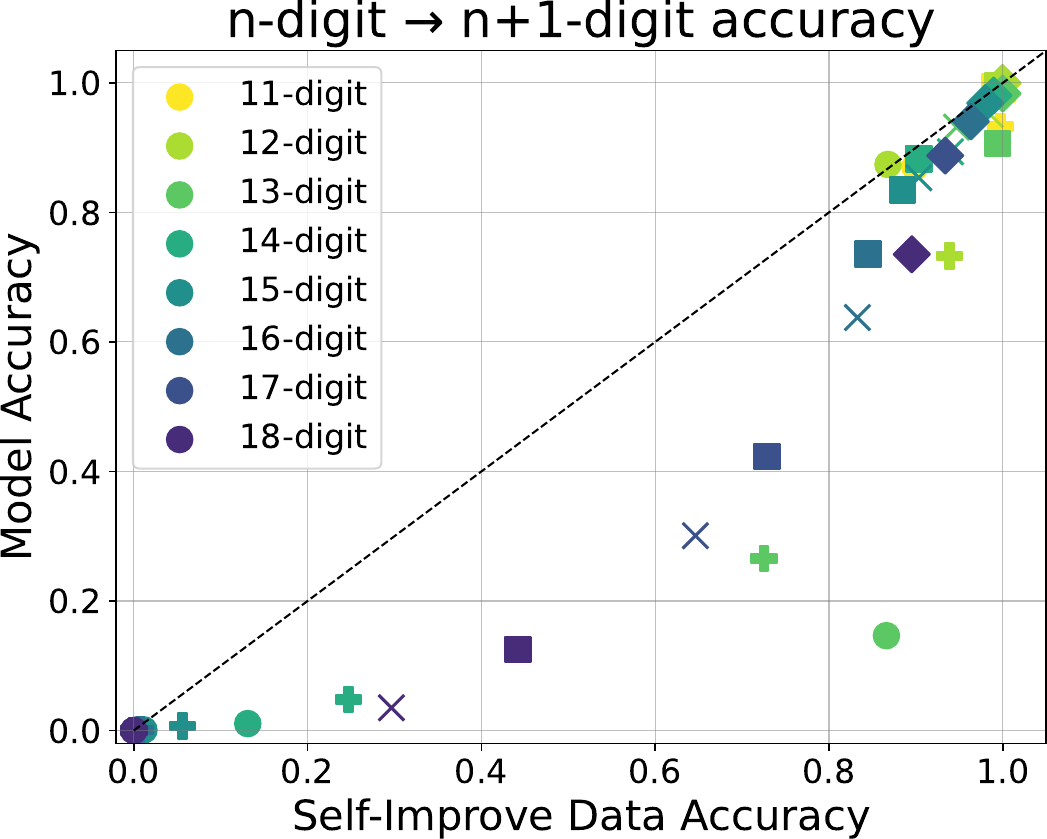}
    \caption{Effect of self-generated data accuracy on length generalization performance in the reverse addition task. Each data point represents the accuracy of the self-improve data $\mathcal{D}_r$ (on $n$ digit addition) generated by model $M_{r-1}$, and the resulting $n+1$-digit performance of the trained model $M_r$ at round $r$. The prevalence of points below the $y=x$ line highlights the critical importance of high-quality data for successful self-improvement.}
    \vspace{-8mm}
    \label{fig:reverse_add_si}
\end{wrapfigure}

The self-improvement framework operates on the principle that the model can generalize slightly beyond the difficulty level it was trained on, incrementally generating self-improve data of increasing difficulty. %
A critical component for this framework to succeed is the quality of the self-generated data. Low-quality data can negatively impact the model’s generalization performance, leading to even lower-quality data in subsequent rounds and ultimately causing a cascading degradation of the self-improvement process.

Figure~\ref{fig:reverse_add_si} demonstrates this effect in the reverse addition task. The x-axis represents the accuracy of the self-improve dataset $\mathcal{D}_r$, generated by model $M_{r-1}$ at round $r$, while the y-axis shows the resulting $n+1$-digit performance of model $M_r$. The prevalence of data points below the $y=x$ line indicates that low-quality data diminishes performance, underscoring the need for maintaining high-quality data throughout the self-improvement process. These cascading error effects are analyzed in greater detail in Section~\ref{sec:error_analysis}.

\begin{figure}[ht]
    \centering
    \includegraphics[width=0.3\linewidth]{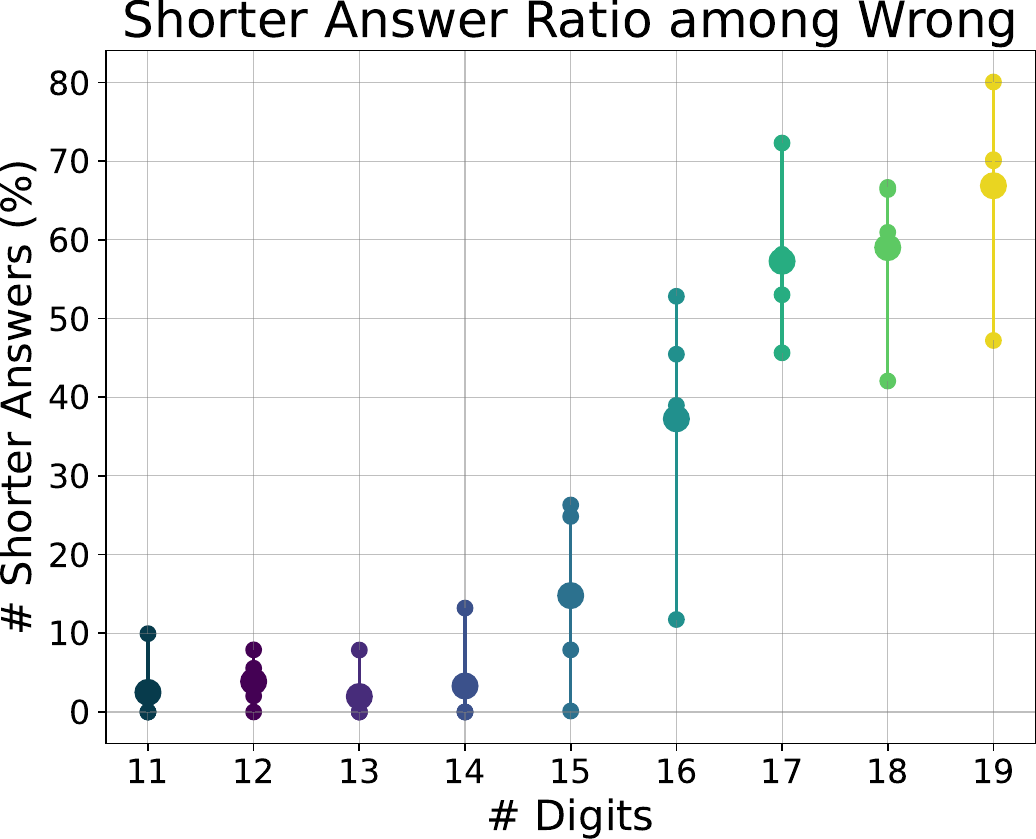}
    \includegraphics[width=0.33\linewidth]{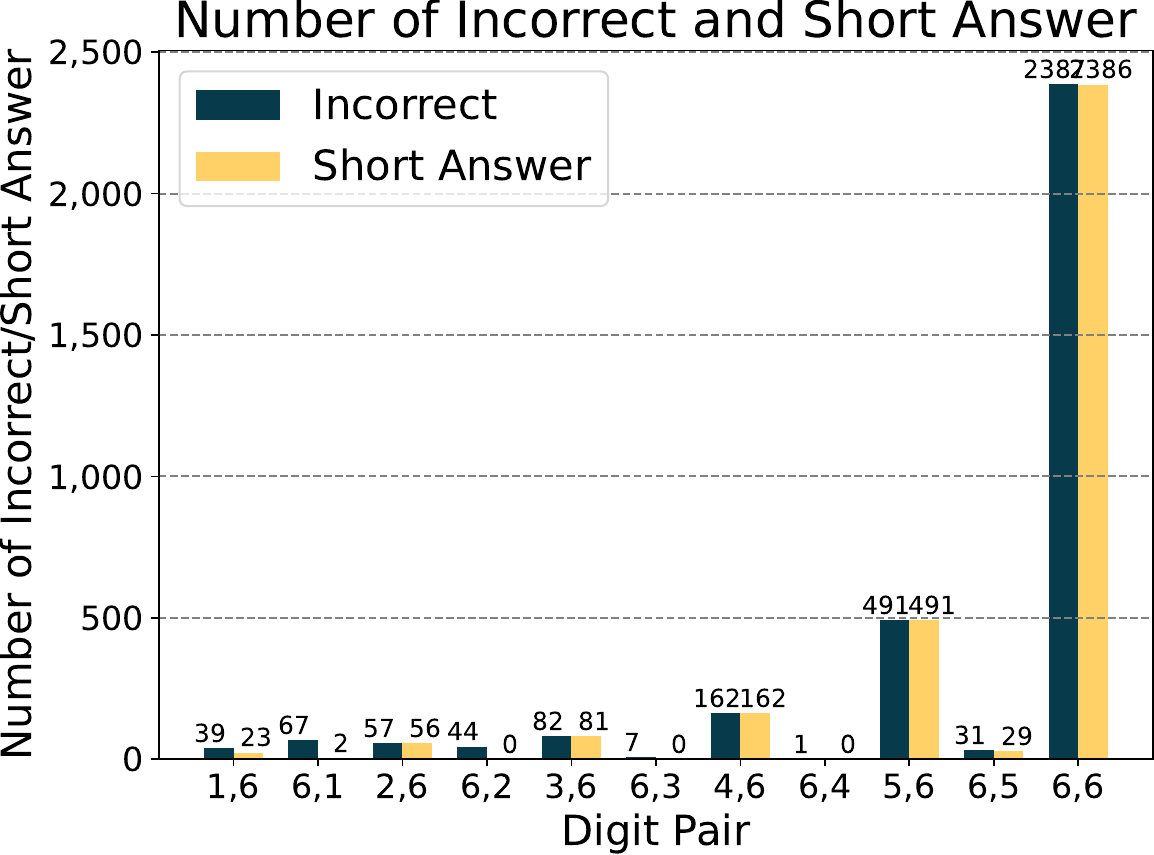}
    \includegraphics[width=0.33\linewidth]{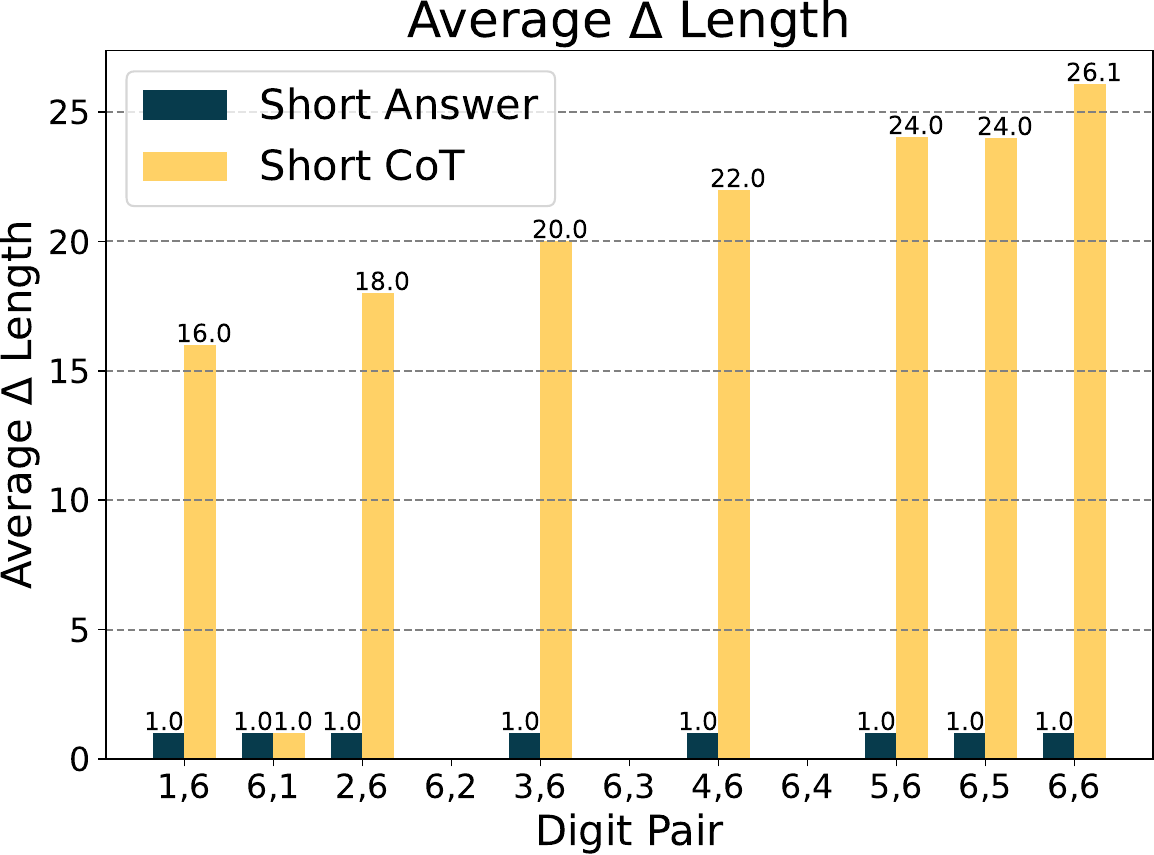}
    \caption{OOD results are often short. (Left) Reverse addition task: the proportion of shorter answers among incorrect predictions increases with each round. (Mid \& Right) CoT-multiplication task with majority voting: (Mid) The majority of incorrect answers are short. (Right) The average length discrepancy of short answers compared to the correct answer or the CoT reasoning part.}
    \label{fig:num_shorter_answers}
\end{figure}

Additionally, we make the following observations: 
\paragraph{OOD Results are often Short. } A common error in model-generated data for tasks with higher difficulty than the training data is that the generated labels are often shorter than the correct answers. Figure~\ref{fig:num_shorter_answers} illustrates this phenomenon for both the reverse addition and CoT-multiplication tasks. In reverse addition (Left), as the number of digits in the training data increases (or as self-improvement rounds progress), the proportion of incorrect self-generated data where the answer is shorter than the correct label length also increases. Similarly, for CoT-multiplication (Mid and Right), most incorrect answers are shorter than the correct ones. Furthermore, in cases where the answers are shorter, the outputs often miss one or more reasoning steps in the chain-of-thought (CoT) reasoning process.

\paragraph{Leveraging Label Diversity with Majority Voting. }
Self-improvement relies on the model's ability to generalize to slightly harder problems. However, this generalization is not always robust and can vary significantly across different training instances~\citep{zhou2024transformers}. Majority voting mitigates this variability by aggregating predictions across multiple independently trained models, thereby improving the reliability of self-generated labels.

\begin{figure}
    \centering
    \includegraphics[width=0.6\linewidth]{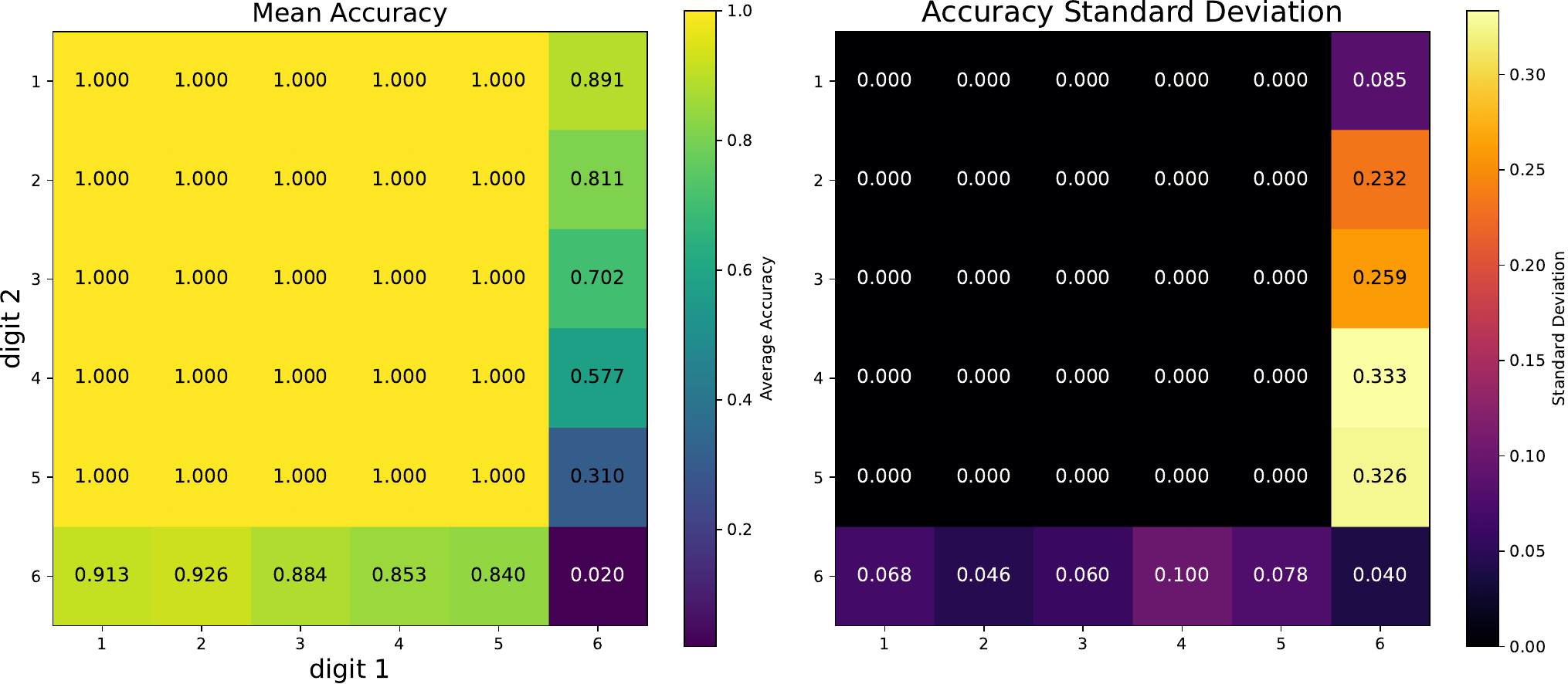}
    \hspace{1mm}
    \includegraphics[width=0.3\linewidth]{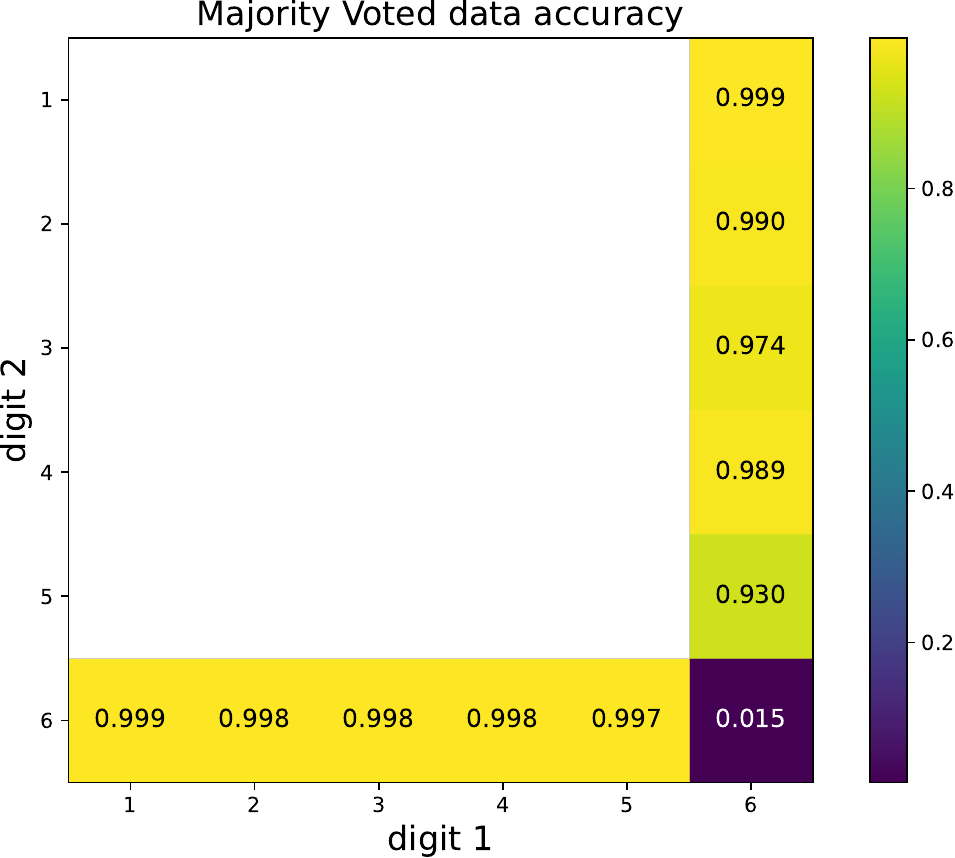}
    \caption{ Majority voting leverages label diversity. (Left \& Mid): Mean and standard deviation of accuracy among five models trained with different seeds on the initial training round. (Right): Accuracy of majority-voted data points. Majority voting significantly boosts data quality, with 5-by-6 multiplication data accuracy increasing from an average of 31\% to 93.3\% }
    \label{fig:majority_vote}
\end{figure}

Figure~\ref{fig:majority_vote} demonstrates the effectiveness of majority voting in the multiplication task across five models trained with different seeds during the initial training phase on data \( \mathcal{D}_0 \), which consists of up to 5-by-5 multiplication problems. The mean accuracy (Left) is relatively low, with a high standard deviation (Mid), indicating substantial variability among the models. By applying majority voting with a consensus on at least 4 out of 5 model outputs, the generated dataset quality improves significantly (Right). For example, while the 5-by-6 multiplication task achieves an average accuracy of 31\% across models, the majority-voting strategy generates a dataset with 93.3\% accuracy\footnote{
In practice, datasets for larger multiplications, such as 5-by-6 digits, are created after multiple rounds of self-improvement training, gradually incorporating \( m \)-by-6 and 6-by-\( m \) data with incrementally increasing \( m \) at each round.}.

\subsection{Data Filtering Methods}

\begin{figure}[ht]
    \centering
    \includegraphics[width=0.46\linewidth]{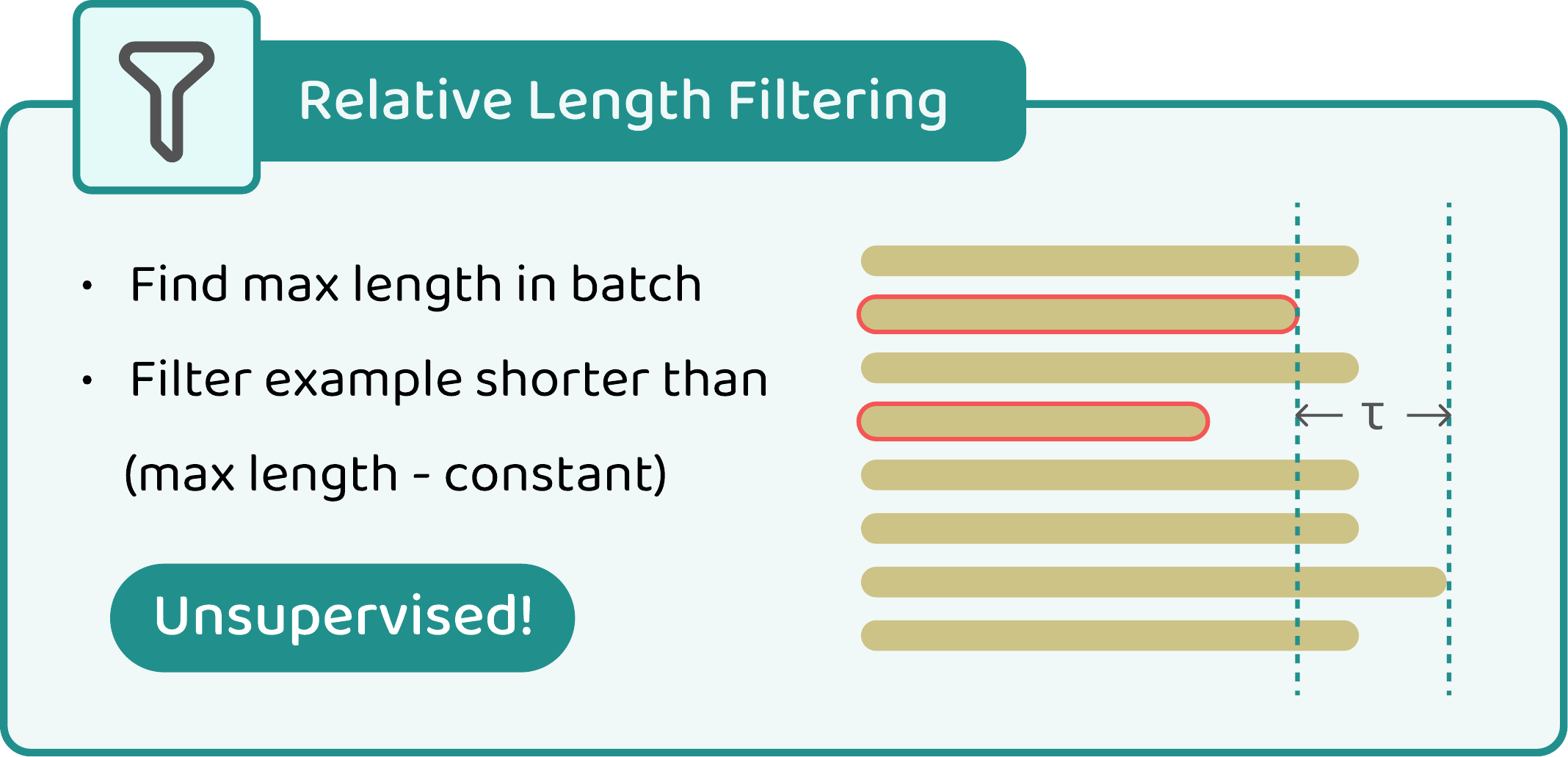}
    \hspace{0.03\linewidth}
    \includegraphics[width=0.46\linewidth]{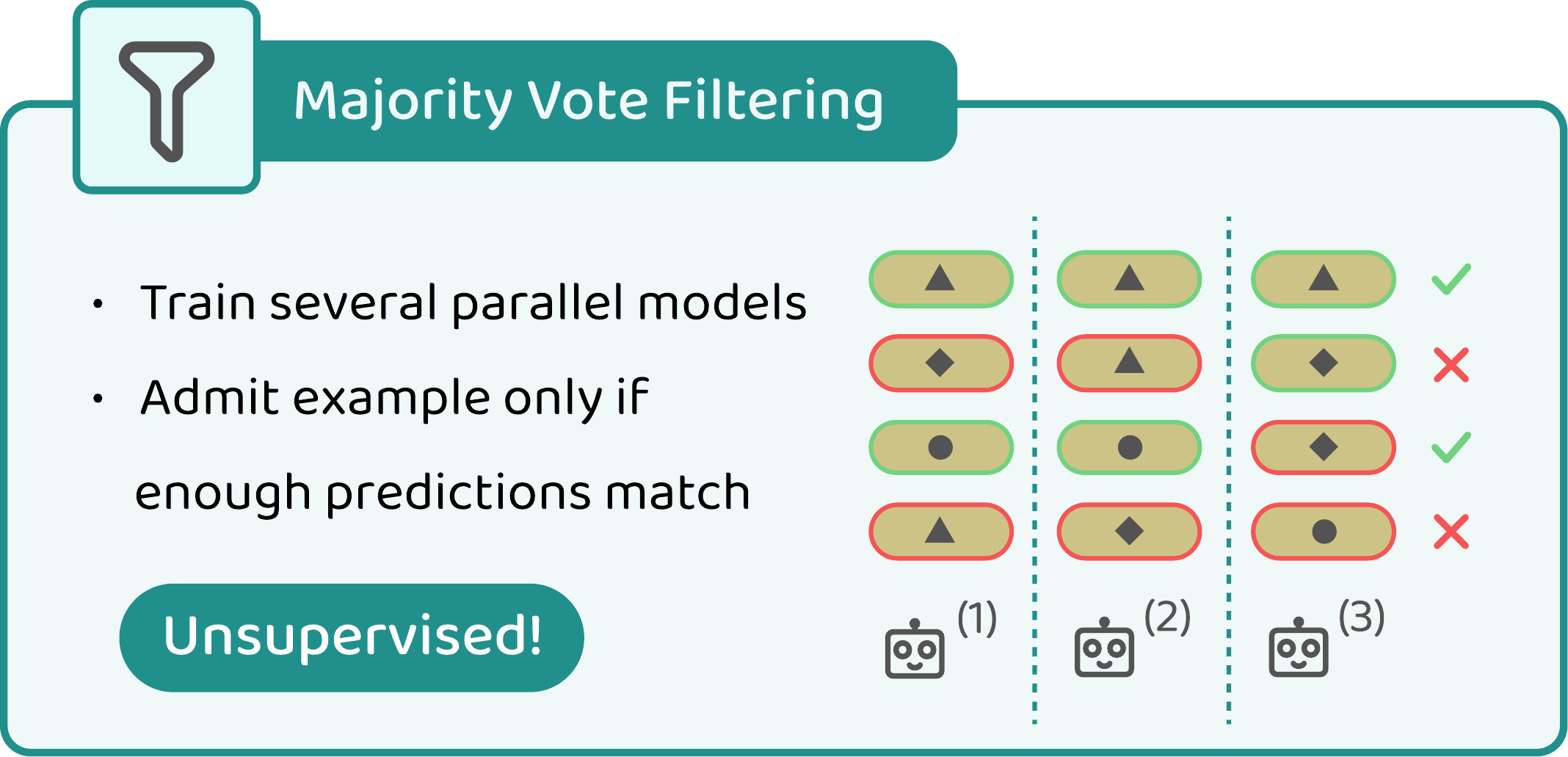}
    \caption{Overview of the two data-filtering methods employed. Length filtering removes data points with outputs shorter than a predefined threshold, relative to the maximum output length in the batch. Majority voting filters data based on consensus among predictions from multiple models trained with different random seeds. Both data filtering methods are \textit{unsupervised}. }
    \label{fig:data_filtering}
\end{figure}

We focus on two key data-filtering methods used in this work: length filtering and majority voting (illustrated in Figure~\ref{fig:data_filtering}).  For a given self-improved dataset \( \mathcal{D}_r = \{(x_i, M_{r-1}(x_i))\}_{i=1}^{N_r} \) at round \( r \), data is filtered based on specific criteria applied to the model-generated outputs \( M_{r-1}(x_i) \). This process produces a smaller, refined dataset \( \tilde{\mathcal{D}}_r = \{(x_i, M_{r-1}(x_i))\}_{i=1}^{\tilde{N}_r} \), which is then used for training in subsequent rounds. 

These filtering strategies are essential for sustaining the self-improvement framework and enabling effective generalization across diverse tasks, including forward addition, multiplication, and maze-solving.

\paragraph{Relative Length Filtering. } 
The observation that OOD results are often short motivates a filtering method based on the relative lengths of model-generated predictions. Specifically, predictions shorter than a predefined threshold—calculated relative to the maximum prediction length within their batch—are filtered out. For a batch of model-predicted outputs, we identify the maximum length of the output $L=\max |M_{r-1}(x_i)|$ and filter out predictions $M_{r-1}(x_i)$ with lengths shorter than a predefined threshold $\tau$. This method is \textit{unsupervised}, as it relies solely on comparing lengths within model-generated outputs rather than referencing ground-truth labels. While particularly suited to length generalization tasks, where harder problems are expected to yield longer answers, length-based filtering demonstrates broader potential for addressing similar challenges in other tasks.

\paragraph{Majority Voting. } 
Generating multiple candidate answers to ensure self-consistency is a widely used approach for enhancing data quality~\citep{huang2022large,wang2022self,qu2024recursive,peng2024regenesis}. However, unlike the common practice of sampling multiple reasoning paths by generating outputs with a non-zero temperature, our task of interest requires a single correct answer for each instance. To address this, we train \( k \) models (\( M_{r-1}^{(1)}, \cdots , M_{r-1}^{(k)} \)) using different random seeds and self-improvement data, then apply a majority-voting mechanism with a threshold \( \tau \). 

Specifically, for each self-improved dataset \( \mathcal{D}_r^{s} = \{(x_i, M_{r-1}^{(s)}(x_i))\}_{i=1}^{N_r} \) where $s$ is the seed index, we filter the data such that only pairs \( \{(x_i, M_{r-1}^{(s)}(x_i))\} \) where \( M_{r-1}^{(s)}(x_i) \) matches at least \( \lceil \tau \times k \rceil \) outputs among the \( k \) models are retained. This ensures that only high-consensus data are preserved for training in subsequent rounds, thereby significantly improving overall data quality and model performance. This approach is conceptually similar to an iterative version of the bagging algorithm~\citep{breiman1996bagging}, where reliable pseudo-labels are generated through ensemble agreement to handle harder instances. Ablation on majority voting is provided in Section~\ref{sec:mv_ablations}.

\section{Length and Difficulty Generalization on Forward Addition, Multiplication, Maze}~\label{sec:harder_tasks}

\begin{finding}
    Augmenting self-improvement training with label filtering based on length and majority voting, transformer models can achieve length and difficulty generalization on forward addition, COT-multiplication and maze solving. 
\end{finding}

We extend our evaluation to a class of harder tasks, including forward addition, multiplication, and maze-solving. %
Our results demonstrate that the framework is not limited to length generalization but extends to \textbf{difficulty generalization}, where the model incrementally learns to solve increasingly difficult problems. By employing controlled sampling of problem difficulty and data filtering techniques for each round, the model successfully adapts to harder tasks, highlighting the versatility and robustness of the self-improvement approach.

\subsection{Forward Addition}\label{sec:forward_addition}
Forward addition is a straightforward yet challenging task for transformer models to length generalize effectively. ~\citet{zhou2023algorithms} hypothesize that Transformers are more likely to length generalize on tasks with small RASP-L complexity. They demonstrate that tasks such as reverse addition and copying have low RASP-L complexity, making them easier to length generalize, whereas forward addition poses a greater challenge. 
In reverse addition, each step only requires processing a fixed-size subset of the input. However, in the forward addition, the size of the relevant input required to generate correct tokens increases, making the problem more complex. 

\begin{figure}
    \centering
    \includegraphics[width=0.42\linewidth]{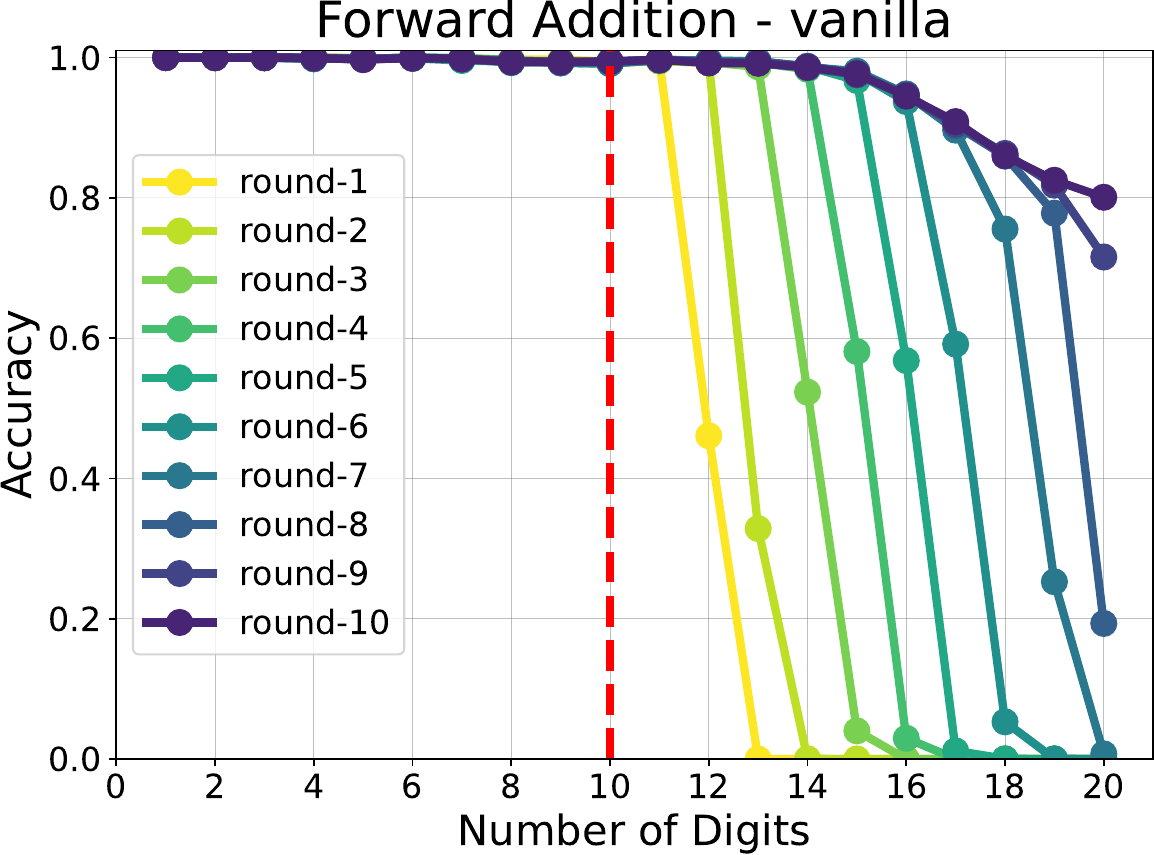}
    \hspace{2mm}
    \includegraphics[width=0.42\linewidth]{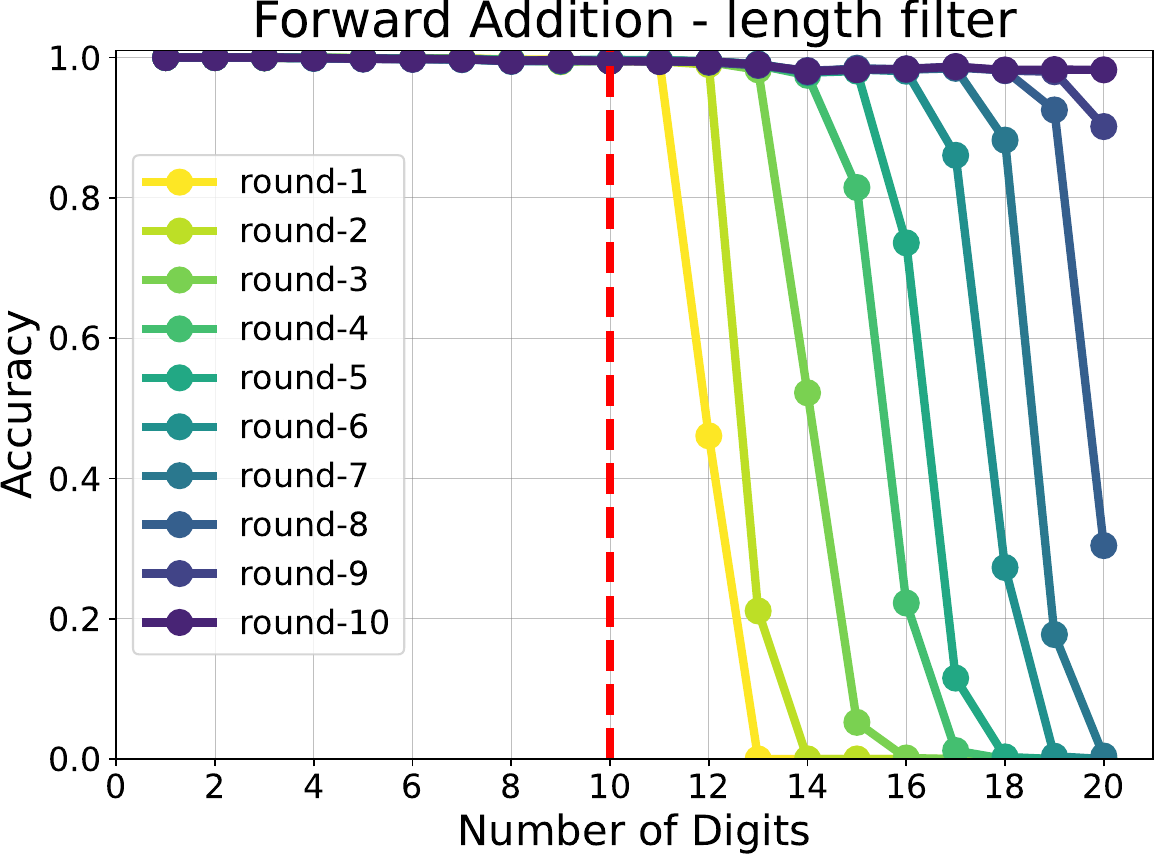}
    \caption{Models trained on forward addition over 10 self-improvement rounds. (Left) Without data filtering. (Right) With length-based filtering using a threshold of 2. Data filtering significantly enhances length generalization performance.}
    \label{fig:filtering_forward_add}
\end{figure}

\paragraph{Setting.}
The models are initially trained on a dataset $\mathcal{D}_0$ containing 2 million labeled examples of forward addition, with operand lengths ranging from 1 to 10 digits. This initial training phase spans 10,000 steps. In each subsequent self-improvement round, we generate 50,000 additional training examples, incrementally extending the operand length by one digit. Specifically, at self-improvement round \( r \), the self-generated dataset $\mathcal{D}_r$ contains length-\( 10+r \) examples produced by the model \( M_r \). The model is then fine-tuned for 3,000 steps on the combined dataset $\mathcal{D}_0 \cup \mathcal{D}_1 \cup \dots \cup \mathcal{D}_r$, resulting in an updated model \( M_{r+1} \).

\paragraph{Results.}

\begin{wrapfigure}{r}{0.5\textwidth}
    \vspace{-8mm}
    \centering
    \includegraphics[width=0.5\textwidth]{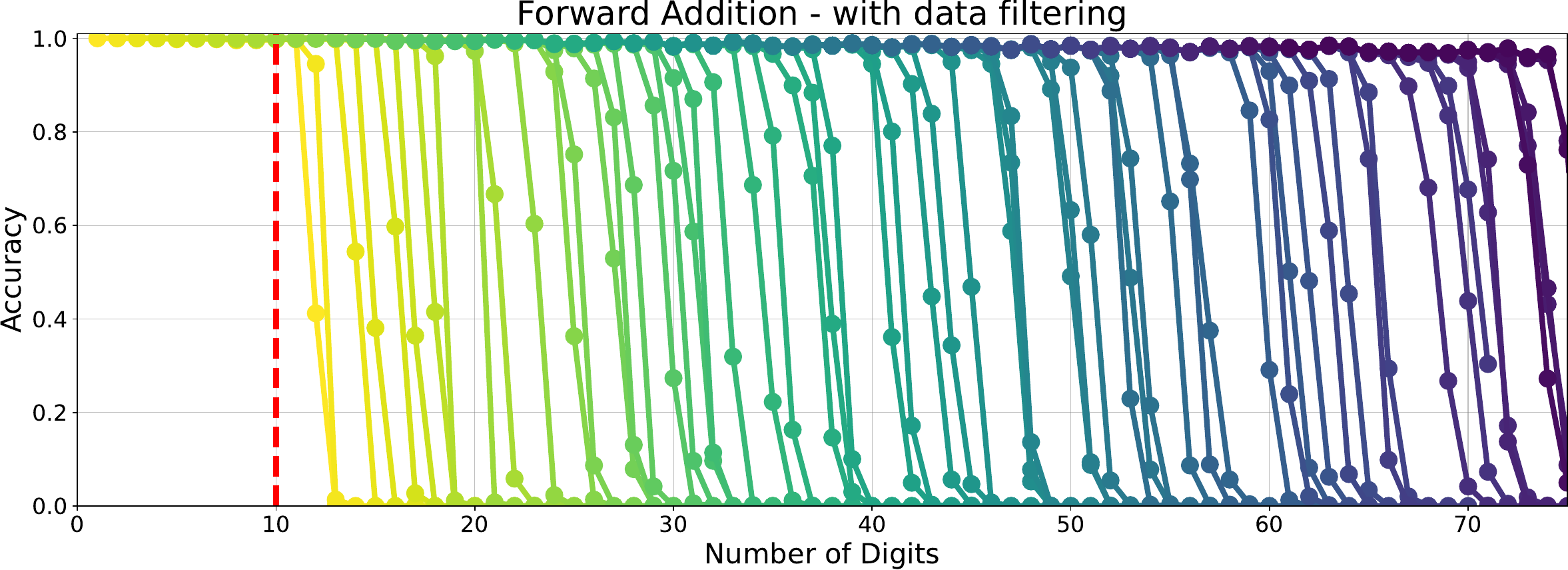}
    \caption{Results on the forward addition task with length filtering. The model is initially trained on labeled forward addition data of lengths 1 to 10. Using the self-improvement framework over 60 rounds, with incremental increases in digit length by 1 per round, the model achieves strong generalization to lengths up to 75.}
    \vspace{-5mm}
    \label{fig:forward_add}
\end{wrapfigure}

Figure~\ref{fig:filtering_forward_add} shows the results of forward addition experiments, where the model is initially trained on labeled data of up to 10 digits and then undergoes 10 rounds of self-improvement.
Without any data filtering (Left), the model's performance begins to deteriorate after a few rounds of training, leading to a collapse in generalization. However, applying the length-based filtering approach with a threshold length of 2 results in significant improvements in length generalization performance (Right). By refining the self-improvement dataset at each round, the self-improvement framework remains robust across multiple rounds.

With continued training over 60 self-improvement rounds, the model maintains performance exceeding 98\% accuracy for sequences up to length 70 (Figure~\ref{fig:forward_add}). This demonstrates the effectiveness of length-based filtering in sustaining the self-improvement process and enabling models to generalize to much longer sequences.

\paragraph{Relevance to Prior Work.} Our self-improvement framework and results are closely related to those of~\citet{zhang2023chain}, where self-training enables forward addition generalization from 6-digit examples to 24-digit addition. Like their approach, we iteratively apply self-training on progressively harder problems. However, a key distinction is that their method follows a two-step process in each round: first generating solutions using chain-of-thought (CoT) reasoning, then fine-tuning the model to produce direct answers without CoT.

\subsection{Multiplication}\label{sec:mult}
We also extend our approach on multiplication, which is a challenging task even in-distribution, as noted in~\citet{dziri2024faith}. Fine-tuning large language models on datasets with chain-of-thought(CoT) steps has shown limited success. We adopt a data format similar to~\citep{deng2024explicit}, where the input prompt is structured as \texttt{9172*9431=}, and the label expands the multiplication into stepwise additions, such as:
\texttt{17442+067801(132331)+0075180(1398490)+00091720=13976630}.
Each step includes the intermediate results (in parentheses) representing partial products formed by multiplying the first operand with each digit of the second operand. 

\begin{figure}
    \centering
    \includegraphics[width=0.24\linewidth]{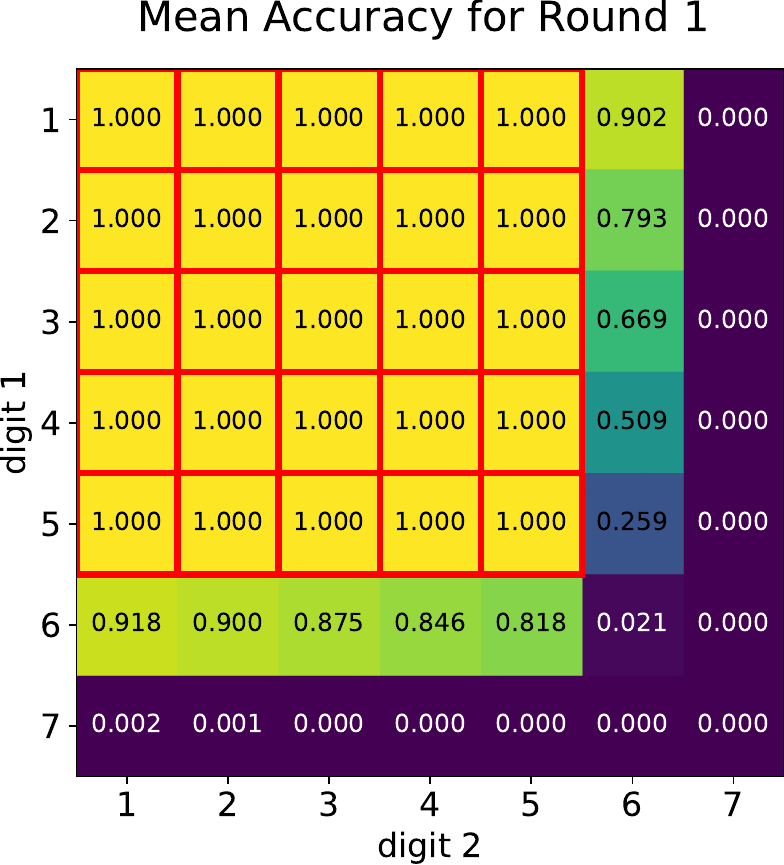}
    \includegraphics[width=0.24\linewidth]{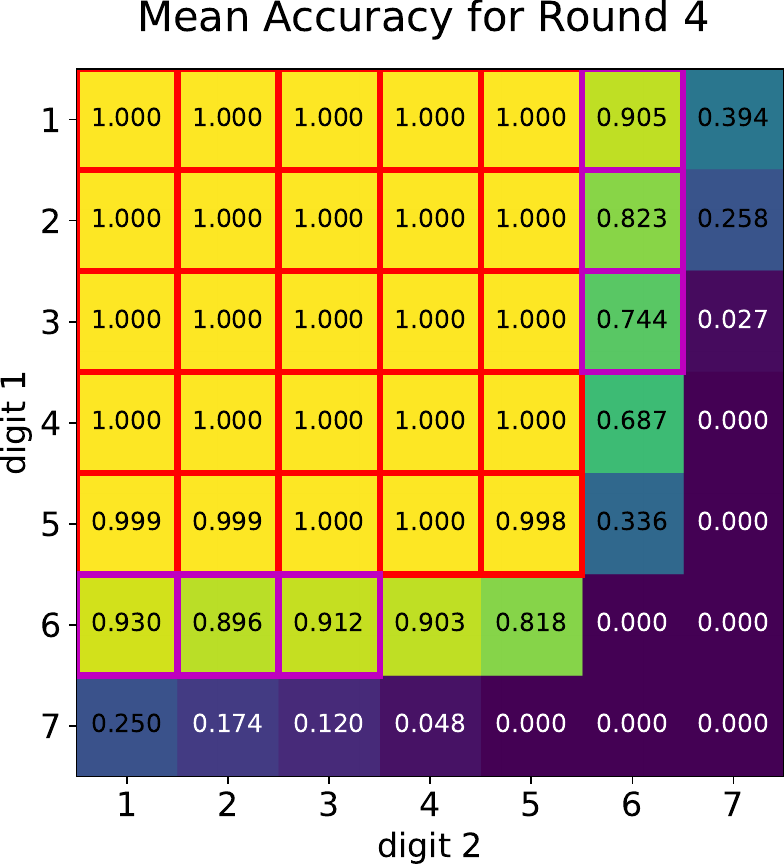}
    \includegraphics[width=0.24\linewidth]{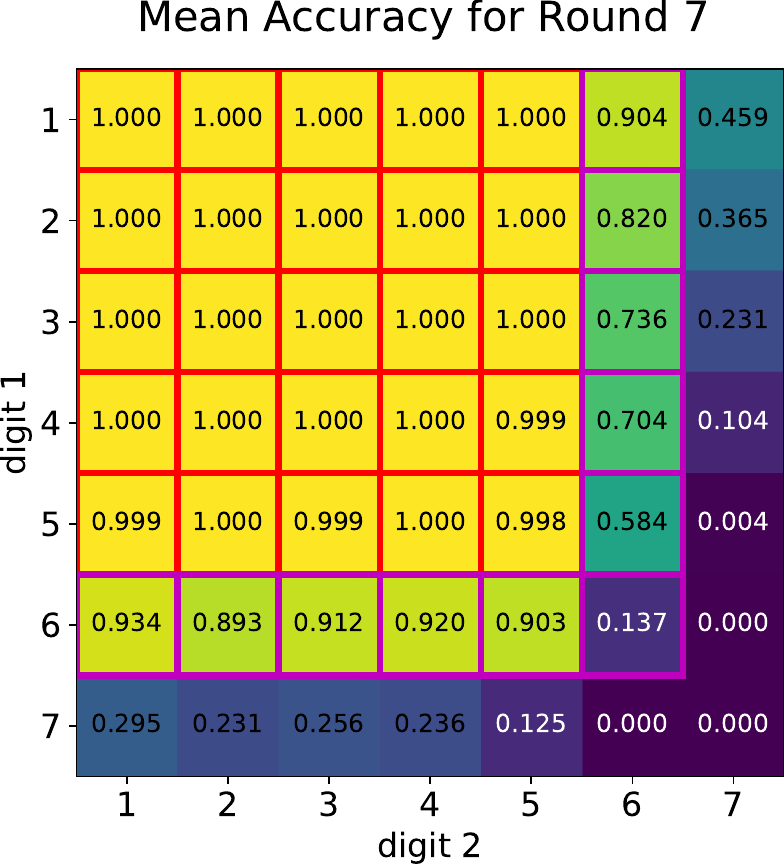}    
    \caption{Results on multiplication without filtering (round $1,4,7$). Each cell represents the accuracy on $n$-digit by $m$-digit multiplication. Red boxes indicate labeled in-distribution examples, while magenta boxes indicate evaluations after training on self-improve data. The model is initially trained on up to $5$-by-$5$ multiplication. Without filtering, generalizing to larger multiplications is challenging.}
    \label{fig:multiplication_vanilla}
    \vspace{-3mm}
\end{figure}

The data format is inherently asymmetrical. For example, an $m$-by-$n$ multiplication requires $n$ intermediate steps, where each step corresponds to multiplying the $m$-digit number by one digit of the $n$-digit number. Conversely, an $n$-by-$m$ multiplication involves $m$ intermediate steps of multiplying the $n$-digit number by each digit of the $m$-digit number.

\begin{wrapfigure}{r}{0.51\textwidth}
    \vspace{-8mm}
    \centering
    \includegraphics[width=0.49\linewidth]{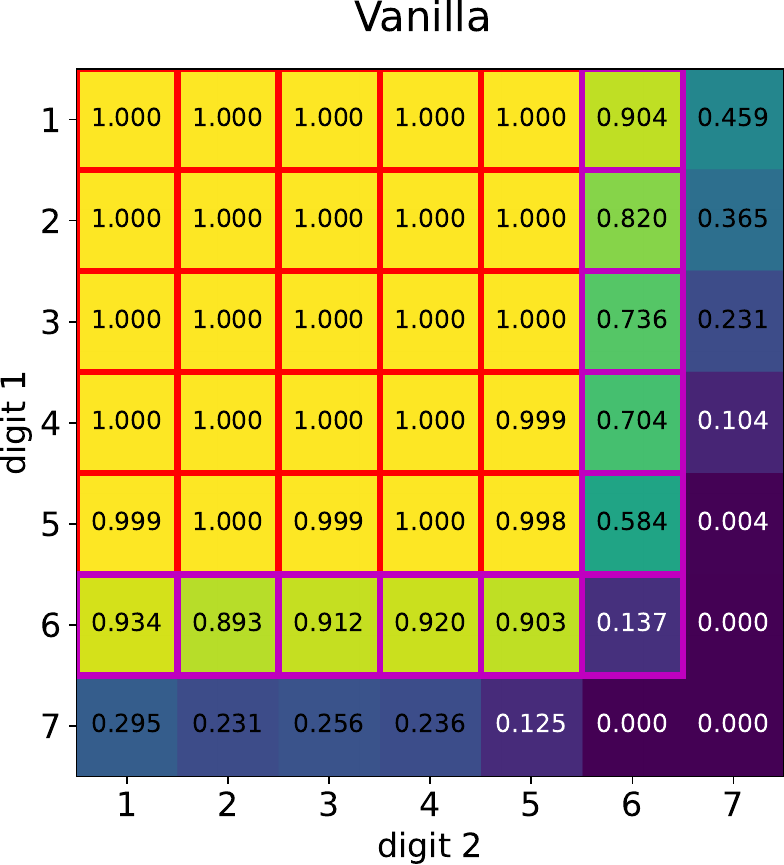}
    \includegraphics[width=0.49\linewidth]{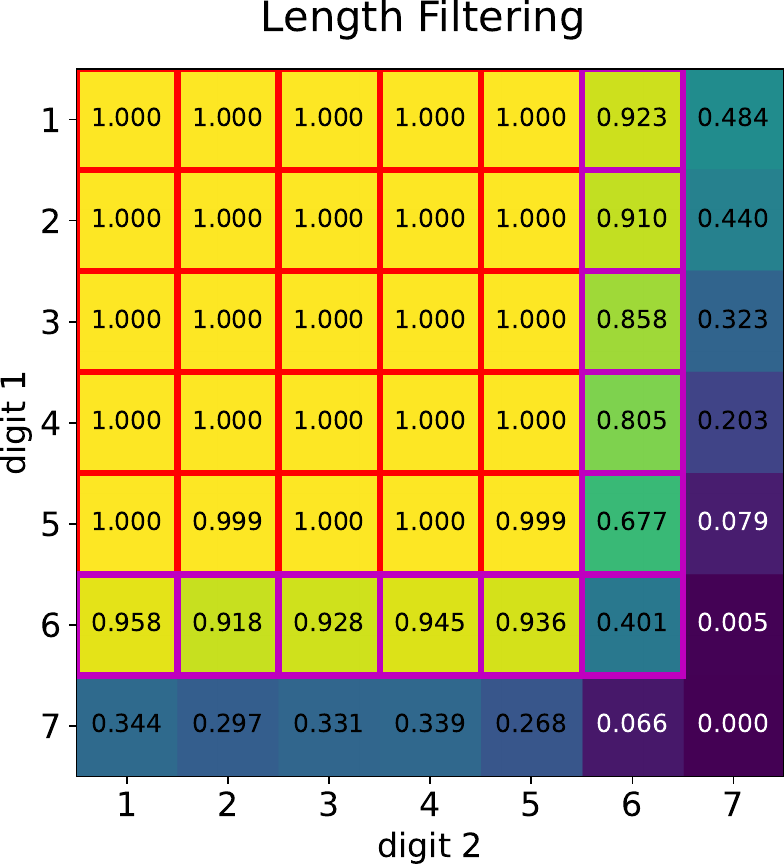}
    \includegraphics[width=0.49\linewidth]{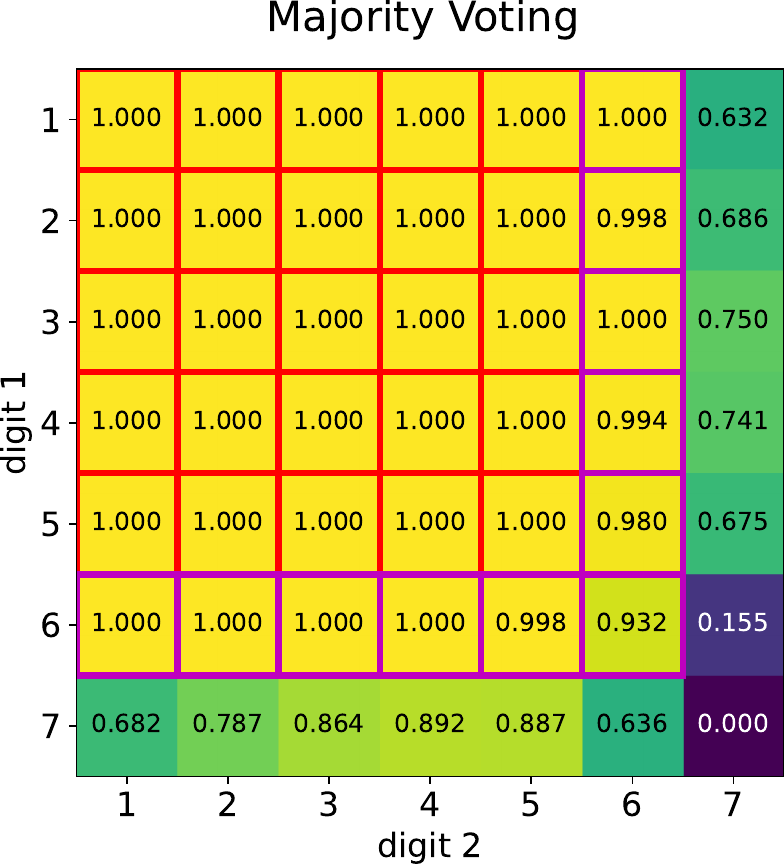}
    \includegraphics[width=0.49\linewidth]{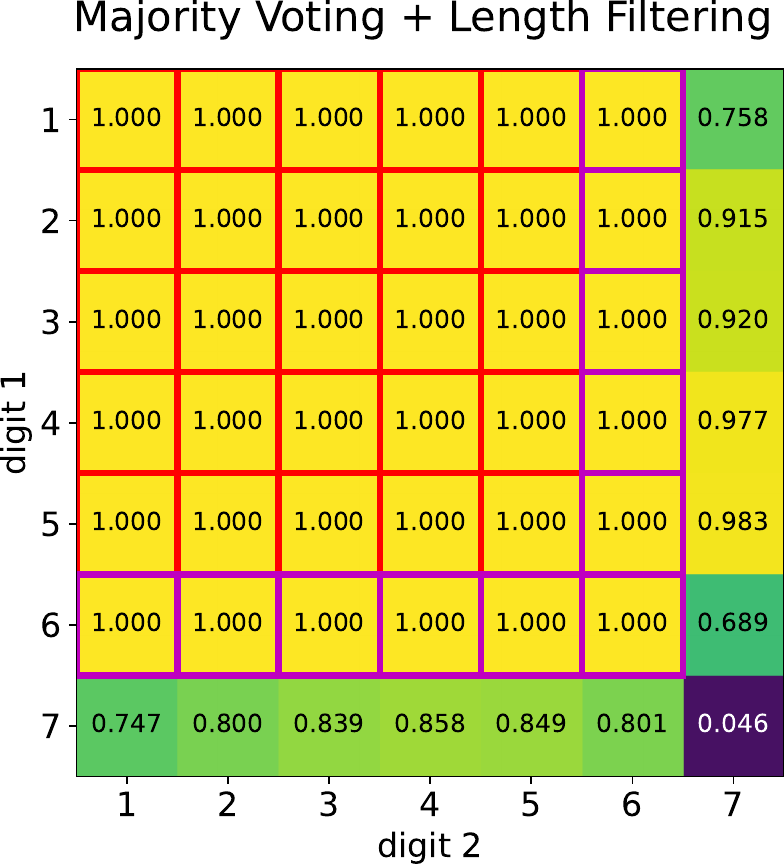}    
    \caption{Comparison of data filtering methods at round 7. (Top-Left) no filtering (Top-Right) length filtering (Bottom-Left) majority voting (Bottom-Right) and a combination of majority voting and length filtering. Data filtering significantly improves self-improvement performance, with the combined approach achieving the best results.}
    \vspace{-12mm}
    \label{fig:multiplication_filtering_comparison}
\end{wrapfigure}

\paragraph{Setting. }
The model is initially trained on $n$-by-$n$ multiplication examples with $n=5$. Directly introducing $n+1$-by-$n+1$ examples results in poor performance, hence, we adopt a more fine-grained difficulty schedule. In each self-improvement round, we incrementally increase one operand by one digit, sampling $n+1$-by-$m$ and $m$-by-$n+1$ examples, where $m$ grows from 1 to $n+1$. This gradual progression allows the model to adapt incrementally to larger operand sizes, making the transition to harder examples more manageable.

\paragraph{Results without Data Filtering. }
Despite being trained on CoT data with explicit intermediate steps, the model exhibits limited extrapolation performance for larger operand sizes. As shown in Figure~\ref{fig:multiplication_vanilla}, the model struggles to generalize beyond its training distribution without any refinement in the generated self-improvement data. After 7 self-improvement rounds, the accuracy for $6$-by-$6$ multiplication reaches only 13.7\%.
Additionally, performance varies significantly across models trained with different random seeds (Appendix Figure~\ref{fig:mult_has_high_variance1}), intermediate steps are often dropped. resulting in outputs shorter than the correct answer (Figure~\ref{fig:num_shorter_answers} (Right)). This highlights the need for data refinement strategies to improve both the quality and consistency of the self-generated data.

\paragraph{Results with Data Filtering. }
To improve the quality of self-generated training data, we apply three data refinement techniques: length filtering, majority voting, and a combination of both. For length filtering, we remove self-generated samples where the output length is shorter than the longest output in the batch by more than 10 tokens. This helps eliminate incorrect solutions that omit intermediate steps. For majority voting, we train five models in parallel using different random seeds and retain only those data points where at least four out of the five models produce the same output. This strategy ensures that only high-consensus, reliable data points are used for training.

Figure~\ref{fig:multiplication_filtering_comparison} compares the effectiveness of these filtering methods at round 7, where models are trained on self-generated data for up to $6$-by-$6$ multiplication. All three filtering methods enhance self-improvement, with majority voting outperforming length filtering. The combined approach—applying both majority voting and length filtering—achieves near-perfect generalization to $6$-by-$6$ multiplication.

\begin{figure}
    \centering
    \includegraphics[width=0.3\linewidth]{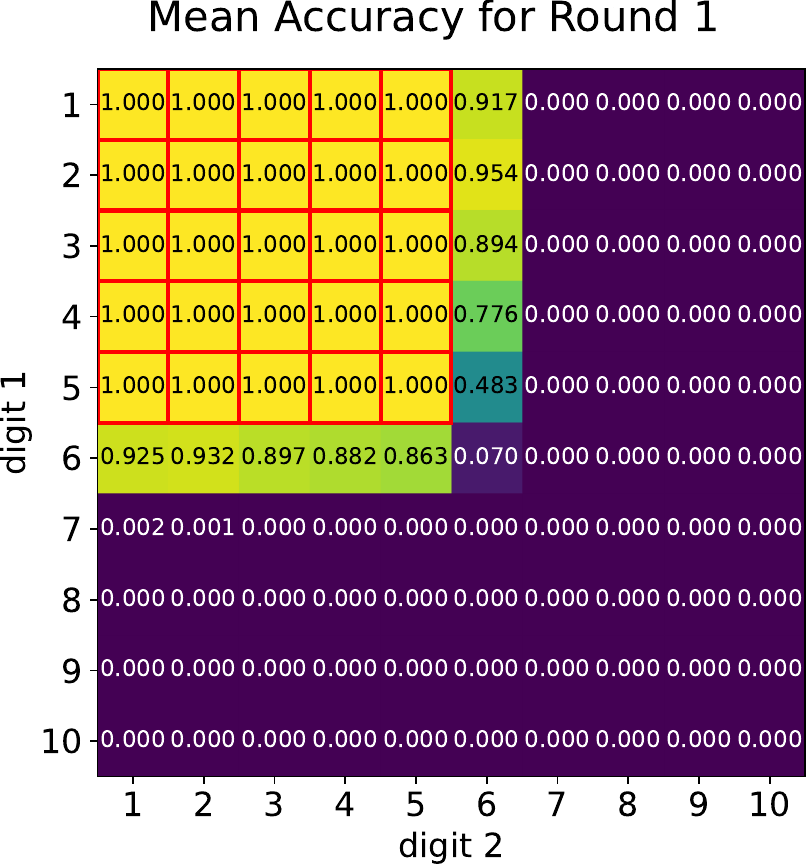}
    \includegraphics[width=0.3\linewidth]{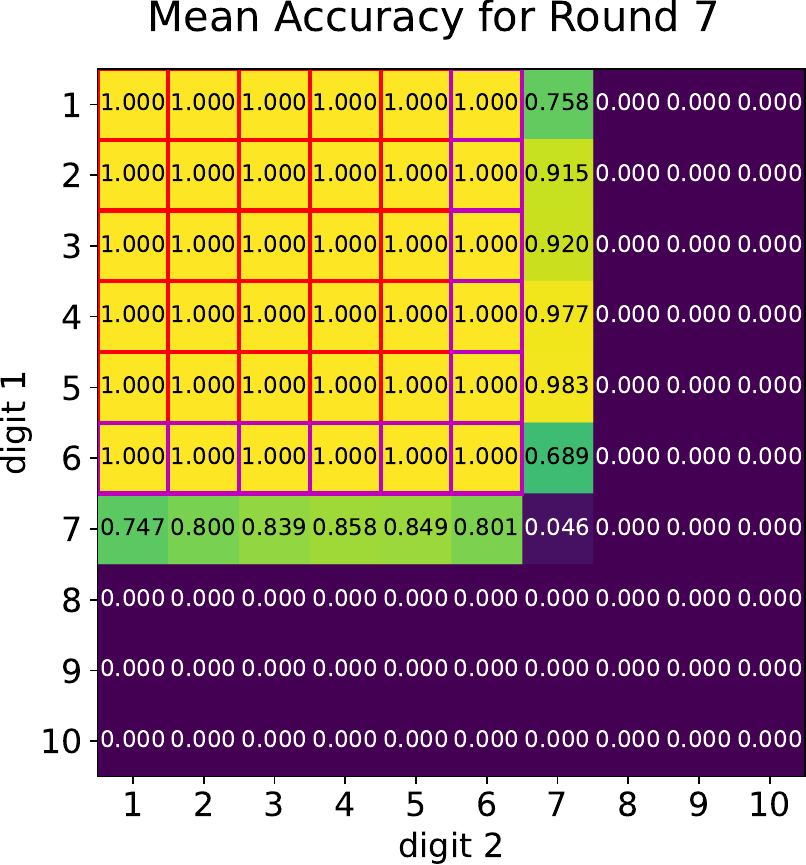}
    \includegraphics[width=0.3\linewidth]{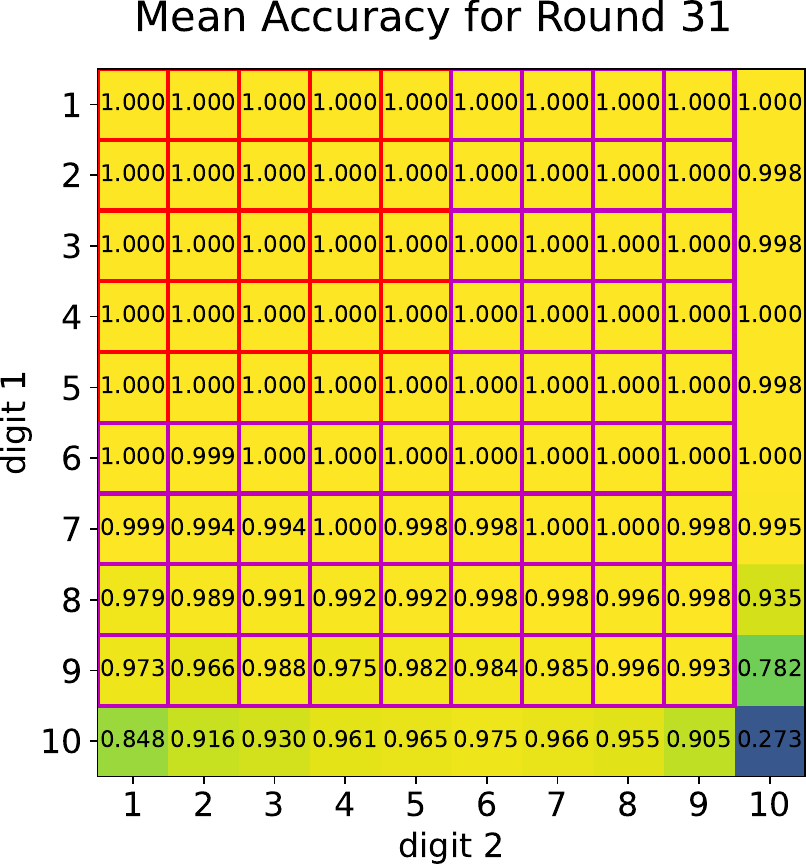}
    \caption{Results on combining majority voting with length filtering (at round $1,7,31$) for the multiplication task.
    This approach achieves near-perfect length generalization up to $9 \times 9$, demonstrating the effectiveness of filtering strategies in improving self-improvement.}
    \label{fig:multiplication_mv_len_n10}
    \vspace{-3mm}
\end{figure}

Training for additional rounds further extends this generalization. As shown in Figure~\ref{fig:multiplication_mv_len_n10}, the combined filtering strategy continues to yield near-perfect accuracy up to $9$-by-$9$ multiplication at round 31, with the potential for even further generalization in subsequent rounds. In Section~\ref{sec:accelerate}, we demonstrate that a carefully crafted self-improvement schedule can accelerate this process, achieving perfect performance on $10$-by-$10$ multiplication in just 19 rounds. Detailed results for all filtering methods are provided in Appendix Section~\ref{sec:mult_full_results}.

\paragraph{Relevance to Prior Work. }
These results have relevance with findings by~\citet{jelassi2023length}, who showed that dataset priming (adding a small number of labeled long-sequence examples) can enable length generalization\footnote{they consider encoder-decoder architecture which differs for our decoder-only model} for multiplication (although this is not strictly out-of-distribution). Our approach of incorporating accurate, self-generated out-of-distribution data via filtering can be seen as an automated form of dataset priming.
Furthermore, while our approach uses chain-of-thought (CoT) data for multiplication, we believe it is possible to length generalize on non-COT multiplication as well, by incorporating methods like~\citet{deng2024explicit} to help the model iteratively internalize the CoT steps.

\subsection{Maze-Solving}\label{sec:maze}

While arithmetic tasks and string manipulations provide valuable testbeds for studying language model generalization, we extend our evaluation to a more complex problem: finding the shortest path in a maze. Pathfinding presents significant challenges for autoregressive models~\citep{bachmann2024pitfalls}. Our mazes can be represented by a tree graph in a 2-dimensional space and they do not have loops. Figure~\ref{fig:maze_data} provides a visualization of this task and the corresponding input and output data format. 

Each tree graph consists of \(N\) nodes, which are randomly labeled with numbers between 0 and 99. The input format follows the structure: 
\texttt{start\_node>end\_node}, followed by a graph adjacency list formatted as \texttt{random\_node:adjacent\_nodes}. Here, \texttt{adjacent\_nodes} are separated by commas (\texttt{,}), and each \texttt{random\_node} is separated by a hyphen (\texttt{-}). The target output is a sequence of hops from the start node to the end node, separated by \texttt{>}. Additional details on maze generation are provided in Appendix~\ref{sec:data_gen-maze}. 

We evaluate two generalization settings: 1) increasing the number of hops while keeping the number of nodes fixed, and 2) increasing the number of nodes while keeping the number of hops fixed. In the first setting, the input graph description remains constant in size, but the output length grows as the difficulty increases. In the second setting, the input graph expands with more nodes, while the output remains of fixed length.

\subsubsection{Increasing the Number of Hops}
The difficulty of the maze-solving task increases with the number of hops required to traverse from the start node to the end node, which directly corresponds to the number of \texttt{>} symbols in the output. We begin by training the model on a labeled dataset containing paths of up to \( h=9 \) hops. In each self-improvement round, we increase \( h \) by one, progressively introducing longer paths.

\paragraph{Setting.}
The model is first trained on a dataset \( \mathcal{D}_0 \) containing 5 million labeled maze-solving examples, where the number of nodes is fixed at \( N=30 \) and paths range from \( h=1 \) to \( h=9 \) hops. This initial training phase spans 25,000 steps. In subsequent self-improvement rounds, we generate 50,000 additional training examples, increasing \( h \) by 1, and fine-tune the model for 5,000 steps per round. We experiment with both unfiltered training data and majority voting, where only outputs agreed upon by all 3 models are retained.

\begin{figure}
    \centering
    \includegraphics[width=0.32\linewidth]{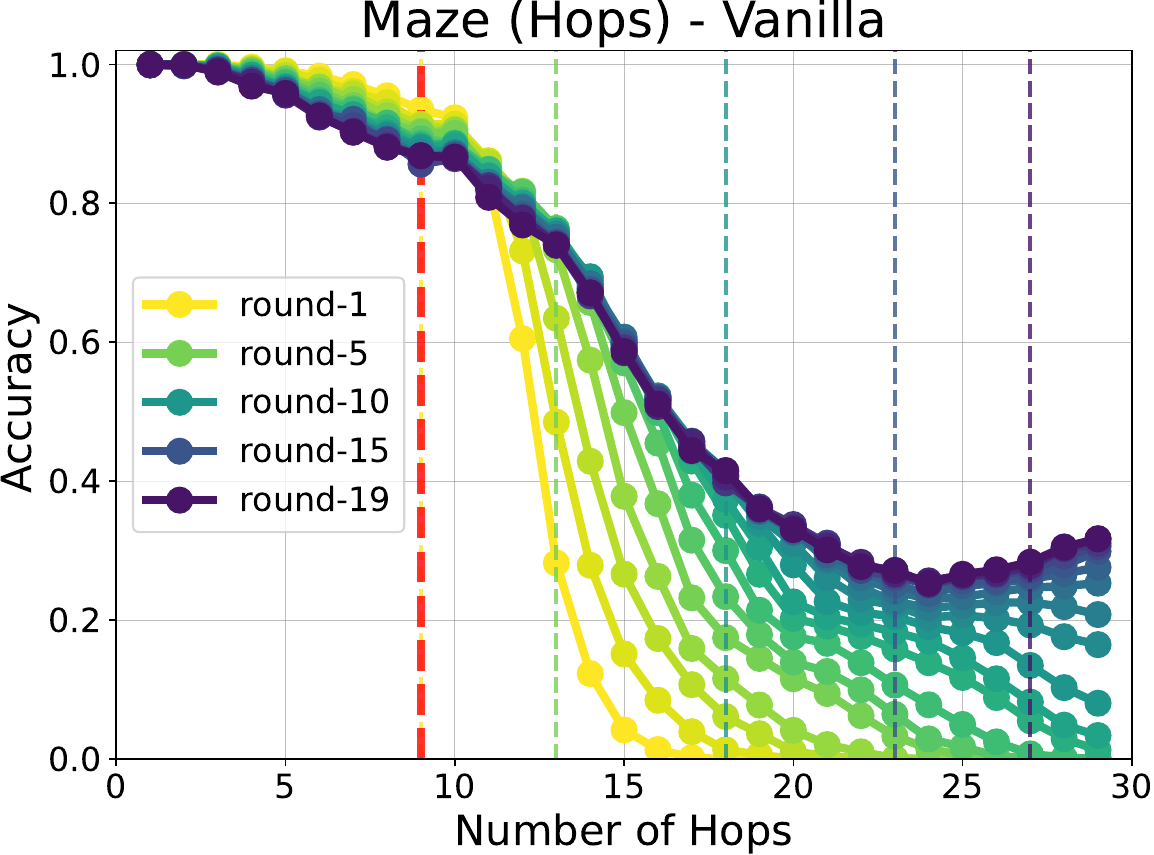}
    \hspace{1mm}
    \includegraphics[width=0.32\linewidth]{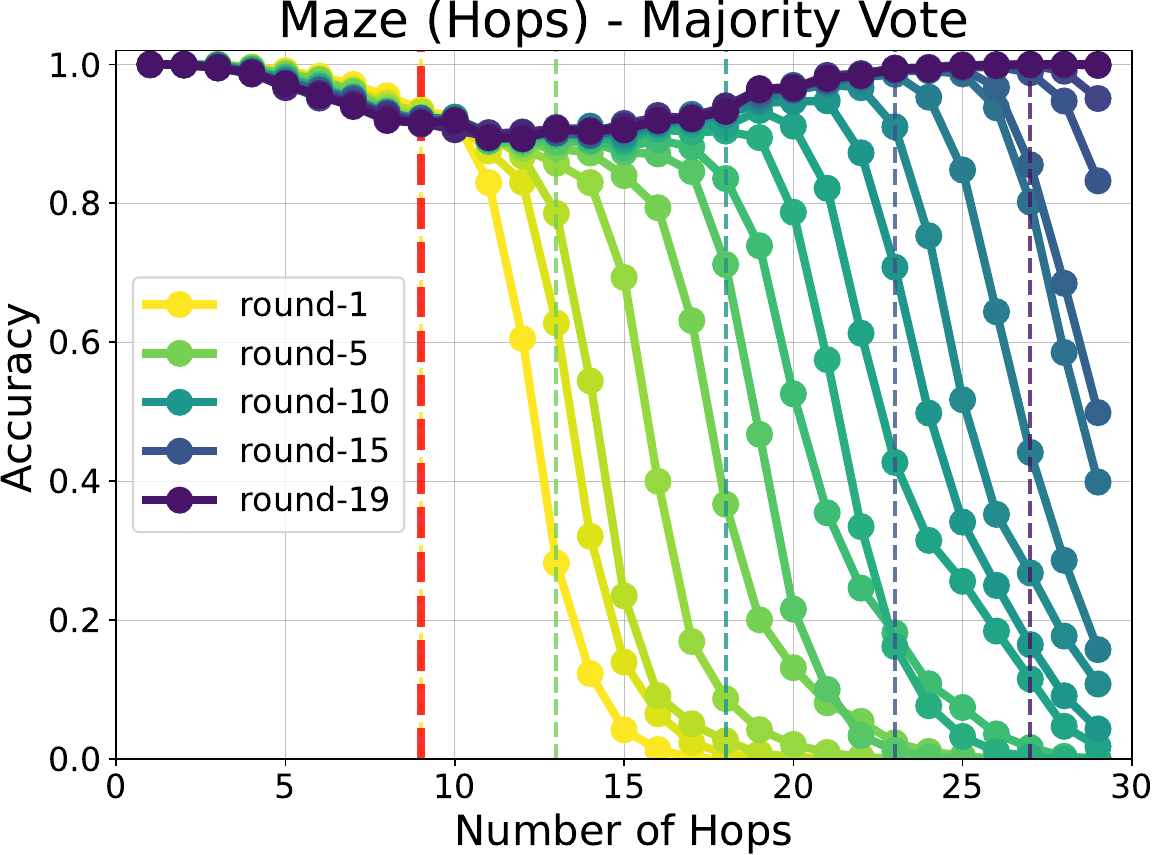}
    \hspace{1mm}
    \includegraphics[width=0.32\linewidth]{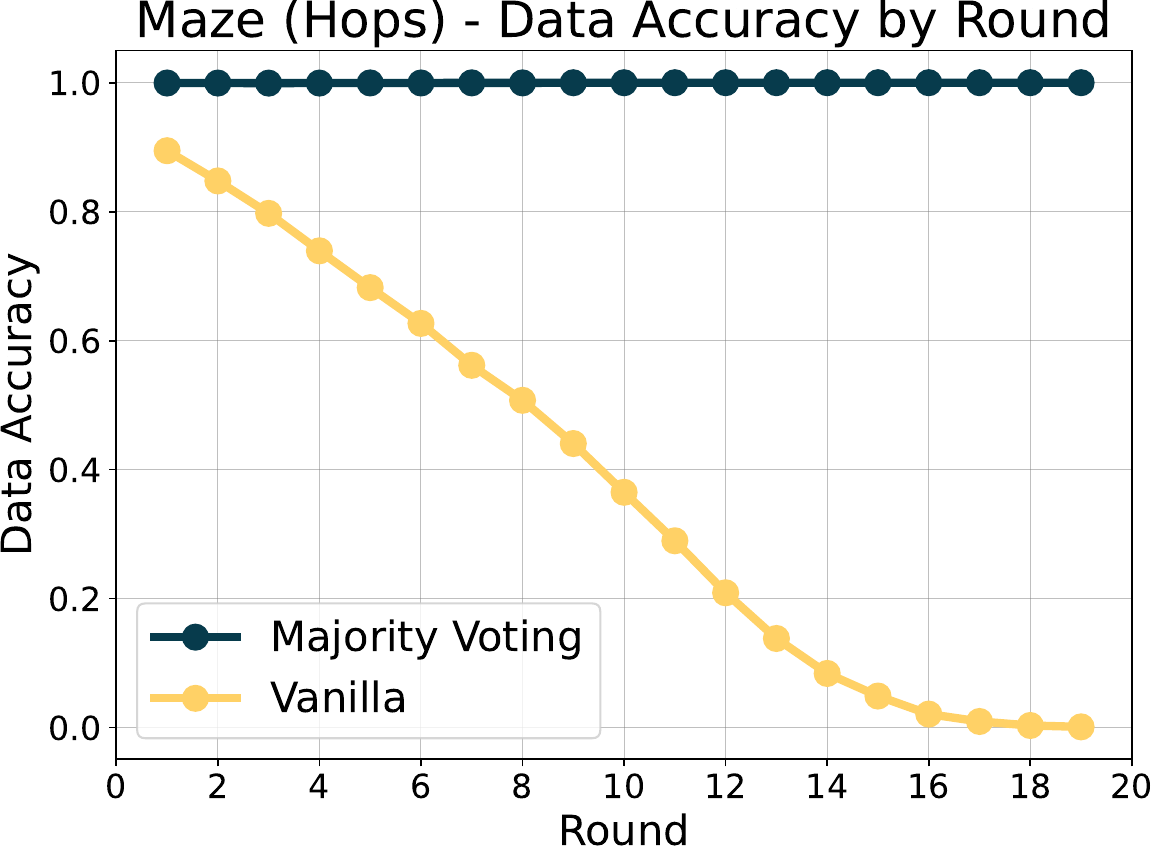}
    \caption{Maze-solving task with increasing hops (\( N=30 \) nodes). Models are trained on graphs with up to 9 hops and generalized by incrementally increasing hops by 1 in each self-improvement round. Results show mean accuracy across 3 seeds. (Left) No filtering. (Middle) Majority voting. (Right) Self-improve data accuracy per round. Filtering significantly enhances data accuracy and improves generalization.}
    \label{fig:maze_hops_len_gen_result}
\end{figure}

\paragraph{Results.}
As shown in Figure~\ref{fig:maze_hops_len_gen_result}, without filtering, self-generated training data degrades over successive rounds, leading to a collapse in the self-improvement process. In contrast, majority voting stabilizes data quality, allowing the model to continue generalizing successfully to paths up to 30 hops.

\subsubsection{Increasing the Number of Nodes}
Another approach to increasing task difficulty is to expand the number of nodes in the graph while keeping the number of hops fixed at \( h=9 \).

\paragraph{Setting.}
The model is first trained on a dataset \( \mathcal{D}_0 \) containing 5 million labeled maze-solving examples, with a fixed hop count \( h=9 \) and node counts ranging from \( N=10 \) to \( N=30 \). This initial training lasts 12,000 steps. In self-improvement rounds, the number of nodes \( N \) is increased by 3 per round, generating 50,000 additional training examples at each step and fine-tuning for 4,000 steps. We compare training without filtering against majority voting, where only outputs agreed upon by all 3 models are kept.

\begin{figure}
    \centering
    \includegraphics[width=0.32\linewidth]{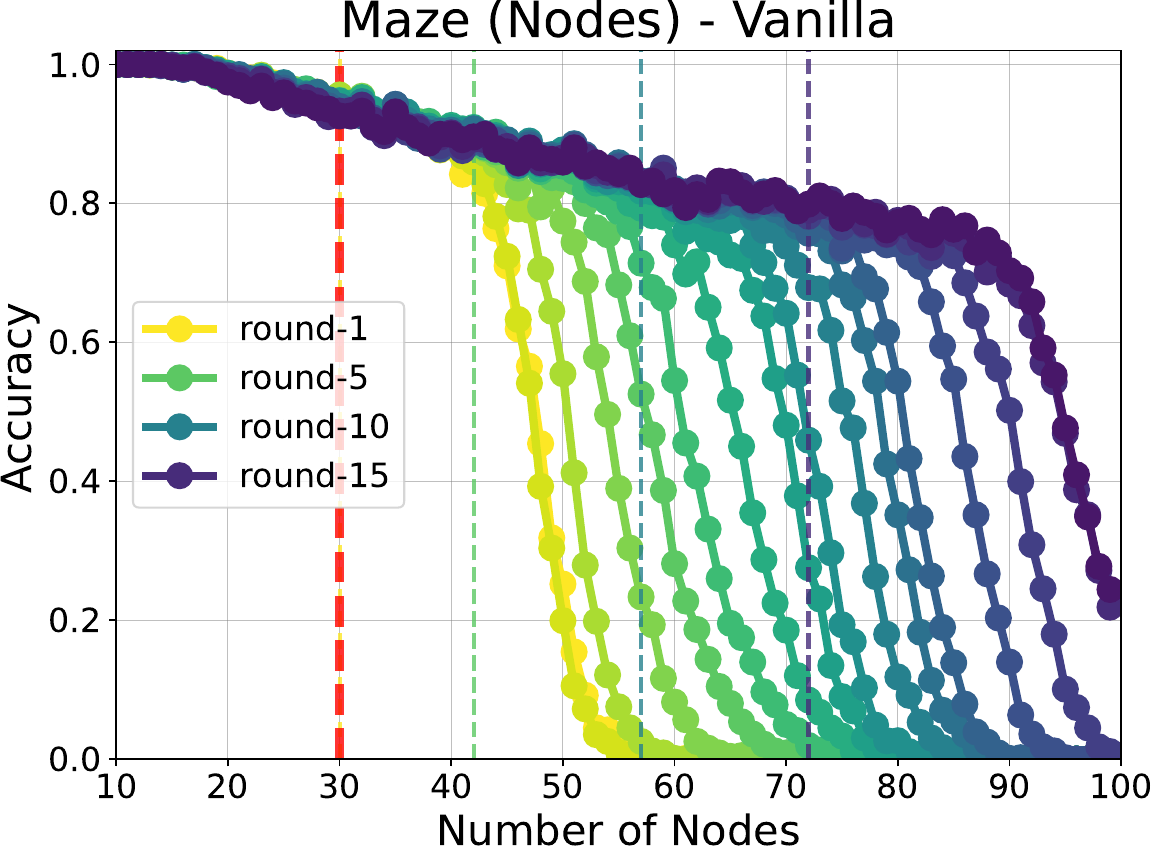}
    \hspace{1mm}
    \includegraphics[width=0.32\linewidth]{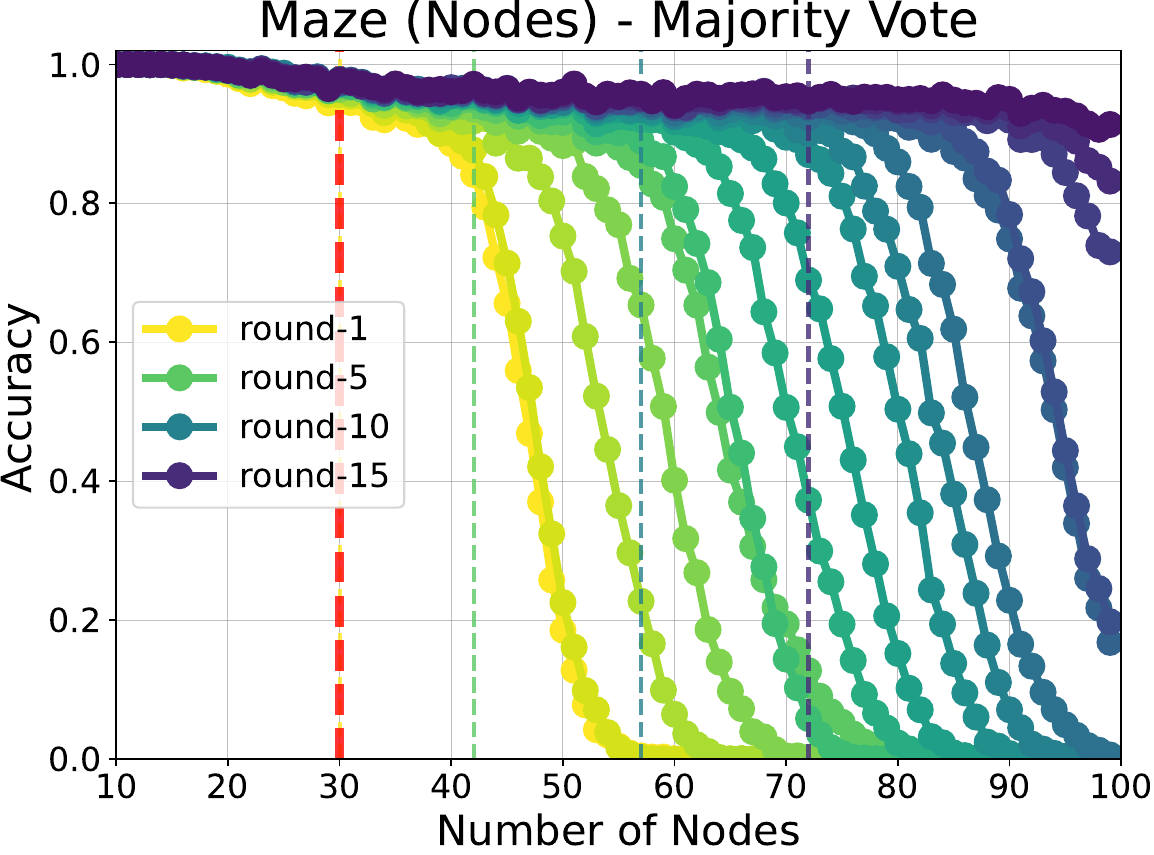}
    \hspace{1mm}
    \includegraphics[width=0.32\linewidth]{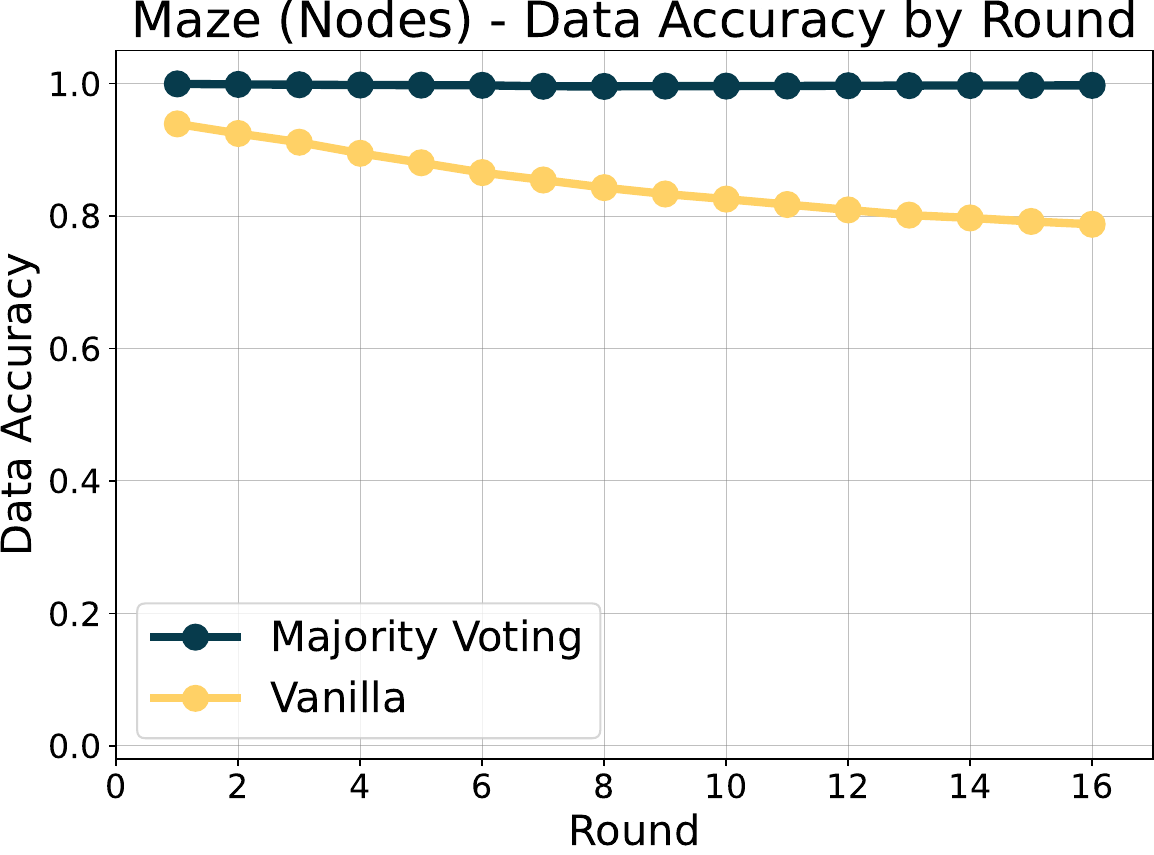}
    \caption{Maze-solving task with increasing nodes (fixed at 9 hops). Models are trained on graphs with up to 30 nodes and generalized by incrementally increasing the number of nodes by 3 per round. (Left) No filtering. (Middle) Majority voting. (Right) Self-improve data accuracy. Majority voting improves generalization to larger graphs.}
    \label{fig:maze_nodes_len_gen_result}
\end{figure}

\paragraph{Results.}
As shown in Figure~\ref{fig:maze_nodes_len_gen_result}, training without filtering leads to gradual performance degradation, whereas majority voting preserves high-quality data, maintaining a self-improvement accuracy above 99.7\% and enabling generalization to larger graphs with 9 hops.

While this experiment focuses on fixing one dimension (number of hops or number of nodes) and increasing the other, alternating between increasing the difficulty in both dimensions is expected to generalize the maze-solving task to handle larger graphs and longer paths simultaneously.

\subsubsection{Using Verifiers for Data Filtering. }
Solving the shortest path problem can be computationally expensive, but verifying the correctness of a given solution is significantly simpler. A valid path can be verified by traversing the sequence and ensuring three conditions: 1) each move is valid, meaning the path follows adjacency constraints; 2) the final destination matches the intended goal; and 3) no nodes are repeated, confirming that the solution is indeed the shortest path.

Self-improvement frameworks commonly incorporate verifiers to filter self-generated data, often leveraging trained models or reward models~\citep{zelikman2022star,singh2023beyond,hosseini2024v,lightman2023let}. While our primary focus is not on training or designing an additional verification mechanism, we investigate the effectiveness of using an external verifier as a data-filtering method.

To this end, we evaluate an oracle verifier that enforces two essential constraints: 1) move validity, ensuring that every transition in the generated solution adheres to the adjacency constraints of the maze, and 2) end validity, confirming that the final node in the solution corresponds to the correct destination. We compare the effectiveness of this oracle-based filtering against self-improvement without data filtering and majority-voting-based filtering to assess its impact on performance and stability.

\begin{figure}
    \centering
    \includegraphics[width=0.32\linewidth]{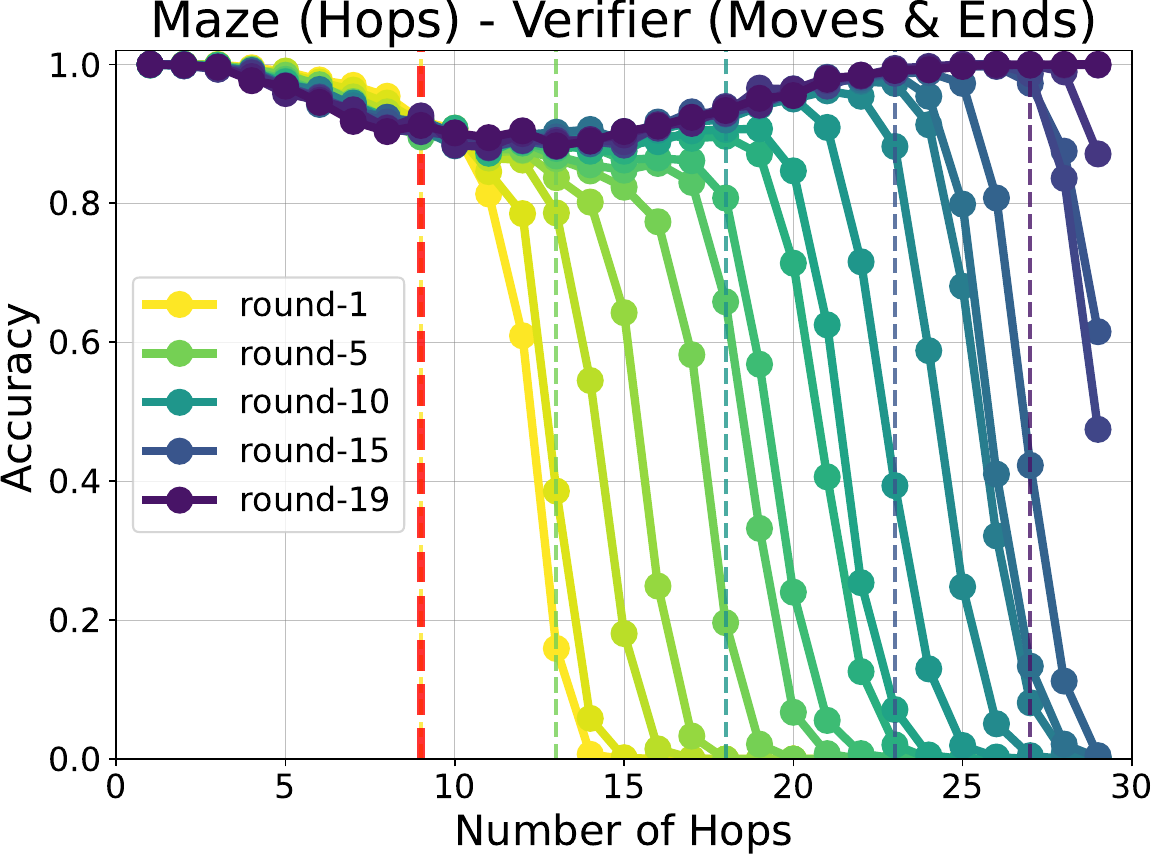}
    \hspace{1mm}
    \includegraphics[width=0.32\linewidth]{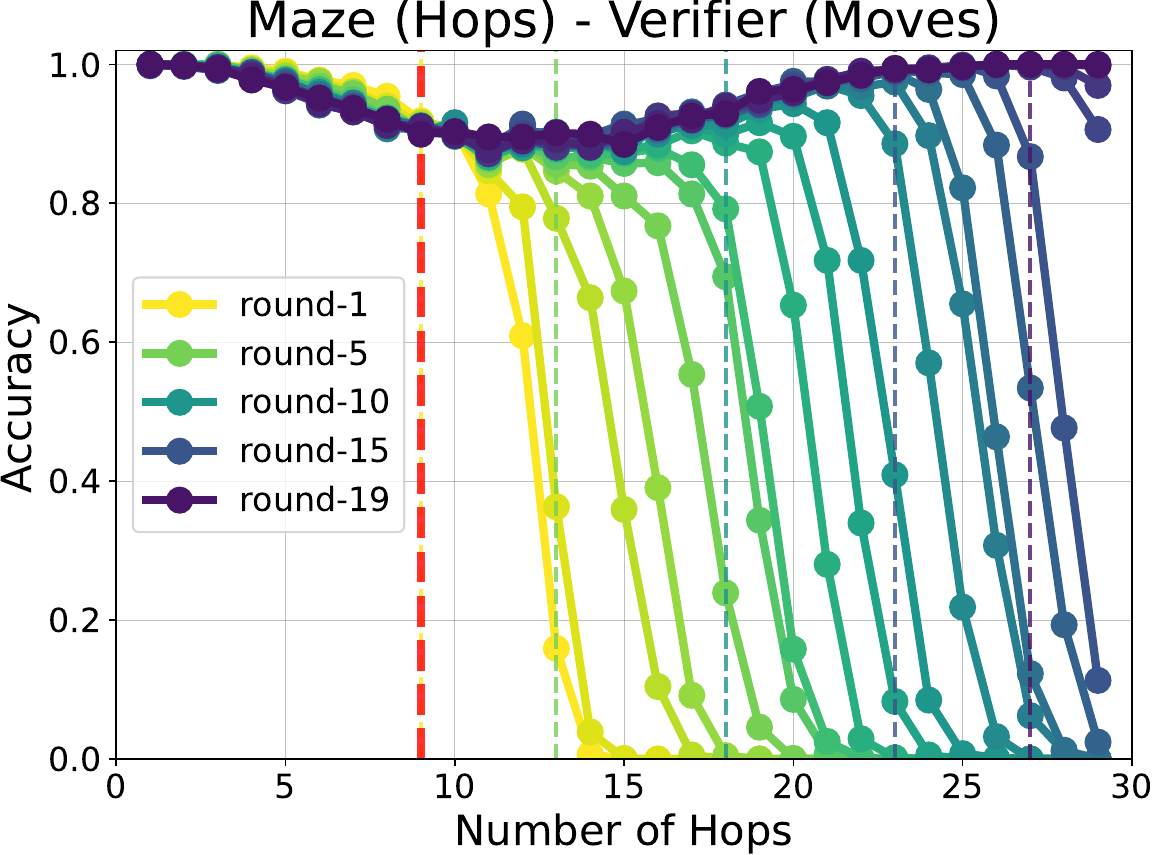}
    \hspace{1mm}
    \includegraphics[width=0.32\linewidth]{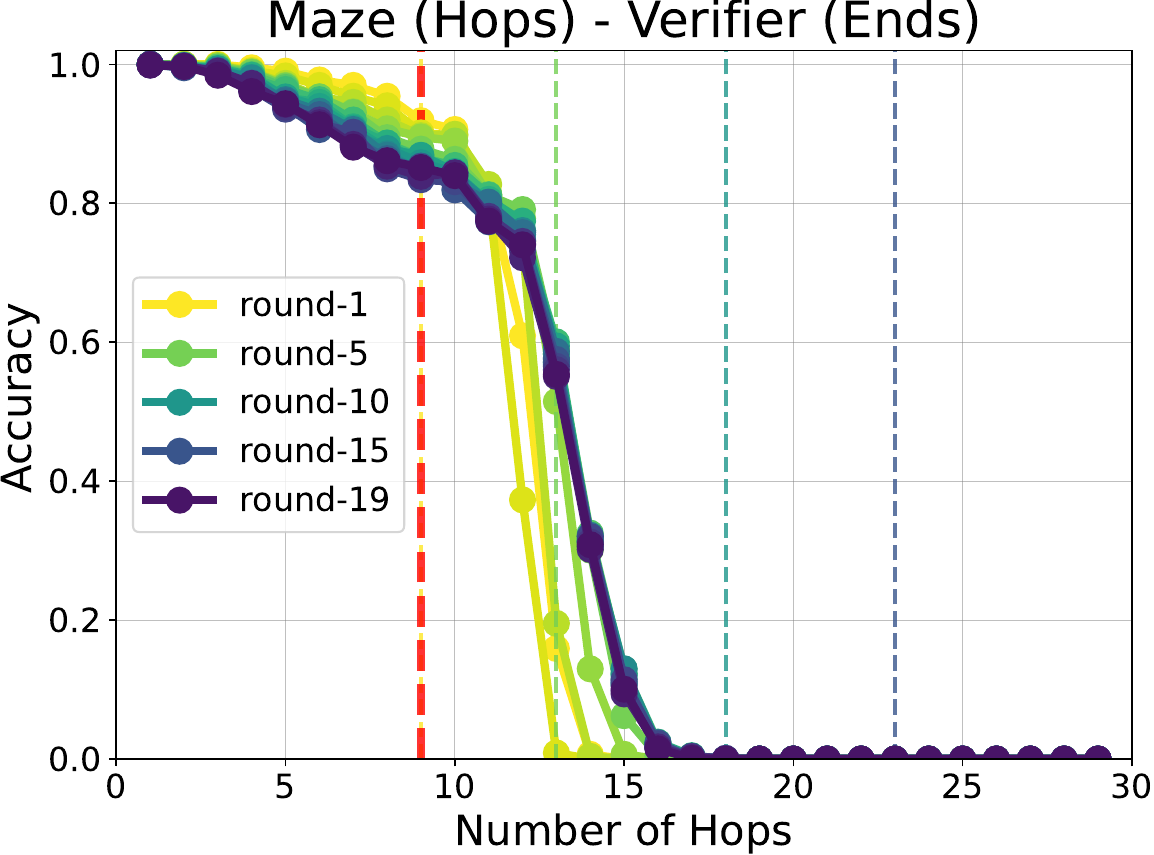}    
    \\
    \includegraphics[width=0.32\linewidth]{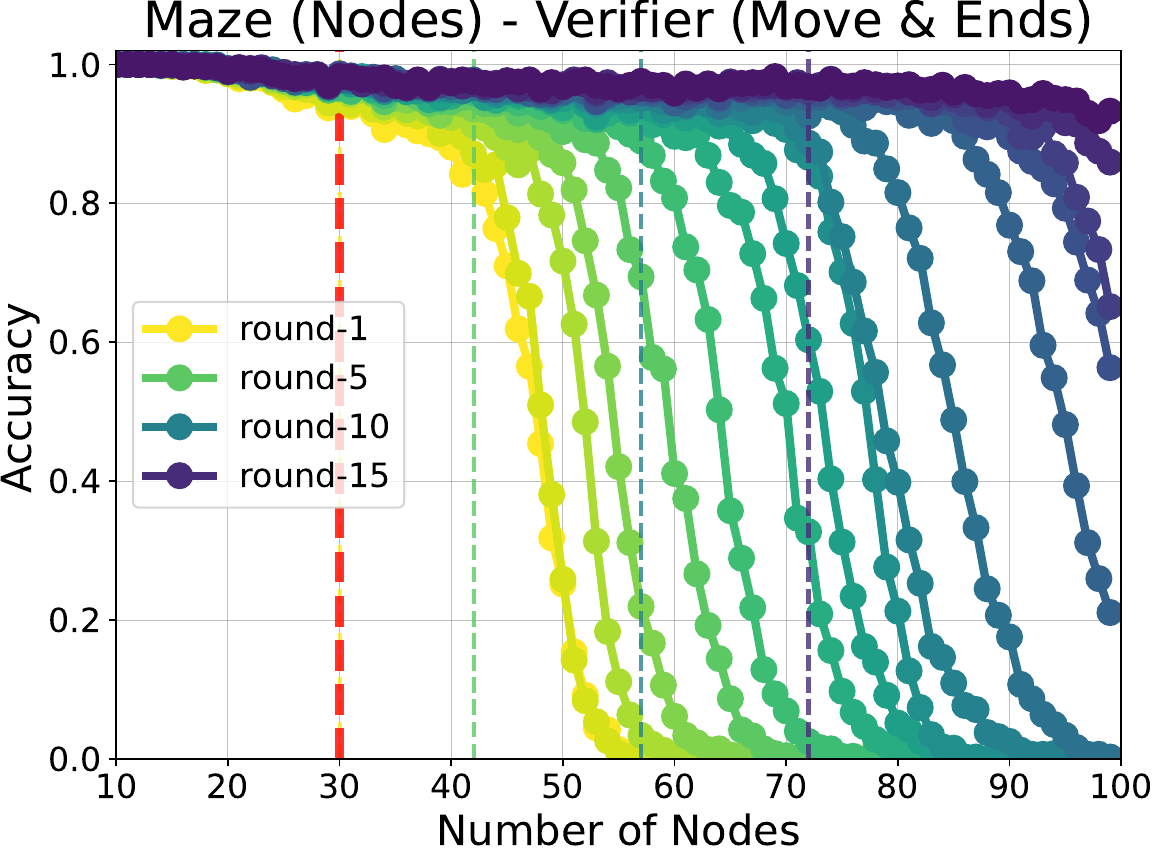}
    \hspace{1mm}
    \includegraphics[width=0.32\linewidth]{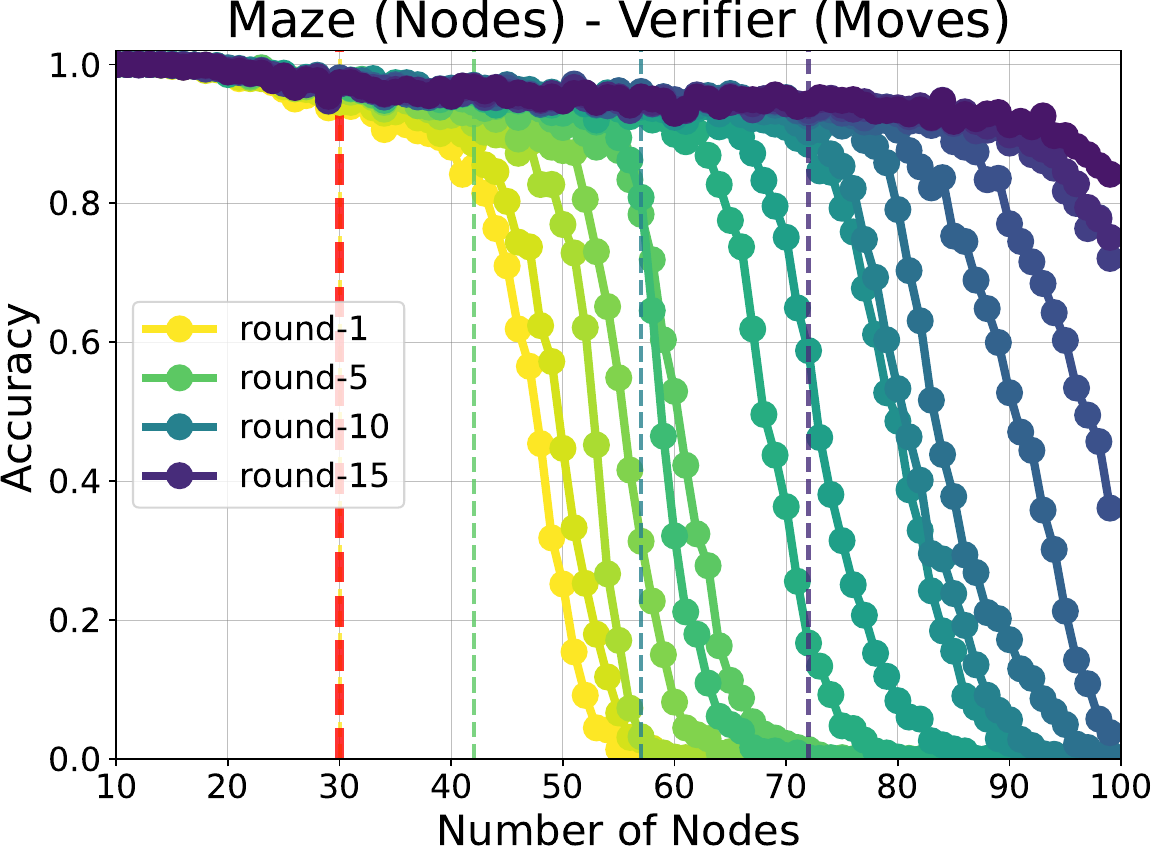}
    \hspace{1mm}
    \includegraphics[width=0.32\linewidth]{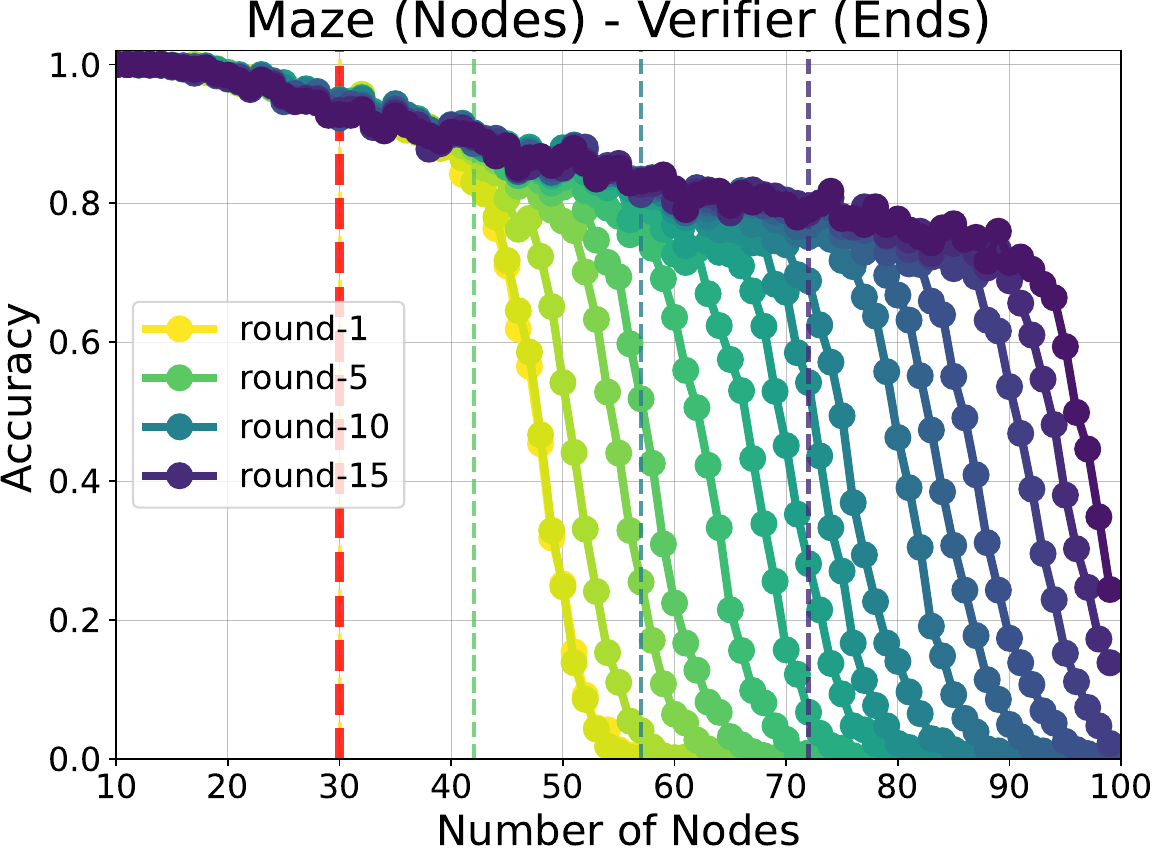}
    \caption{Results on using verifier for data filtering. (Top) Increasing hops. (Bottom) Increasing nodes. (Left) Verifier on both moves and ends. (Middle) Verifier on moves only. (Right) Verifier on ends only. Verifier-based filtering improves self-improvement performance, with move validation proving more effective than end validation alone. Interestingly, majority voting performs on par with oracle verification, suggesting that self-consistency mechanisms can serve as effective alternatives to explicit verification.}
    \label{fig:maze_verifier}
\end{figure}

\paragraph{Results. }
Figure~\ref{fig:maze_verifier} shows results for mazes with increasing hops, increasing nodes, and three different verification strategies: checking moves, checking end validity, and checking both. As expected, verification improves data quality and serves as an effective filtering technique in self-improvement. Notably, verifying move validity proves to be significantly more effective than verifying only the correctness of the end node. Interestingly, however, majority voting—a strategy that does not rely on an external verifier—performs comparably to verification-based filtering. This suggests that self-consistency mechanisms alone can be sufficient for maintaining high-quality training data. 

Additional results, including finer-grained analysis of move validity and end validity beyond exact match accuracy, are provided in Appendix~\ref{sec:maze_full_results}.

\section{Ablations}
\subsection{OOD generalization Increases with More Self-Improvement}\label{sec:ood_increases}

\begin{finding}
    The amount of out-of-distribution (OOD) extrapolation increases with more rounds of self-improvement in the addition, copying, and multiplication tasks.
\end{finding}

\begin{figure}
    \centering
    \includegraphics[width=0.7\linewidth]{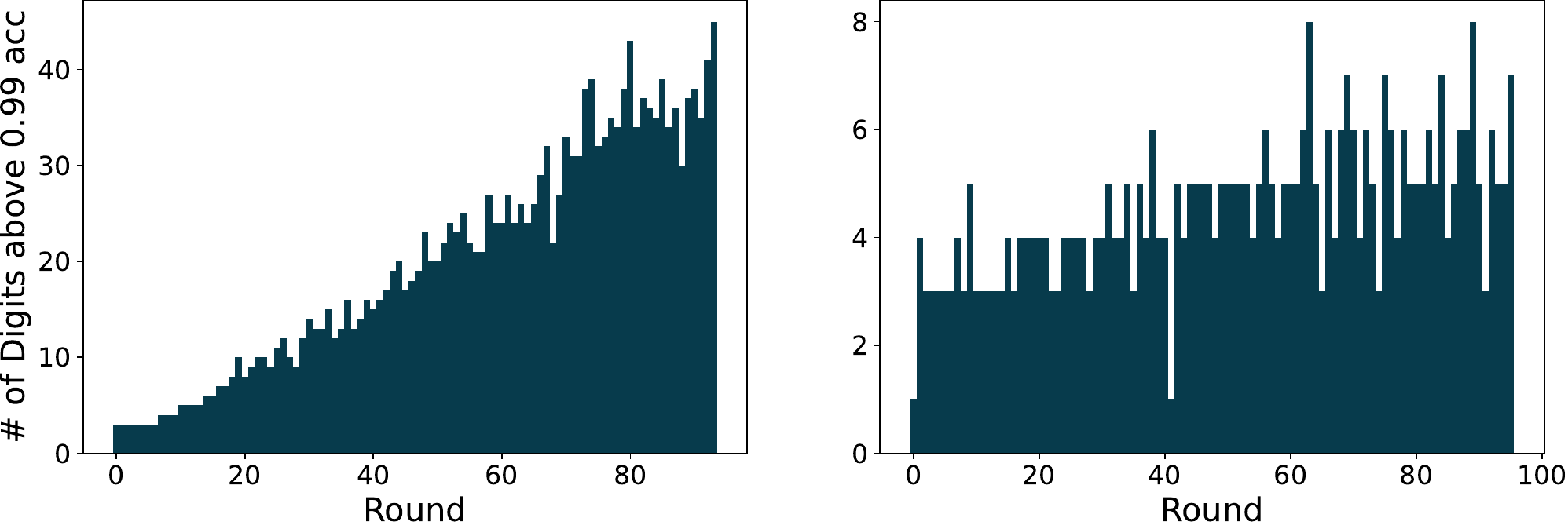}
    \caption{Number of \textbf{extra} OOD digit lengths achieving over 99\% accuracy when self-improving with one additional digit per round, on (Left) copy and (Right) reverse addition. The growing OOD capability suggests the potential to sample more digits per round as self-improvement progresses.}
    \label{fig:acc_ood_digits}
\end{figure}

Out-of-distribution (OOD) generalization is a critical measure of a model's ability to extrapolate beyond its training data. In tasks like reverse addition and copy, we observe that OOD extrapolation capabilities improve progressively as the model undergoes more rounds of self-improvement. Figure~\ref{fig:acc_ood_digits} illustrates how the number of additional OOD lengths achieving over 99\% accuracy grow with each round when the model is self-improved using only one additional digit per round. The model's OOD extrapolation capabilities expand as it is trained on longer sequences. 

\paragraph{``Safe Range'' of Sampling of Next Round Difficulty. } 
At the core of the self-improvement framework is the observation that the models perform well on samples slightly harder than those in the training set. Sampling instances that are too difficult for the current model is detrimental to the quality of self-improvement data, which causes downstream performance to break down. Here, we have shown that the ``safe range'' for sampling next-round difficulty becomes more forgiving.  Nevertheless, in real-world datasets, difficulty measures are not strictly quantifiable, and sampling by difficulty often involves noise. It is important to develop precise and controllable notions of difficulty for real-world tasks.

\subsection{Self-Improvement can be Accelerated}\label{sec:accelerate}
We observe that for addition and string manipulation tasks, the amount of extra OOD generalization increases roughly lineraly with each additional round of self-improvement (Figure~\ref{fig:acc_ood_digits}). This observation suggests that sampling multiple difficulty levels within the safe range per round could lead to exponential improvements in performance.

It is important to note that the sampling schedule in this proof-of-concept experiment leverages information about test set accuracy to determine the extra OOD lengths to sample, which is not typically available in practical scenarios. While this approach demonstrates the potential of an accelerated self-improvement schedule, practical implementation would require using proxy metrics to estimate suitable OOD lengths for sampling.

\begin{figure}
    \centering
    \includegraphics[width=0.45\linewidth]{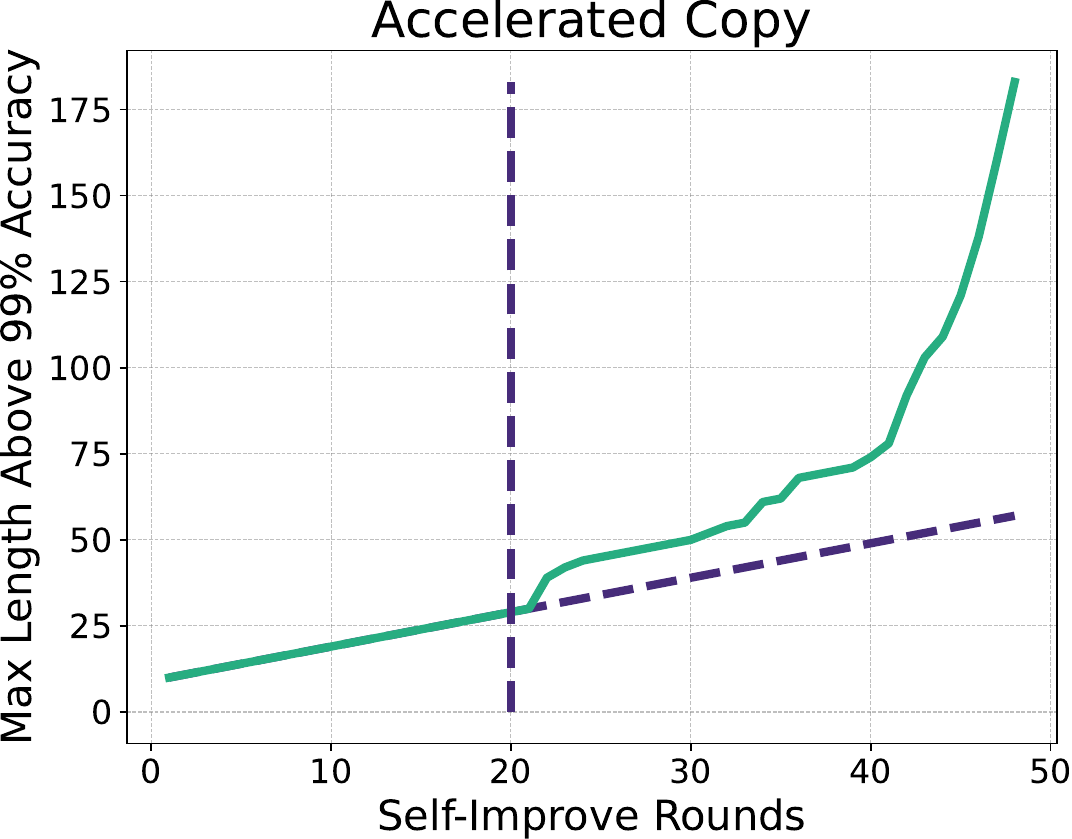}
    \includegraphics[width=0.431\linewidth]{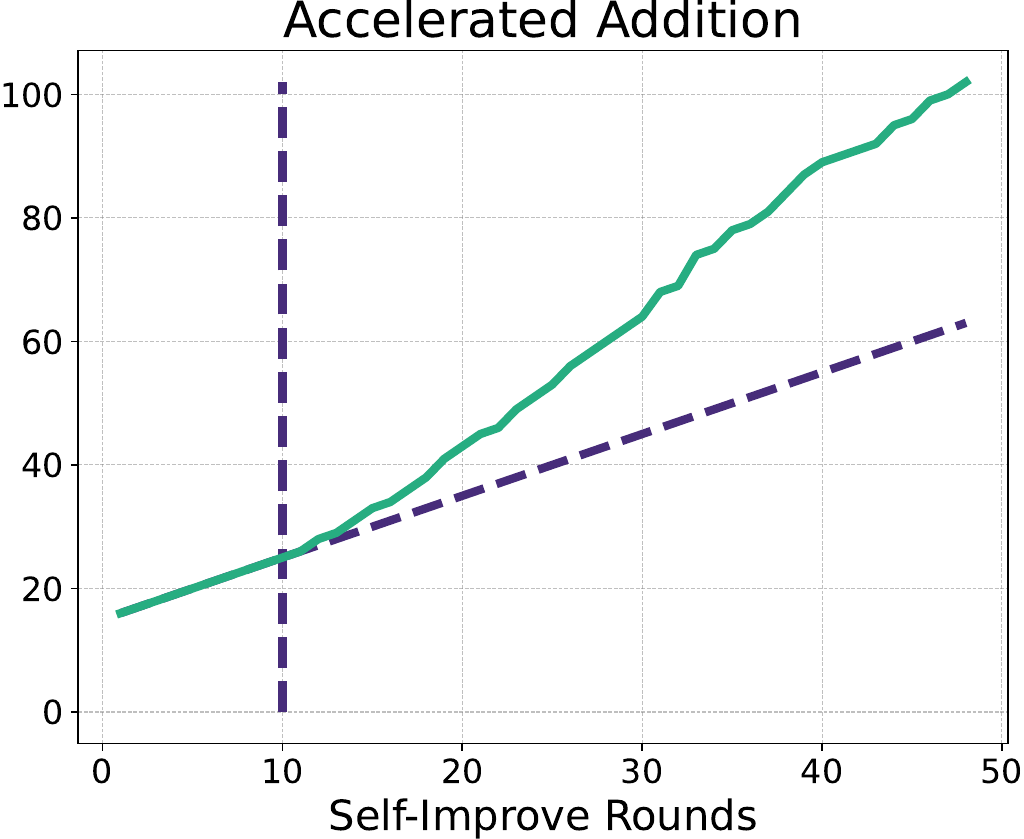}
    \caption{Maximum input length achieving over 99\% accuracy at different self-improvement rounds for (Left) Copy task and (Right) Reverse addition. The dashed linear line represents the standard schedule of sampling one additional length per round. The vertical line is when we start allowing accelerated schedule. Faster self-improvement schedules allow the model to generalize to longer inputs with fewer rounds.}
    \label{fig:fast_copy_add}
\end{figure}

\paragraph{Accelerated Self-Improvement Schedule.}
Building on the observations in Figure~\ref{fig:acc_ood_digits}, we propose an accelerated self-improvement schedule where the model samples multiple difficulty levels at each round, instead of incrementally increasing by only one additional length (for string manipulation) or digit (for reverse addition). As shown in Figure~\ref{fig:fast_copy_add}, this approach significantly speeds up performance in copying and reverse addition tasks, allowing the model to achieve high accuracy with fewer training rounds and training steps. At each round, the self-improvement dataset is uniformly sampled from all difficulty levels achieving over 99\% evaluation accuracy. All other hyperparameters remain unchanged.

\begin{wrapfigure}{r}{0.61\textwidth}
    \vspace{-9mm}
    \centering
    \includegraphics[width=0.25\textwidth]{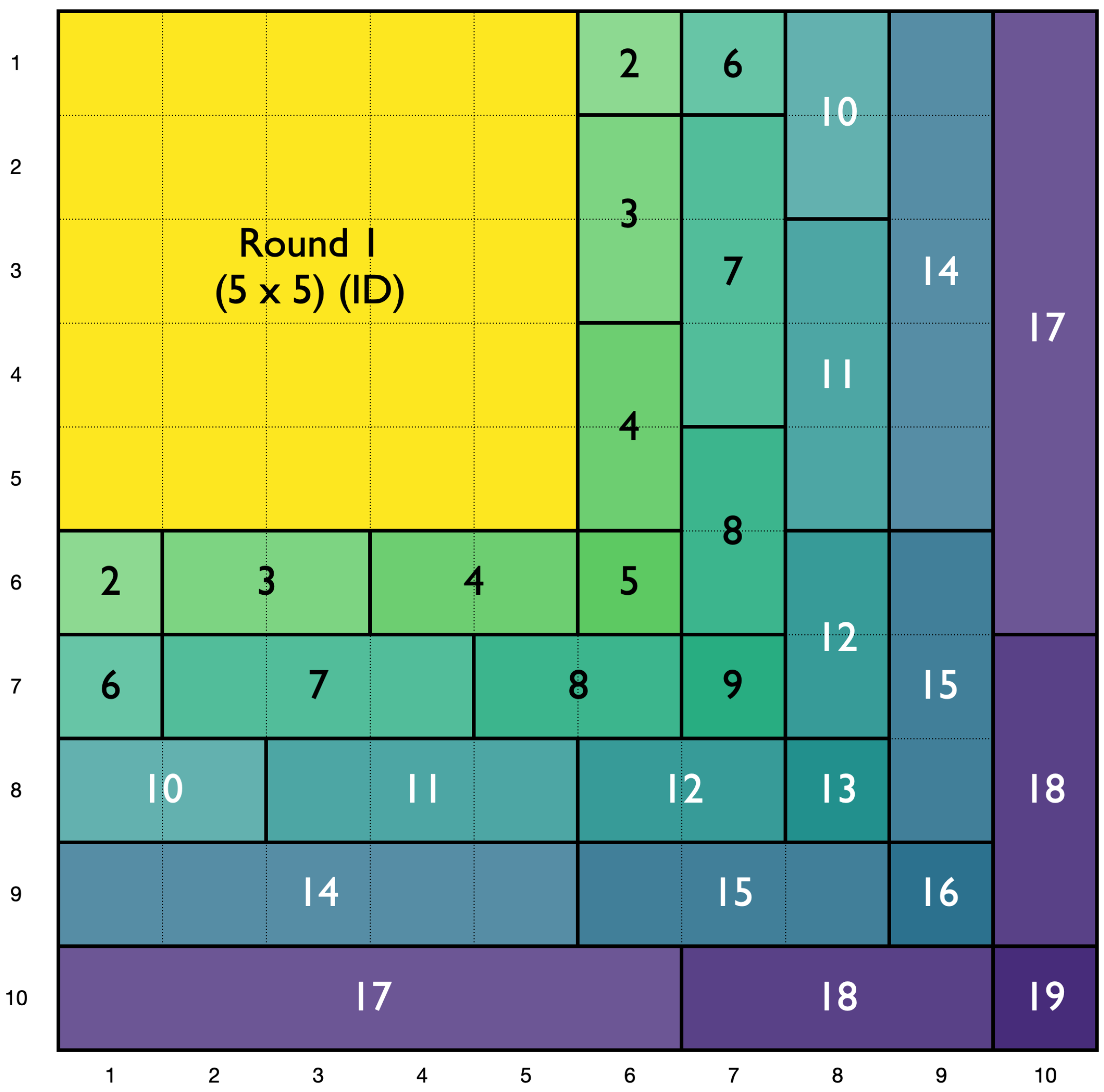}
    \includegraphics[width=0.25\textwidth]{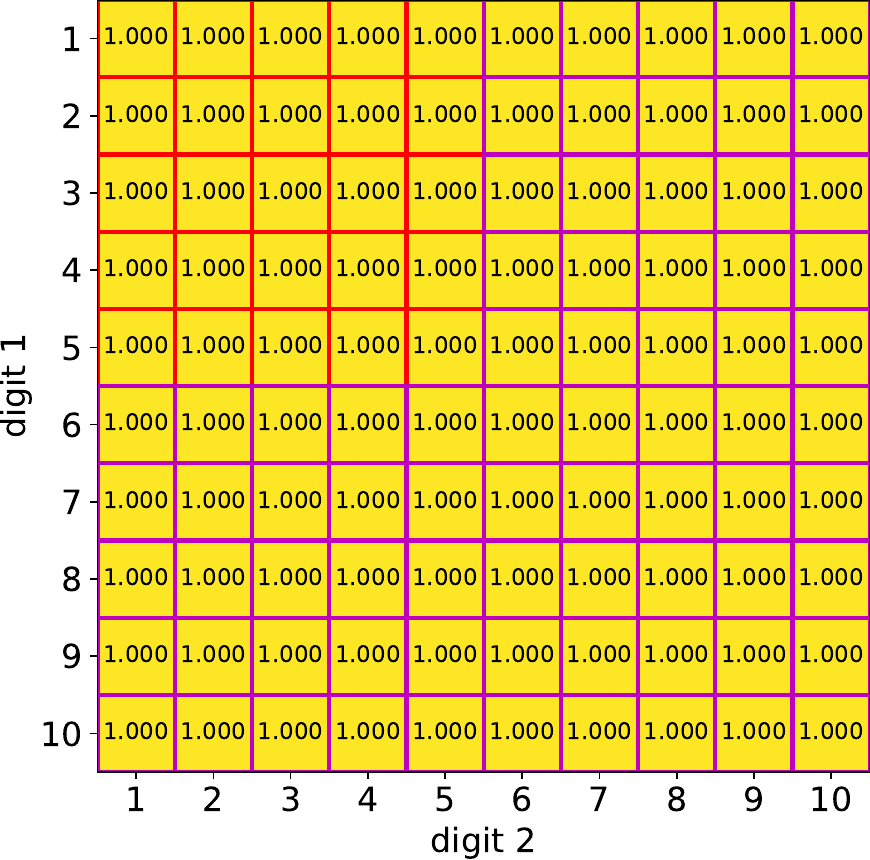}
    \caption{Accelerated self-improvement in multiplication. (Left) Accelerated schedule for multiplication. The rows and columns represent the number of digits in the two operands of the multiplication task. The number within each cell indicates the self-improvement round in which the corresponding digit pair is included for training. (Right) Results at round 19. Controlled scheduling progressively incorporates more digit pairs in each round, accelerating the self-improvement process.}
    \label{fig:mult_accelerated}
    \vspace{-8mm}
\end{wrapfigure}

We observe similar improvements using an accelerated schedule for multiplication task as well, as depicted in Figure~\ref{fig:mult_accelerated}. Under the standard schedule, reaching 10-by-10 multiplication from 5-by-5 requires 41 self-improvement rounds, incrementally increasing one operand by 1 at a time. With the accelerated schedule, we progressively sample more operand pairs as self-improvement proceeds, reducing the required rounds to 19 while achieving perfect test performance (see Appendix Figure~\ref{fig:multiplication_accelerated_len_n10_full} for full results). The settings for multiplication follow the setting in Section~\ref{sec:harder_tasks}.

\subsection{Pretrained Models achieve Faster Acceleration}\label{sec:pretrained}
\begin{finding}
    Self-improvement is more effective for larger pretrained models, and these models achieve faster acceleration.
\end{finding}

We extend our self-improvement framework to pretrained models, specifically Llama-1B and Llama-3B~\citep{llama3modelcard}, to explore scaling effects and the impact of finetuning on larger models. Our results indicate that larger models exhibit superior extrapolation performance and benefit from accelerated self-improvement.

\begin{figure}
    \centering
    \includegraphics[width=0.49\linewidth]{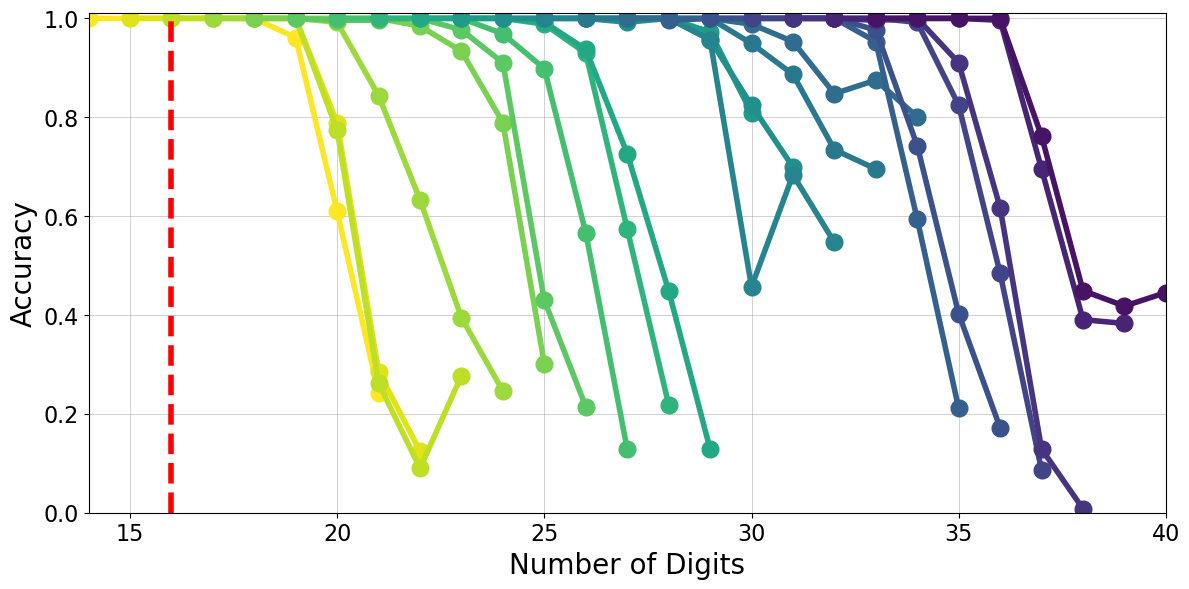}
    \includegraphics[width=0.49\linewidth]{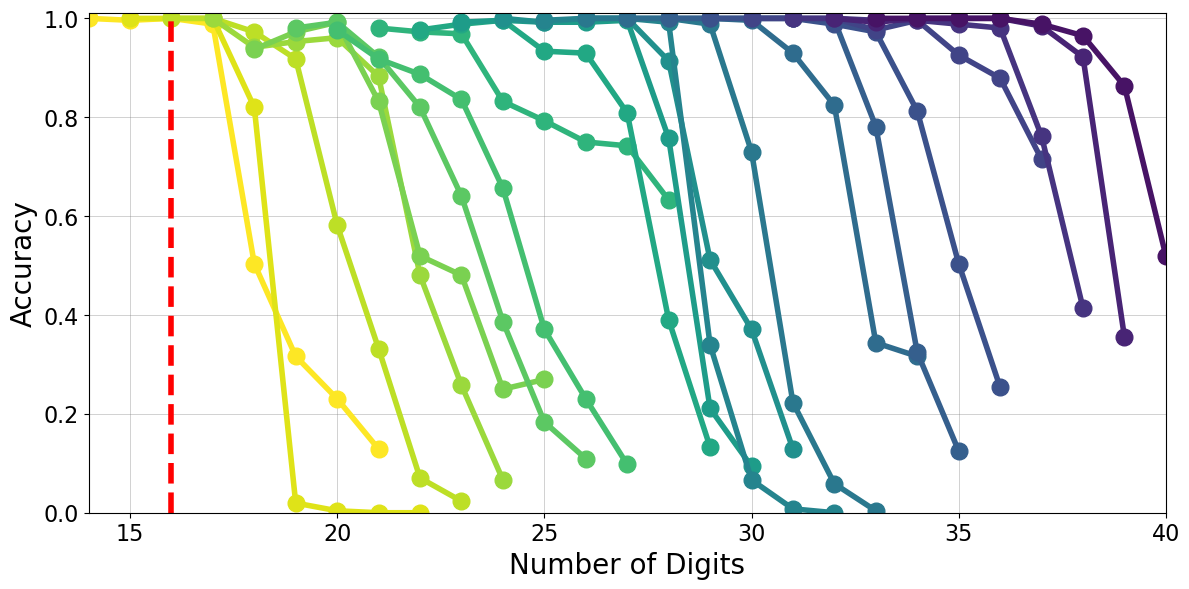}
    \caption{Reverse addition results for pretrained models. (Left) Llama-1B model. (Right) Llama-3B model. Larger models exhibit better extrapolation performance across rounds of self-improvement. }
    \label{fig:pretrained_model}
\end{figure}

\begin{wrapfigure}{r}{0.44\textwidth}
    \vspace{-6mm}
    \centering
    \includegraphics[width=0.44\textwidth]{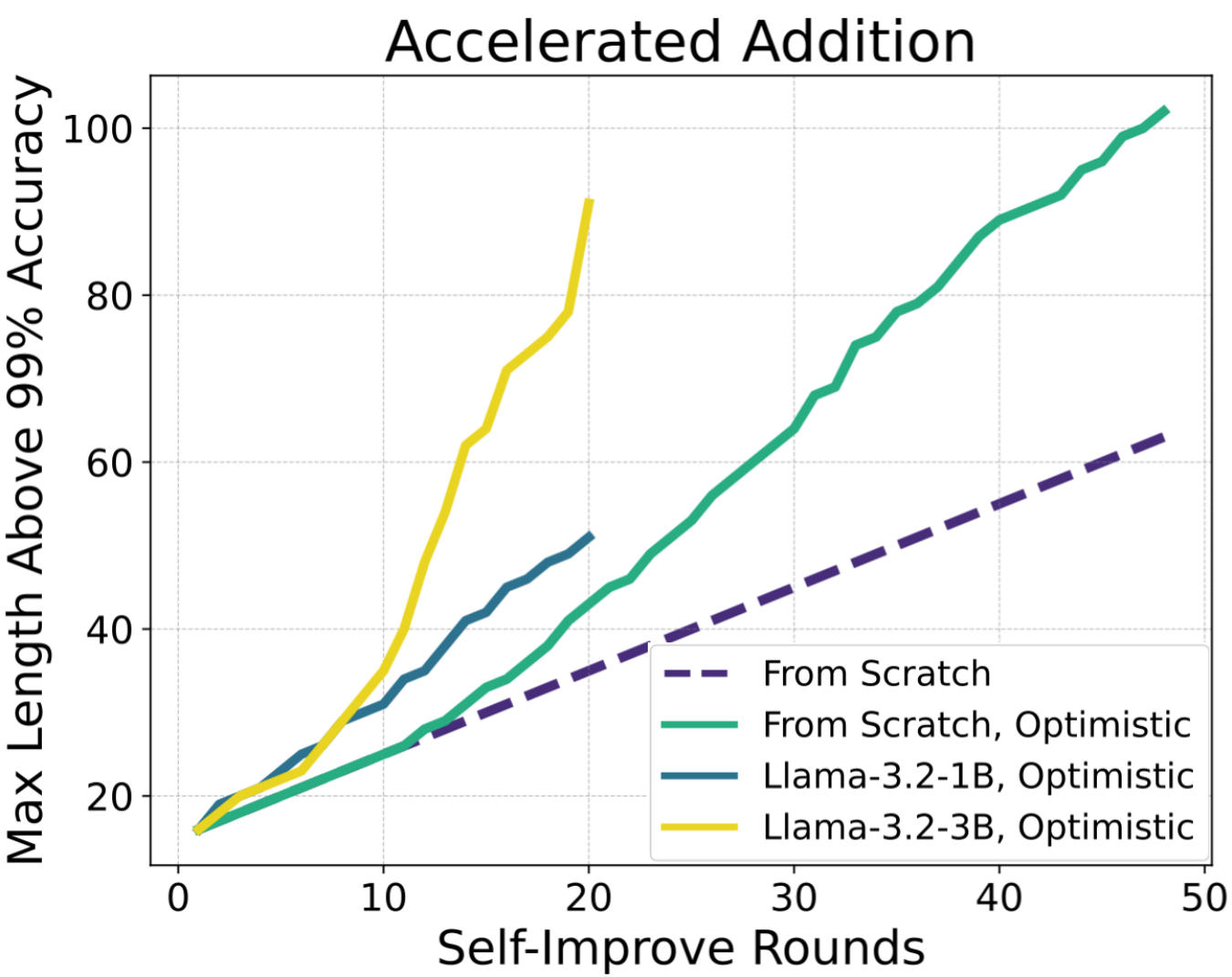}
    \caption{
    Accelerated reverse addition with pretrained models.
    Comparison of self-improvement acceleration for Llama-3B, Llama-1B, and a smaller 14M parameter model trained from scratch. Larger pretrained models demonstrate faster and more robust self-improvement.}
    \vspace{-5mm}
    \label{fig:pretrained_model_accelerate}
\end{wrapfigure}

\paragraph{Setting.}
To maintain consistency in tokenization, we use character-level tokenization instead of the default tokenizer of the Llama models. All other model components, including initial weights and positional embeddings, remain unchanged. We use LoRA~\citep{Hu2021LoRALA} with $r=64$ and $\alpha=128$ for Llama-1B, and $r=32$ and $\alpha=128$ for Llama-3B. 
In the initial round, we train for 1200 steps with a learning rate schedule that includes 10\% warm-up steps to a constant learning rate of \( 1\text{e-}4 \), followed by 20\% cosine decay steps to a final learning rate of \( 1\text{e-}6 \). For subsequent rounds, we train for 600 steps per round using a cosine decay learning rate schedule without warm-up, starting at \( 1\text{e-}4 \) and decaying to \( 1\text{e-}6 \).

\paragraph{Results.}
Figure~\ref{fig:pretrained_model} illustrates that larger models, such as Llama-3B, achieve better extrapolation performance in the reverse addition task compared to smaller models. Additionally, Figure~\ref{fig:pretrained_model_accelerate} highlights the potential for faster acceleration with larger models. By leveraging their pretraining capabilities, these models can generalize to longer sequences with fewer rounds of self-improvement. 

Figure~\ref{fig:pretrained_model_accelerate} compares self-improvement acceleration between Llama-3B, Llama-1B, and a smaller 14M parameter model trained from scratch. The results demonstrate that pretrained models significantly outperform the smaller model, both in terms of extrapolation and speed of self-improvement, showcasing the scalability of our approach.

\section{Analysis on Errors}\label{sec:error_analysis}
\vspace{0.5em}
\begin{finding}
During self-improvement failure cases, we see an \textbf{error avalanche} phenomenon where errors build up, eventually causing performance to crash. By simulating model errors at a particular round, we find that the model tolerates errors up to a certain point before crashing in accuracy.
\end{finding}

\begin{figure}
    \centering
    \includegraphics[width=0.45\linewidth]{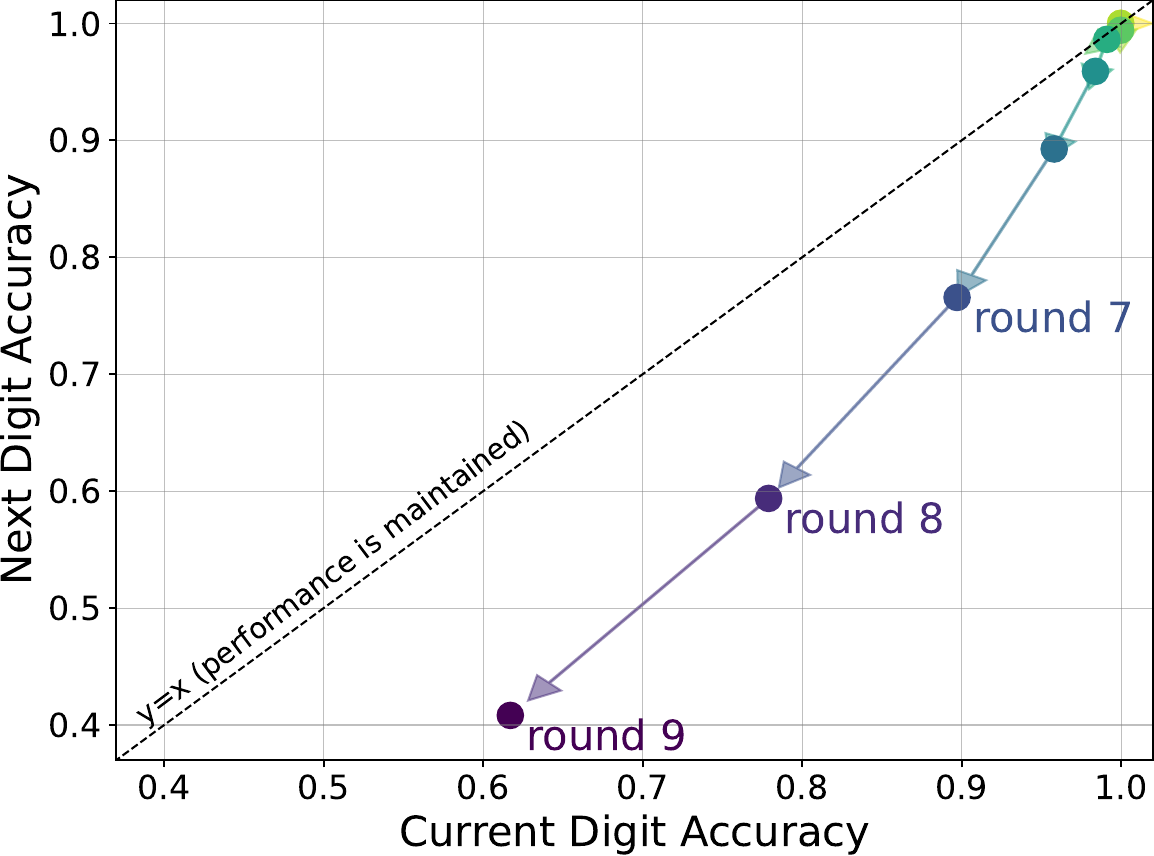}
    \hspace{1mm}
    \includegraphics[width=0.45\linewidth]{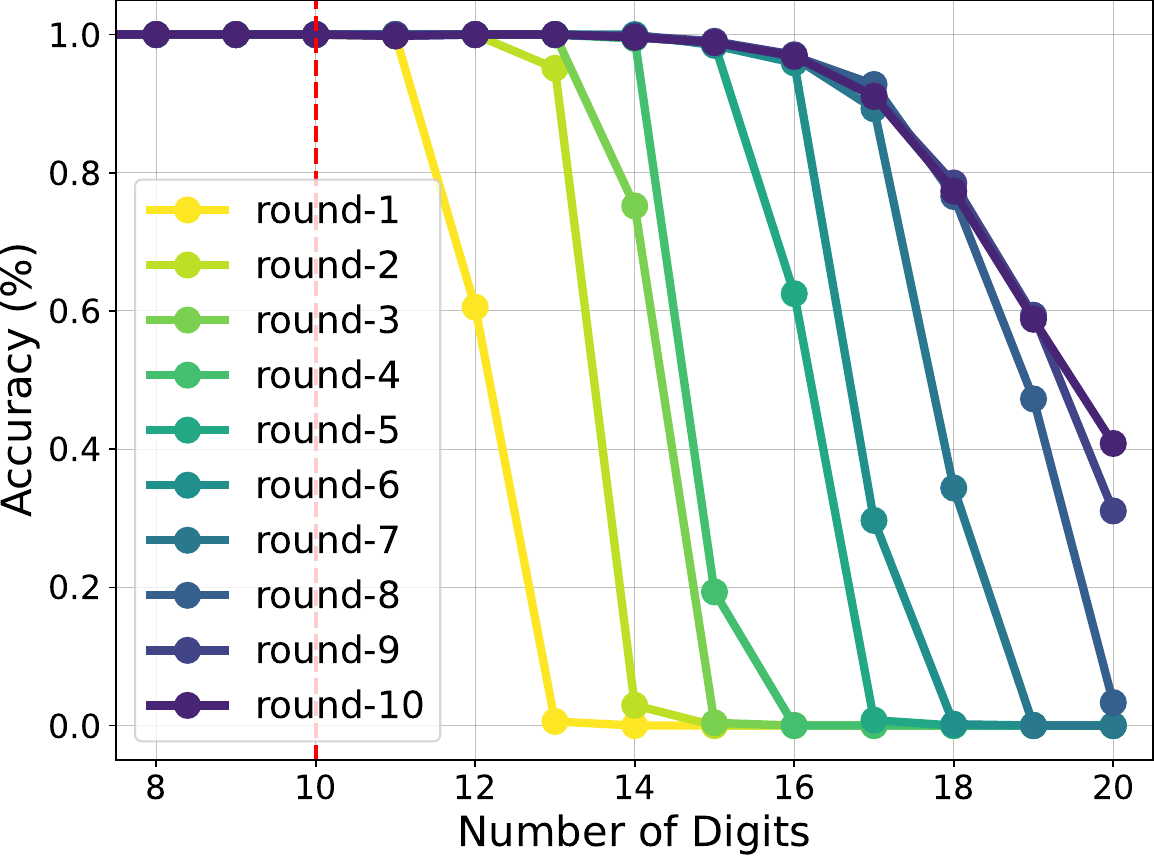}
    \caption{\textit{Error avalanche} is a common failure case for self-improvement. As inaccuracies in self-generated data accumulate, they degrade future rounds of training, leading to eventual failure. (Left) The impact of inaccuracies in $n$-digit data on $n+1$-digit generalization. (Right) Gradual performance degradation over successive self-improvement rounds}
    \label{fig:error_avalanche}
\end{figure}

\subsection{Error Avalanches in Self-Improvement}

Out-of-distribution (OOD) generalization is highly sensitive to inaccuracies in self-generated data. Figure~\ref{fig:error_avalanche} highlights a key challenge in this setting: errors in $n$-digit training data propagate to $n+1$-digit examples, degrading performance in later rounds. This is evident from data points falling below the $y = x$ line, indicating that self-improvement data is becoming progressively less reliable.

This cascading effect, known as an \textit{error avalanche}, compounds over successive self-improvement rounds, leading to a gradual collapse of the training process. As inaccuracies accumulate, the model's self-generated labels become increasingly erroneous, reducing the effectiveness of future training. Without effective data filtering or correction mechanisms, this process eventually causes the model to fail entirely.

\subsection{Simulating the Error Avalanche}
A natural question to ask at this point is,\textit{ how much error the model must accumulate to trigger an avalanche?} We investigate this question by first characterizing the model mistakes, and then injecting synthetic wrong examples in the self-improvement data.

\begin{figure}
    \centering
    \includegraphics[height=10.0625em]{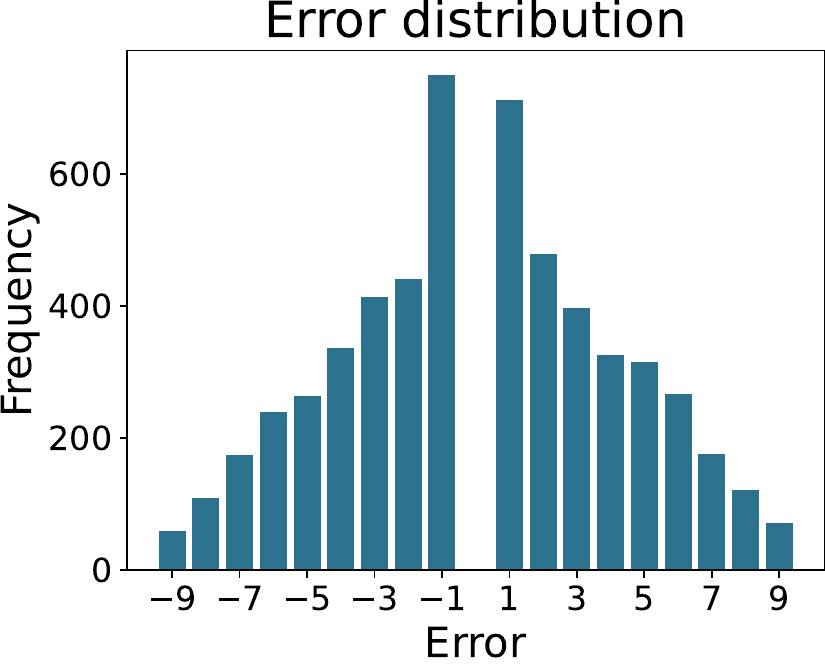}
    \includegraphics[height=10.0625em]{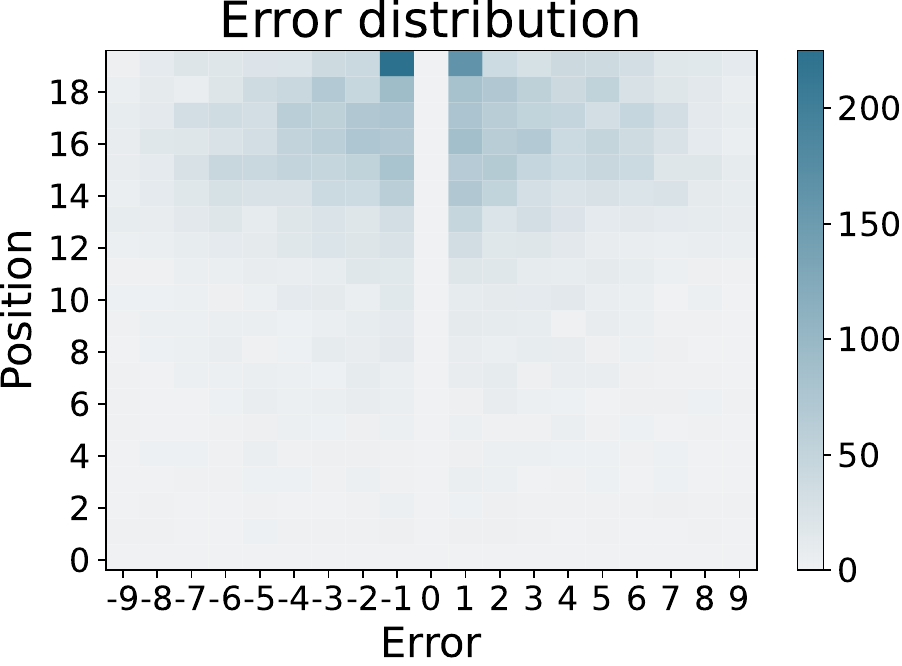}
    \includegraphics[height=10.0625em]{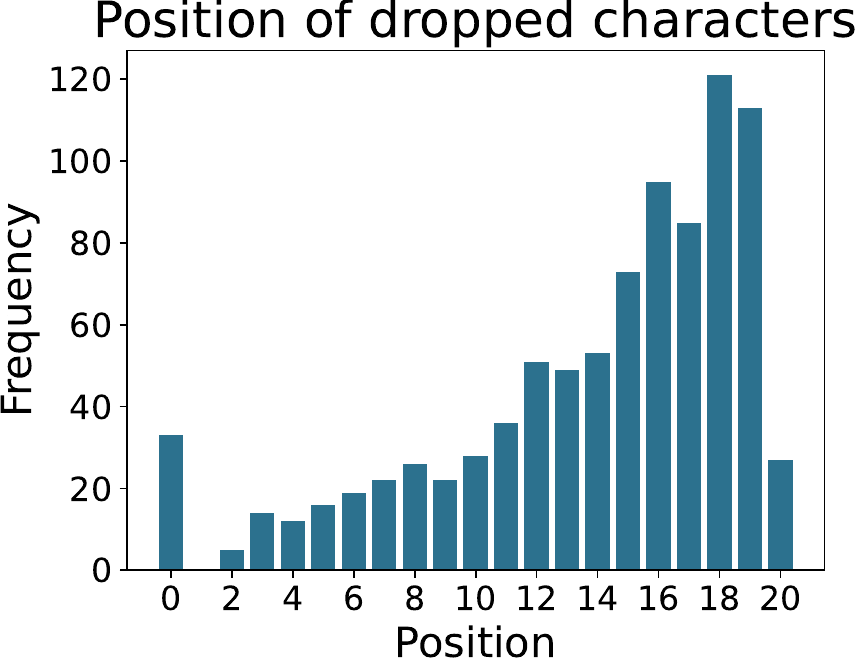}
    \caption{Patterns in model errors. (Left) Most incorrect digits are off by 1. (Middle) Errors cluster near the end of the sequence. (Right) Digit drop errors are strongly location-dependent.}
    \label{fig:mistake_patterns}
\end{figure}

\paragraph{Patterns in Model Mistakes.} 
We can categorize all mistakes into two bins. At each digit position, either the model drop the digit, or outputs a wrong digit. Since these two kinds of mistakes are entangled in practice, we use a string matching algorithm to compare the model output and predictions and obtain the best guess. In figure~\ref{fig:mistake_patterns}, we find that digit drops by the model are concentrated near the end of the sequence, and wrong digits are most often off by 1. 

\paragraph{Injecting Synthetic Errors.} 
Having characterized the model mistakes, we simulate them by constructing four kinds of noises:

\begin{itemize}
    \item Uniform: Replaces the label with a random number of the same length. 
    \item Perturb: Randomly modifies the last three digits by $\pm 1$.  
    \item Drop-Digits: Randomly removes 1, 2, or 3 digits from the last three positions. 
    \item Drop-Perturb: Combines "perturb" and "drop-digits" by first modifying digits and then randomly deleting some.
\end{itemize}

\begin{figure}
    \centering
    \includegraphics[width=0.49\linewidth]{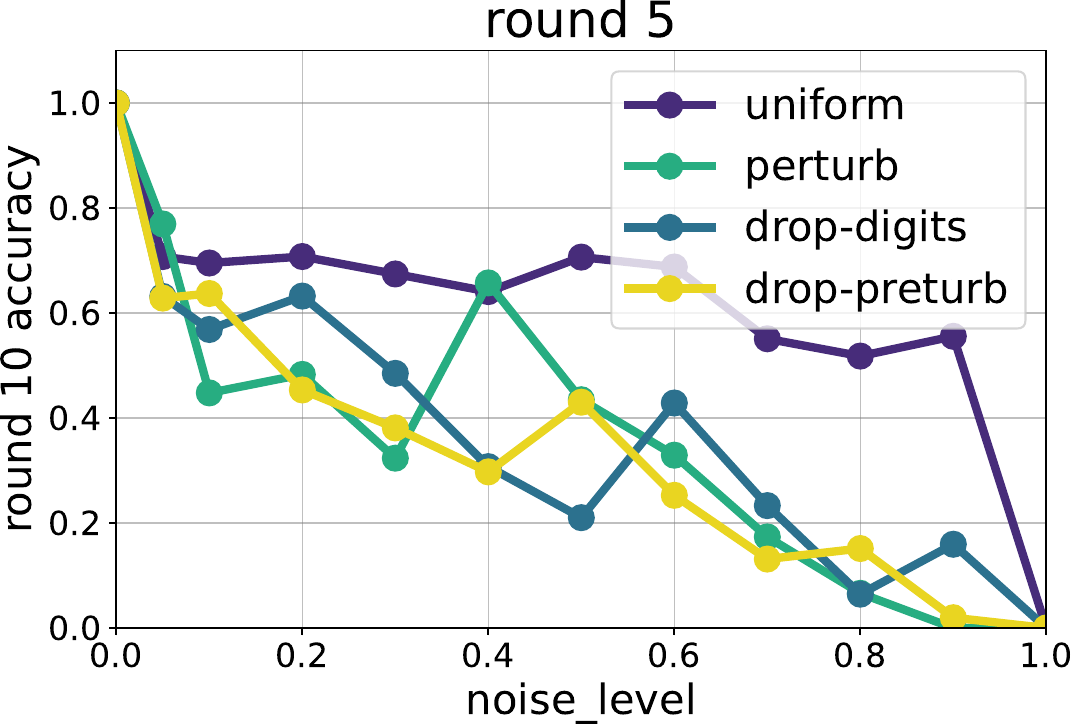}
    \includegraphics[width=0.49\linewidth]{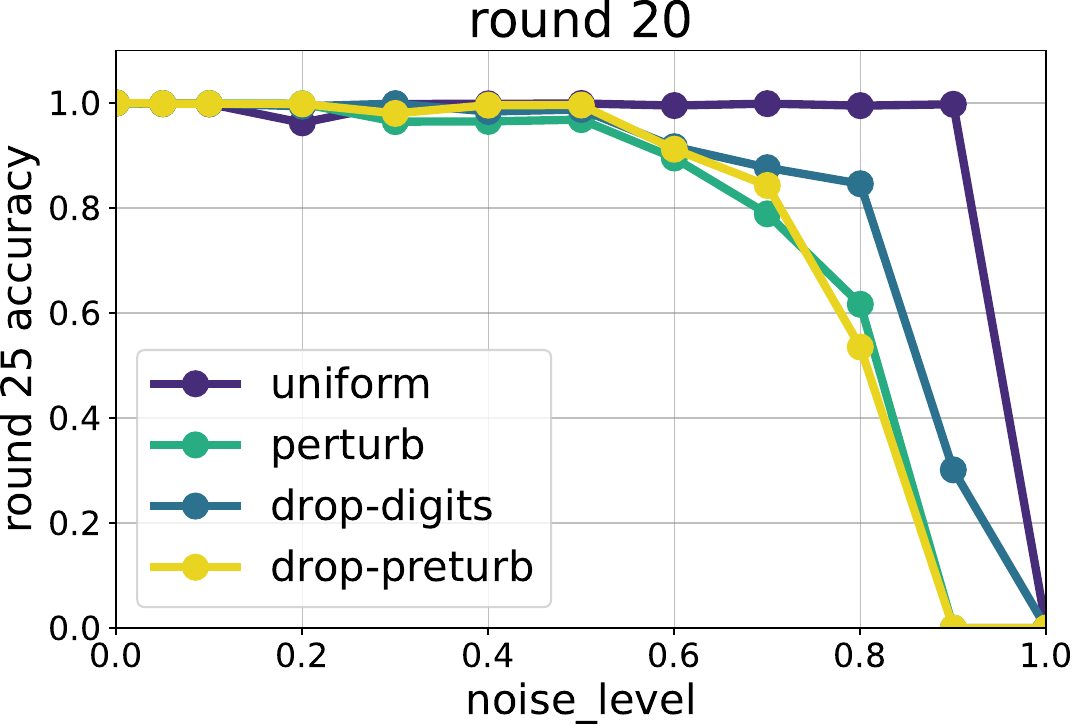}
    \caption{Simulating error avalanche. Synthetic mistakes of varying noise levels are injected at the end of rounds 5 and 20. The self-improvement process continues for 5 more rounds, and the resulting accuracy is recorded. The model tolerates errors up to a certain threshold, with greater tolerance observed in later self-improvement rounds.}
    \label{fig:simulated_error}
\end{figure}

We inject these errors of varying noise levels in rounds 5 and 20 of the reverse addition task and track their effects after five subsequent self-improvement rounds. As shown in Figure~\ref{fig:simulated_error}, injecting sufficient noise into the training data causes performance on the next difficulty to crash. In particular, we find that 1) structured noises (digit drops and perturbations) are more harmful than uniform noise and 2) more rounds of self-improvement improve robustness against label noise. Additional results on uniform errors are provided in Appendix~\ref{sec:appdx_label_noise}.

\paragraph{Models can Generalize Despite Memorizing Past Mistakes.}

\begin{wrapfigure}{r}{0.43\textwidth}
    \vspace{-3mm}
    \centering
    \includegraphics[width=0.42\textwidth]{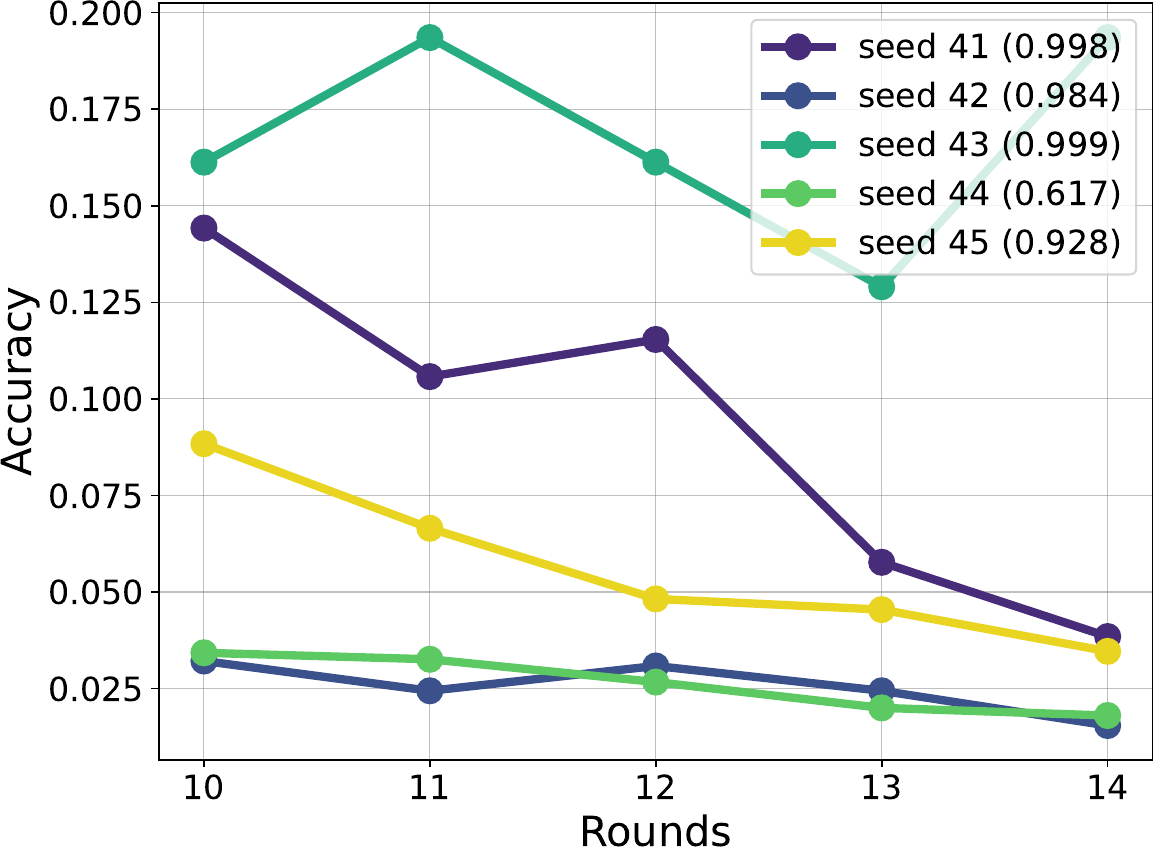}
    \caption{Models memorize their mistakes. Accuracy on incorrect training examples decreases with additional self-improvement rounds, indicating that repeated exposure reinforces errors instead of correcting them.}
    \vspace{-5mm}
    \label{fig:memorizing_mistakes}
\end{wrapfigure}

Since self-improvement involves recycling model predictions into training data, an important question is whether the model continues making mistakes on previously incorrect examples. To investigate this, we isolate incorrect self-generated samples and evaluate the model’s performance on them. 
As shown in Figure~\ref{fig:memorizing_mistakes}, the model struggles to rectify these errors. Accuracy on incorrect examples from the 19-digit self-generated training data decreases over additional self-improvement rounds, suggesting that repeated exposure to errors reinforces them rather than correcting them. %

However, memorizing past mistakes does not necessarily cause an error avalanche. The model under self-improvement often generalize to higher difficulties while treating the incorrect samples as outliers. For example, Figure~\ref{fig:simulated_error} shows that after 20 rounds of self-improvement, the model can tolerate a surprisingly large amount of label noise, from both uniform noise and structured noise. This suggests that while individual mistakes persist, they do not necessarily hinder overall generalization.

These findings emphasize the critical need for maintaining high-quality self-generated data to sustain effective and persistent self-improvement.

\paragraph{Relevance to Prior Work. }
Our results align with findings from~\citet{rolnick2017deep}, which demonstrate that deep neural networks are robust to significant label noise in image classification tasks. Additionally,~\citet{Bayat2024ThePO} recently emphasized that memorization alone does not harm generalization; rather, the combination of memorization with spurious correlations is what undermines learning. Our results suggest that despite memorizing past mistakes, the self-improvement framework remains effective, provided that incorrect samples do not dominate the training distribution.

\subsection{Other analysis}\label{sec:si_data_size}

\paragraph{Effect of Self-Improvement Dataset Size. }
We investigate how the quantity of self-generated training data impacts model performance. We first train 10 base models \( M_0^{(s)} \) (\( s=1, \dots, 10 \)) on a supervised 1-10 digit reverse addition dataset \( \mathcal{D}_0^s \), each using a different random seed. These models are categorized based on their accuracy on 11-digit addition: low-performing models (less than 98\% accuracy) are represented with yellow colors, while high-performing models (more than 98\% accuracy) are depicted with blue colors. 

To study the effect of dataset size, we generate self-improvement datasets \( \mathcal{D}_1^s = \{(x_i, M_0^{(s)}(x_i))\}_{i=1}^{N_1} \) of varying sizes (\( N_1 = 10,000, 50,000, 100,000, 500,000, 1,000,000 \)). Each model is then trained on the combined dataset \( \mathcal{D}_0^s \cup \mathcal{D}_1^s \). The number of incorrect examples in each self-generated dataset is approximately \( N_1 \times (1 - (\text{11-digit accuracy of } M_0)) \).

\begin{figure}
    \centering
    \includegraphics[width=0.4\linewidth]{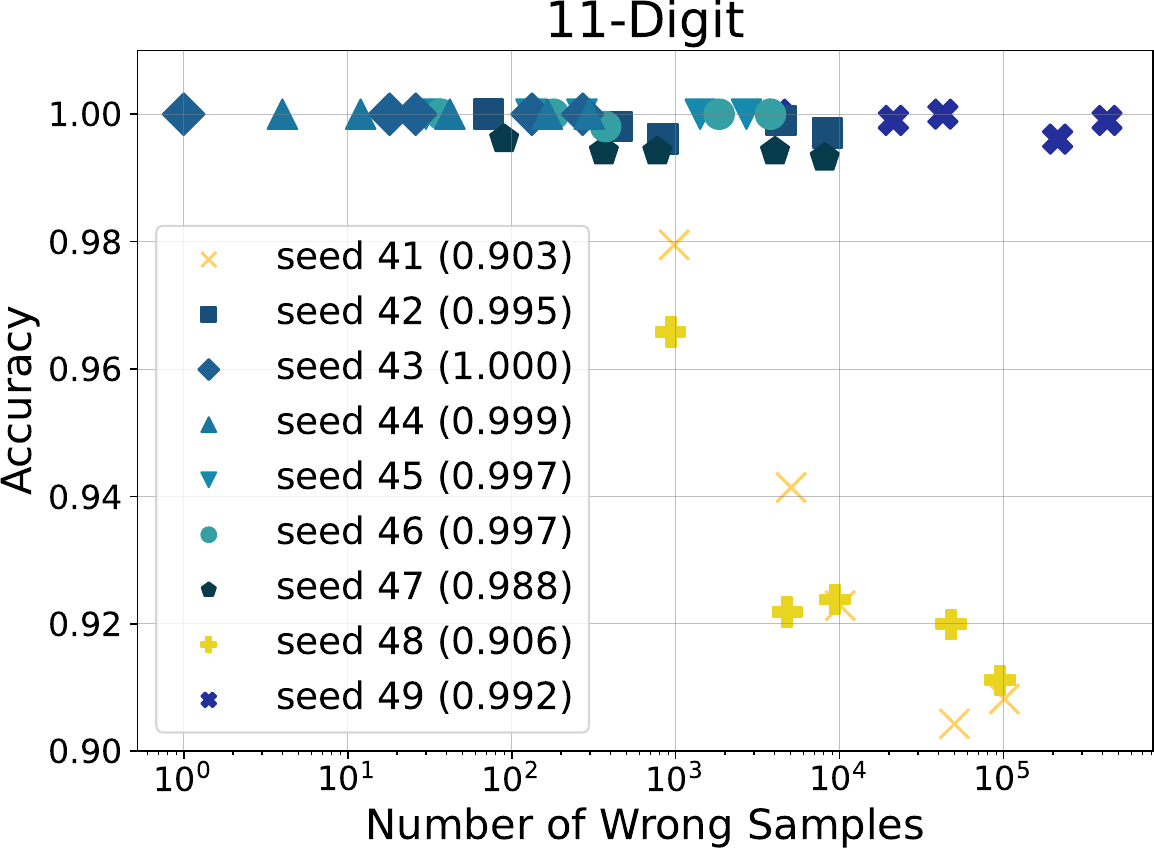}
    \hspace{1mm}
    \includegraphics[width=0.4\linewidth]{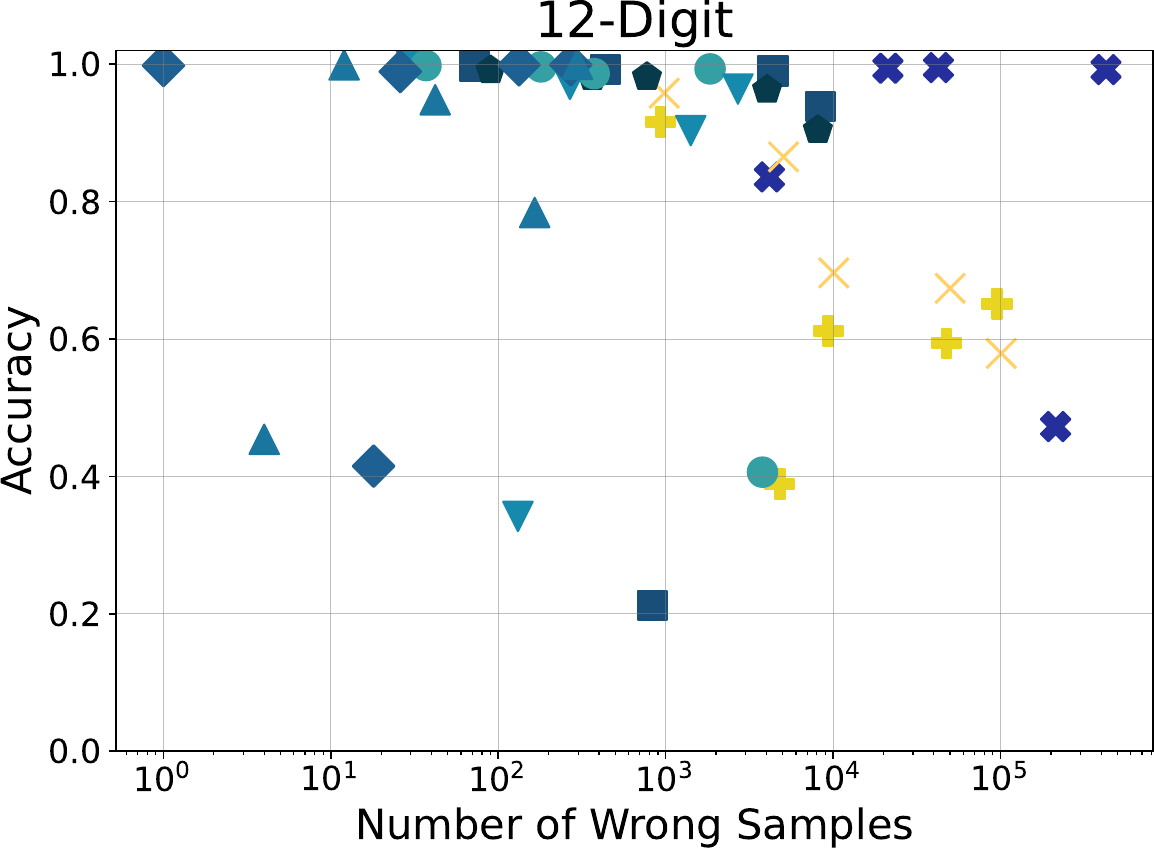}

    \caption{Effect of self-generated training data quantity and quality on model performance. Each model is trained on \( \mathcal{D}_0 \) (1-10 digit addition) and self-generated \( \mathcal{D}_1 \) (11-digit addition), then evaluated on 11-digit (in-distribution) and 12-digit (out-of-distribution) test performance. For low-performing models, increasing the quantity of self-generated data leads to degraded performance. For high-performing models, the impact of dataset size is less clear.}
    \label{fig:si_performance_by_num_samples}
\end{figure}

Results in Figure~\ref{fig:si_performance_by_num_samples} show that for low-performing models, increasing the quantity of self-generated data (which is of lower quality) degrades performance on both in-distribution (11-digit) and out-of-distribution (12-digit) addition. In contrast, for high-performing models, the relationship between the number of self-generated examples and performance is less clear. The total number of 11-digit examples seen during training remains constant across experiments, with smaller datasets being repeated more often. This suggests that exposure to a greater diversity of incorrect examples can bias the model more negatively.

\section{Limitations}\label{sec:discussion}
In our framework, the model does not generate new input instances during self-improvement; it only generates solutions (labels) for training. When it is unfeasible to generate the problems themselves, modeling the input distribution conditioned on task difficulty becomes an additional challenge.

A key consideration in self-improvement is defining and quantifying task difficulty. In real-world domains such as mathematics and natural language tasks, formalizing "difficulty" remains an open question. Our experiments demonstrate that careful difficulty scheduling is crucial for effective self-improvement. However, we also find that models exhibit some robustness to difficulty slack—especially when trained on harder tasks (Section~\ref{sec:ood_increases}) and when leveraging pretrained models (Section~\ref{sec:pretrained}). %

Another fundamental assumption in our framework is that models can handle slightly harder tasks than those seen in training. While this holds in many structured tasks, there are cases where such generalization is inherently difficult. For example, training on raw multiplication problems without intermediate steps leads to poor OOD generalization, making self-improvement infeasible. However, we show that breaking down tasks into intermediate steps enables slight OOD generalization, which can be leveraged for self-improvement(Section~\ref{sec:mult}). This highlights the importance of designing task representations that align with a model’s inherent generalization capabilities.

\section{Conclusion}\label{sec:conclusion}

In this work, we have shown self-improvement training enables transformers to gradually generalize from easy to hard problems without access to hard labels. One extension is to incorporate more sophisticated verifiers as well as problem classes that is easy to verify but hard to solve. We expect self-improve to synergize with strong verification to enable transformers to solve harder problems beyond arithmetic or mazes.

\bibliography{ref}

\begin{thebibliography}{70}
\providecommand{\natexlab}[1]{#1}
\providecommand{\url}[1]{\texttt{#1}}
\expandafter\ifx\csname urlstyle\endcsname\relax
  \providecommand{\doi}[1]{doi: #1}\else
  \providecommand{\doi}{doi: \begingroup \urlstyle{rm}\Url}\fi

\bibitem[AI@Meta(2024)]{llama3modelcard}
AI@Meta.
\newblock Llama 3 model card.
\newblock 2024.
\newblock URL \url{https://github.com/meta-llama/llama3/blob/main/MODEL_CARD.md}.

\bibitem[Alemohammad et~al.(2023)Alemohammad, Casco-Rodriguez, Luzi, Humayun, Babaei, LeJeune, Siahkoohi, and Baraniuk]{Alemohammad2023SelfConsumingGM}
Alemohammad, S., Casco-Rodriguez, J., Luzi, L., Humayun, A.~I., Babaei, H.~R., LeJeune, D., Siahkoohi, A., and Baraniuk, R.
\newblock Self-consuming generative models go mad.
\newblock \emph{ArXiv}, abs/2307.01850, 2023.
\newblock URL \url{https://api.semanticscholar.org/CorpusID:259341801}.

\bibitem[Alfarano et~al.(2024)Alfarano, Charton, and Hayat]{alfarano2024global}
Alfarano, A., Charton, F., and Hayat, A.
\newblock Global lyapunov functions: a long-standing open problem in mathematics, with symbolic transformers.
\newblock In \emph{The Thirty-eighth Annual Conference on Neural Information Processing Systems}, 2024.

\bibitem[Anil et~al.(2022)Anil, Wu, Andreassen, Lewkowycz, Misra, Ramasesh, Slone, Gur-Ari, Dyer, and Neyshabur]{anil2022exploring}
Anil, C., Wu, Y., Andreassen, A., Lewkowycz, A., Misra, V., Ramasesh, V., Slone, A., Gur-Ari, G., Dyer, E., and Neyshabur, B.
\newblock Exploring length generalization in large language models.
\newblock \emph{Advances in Neural Information Processing Systems}, 35:\penalty0 38546--38556, 2022.

\bibitem[Bachmann \& Nagarajan(2024)Bachmann and Nagarajan]{bachmann2024pitfalls}
Bachmann, G. and Nagarajan, V.
\newblock The pitfalls of next-token prediction.
\newblock \emph{arXiv preprint arXiv:2403.06963}, 2024.

\bibitem[Bansal et~al.(2022)Bansal, Schwarzschild, Borgnia, Emam, Huang, Goldblum, and Goldstein]{bansal2022end}
Bansal, A., Schwarzschild, A., Borgnia, E., Emam, Z., Huang, F., Goldblum, M., and Goldstein, T.
\newblock End-to-end algorithm synthesis with recurrent networks: Extrapolation without overthinking.
\newblock \emph{Advances in Neural Information Processing Systems}, 35:\penalty0 20232--20242, 2022.

\bibitem[Bansal et~al.(2024)Bansal, Hosseini, Agarwal, Tran, and Kazemi]{bansal2024smaller}
Bansal, H., Hosseini, A., Agarwal, R., Tran, V.~Q., and Kazemi, M.
\newblock Smaller, weaker, yet better: Training llm reasoners via compute-optimal sampling.
\newblock \emph{arXiv preprint arXiv:2408.16737}, 2024.

\bibitem[Bayat et~al.(2024)Bayat, Pezeshki, Dohmatob, Lopez-Paz, and Vincent]{Bayat2024ThePO}
Bayat, R., Pezeshki, M., Dohmatob, E., Lopez-Paz, D., and Vincent, P.
\newblock The pitfalls of memorization: When memorization hurts generalization.
\newblock 2024.
\newblock URL \url{https://api.semanticscholar.org/CorpusID:274610625}.

\bibitem[Bertrand et~al.(2023)Bertrand, Bose, Duplessis, Jiralerspong, and Gidel]{Bertrand2023OnTS}
Bertrand, Q., Bose, A.~J., Duplessis, A., Jiralerspong, M., and Gidel, G.
\newblock On the stability of iterative retraining of generative models on their own data.
\newblock \emph{ArXiv}, abs/2310.00429, 2023.
\newblock URL \url{https://api.semanticscholar.org/CorpusID:263334017}.

\bibitem[Breiman(1996)]{breiman1996bagging}
Breiman, L.
\newblock Bagging predictors.
\newblock \emph{Machine learning}, 24:\penalty0 123--140, 1996.

\bibitem[Briesch et~al.(2023)Briesch, Sobania, and Rothlauf]{Briesch2023LargeLM}
Briesch, M., Sobania, D., and Rothlauf, F.
\newblock Large language models suffer from their own output: An analysis of the self-consuming training loop.
\newblock \emph{ArXiv}, abs/2311.16822, 2023.
\newblock URL \url{https://api.semanticscholar.org/CorpusID:265466007}.

\bibitem[Burns et~al.(2023)Burns, Izmailov, Kirchner, Baker, Gao, Aschenbrenner, Chen, Ecoffet, Joglekar, Leike, et~al.]{burns2023weak}
Burns, C., Izmailov, P., Kirchner, J.~H., Baker, B., Gao, L., Aschenbrenner, L., Chen, Y., Ecoffet, A., Joglekar, M., Leike, J., et~al.
\newblock Weak-to-strong generalization: Eliciting strong capabilities with weak supervision.
\newblock \emph{arXiv preprint arXiv:2312.09390}, 2023.

\bibitem[Charton et~al.(2024)Charton, Ellenberg, Wagner, and Williamson]{charton2024patternboost}
Charton, F., Ellenberg, J.~S., Wagner, A.~Z., and Williamson, G.
\newblock Patternboost: Constructions in mathematics with a little help from ai.
\newblock \emph{arXiv preprint arXiv:2411.00566}, 2024.

\bibitem[Chen et~al.(2023)Chen, Lin, Sch{\"a}rli, and Zhou]{chen2023teaching}
Chen, X., Lin, M., Sch{\"a}rli, N., and Zhou, D.
\newblock Teaching large language models to self-debug.
\newblock \emph{arXiv preprint arXiv:2304.05128}, 2023.

\bibitem[Cho et~al.(2024)Cho, Cha, Awasthi, Bhojanapalli, Gupta, and Yun]{Cho2024PositionCI}
Cho, H., Cha, J., Awasthi, P., Bhojanapalli, S., Gupta, A., and Yun, C.
\newblock Position coupling: Improving length generalization of arithmetic transformers using task structure.
\newblock 2024.
\newblock URL \url{https://api.semanticscholar.org/CorpusID:273695226}.

\bibitem[de~Arcaute et~al.(2023)de~Arcaute, Watson, Reviriego, Hern{\'a}ndez, Ju{\'a}rez, and Sarkar]{Arcaute2023CombiningGA}
de~Arcaute, G. M.~R., Watson, L., Reviriego, P., Hern{\'a}ndez, J.~A., Ju{\'a}rez, M., and Sarkar, R.
\newblock Combining generative artificial intelligence (ai) and the internet: Heading towards evolution or degradation?
\newblock \emph{ArXiv}, abs/2303.01255, 2023.
\newblock URL \url{https://api.semanticscholar.org/CorpusID:257280389}.

\bibitem[Deng et~al.(2024)Deng, Choi, and Shieber]{deng2024explicit}
Deng, Y., Choi, Y., and Shieber, S.
\newblock From explicit cot to implicit cot: Learning to internalize cot step by step.
\newblock \emph{arXiv preprint arXiv:2405.14838}, 2024.

\bibitem[Dohmatob et~al.(2024)Dohmatob, Feng, Yang, Charton, and Kempe]{Dohmatob2024ATO}
Dohmatob, E., Feng, Y., Yang, P., Charton, F., and Kempe, J.
\newblock A tale of tails: Model collapse as a change of scaling laws.
\newblock \emph{ArXiv}, abs/2402.07043, 2024.
\newblock URL \url{https://api.semanticscholar.org/CorpusID:267628004}.

\bibitem[Duan et~al.(2023)Duan, Shi, and Xu]{duan2023interpolation}
Duan, S., Shi, Y., and Xu, W.
\newblock From interpolation to extrapolation: Complete length generalization for arithmetic transformers.
\newblock \emph{arXiv preprint arXiv:2310.11984}, 2023.

\bibitem[Dubois et~al.(2019)Dubois, Dagan, Hupkes, and Bruni]{dubois2019location}
Dubois, Y., Dagan, G., Hupkes, D., and Bruni, E.
\newblock Location attention for extrapolation to longer sequences.
\newblock \emph{arXiv preprint arXiv:1911.03872}, 2019.

\bibitem[Dziri et~al.(2024)Dziri, Lu, Sclar, Li, Jiang, Lin, Welleck, West, Bhagavatula, Le~Bras, et~al.]{dziri2024faith}
Dziri, N., Lu, X., Sclar, M., Li, X.~L., Jiang, L., Lin, B.~Y., Welleck, S., West, P., Bhagavatula, C., Le~Bras, R., et~al.
\newblock Faith and fate: Limits of transformers on compositionality.
\newblock \emph{Advances in Neural Information Processing Systems}, 36, 2024.

\bibitem[Fan et~al.(2024)Fan, Du, Ramchandran, and Lee]{fan2024looped}
Fan, Y., Du, Y., Ramchandran, K., and Lee, K.
\newblock Looped transformers for length generalization.
\newblock \emph{arXiv preprint arXiv:2409.15647}, 2024.

\bibitem[Feng et~al.(2024)Feng, Dohmatob, Yang, Charton, and Kempe]{feng2024beyond}
Feng, Y., Dohmatob, E., Yang, P., Charton, F., and Kempe, J.
\newblock Beyond model collapse: Scaling up with synthesized data requires reinforcement.
\newblock \emph{arXiv preprint arXiv:2406.07515}, 2024.

\bibitem[Gerstgrasser et~al.(2024)Gerstgrasser, Schaeffer, Dey, Rafailov, Sleight, Hughes, Korbak, Agrawal, Pai, Gromov, et~al.]{gerstgrasser2024model}
Gerstgrasser, M., Schaeffer, R., Dey, A., Rafailov, R., Sleight, H., Hughes, J., Korbak, T., Agrawal, R., Pai, D., Gromov, A., et~al.
\newblock Is model collapse inevitable? breaking the curse of recursion by accumulating real and synthetic data.
\newblock \emph{arXiv preprint arXiv:2404.01413}, 2024.

\bibitem[Gillman et~al.(2024)Gillman, Freeman, Aggarwal, Hsu, Luo, Tian, and Sun]{gillman2024self}
Gillman, N., Freeman, M., Aggarwal, D., Hsu, C.-H., Luo, C., Tian, Y., and Sun, C.
\newblock Self-correcting self-consuming loops for generative model training.
\newblock \emph{arXiv preprint arXiv:2402.07087}, 2024.

\bibitem[Gulcehre et~al.(2023)Gulcehre, Paine, Srinivasan, Konyushkova, Weerts, Sharma, Siddhant, Ahern, Wang, Gu, et~al.]{gulcehre2023reinforced}
Gulcehre, C., Paine, T.~L., Srinivasan, S., Konyushkova, K., Weerts, L., Sharma, A., Siddhant, A., Ahern, A., Wang, M., Gu, C., et~al.
\newblock Reinforced self-training (rest) for language modeling.
\newblock \emph{arXiv preprint arXiv:2308.08998}, 2023.

\bibitem[Hase et~al.(2024)Hase, Bansal, Clark, and Wiegreffe]{hase2024unreasonable}
Hase, P., Bansal, M., Clark, P., and Wiegreffe, S.
\newblock The unreasonable effectiveness of easy training data for hard tasks.
\newblock \emph{arXiv preprint arXiv:2401.06751}, 2024.

\bibitem[Hataya et~al.(2023)Hataya, Bao, and Arai]{Hataya_2023_ICCV}
Hataya, R., Bao, H., and Arai, H.
\newblock Will large-scale generative models corrupt future datasets?
\newblock In \emph{Proceedings of the IEEE/CVF International Conference on Computer Vision (ICCV)}, pp.\  20555--20565, October 2023.

\bibitem[Hosseini et~al.(2024)Hosseini, Yuan, Malkin, Courville, Sordoni, and Agarwal]{hosseini2024v}
Hosseini, A., Yuan, X., Malkin, N., Courville, A., Sordoni, A., and Agarwal, R.
\newblock V-star: Training verifiers for self-taught reasoners.
\newblock \emph{arXiv preprint arXiv:2402.06457}, 2024.

\bibitem[Hu et~al.(2021)Hu, Shen, Wallis, Allen-Zhu, Li, Wang, and Chen]{Hu2021LoRALA}
Hu, J.~E., Shen, Y., Wallis, P., Allen-Zhu, Z., Li, Y., Wang, S., and Chen, W.
\newblock Lora: Low-rank adaptation of large language models.
\newblock \emph{ArXiv}, abs/2106.09685, 2021.
\newblock URL \url{https://api.semanticscholar.org/CorpusID:235458009}.

\bibitem[Huang et~al.(2024)Huang, Block, Foster, Rohatgi, Zhang, Simchowitz, Ash, and Krishnamurthy]{huang2024selfimprovementlanguagemodelssharpening}
Huang, A., Block, A., Foster, D.~J., Rohatgi, D., Zhang, C., Simchowitz, M., Ash, J.~T., and Krishnamurthy, A.
\newblock Self-improvement in language models: The sharpening mechanism, 2024.
\newblock URL \url{https://arxiv.org/abs/2412.01951}.

\bibitem[Huang et~al.(2022)Huang, Gu, Hou, Wu, Wang, Yu, and Han]{huang2022large}
Huang, J., Gu, S.~S., Hou, L., Wu, Y., Wang, X., Yu, H., and Han, J.
\newblock Large language models can self-improve.
\newblock \emph{arXiv preprint arXiv:2210.11610}, 2022.

\bibitem[Hupkes et~al.(2020)Hupkes, Dankers, Mul, and Bruni]{hupkes2020compositionality}
Hupkes, D., Dankers, V., Mul, M., and Bruni, E.
\newblock Compositionality decomposed: How do neural networks generalise?
\newblock \emph{Journal of Artificial Intelligence Research}, 67:\penalty0 757--795, 2020.

\bibitem[Jelassi et~al.(2023)Jelassi, d'Ascoli, Domingo-Enrich, Wu, Li, and Charton]{jelassi2023length}
Jelassi, S., d'Ascoli, S., Domingo-Enrich, C., Wu, Y., Li, Y., and Charton, F.
\newblock Length generalization in arithmetic transformers.
\newblock \emph{arXiv preprint arXiv:2306.15400}, 2023.

\bibitem[Kazemnejad et~al.(2024)Kazemnejad, Padhi, Natesan~Ramamurthy, Das, and Reddy]{kazemnejad2024impact}
Kazemnejad, A., Padhi, I., Natesan~Ramamurthy, K., Das, P., and Reddy, S.
\newblock The impact of positional encoding on length generalization in transformers.
\newblock \emph{Advances in Neural Information Processing Systems}, 36, 2024.

\bibitem[Lee et~al.(2023)Lee, Sreenivasan, Lee, Lee, and Papailiopoulos]{lee2023teaching}
Lee, N., Sreenivasan, K., Lee, J.~D., Lee, K., and Papailiopoulos, D.
\newblock Teaching arithmetic to small transformers.
\newblock \emph{arXiv preprint arXiv:2307.03381}, 2023.

\bibitem[Li et~al.(2023)Li, You, Guruganesh, Ainslie, Ontanon, Zaheer, Sanghai, Yang, Kumar, and Bhojanapalli]{li2023functional}
Li, S., You, C., Guruganesh, G., Ainslie, J., Ontanon, S., Zaheer, M., Sanghai, S., Yang, Y., Kumar, S., and Bhojanapalli, S.
\newblock Functional interpolation for relative positions improves long context transformers.
\newblock \emph{arXiv preprint arXiv:2310.04418}, 2023.

\bibitem[Liang et~al.(2024)Liang, Zhang, Qu, Zheng, Guo, Du, Yang, Liu, Lin, Ma, et~al.]{liang2024sheep}
Liang, Y., Zhang, G., Qu, X., Zheng, T., Guo, J., Du, X., Yang, Z., Liu, J., Lin, C., Ma, L., et~al.
\newblock I-sheep: Self-alignment of llm from scratch through an iterative self-enhancement paradigm.
\newblock \emph{arXiv preprint arXiv:2408.08072}, 2024.

\bibitem[Lightman et~al.(2023)Lightman, Kosaraju, Burda, Edwards, Baker, Lee, Leike, Schulman, Sutskever, and Cobbe]{lightman2023let}
Lightman, H., Kosaraju, V., Burda, Y., Edwards, H., Baker, B., Lee, T., Leike, J., Schulman, J., Sutskever, I., and Cobbe, K.
\newblock Let's verify step by step.
\newblock \emph{arXiv preprint arXiv:2305.20050}, 2023.

\bibitem[Madaan et~al.(2024)Madaan, Tandon, Gupta, Hallinan, Gao, Wiegreffe, Alon, Dziri, Prabhumoye, Yang, et~al.]{madaan2024selfrefine}
Madaan, A., Tandon, N., Gupta, P., Hallinan, S., Gao, L., Wiegreffe, S., Alon, U., Dziri, N., Prabhumoye, S., Yang, Y., et~al.
\newblock Self-refine: Iterative refinement with self-feedback.
\newblock \emph{Advances in Neural Information Processing Systems}, 36, 2024.

\bibitem[McLeish et~al.(2024)McLeish, Bansal, Stein, Jain, Kirchenbauer, Bartoldson, Kailkhura, Bhatele, Geiping, Schwarzschild, et~al.]{mcleish2024transformers}
McLeish, S., Bansal, A., Stein, A., Jain, N., Kirchenbauer, J., Bartoldson, B.~R., Kailkhura, B., Bhatele, A., Geiping, J., Schwarzschild, A., et~al.
\newblock Transformers can do arithmetic with the right embeddings.
\newblock \emph{arXiv preprint arXiv:2405.17399}, 2024.

\bibitem[Newman et~al.(2020)Newman, Hewitt, Liang, and Manning]{newman2020eos}
Newman, B., Hewitt, J., Liang, P., and Manning, C.~D.
\newblock The eos decision and length extrapolation.
\newblock \emph{arXiv preprint arXiv:2010.07174}, 2020.

\bibitem[Pang et~al.(2024)Pang, Yuan, Cho, He, Sukhbaatar, and Weston]{pang2024iterativereasoningpreferenceoptimization}
Pang, R.~Y., Yuan, W., Cho, K., He, H., Sukhbaatar, S., and Weston, J.
\newblock Iterative reasoning preference optimization, 2024.
\newblock URL \url{https://arxiv.org/abs/2404.19733}.

\bibitem[Peng et~al.(2024)Peng, Xia, Yang, Xiong, Wu, and Xing]{peng2024regenesis}
Peng, X., Xia, C., Yang, X., Xiong, C., Wu, C.-S., and Xing, C.
\newblock Regenesis: Llms can grow into reasoning generalists via self-improvement.
\newblock \emph{arXiv preprint arXiv:2410.02108}, 2024.

\bibitem[Press et~al.(2021)Press, Smith, and Lewis]{press2021train}
Press, O., Smith, N.~A., and Lewis, M.
\newblock Train short, test long: Attention with linear biases enables input length extrapolation.
\newblock \emph{arXiv preprint arXiv:2108.12409}, 2021.

\bibitem[Qu et~al.(2024)Qu, Zhang, Garg, and Kumar]{qu2024recursive}
Qu, Y., Zhang, T., Garg, N., and Kumar, A.
\newblock Recursive introspection: Teaching language model agents how to self-improve.
\newblock \emph{arXiv preprint arXiv:2407.18219}, 2024.

\bibitem[Quirke \& Barez(2023)Quirke and Barez]{quirke2023understanding}
Quirke, P. and Barez, F.
\newblock Understanding addition in transformers.
\newblock \emph{arXiv preprint arXiv:2310.13121}, 2023.

\bibitem[Rolnick(2017)]{rolnick2017deep}
Rolnick, D.
\newblock Deep learning is robust to massive label noise.
\newblock \emph{arXiv preprint arXiv:1705.10694}, 2017.

\bibitem[Ruoss et~al.(2023)Ruoss, Del{\'e}tang, Genewein, Grau-Moya, Csord{\'a}s, Bennani, Legg, and Veness]{ruoss2023randomized}
Ruoss, A., Del{\'e}tang, G., Genewein, T., Grau-Moya, J., Csord{\'a}s, R., Bennani, M., Legg, S., and Veness, J.
\newblock Randomized positional encodings boost length generalization of transformers.
\newblock \emph{arXiv preprint arXiv:2305.16843}, 2023.

\bibitem[Sabbaghi et~al.(2024)Sabbaghi, Pappas, Hassani, and Goel]{sabbaghi2024explicitly}
Sabbaghi, M., Pappas, G., Hassani, H., and Goel, S.
\newblock Explicitly encoding structural symmetry is key to length generalization in arithmetic tasks.
\newblock \emph{arXiv preprint arXiv:2406.01895}, 2024.

\bibitem[Schwarzschild et~al.(2021)Schwarzschild, Borgnia, Gupta, Huang, Vishkin, Goldblum, and Goldstein]{schwarzschild2021can}
Schwarzschild, A., Borgnia, E., Gupta, A., Huang, F., Vishkin, U., Goldblum, M., and Goldstein, T.
\newblock Can you learn an algorithm? generalizing from easy to hard problems with recurrent networks.
\newblock \emph{Advances in Neural Information Processing Systems}, 34:\penalty0 6695--6706, 2021.

\bibitem[Shen et~al.(2023)Shen, Bubeck, Eldan, Lee, Li, and Zhang]{shen2023positional}
Shen, R., Bubeck, S., Eldan, R., Lee, Y.~T., Li, Y., and Zhang, Y.
\newblock Positional description matters for transformers arithmetic.
\newblock \emph{arXiv preprint arXiv:2311.14737}, 2023.

\bibitem[Shin et~al.(2024)Shin, Cooper, and Sala]{shin2024weak}
Shin, C., Cooper, J., and Sala, F.
\newblock Weak-to-strong generalization through the data-centric lens.
\newblock \emph{arXiv preprint arXiv:2412.03881}, 2024.

\bibitem[Shumailov et~al.(2023)Shumailov, Shumaylov, Zhao, Gal, Papernot, and Anderson]{shumailov2023curse}
Shumailov, I., Shumaylov, Z., Zhao, Y., Gal, Y., Papernot, N., and Anderson, R.
\newblock The curse of recursion: Training on generated data makes models forget.
\newblock \emph{arXiv preprint arXiv:2305.17493}, 2023.

\bibitem[Shumailov et~al.(2024)Shumailov, Shumaylov, Zhao, Papernot, Anderson, and Gal]{shumailov2024ai}
Shumailov, I., Shumaylov, Z., Zhao, Y., Papernot, N., Anderson, R., and Gal, Y.
\newblock Ai models collapse when trained on recursively generated data.
\newblock \emph{Nature}, 631\penalty0 (8022):\penalty0 755--759, 2024.

\bibitem[Singh et~al.(2023)Singh, Co-Reyes, Agarwal, Anand, Patil, Garcia, Liu, Harrison, Lee, Xu, et~al.]{singh2023beyond}
Singh, A., Co-Reyes, J.~D., Agarwal, R., Anand, A., Patil, P., Garcia, X., Liu, P.~J., Harrison, J., Lee, J., Xu, K., et~al.
\newblock Beyond human data: Scaling self-training for problem-solving with language models.
\newblock \emph{arXiv preprint arXiv:2312.06585}, 2023.

\bibitem[Song et~al.(2024)Song, Zhang, Eisenach, Kakade, Foster, and Ghai]{song2024mind}
Song, Y., Zhang, H., Eisenach, C., Kakade, S., Foster, D., and Ghai, U.
\newblock Mind the gap: Examining the self-improvement capabilities of large language models.
\newblock \emph{arXiv preprint arXiv:2412.02674}, 2024.

\bibitem[Sun et~al.(2024)Sun, Yu, Shen, Liu, Yang, Welleck, and Gan]{sun2024easy}
Sun, Z., Yu, L., Shen, Y., Liu, W., Yang, Y., Welleck, S., and Gan, C.
\newblock Easy-to-hard generalization: Scalable alignment beyond human supervision.
\newblock \emph{arXiv preprint arXiv:2403.09472}, 2024.

\bibitem[Vaswani et~al.(2017)Vaswani, Shazeer, Parmar, Uszkoreit, Jones, Gomez, Kaiser, and Polosukhin]{vaswani2017attention}
Vaswani, A., Shazeer, N., Parmar, N., Uszkoreit, J., Jones, L., Gomez, A.~N., Kaiser, {\L}., and Polosukhin, I.
\newblock Attention is all you need.
\newblock \emph{Advances in neural information processing systems}, 30, 2017.

\bibitem[Wang et~al.(2022{\natexlab{a}})Wang, Wei, Schuurmans, Le, Chi, and Zhou]{wang2022self}
Wang, X., Wei, J., Schuurmans, D., Le, Q., Chi, E., and Zhou, D.
\newblock Self-consistency improves chain of thought reasoning in language models.
\newblock \emph{arXiv preprint arXiv:2203.11171}, 2022{\natexlab{a}}.

\bibitem[Wang et~al.(2022{\natexlab{b}})Wang, Kordi, Mishra, Liu, Smith, Khashabi, and Hajishirzi]{wang2022selfinstruct}
Wang, Y., Kordi, Y., Mishra, S., Liu, A., Smith, N.~A., Khashabi, D., and Hajishirzi, H.
\newblock Self-instruct: Aligning language models with self-generated instructions.
\newblock \emph{arXiv preprint arXiv:2212.10560}, 2022{\natexlab{b}}.

\bibitem[Wen et~al.(2024)Wen, Li, Wang, Hall, Liang, and Ma]{wen2024understanding}
Wen, K., Li, Z., Wang, J., Hall, D., Liang, P., and Ma, T.
\newblock Understanding warmup-stable-decay learning rates: A river valley loss landscape perspective.
\newblock \emph{arXiv preprint arXiv:2410.05192}, 2024.

\bibitem[Yehudai et~al.(2021)Yehudai, Fetaya, Meirom, Chechik, and Maron]{yehudai2021local}
Yehudai, G., Fetaya, E., Meirom, E., Chechik, G., and Maron, H.
\newblock From local structures to size generalization in graph neural networks.
\newblock In \emph{International Conference on Machine Learning}, pp.\  11975--11986. PMLR, 2021.

\bibitem[Yuan et~al.(2024)Yuan, Pang, Cho, Sukhbaatar, Xu, and Weston]{yuan2024self}
Yuan, W., Pang, R.~Y., Cho, K., Sukhbaatar, S., Xu, J., and Weston, J.
\newblock Self-rewarding language models.
\newblock \emph{arXiv preprint arXiv:2401.10020}, 2024.

\bibitem[Zelikman et~al.(2022)Zelikman, Wu, and Goodman]{zelikman2022star}
Zelikman, E., Wu, Y., and Goodman, N.~D.
\newblock Star: Bootstrapping reasoning with reasoning.
\newblock 2022.
\newblock URL \url{https://api.semanticscholar.org/CorpusID:247762790}.

\bibitem[Zhang et~al.(2024)Zhang, Zhu, Saphra, Kleiman, Edelman, Tambe, Kakade, and Malach]{zhang2024transcendence}
Zhang, E., Zhu, V., Saphra, N., Kleiman, A., Edelman, B.~L., Tambe, M., Kakade, S.~M., and Malach, E.
\newblock Transcendence: Generative models can outperform the experts that train them.
\newblock \emph{arXiv preprint arXiv:2406.11741}, 2024.

\bibitem[Zhang \& Parkes(2023)Zhang and Parkes]{zhang2023chain}
Zhang, H. and Parkes, D.~C.
\newblock Chain-of-thought reasoning is a policy improvement operator.
\newblock \emph{arXiv preprint arXiv:2309.08589}, 2023.

\bibitem[Zhang et~al.(2019)Zhang, Song, Gao, Chen, Bao, and Ma]{zhang2019your}
Zhang, L., Song, J., Gao, A., Chen, J., Bao, C., and Ma, K.
\newblock Be your own teacher: Improve the performance of convolutional neural networks via self distillation.
\newblock In \emph{Proceedings of the IEEE/CVF international conference on computer vision}, pp.\  3713--3722, 2019.

\bibitem[Zhou et~al.(2023)Zhou, Bradley, Littwin, Razin, Saremi, Susskind, Bengio, and Nakkiran]{zhou2023algorithms}
Zhou, H., Bradley, A., Littwin, E., Razin, N., Saremi, O., Susskind, J., Bengio, S., and Nakkiran, P.
\newblock What algorithms can transformers learn? a study in length generalization.
\newblock \emph{arXiv preprint arXiv:2310.16028}, 2023.

\bibitem[Zhou et~al.(2024)Zhou, Alon, Chen, Wang, Agarwal, and Zhou]{zhou2024transformers}
Zhou, Y., Alon, U., Chen, X., Wang, X., Agarwal, R., and Zhou, D.
\newblock Transformers can achieve length generalization but not robustly.
\newblock \emph{arXiv preprint arXiv:2402.09371}, 2024.

\end{thebibliography}
\bibliographystyle{icml2023}

\newpage
\tableofcontents

\appendix
\addcontentsline{toc}{section}{Appendix} %
\part{Appendix}
\parttoc %
\newpage
\section{Detailed Discussion of Related Work}\label{sec:related_work_extended}

\paragraph{Length Generalization.}
While Transformers~\citep{vaswani2017attention} have achieved remarkable success, they often struggle with length generalization—where a model trained on problems of fixed length fails to extrapolate to longer sequences~\citep{dubois2019location,hupkes2020compositionality,newman2020eos,anil2022exploring}. Addressing this limitation is crucial, as poor length generalization indicates that language models may not fully understand the underlying task. 
~\citet{zhou2023algorithms} hypothesize that Transformers are more likely to length generalize on tasks with small RASP-L complexity. They demonstrate that tasks such as reverse addition and copying have low RASP-L complexity, making them easier to length generalize, whereas forward addition poses a greater challenge. 

Several approaches have been proposed to improve length generalization, particularly in arithmetic tasks. These include modifications to positional embeddings, such as Abacus embeddings~\citep{mcleish2024transformers}, NoPE~\citep{kazemnejad2024impact}, FIRE~\citep{li2023functional}, and pairwise positional encodings~\citep{sabbaghi2024explicitly,Cho2024PositionCI}, randomized positional encodings~\citep{ruoss2023randomized,zhou2024transformers}. Other methods focus on architectural changes, such as introducing looping mechanisms~\citep{fan2024looped} or incorporating hand-crafted bias corrections in attention score matrices~\citep{duan2023interpolation}. Additionally, input modifications, such as index hinting, have been explored to enhance generalization~\citep{zhou2023algorithms,zhou2024transformers}. 
Beyond arithmetic, length generalization has also been studied in the context of size generalization in graph-based tasks~\citep{yehudai2021local}.

In contrast, our approach adheres to the standard transformer architecture without introducing modifications to architecture, positional encodings, or input structure. A key distinction lies in the training methodology. While prior approaches typically rely on fixed-length training datasets without further updates to model weights, we iteratively update model weights on self-generated datasets, enabling the model to progressively improve and extend its generalization capabilities.

Our self-improvement framework and results on forward addition (Section~\ref{sec:forward_addition}) are closely related to those of~\citet{zhang2023chain}, where self-training enables forward addition generalization from 6-digit examples to 24-digit addition. Like their approach, we iteratively apply self-training on progressively harder problems. However, a key distinction is that their method follows a two-step process in each round: first generating solutions using chain-of-thought (CoT) reasoning, then fine-tuning the model to produce direct answers without CoT. 

Our multiplication results in Section~\ref{sec:mult} have relevance with findings by~\citet{jelassi2023length}, who showed that dataset priming (adding a small number of labeled long-sequence examples) can enable length generalization for multiplication (although this is not strictly out-of-distribution). Our approach of incorporating accurate, self-generated out-of-distribution data via filtering can be seen as an automated form of dataset priming.
Furthermore, while our approach uses chain-of-thought (CoT) data for multiplication, we believe it is possible to length generalize on non-COT multiplication as well, by incorporating methods like~\citet{deng2024explicit} to help the model iteratively internalize the CoT steps.

\paragraph{Easy-to-hard Generalization.} 
Our self-improvement framework operates in a setting where human annotation is provided for easier tasks, enabling generalization to harder tasks with no additional supervision. This paradigm, often referred to as easy-to-hard generalization~\citep{schwarzschild2021can,bansal2022end,burns2023weak,hase2024unreasonable,sun2024easy}, leverages the transfer of learned policies or reward models from simpler problems to more challenging ones.
For instance, ~\citet{zhang2024transcendence} study this phenomenon in chess, showing that chess transformers can sometimes outperform all players in the training dataset. Similarly, ~\citet{sun2024easy} finds that a reward model trained on easier mathematical problems can be effectively transferred to harder problems, facilitating generalization through reinforcement learning. ~\citet{shin2024weak} identifies overlap data points—instances containing both easy and hard patterns—as a key mechanism for weak-to-strong generalization, allowing weak models to pseudolabel easier patterns while stronger models use these labels to learn harder patterns. Our work shows that a similar mechanism emerges naturally within self-improvement, where progressively increasing difficulty enables models to generate useful supervision signals for harder tasks without explicit human intervention.

\paragraph{Self-Improvement.}

When high quality training labels are not available, training on self-generated labels is an efficient way to extract more capabilities from the model. Usually, this involves generating candidate labels, pruning wrong labels through verification or filtering, and retraining with self-generated data. 
ReST~\citep{gulcehre2023reinforced} and I-SHEEP~\citep{liang2024sheep} propose self-improvement as a general purpose alternative to reinforcement learning from human feedback (RLHF), while ~\citet{yuan2024self} propose "self-rewarding" model that generates its own instruction tuning set.

The self-improvement framework has been applied to a wide range of tasks. For example, ~\citet{zhang2019your} replaces an expensive teacher distillation with self-distillation for image recognition tasks. In LLM reasoning domains, \citet{huang2022large,singh2023beyond,pang2024iterativereasoningpreferenceoptimization}, and STaR~\citep{zelikman2022star} bootstrap complex reasoning capabilities by asking models to generate rationales for unlabeled questions and training on self-generated rationals that yielded correct answers. Similarly,~\citet{zhang2023chain} shows self-improving using chain-of-thought (COT) data sampled from the model allows generalization of the integer addition task to more digits. 
For coding tasks, ~\citet{chen2023teaching} teaches LLMs to self-debug with feedback using self-generated code explanation and unit test execution results. 
In mathematics, PatternBoost~\citep{charton2024patternboost} shows that transformers can discover unsolved mathematical constructions of various problems using an algorithm that alternates between sampling constructions from the model (local search) and training on self-generated data (global learning). Similarly, \citet{alfarano2024global} generate synthetic training samples to train transformers to discover new Lyapunov functions.
Recent works also investigate theoretical and empirical aspects of self-improvement. \citet{bansal2024smaller} highlight the effectiveness of smaller models in self-improvement, while \citet{song2024mind} identify the generation-verification gap as a key factor governing the self-improvement process. \citet{huang2024selfimprovementlanguagemodelssharpening} introduce the "sharpening mechanism," where training on best-of-N responses from the model amortizes maximum likelihood inference and improves output quality.

Our work is greatly inspired by ReST~\citep{gulcehre2023reinforced} and STaR~\citep{zelikman2022star}, in which models iteratively generate predictions, filter high-quality responses, and fine-tune on the self-generated dataset.

\paragraph{Model Collapse.}
Recent research has extensively investigated the phenomenon of model collapse, where repeated training on a model's own outputs leads to performance degeneration and a loss of the true underlying data distribution~\citep{shumailov2024ai,Hataya_2023_ICCV,Arcaute2023CombiningGA,shumailov2023curse,Alemohammad2023SelfConsumingGM,Briesch2023LargeLM}.

\citet{shumailov2024ai} provide evidence that iterative training on model-generated data, without filtering, results in rapid degeneration and forgetting of the true data distribution. They emphasize the importance of preserving original data sources over time. Similarly, \citet{shumailov2023curse} show that the tails of the original content distribution diminish after repeated self-training, while \citet{zhang2023chain} highlight the error avalanching effect, where errors compound as models are trained on their own generated data.

Despite its apparent inevitability, several strategies have been proposed to mitigate model collapse. Research shows that the risk of collapse diminishes when the initial model closely approximates the true data distribution~\citep{Bertrand2023OnTS}, or when real data is retained throughout training rather than being fully replaced~\citep{gerstgrasser2024model,Dohmatob2024ATO,Briesch2023LargeLM}. Additionally, ~\citet{gillman2024self,feng2024beyond} suggest using reliable verifiers during self-training to ensure high-quality self-generated data, further reducing the likelihood of collapse.

Our approach addresses these challenges by maintaining a core labeled dataset throughout training, consisting of examples of limited length or difficulty. Synthetic data, generated incrementally by the model, is added in a controlled manner. By incorporating unsupervised filtering techniques such as length filtering and majority voting, we ensure the quality of self-generated data. Our framework builds upon prior findings by preserving clean data and selectively incorporating synthetic data.

Additionally, our results in Section~\ref{sec:error_analysis} align with findings from~\citet{rolnick2017deep}, which demonstrate that deep neural networks are robust to significant label noise in image classification tasks. Additionally,~\citet{Bayat2024ThePO} recently emphasized that memorization alone does not harm generalization; rather, the combination of memorization with spurious correlations is what undermines learning. Our results suggest that despite memorizing past mistakes, the self-improvement framework remains effective, provided that incorrect samples do not dominate the training distribution.

\newpage
\section{Additional Results}\label{sec:additional_results}

\subsection{Does the Model Truly Learn Addition?}
\begin{figure}[ht!]
    \centering
    \includegraphics[width=0.35\linewidth]{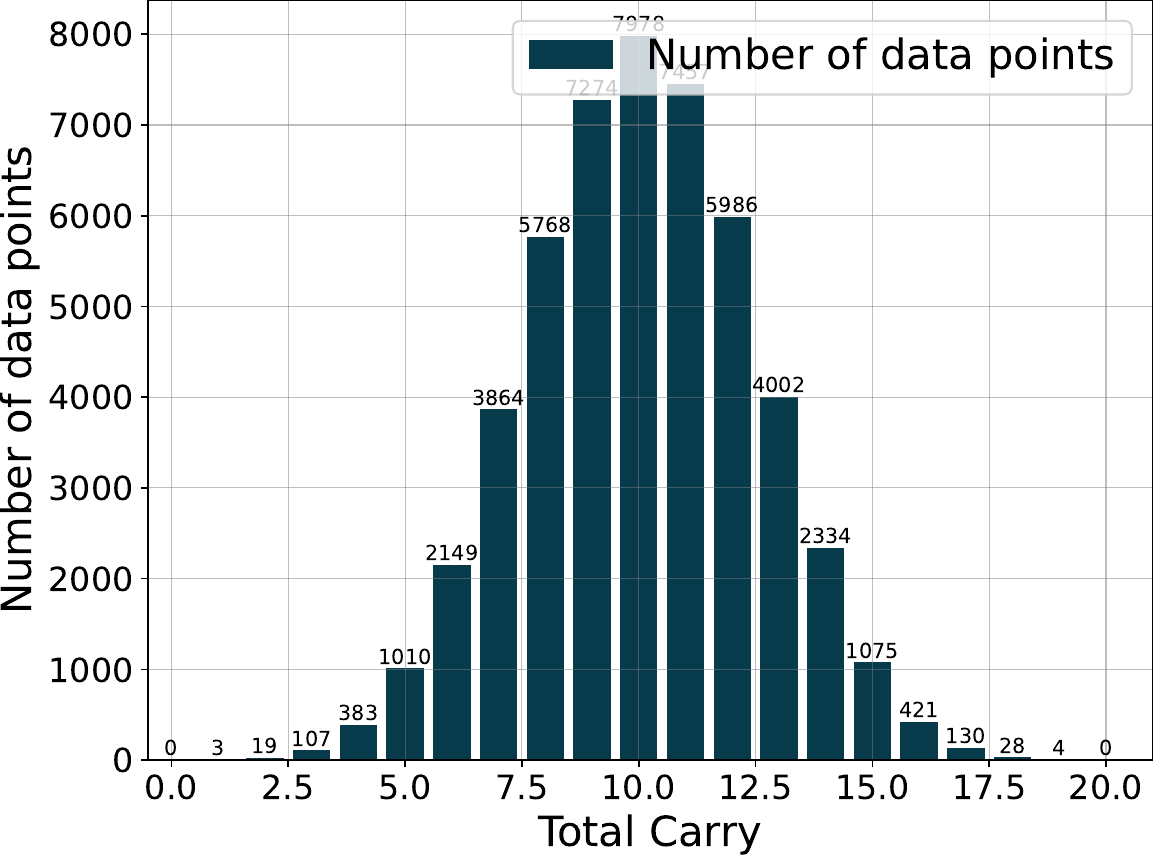}
    \hspace{1mm}
    \includegraphics[width=0.35\linewidth]{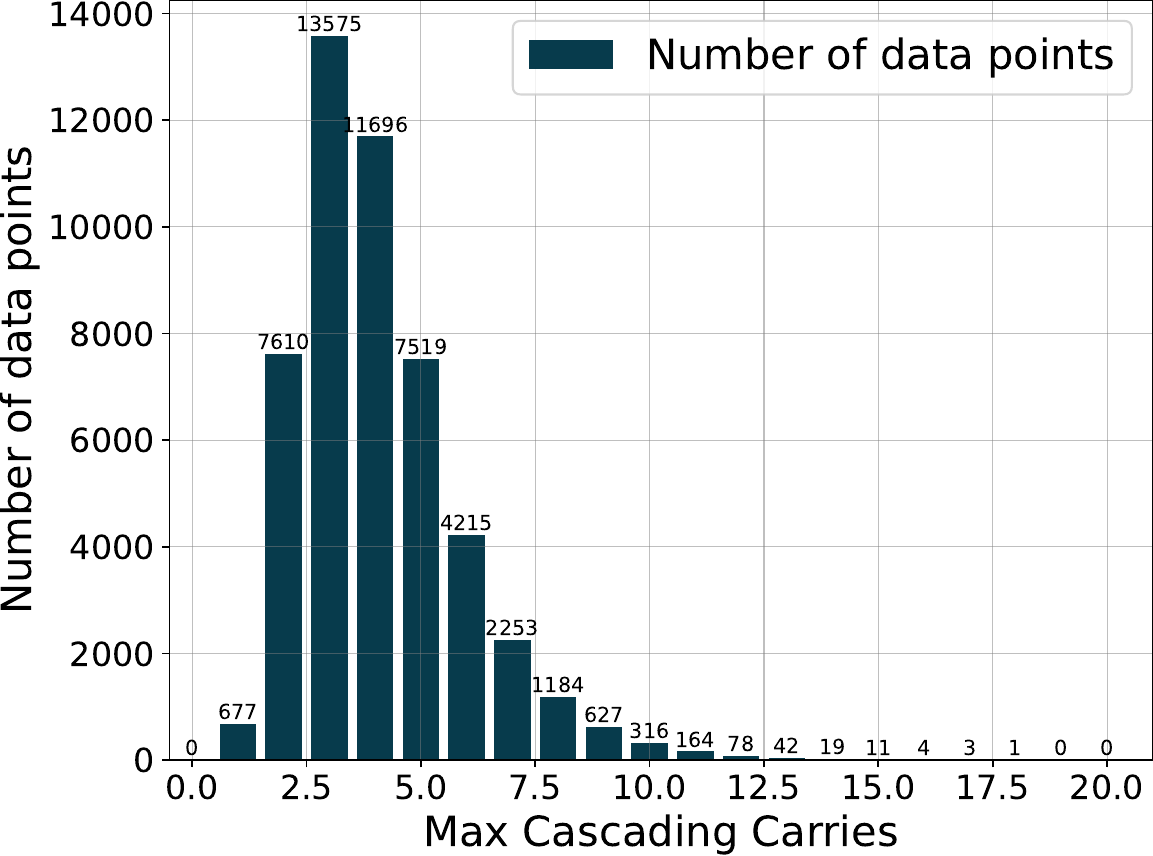}
    \caption{Distribution of carry occurrences in the standard 20-digit self-improve dataset. Models rarely encounter examples with large numbers of carries during training. }
    \label{fig:num_carries}
\end{figure}

When the two operands of length $N$ are sampled randomly, the probability of encountering an instance with a carry chain length of $N$ decreases exponentially with $N$. Under this sampling strategy, the model rarely, if ever, see ``hard\footnote{we define hard instance of addition to be cases with multiple numbers of cascading carries~\citep{quirke2023understanding}}'' instances of addition, as shown in Figure~\ref{fig:num_carries}. To address this, we manually construct a test dataset to include at least 500 examples for each maximum cascading carry length. This ensures that the evaluation captures the model's ability to handle harder instances of addition.

\begin{figure}[ht!]
    \centering
    \includegraphics[width=0.37\linewidth]{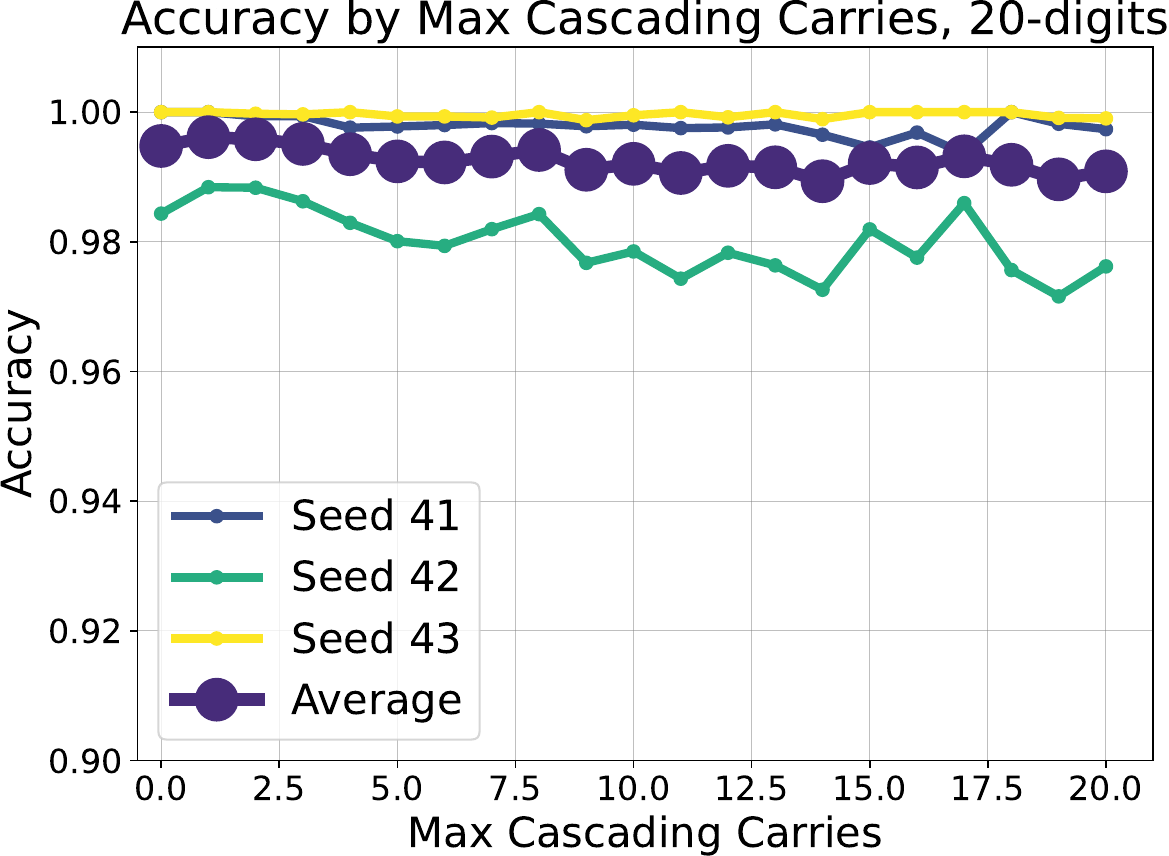}
    \hspace{1mm}
    \includegraphics[width=0.35\linewidth]{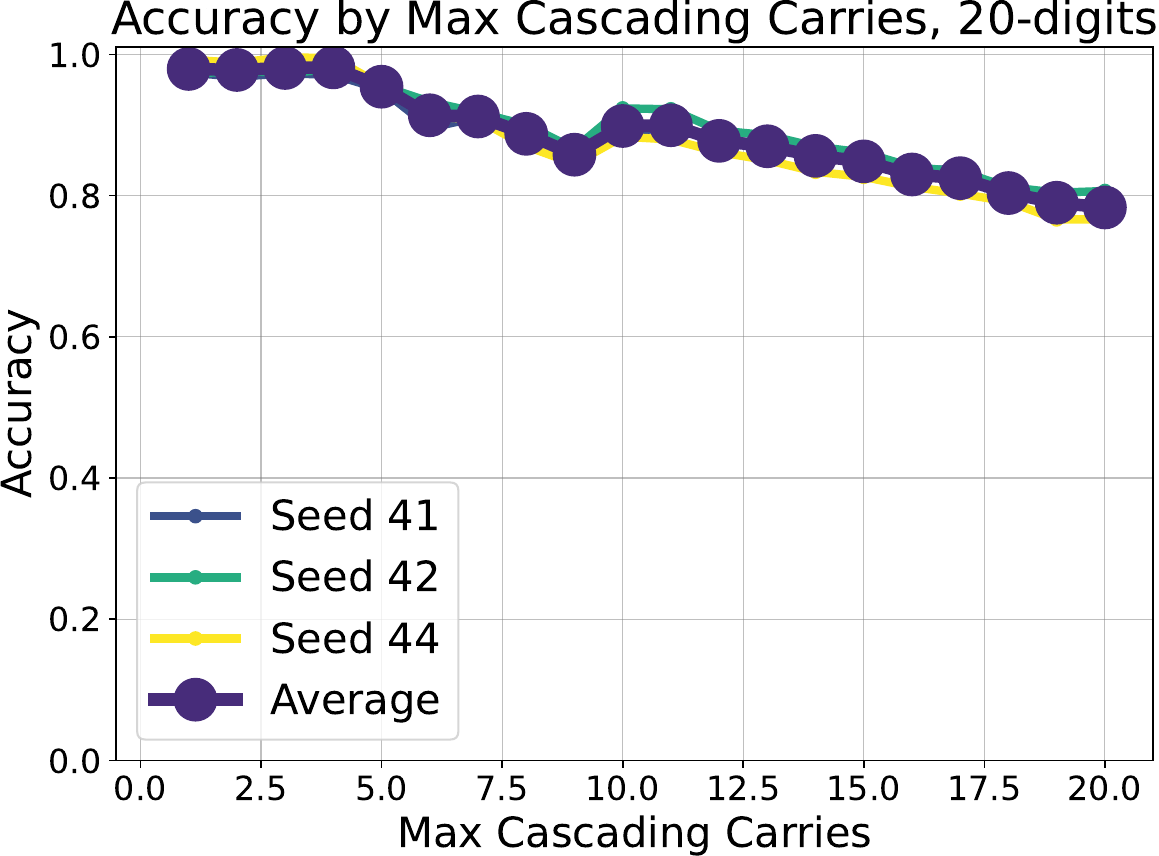}
    \caption{Performance of the model at round 10 (trained with self-generated data up to 19 digits). Accuracy as a function of the maximum cascading carries. (Left) Reverse addition. (Right) Forward addition. Despite limited exposure to high-carry cases in training, models achieve strong generalization to these harder instances.}
    \label{fig:carry_acc}
\end{figure}

We evaluate models trained on 1-10 digit addition that undergo 10 rounds of self-improvement, encountering self-generated data up to 19 digits in length. The results in Figure~\ref{fig:carry_acc} indicate that the model successfully performs additions involving up to 20 cascading carries, despite never encountering such cases during training. Reverse addition maintains near-perfect accuracy across all carry lengths, whereas forward addition performance drops to 80\% at 20 cascading carries—where the carry effect propagates across the entire sequence. Nonetheless, the model's ability to handle these challenging cases demonstrates its capacity to generalize to harder instances beyond its training distribution.

\subsection{Additional Results on Majority Voting}
\subsubsection{Majority Voting Leverages Label Diversity}\label{sec:mv_diversity}

Self-improvement relies on the model's ability to generalize to slightly harder problems. However, this generalization is not always robust and can vary significantly across different training instances~\citep{zhou2024transformers}. Majority voting mitigates this variability by aggregating predictions across multiple independently trained models, thereby improving the reliability of self-generated labels.

\begin{figure}[ht!]
    \centering
    \resizebox{0.8\linewidth}{!}{
    \begin{minipage}{\linewidth} \centering
    \includegraphics[width=0.3\linewidth]{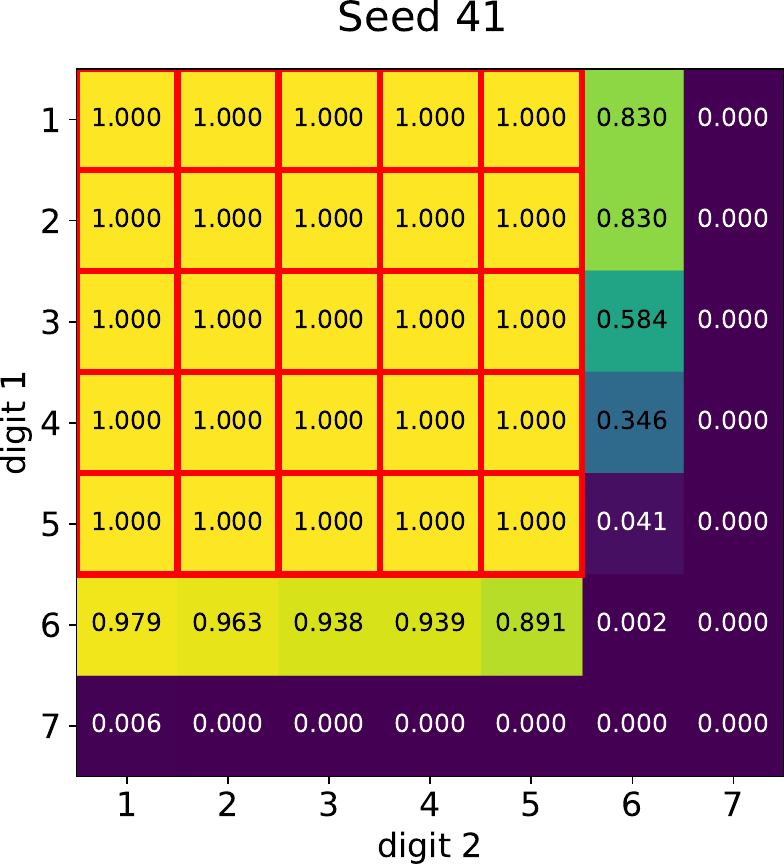}
    \includegraphics[width=0.3\linewidth]{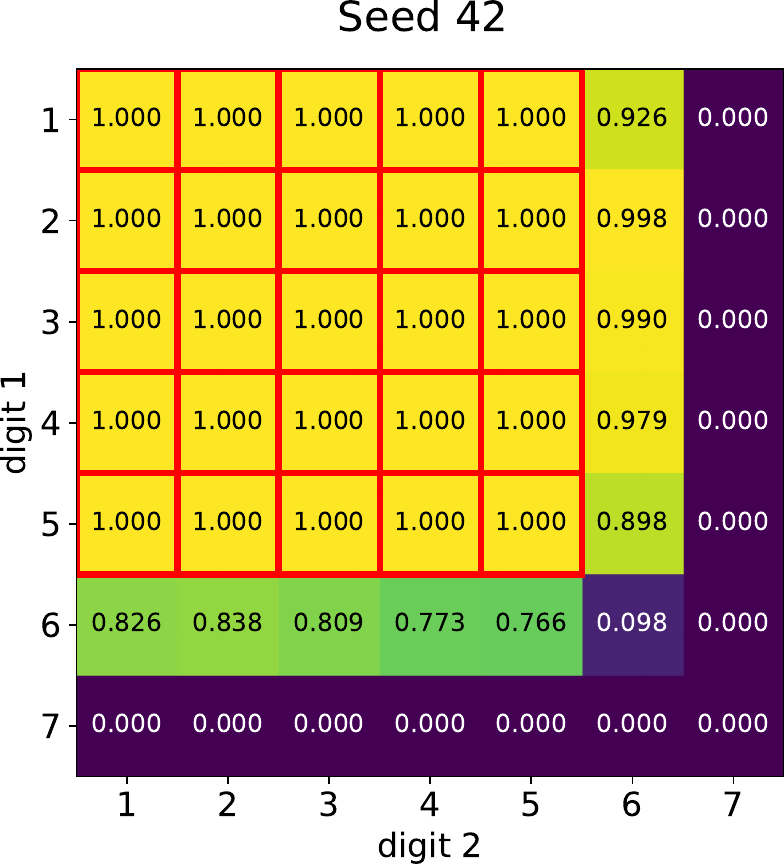}
    \includegraphics[width=0.3\linewidth]{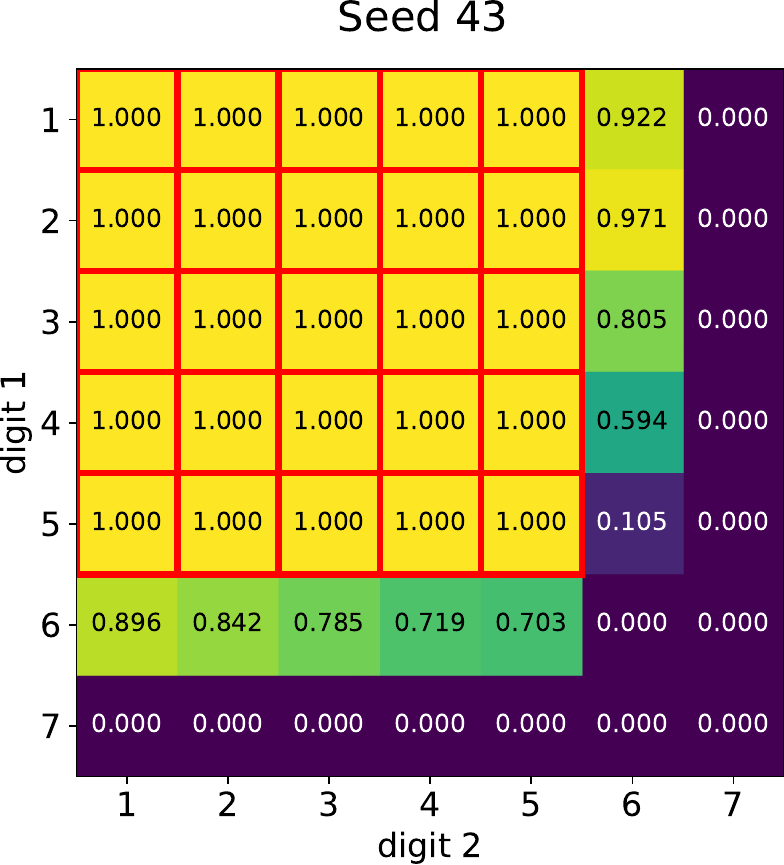}\\
    \includegraphics[width=0.3\linewidth]{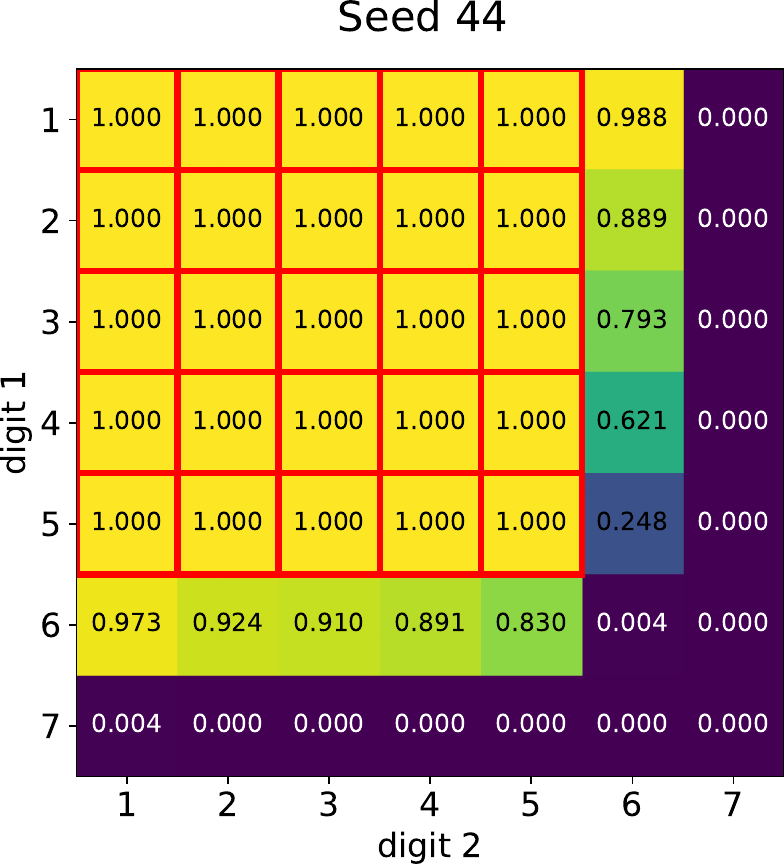}
    \includegraphics[width=0.3\linewidth]{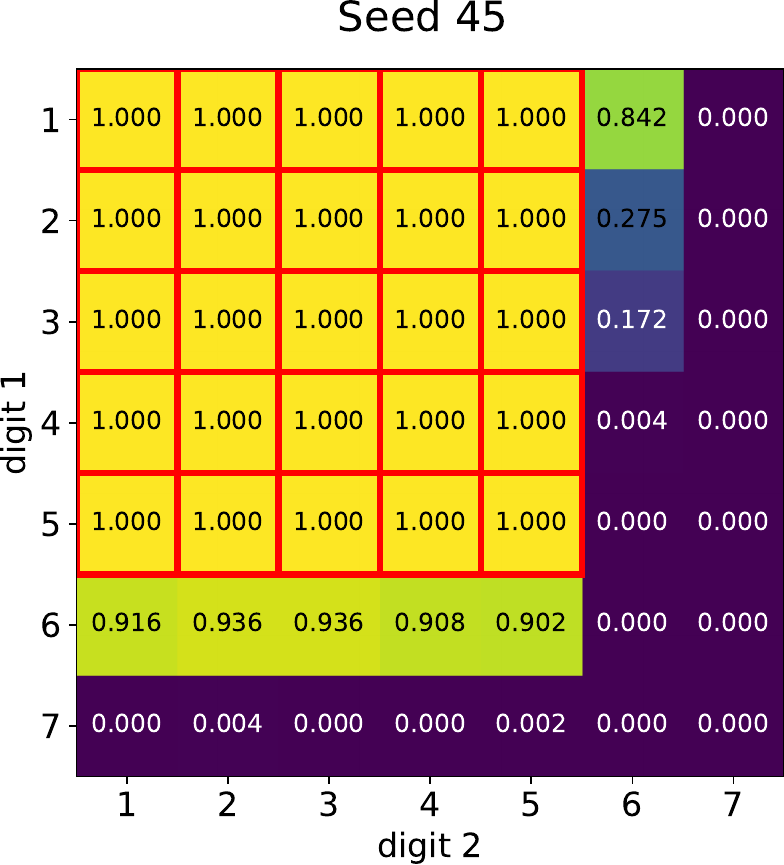}
    \end{minipage}
    }
    \hspace{1mm}
    \caption{Test accuracy on 5 different seeds during the initial training phase. Models exhibit high variance in performance.  }
    \label{fig:mult_has_high_variance1}
\end{figure}

\begin{figure}[ht!]
    \centering
    \resizebox{0.8\linewidth}{!}{
    \begin{minipage}{\linewidth}    \centering
    \includegraphics[width=0.3\linewidth]{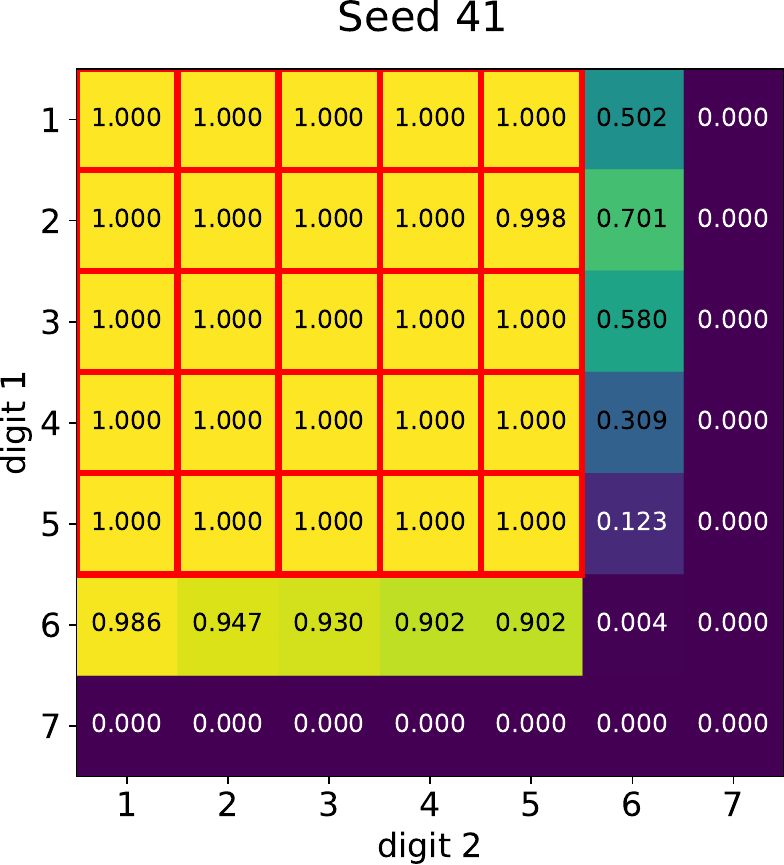}
    \includegraphics[width=0.3\linewidth]{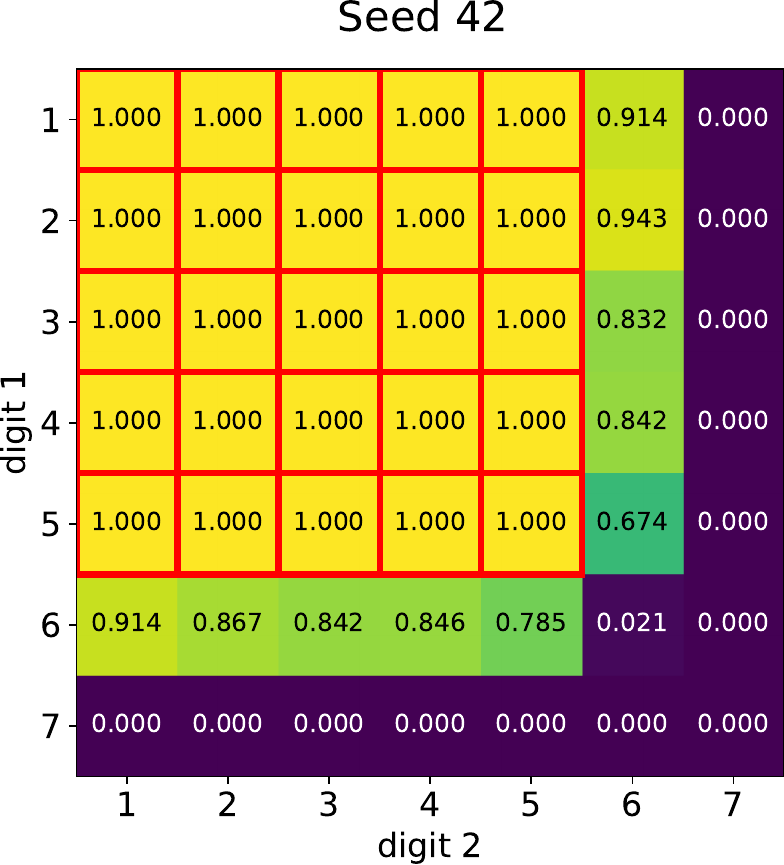}
    \includegraphics[width=0.3\linewidth]{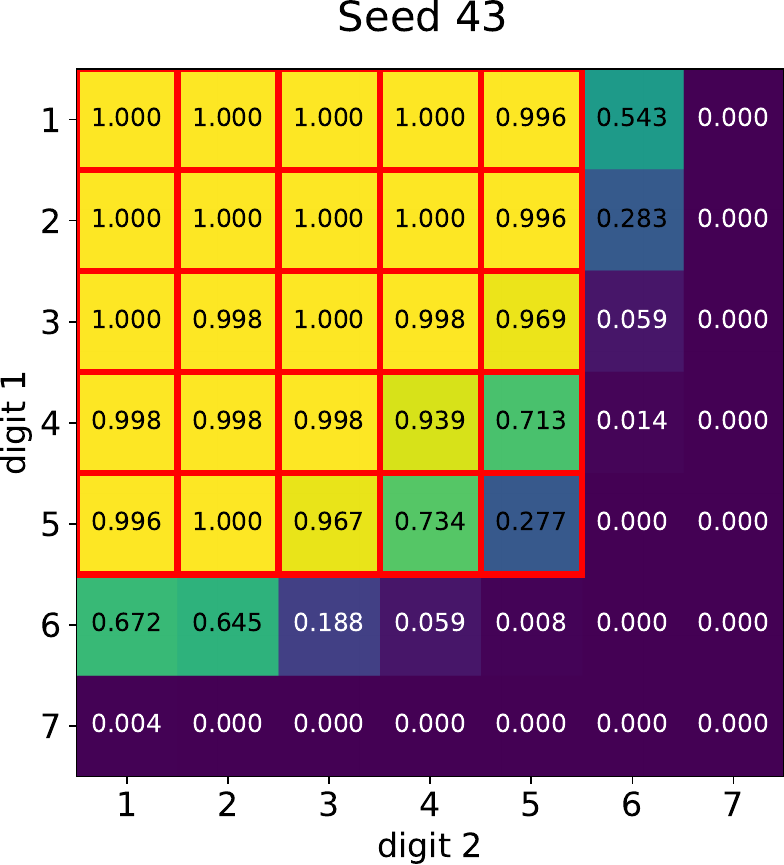}
    \includegraphics[width=0.3\linewidth]{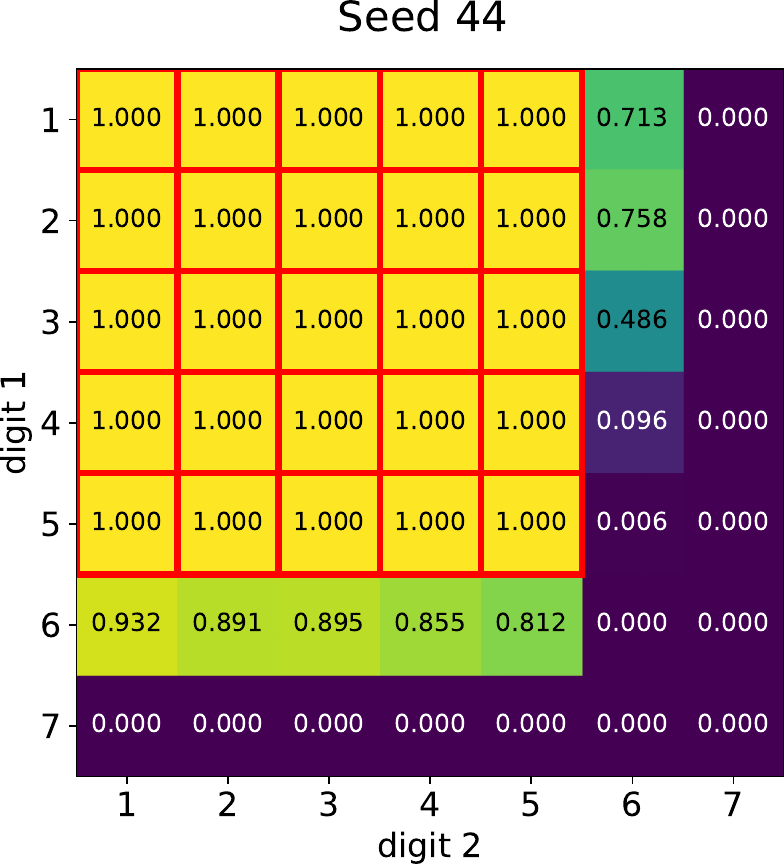}
    \includegraphics[width=0.3\linewidth]{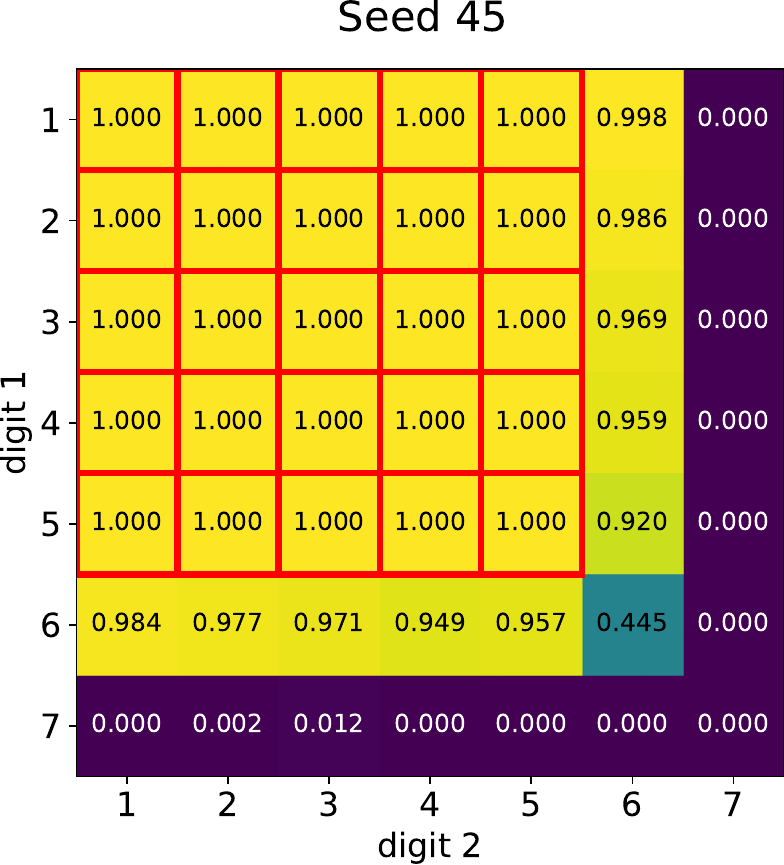}
    \end{minipage}
    }
    \hspace{1mm}
    \caption{Test accuracy on models trained with the same seed data but different training seeds. Despite identical training data, models exhibit large variability. }
    \label{fig:mult_has_high_variance2}
\end{figure}

To illustrate this variability, Figure~\ref{fig:mult_has_high_variance1} shows test accuracy across five models trained with different random seeds on the initial training dataset containing up to 5-by-5 multiplication. Even when trained on identical training data, models exhibit substantial performance differences in extrapolation. Similarly, Figure~\ref{fig:mult_has_high_variance2} demonstrates that this variability persists even when models are trained from the same seed data.

\subsubsection{Ablations for Majority Voting}\label{sec:mv_ablations}

Our majority voting method requires training multiple models in parallel. In our primary setting, we train $k$ models with different random seeds, allowing each to generate and train on its own independent self-improved dataset at every round. 

To evaluate the necessity of training multiple independent models and generating separate self-improvement datasets, we compare our approach against the following baselines:
\begin{enumerate}
    \item No majority voting, but larger self-improve data: Instead of using multiple models, we train a single model while sampling $k$ times more self-improve data per round, ensuring that the total amount of generated data matches our main setting.
    \item Shared self-improve data: We train $k$ models with different initial seeds but subsequently train all models on the same self-improved dataset.
    \item Shared initial training seed: All models are initialized from the same seed but then trained on separate self-improved datasets.
    \item Our main setting: Each model is initialized with a different seed and trained on its own independently generated self-improve dataset.
\end{enumerate}

\begin{figure}[ht!]
    \centering
    \includegraphics[width=0.24\linewidth]{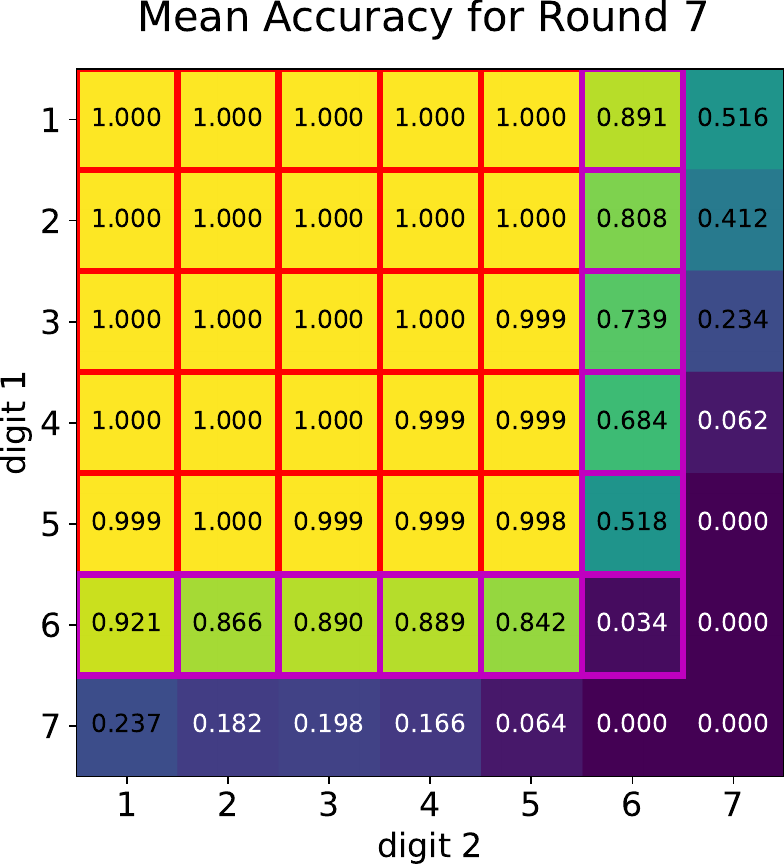}
    \includegraphics[width=0.24\linewidth]{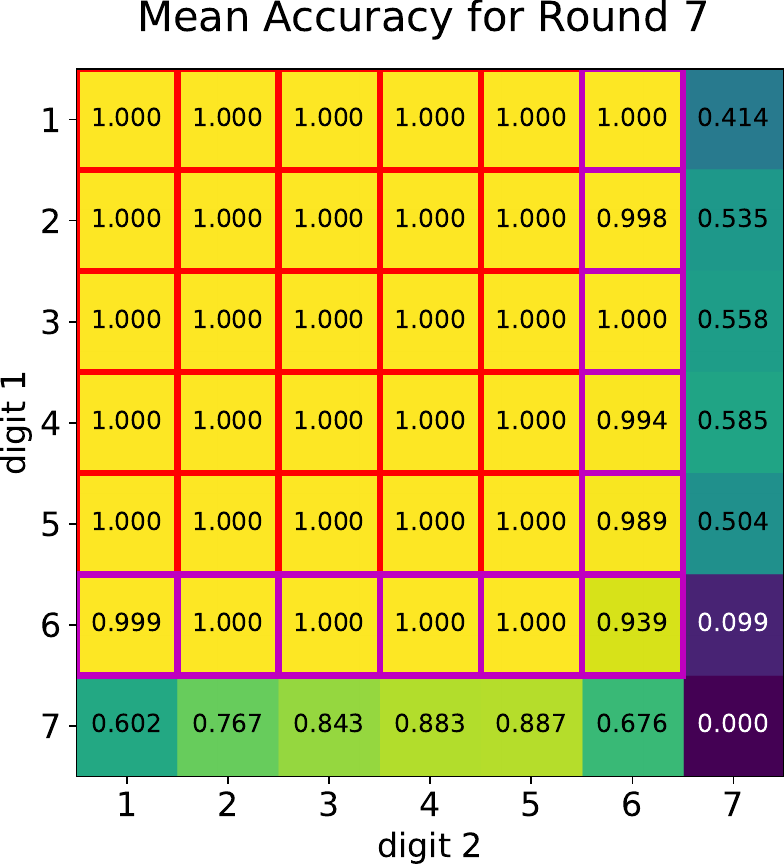}
    \includegraphics[width=0.24\linewidth]{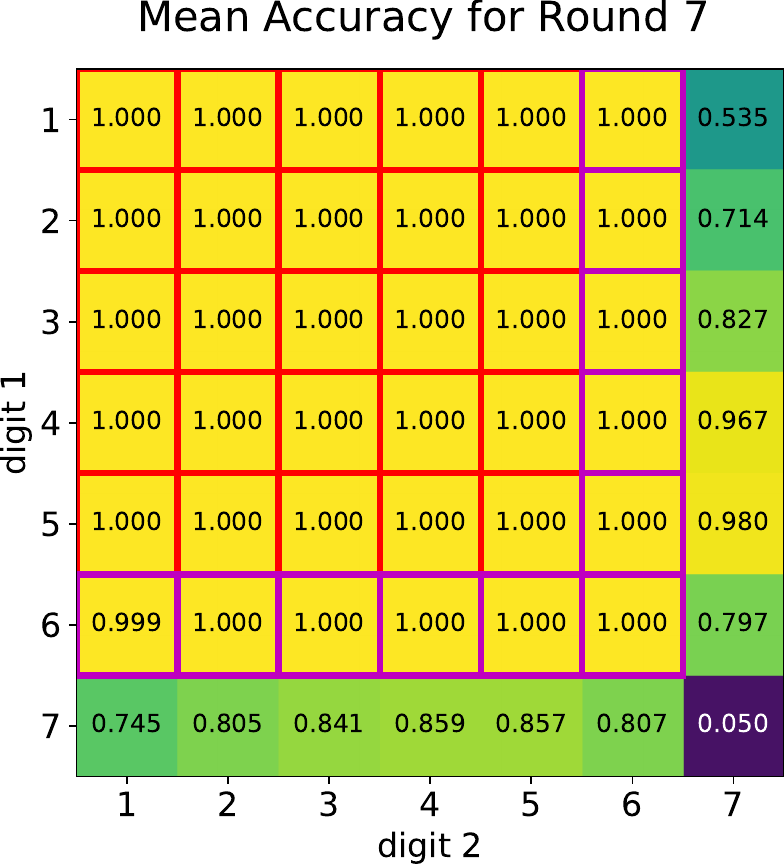}
    \includegraphics[width=0.24\linewidth]{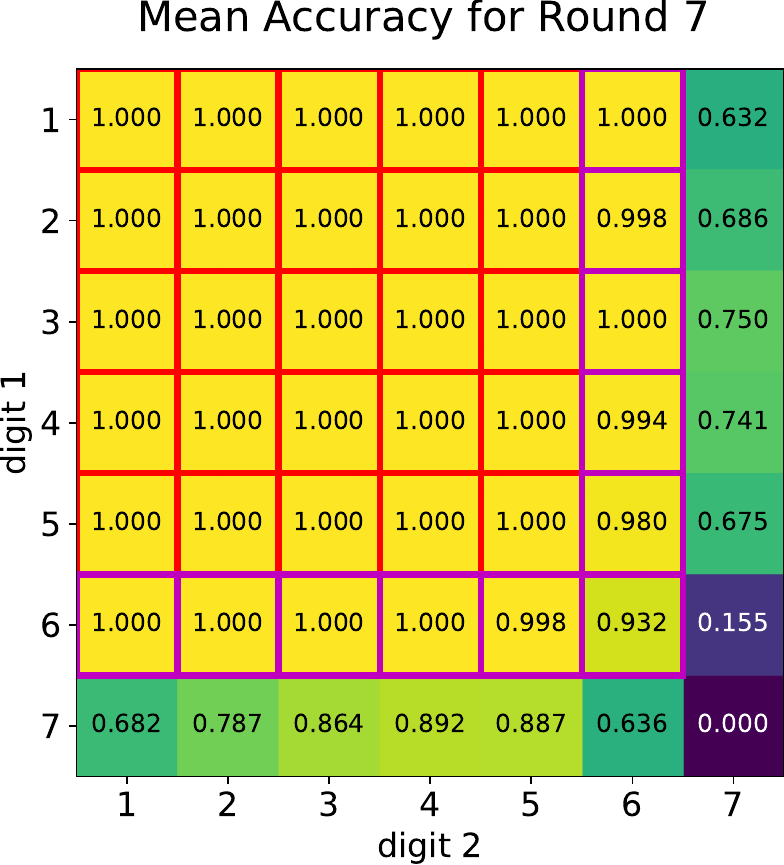}
    \caption{Ablations on majority voting. (Left) No majority voting, but larger self-improve data. (Left-Center) Majority voting with shared self-improve data. (Right-Center) Majority voting with shared initial training seed. (Right) Our primary setting with fully independent training and self-improve datasets.}
    \label{fig:mv_baseline}
\end{figure}

Figure~\ref{fig:mv_baseline} presents the performance of these variations, highlighting the importance of training on independently generated self-improve datasets rather than simply increasing dataset size or sharing training trajectories across models.

\begin{table}[ht!]
    \centering
    \caption{Comparison of Data Cost Across Majority Voting Variants}
    \footnotesize
    \small
    \setlength{\tabcolsep}{4pt}
    \renewcommand{\arraystretch}{0.5}
    {

    \label{tab:mv_data_cost}
    \begin{tabular}{lcc}
        \toprule
        \textbf{Method} & \textbf{Initial Training Data Cost} & \textbf{Self-Improve Data Cost (Per Round)} \\
        \midrule
        No Majority Voting, Larger Data & 1 & $k$ \\
        Shared Self-Improve Data & $k$ & 1 \\
        Shared Initial Training Seed & 1 & $k$ \\
        Full Majority Voting (Ours) & $k$ & $k$ \\
        \bottomrule
    \end{tabular}
    }
\end{table}

We set $k=5$ and report the average performance across five models. Figure~\ref{fig:mv_baseline} shows that simply increasing the amount of self-improvement data without filtering leads to poor performance. Surprisingly, using $5\times$ more self-improvement data per round performs even worse than using less data (Figure~\ref{fig:multiplication_vanilla}), consistent with our findings in Section~\ref{sec:si_data_size}.

Additionally, majority voting with shared self-improve data (second panel from the left) underperforms in OOD compared to models trained on separate self-improve datasets. This suggests that model diversity—enabled by training on different self-improve data—may be important for majority voting to be effective. 

On the other hand, comparing the right two panels in Figure~\ref{fig:mv_baseline}, where the difference lies in whether the base models were trained on different labeled data $\mathcal{D}_0$, we find minimal differences in OOD performance. This may be due to the large size of the initial training set (5M examples), which provides sufficient diversity. Furthermore, as Figure~\ref{fig:mult_has_high_variance2} shows, models trained on the same initial dataset but with different training seeds still exhibit substantial variability, suggesting that model diversity can emerge from different training trajectories alone.

\subsection{Additional Error Analysis on Reverse Addition}\label{sec:appdx_error_patterns}

\subsubsection{Patterns in Model Mistakes. } 

\begin{figure}[ht!]
    \centering
    \includegraphics[width=0.4\linewidth]{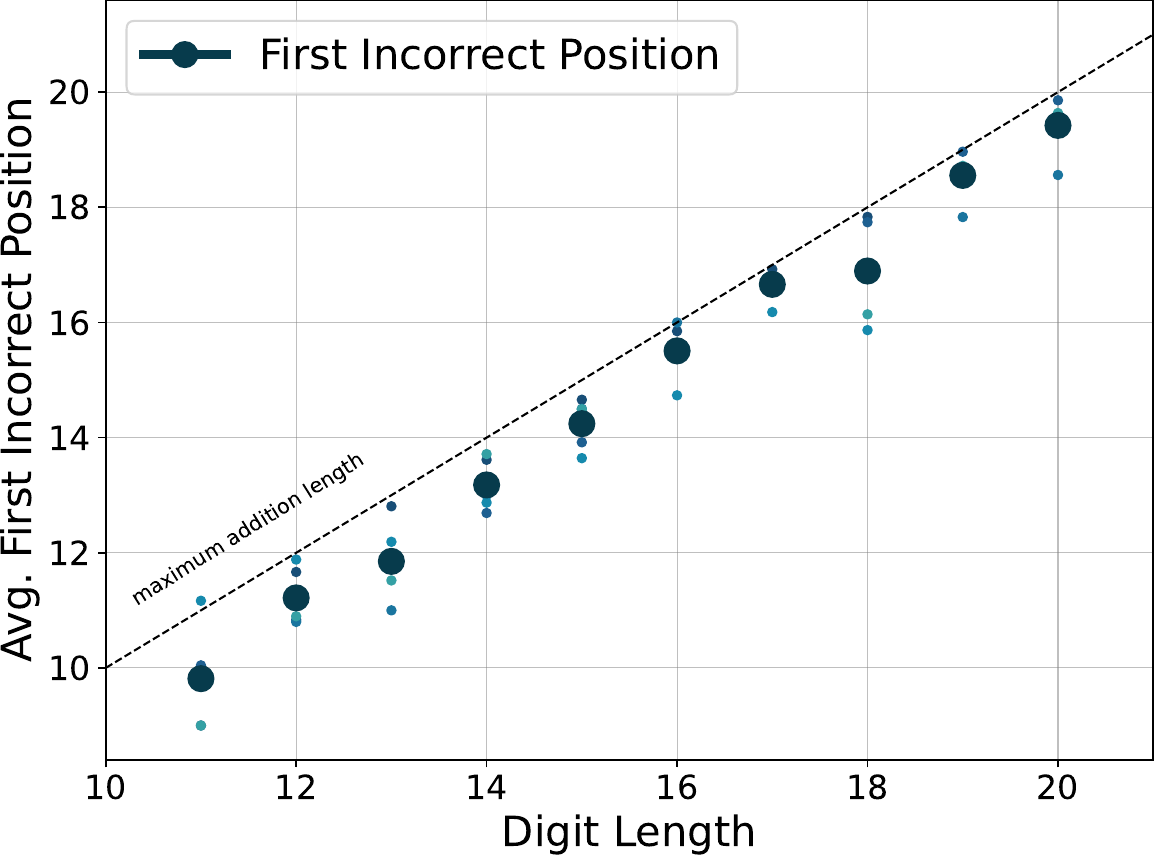}
    \caption{he first incorrect digit in model outputs tends to occur near the most significant digit in reverse addition.}
    \label{fig:first_wrong_location}
\end{figure}

Figure~\ref{fig:first_wrong_location} shows that when models generate incorrect answers, the first mismatch with the ground truth typically occurs near the final digits of the sequence (i.e., near the most significant digit in reverse addition). These observations inform our systematic error simulations, which are used to analyze the error avalanche phenomenon in Section~\ref{sec:error_analysis}.

\subsection{Additional Experiments on Label Noise and Robustness}\label{sec:appdx_label_noise}

\paragraph{Robustness against Random Labels. }
To further examine the model’s resilience to errors in data, we introduce randomization into the labels during training. Correct labels are replaced with random numbers of the \textit{same length} with probabilities 1, 0.8, 0.5, 0.2, 0.1, and 0. A probability of 1 corresponds to entirely incorrect labels, while 0 indicates fully correct data.

The model is initially trained on 1-10 digit reverse addition and further trained across 8 self-improvement rounds, using self-generated data of lengths 11-18 digits. We then construct a dataset of 19-digit data with randomized labels, denoted as $\mathcal{D}^\text{rand}_9$. The model is fine-tuned on a combined dataset consisting of the original dataset $\mathcal{D}_0$, self-improved datasets $\mathcal{D}_1, \dots, \mathcal{D}_8$, and $\mathcal{D}^\text{rand}_9$.

\begin{figure}[ht!]
    \centering
    \includegraphics[width=0.4\linewidth]{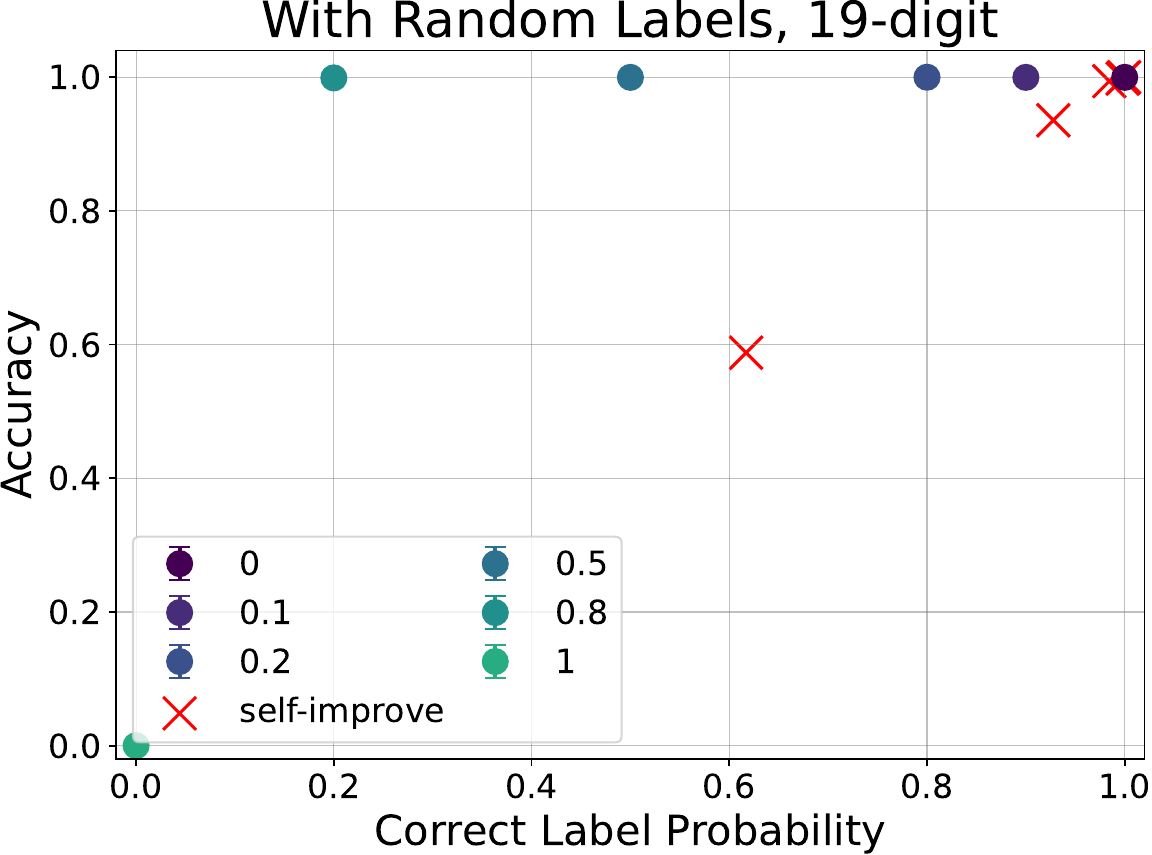}
    \hspace{1mm}
    \includegraphics[width=0.4\linewidth]{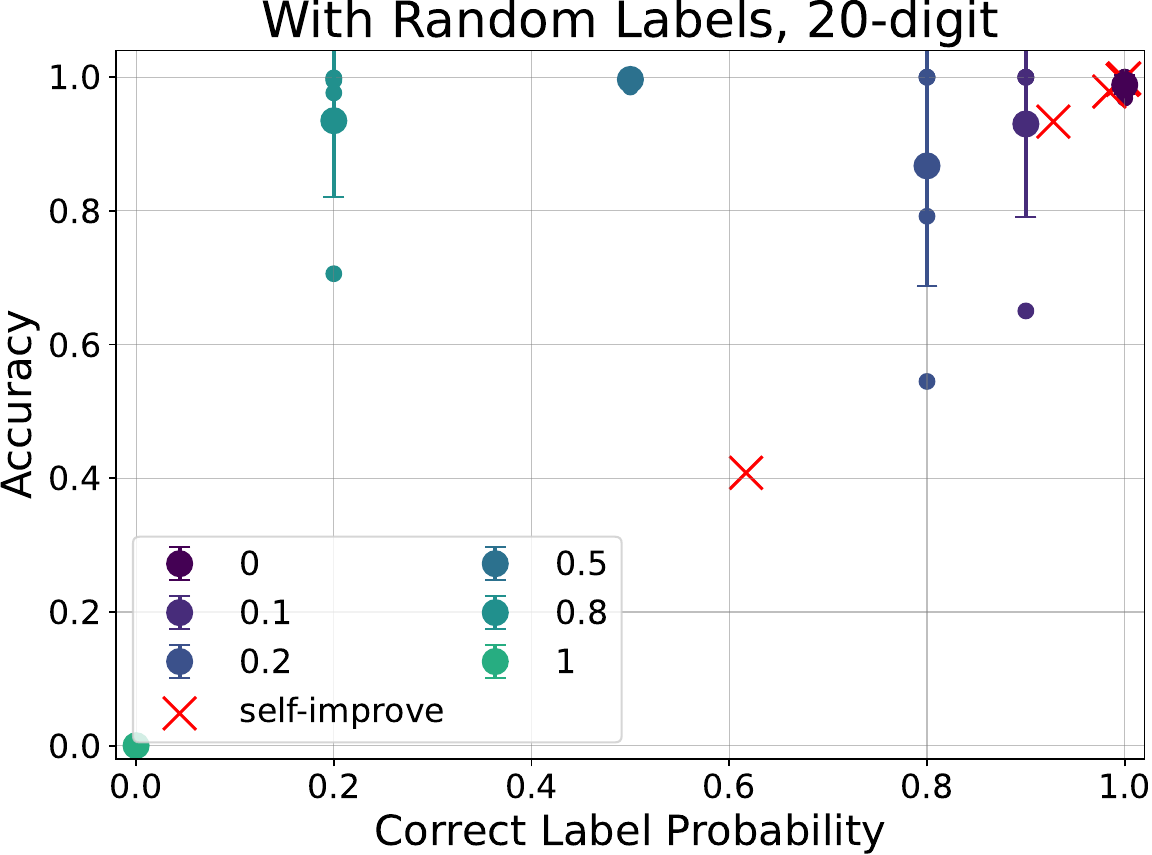}
    \caption{Effect of training on randomized labels. The model is trained on 1-10 digit data, further fine-tuned on 11-18 digit self-generated data over 8 self-improvement rounds, and additionally fine-tuned on 19-digit data with varying probabilities of random label replacement. (Left) Accuracy on 19-digit data. (Right) Accuracy on 20-digit data. The results demonstrate that while the model can self-correct random errors, biases from self-improved data can result in worse performance compared to models trained on random-labeled data of similar accuracy.}
    \label{fig:random_label}
\end{figure}

Results in Figure~\ref{fig:random_label} show that the models can tolerate substantial random label noise, maintaining robust performance even when up to 80\% of the training data is corrupted. This demonstrates the model’s resilience to random errors in the training data and its ability to self-correct such mistakes during learning.

\paragraph{Model Bias vs. Random Labels.}  
Interestingly, biases in self-generated data are more detrimental than uniformly random label noise.  As shown in Figure~\ref{fig:random_label}, models trained with self-improved data perform worse than random-labeled data of comparable accuracy, given the same dataset size and fine-tuning steps. 
This suggests that the inherent biases in self-generated data hinder generalization more than randomly introduced noise.

\section{Experimental Setup}\label{sec:exp_setup}

\subsection{Model}

For all experiments, we use a Decoder-only Transformer architecture. Specifically, for all experiments except for pretrained models settings, we use the Llama architecture~\citep{llama3modelcard}, except we remove the rotary positional encoding. 
For the inputs format, we have one example per line, and stack all example on the batch dimension. Since the examples can have variable length, we pad each line on the right to the maximum length in the batch. 
We exclusively use a character level tokenizer. For pretrained models, we replace the default tokenizer with our character tokenizer, while keeping the embedding component of the pretrained model unchanged. 

\begin{table}[ht!]
\caption{Model Parameters}
\footnotesize
\vspace{1mm}
\centering
\small
\setlength{\tabcolsep}{4pt} %
\renewcommand{\arraystretch}{0.5}
     {
        \begin{tabular}{ccccc} %
        \toprule
        Model & Self-Attn Layers & Num Heads & Embedding Dim \\
        \midrule 
        From-Scratch &  6 &  6 &  384 \\
        Llama 3 1B &  24 &  16 &  1024 \\
        Llama 3 3B &  32 &  32 &  2048 \\
        \bottomrule
        \end{tabular}
    }
\label{table:model_config}
\end{table}

\subsection{Data Formats and Data Sampling}\label{sec:data_gen}
\subsubsection{Maze-Solving}\label{sec:data_gen-maze}

\begin{figure}[ht!]
    \centering
    \includegraphics[width=0.30\linewidth]{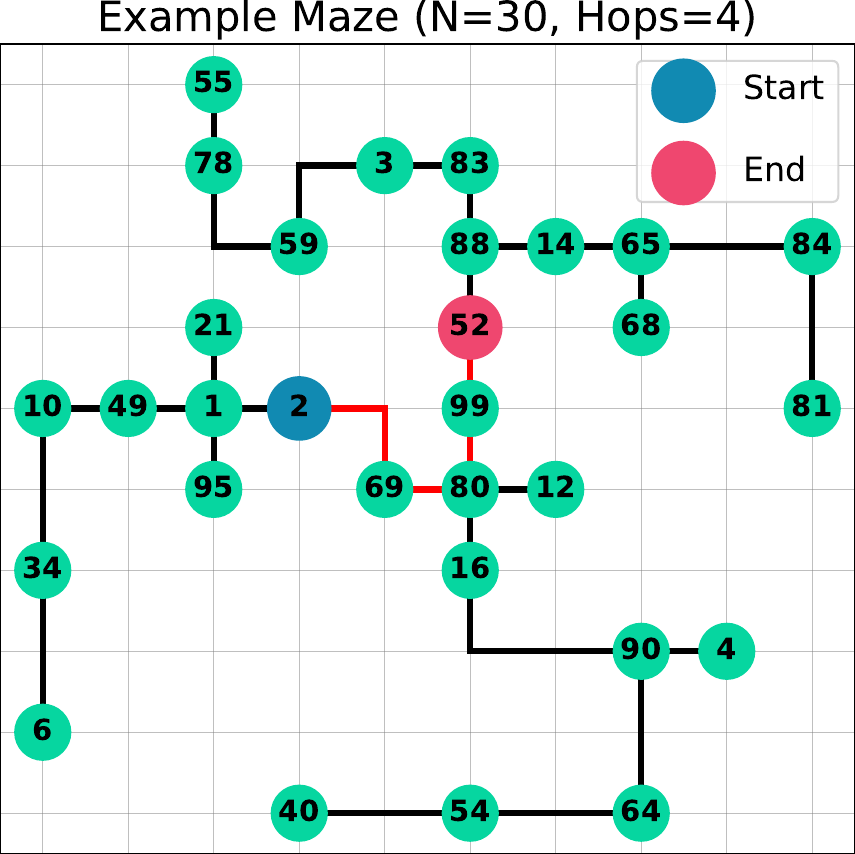}
    \hspace{1mm}
    \includegraphics[width=0.30\linewidth]{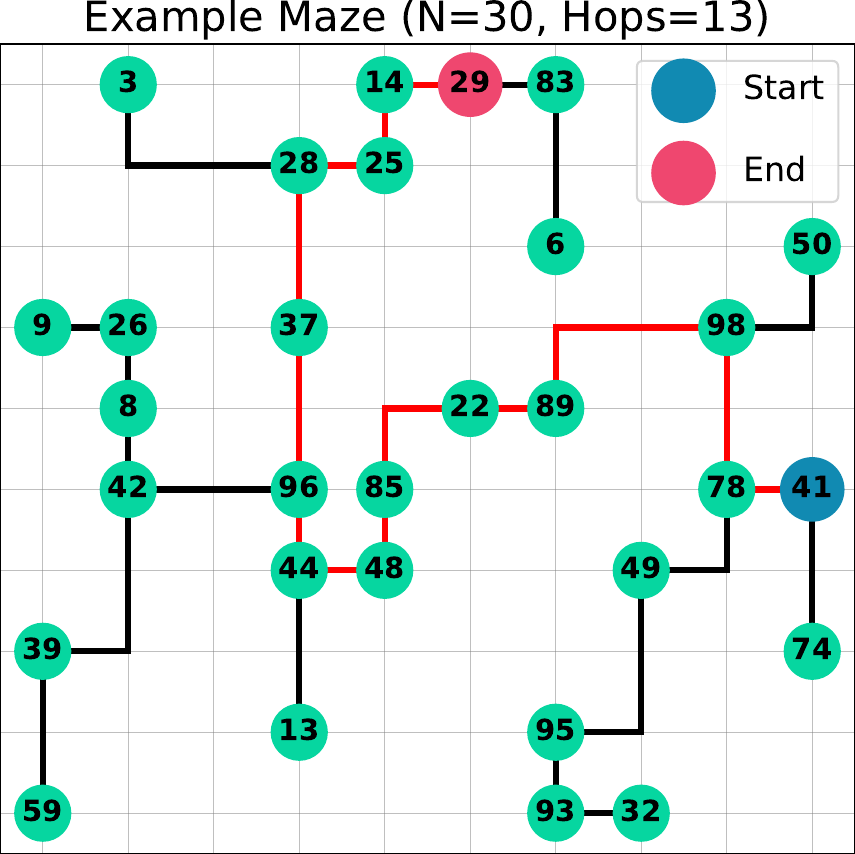}
    \hspace{1mm}
    \includegraphics[width=0.33\linewidth]{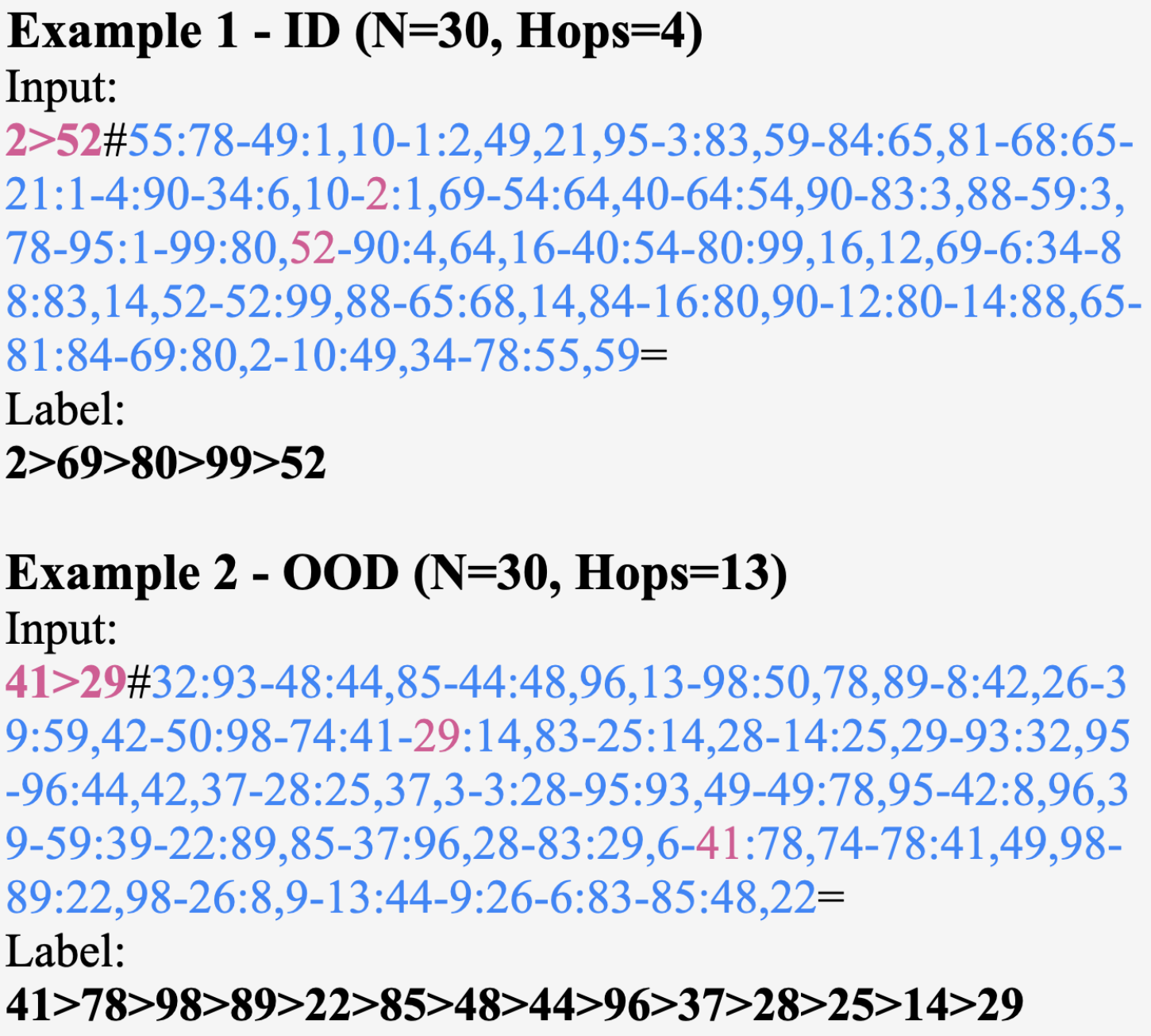}
    \caption{Maze-solving task with \( N=30 \) nodes. (Left \& Middle) Visualization of the maze task with 4 hops (ID) and 13 hops (OOD). (Right) Example of the data format: the input specifies the start and end nodes along with the graph structure, and the output lists the shortest path as hops. The labeled training dataset includes paths of up to 9 hops, with difficulty increased by adding one hop in each subsequent round.}
    \label{fig:maze_data}
\end{figure}

\begin{listing}
\begin{minted}[
framesep=2mm,
baselinestretch=0.9,
bgcolor=LightGray,
fontsize=\scriptsize,
linenos
]{python}
def create_tree_with_hops_wilson(total_nodes, num_hops):
    import networkx as nx

    # Step 1: Create the main path with num_hops
    graph = nx.path_graph(num_hops + 1) 

    # Step 2: Add extra nodes to the tree with random walk
    current_nodes = list(graph.nodes())
    new_nodes = list(range(num_hops + 1, total_nodes))

    while new_nodes:
        new_node = new_nodes.pop()
        # random walk to reach graph
        walk = [new_node]
        while walk[-1] not in current_nodes:
            # choose random node from current & new nodes
            random_node = random.choice(current_nodes + new_nodes)
            walk.append(random_node)
            if random_node in new_nodes:
                new_nodes.remove(random_node)
        # add edges
        for i in range(len(walk) - 1):
            graph.add_edge(walk[i], walk[i + 1])
        current_nodes.append(new_node) 

    # Step 3: Set the start and end nodes for the main path
    start_node = 0
    end_node = num_hops

    return graph, start_node, end_node

def format_graph(graph, start_node, end_node):
    # Assign random labels to nodes
    node_labels = assign_labels(graph.nodes(), label_range=(1, 99))
    
    # Get the shortest path (in terms of edge count) from start_node to end_node
    shortest_path = nx.shortest_path(graph, source=start_node, target=end_node)

    # Format the path as a string
    path_labels = [node_labels[node] for node in shortest_path]
    path_string = ">".join(map(str, path_labels))
    
    # Format start and end nodes
    start_label = node_labels[start_node]
    end_label = node_labels[end_node]
    start_end_str = f"{start_label}>{end_label}#"

    # Build graph_str with end_node connections at the end
    graph_str = ""
    start_node_str = ""  # Temporary storage for the start_node part
    end_node_str = ""  # Temporary storage for the end_node part
    
    # randomize the order of nodes
    random_nodes = list(graph.nodes())
    random.shuffle(random_nodes)
    for node in random_nodes:
        node_label = node_labels[node]
        # randomize the order of neighbors
        random_neighbors = list(graph.adj[node])
        random.shuffle(random_neighbors)
        neighbor_labels = [node_labels[neighbor] for neighbor in random_neighbors]
        graph_str += f"{node_label}:" + ",".join(map(str, neighbor_labels)) + "-"

    # Combine everything, placing the end_node last
    graph_str = start_node_str + graph_str + end_node_str

    return start_end_str + graph_str[:-1] + "=", path_string, node_labels
    
\end{minted}
\caption{Code for the maze format generation used}
\label{listing:code_maze_generation}
\end{listing}

\newpage
\subsection{Experimental Settings}\label{sec:appendix_setting}

\subsubsection{Hyperparameter Configurations}

In this section, we provide a detailed overview of the hyperparameter configuration used in our experiments in Table~\ref{table:hyperparam_base} and \ref{table:hyperparam_si}. To enhance memory efficiency and training speed, we employ flash attention and tf32, bfloat16. Our experiments are run using PyTorch 2.4 and CUDA 12.1. Detailed dependencies are provided in our github repository\footnote{\url{https://github.com/JackCai1206/arithmetic-self-improve/}}. We use Warmup stable decay~\citep{wen2024understanding} as the learning rate schedule. In table~\ref{table:hyperparam_base} and ~\ref{table:hyperparam_si}, the number of constant LR steps is equal to the total training steps minus the sum of warmup and decay steps. We use AdamW optimizer with betas (0.9, 0.99) and epsilon $1e-12$. Weight decay is fixed to 0.1 and we do not use dropout. 

Table~\ref{table:hyperparam_base} shows the training hyperparameters for the initial training phase on labeled data $D_0$. Table~\ref{table:hyperparam_si} shows the hyperparameters for each the self-improve training rounds on $D_{1,\dots,R}$. 

\begin{table}[ht]
\caption{Hyperparameters for initial training on labeled data}
\footnotesize
\vspace{1mm}
\centering
\small
\setlength{\tabcolsep}{4pt} %
\renewcommand{\arraystretch}{0.5}
{
\begin{tabular}{cccccc}
\toprule
Task & Batch Size & LR & Iterations & Warmup Iter & Decay Iter \\
\midrule 
Reverse Addition &  1024 &  5e-4 & 10000 & 1000 & 2000  \\
Reverse Addition (Llama 3 3B) &  128 &  1e-4 & 1200 & 120 & 600  \\
Reverse Addition (Llama 3 1B) &  128 &  1e-4 & 1200 & 120 & 600  \\
Copy/Reverse & 1024 &  5e-4 & 5000 & 500 & 1000  \\
Forward Addition & 1024 &  5e-4 & 10000 & 1000 & 1000  \\
Multiplication & 1024 &  5e-5 & 10000 & 1000 & 2000  \\
Maze (hops) & 1024 &  5e-4 & 25000 & 2500 & 3500  \\
Maze (nodes) & 512 &  5e-4 & 12000 & 1200 & 2800  \\

\bottomrule
\end{tabular}
}
\label{table:hyperparam_base}
\end{table}

\begin{table}[ht]
\caption{Hyperparameters for self-improvement rounds}
\footnotesize
\vspace{1mm}
\centering
\small
\setlength{\tabcolsep}{4pt} %
\renewcommand{\arraystretch}{0.5}
{
\begin{tabular}{cccccc}
\toprule
Input Format & Batch Size & LR & Iterations & Warmup Iter & Decay Iter \\
\midrule 
Reverse Addition &  1024 &  5e-4 & 1500 & 0 & 1500  \\
Reverse Addition (Llama 3 3B) &  128 &  1e-4 & 600 & 0 & 600  \\
Reverse Addition (Llama 3 1B) &  128 &  1e-4 & 600 & 0 & 600  \\
Copy/Reverse & 1024 &  5e-4 & 500 & 0 & 500  \\
Forward Addition & 1024 &  5e-4 & 3000 & 0 & 1000  \\
Multiplication & 1024 &  5e-5 & 3000 & 0 & 1000  \\
Maze (hops) & 1024 &  2e-4 & 5000 & 500 & 1000  \\
Maze (nodes) & 512 &  2e-4 & 4000 & 400 & 1000  \\
\bottomrule
\end{tabular}
}
\label{table:hyperparam_si}
\end{table}

\subsubsection{Self-Improvement Setting for each Task}

\paragraph{Reverse Addition.} 
The initial supervised dataset $\mathcal{D}_0$ contains 2 million examples of reverse addition, with operand lengths ranging from 1 to 16 digits. This dataset is used to train the model for 10,000 steps. In subsequent self-improvement rounds, we sample 50,000 additional training examples at each round, extending the operand length by one digit. Specifically, at self-improvement round \( r \), the self-generated data $\mathcal{D}_r$ consists of length-\( 16+r \) examples produced by the model \( M_r \). The model is fine-tuned on the combined dataset $\mathcal{D}_0 \cup \mathcal{D}_1 \cup \dots \cup \mathcal{D}_r$ for 1,500 steps, resulting in an improved model \( M_{r+1} \).

\paragraph{String Copy \& String Reverse.} 
The initial training set $\mathcal{D_0}$ consists of 2 million examples of strings of length 1 to 10. The vocabulary of the string is the digits $0$ to $9$. For each subsequent round $r$, we sample $D_{r}$ consisting of $50,000$ examples of length $10+r$ from the model $M_r$. Then we continue training $M_r$ on the combined dataset $D_1\cup\dots\cup D_r$ for 500 steps to obtain $M_{r+1}$.

\paragraph{Forward Addition.}
The models are initially trained on a dataset $\mathcal{D}_0$ containing 2 million labeled examples of forward addition, with operand lengths ranging from 1 to 10 digits. This initial training phase spans 10,000 steps. In each subsequent self-improvement round, we generate 50,000 additional training examples, incrementally extending the operand length by one digit. Specifically, at self-improvement round \( r \), the self-generated dataset $\mathcal{D}_r$ contains length-\( 10+r \) examples produced by the model \( M_r \). The model is then fine-tuned for 3,000 steps on the combined dataset $\mathcal{D}_0 \cup \mathcal{D}_1 \cup \dots \cup \mathcal{D}_r$, resulting in an updated model \( M_{r+1} \).

\paragraph{Multiplication.}
The model is initially trained on 5 million  $n$-by-$n$ multiplication examples with $n=5$. Directly introducing $n+1$-by-$n+1$ examples results in poor performance, hence, we adopt a more fine-grained difficulty schedule. In each self-improvement round, we incrementally increase one operand by one digit, sampling $n+1$-by-$m$ and $m$-by-$n+1$ examples, where $m$ grows from 1 to $n+1$. This gradual progression allows the model to adapt incrementally to larger operand sizes, making the transition to harder examples more manageable.

For data filtering, we use the following setting:
for length filtering, we remove self-generated samples where the output length is shorter than the longest output in the batch by more than 10 tokens. This helps eliminate incorrect solutions that omit intermediate steps. For majority voting, we train five models in parallel using different random seeds and retain only those data points where at least 4 out of the 5 models produce the same output. This strategy ensures that only high-consensus, reliable data points are used for training.

\paragraph{Maze Solving - Increasing Hops.}
The model is first trained on a dataset \( \mathcal{D}_0 \) containing 5 million labeled maze-solving examples, where the number of nodes is fixed at \( N=30 \) and paths range from \( h=1 \) to \( h=9 \) hops. This initial training phase spans 25,000 steps. In subsequent self-improvement rounds, we generate 50,000 additional training examples, increasing \( h \) by 1, and fine-tune the model for 5,000 steps per round. We experiment with both unfiltered training data and majority voting, where only outputs agreed upon by all 3 models are retained.

\paragraph{Maze Solving - Increasing Nodes.}
The model is first trained on a dataset \( \mathcal{D}_0 \) containing 5 million labeled maze-solving examples, with a fixed hop count \( h=9 \) and node counts ranging from \( N=10 \) to \( N=30 \). This initial training lasts 12,000 steps. In self-improvement rounds, the number of nodes \( N \) is increased by 3 per round, generating 50,000 additional training examples at each step and fine-tuning for 4,000 steps. We compare training without filtering against majority voting, where only outputs agreed upon by all 3 models are kept.

\paragraph{Ablation Task - Pretrained Models.}
To maintain consistency in tokenization, we use character-level tokenization instead of the default tokenizer of the Llama models. We use LoRA~\citep{Hu2021LoRALA} with $r=64$ and $\alpha=128$ for Llama-1B, and $r=32$ and $\alpha=128$ for Llama-3B. 
In the initial round, we train for 1200 steps with a learning rate schedule that includes 10\% warm-up steps to a constant learning rate of \( 1\text{e-}4 \), followed by 20\% cosine decay steps to a final learning rate of \( 1\text{e-}6 \). For subsequent rounds, we train for 600 steps per round using a cosine decay learning rate schedule without warm-up, starting at \( 1\text{e-}4 \) and decaying to \( 1\text{e-}6 \).

\newpage
\section{Full Results}\label{sec:full_results}

\subsection{Results on Multiplication}\label{sec:mult_full_results}

Each figure represents the average over 5 different models.

\begin{figure}[ht!]
    \centering
    \resizebox{0.8\linewidth}{!}{
    \begin{minipage}{\linewidth}
    \centering
    \includegraphics[width=0.24\linewidth]{fig/mult/vanilla/multiplication_vanilla_1_acc.pdf}
    \includegraphics[width=0.24\linewidth]{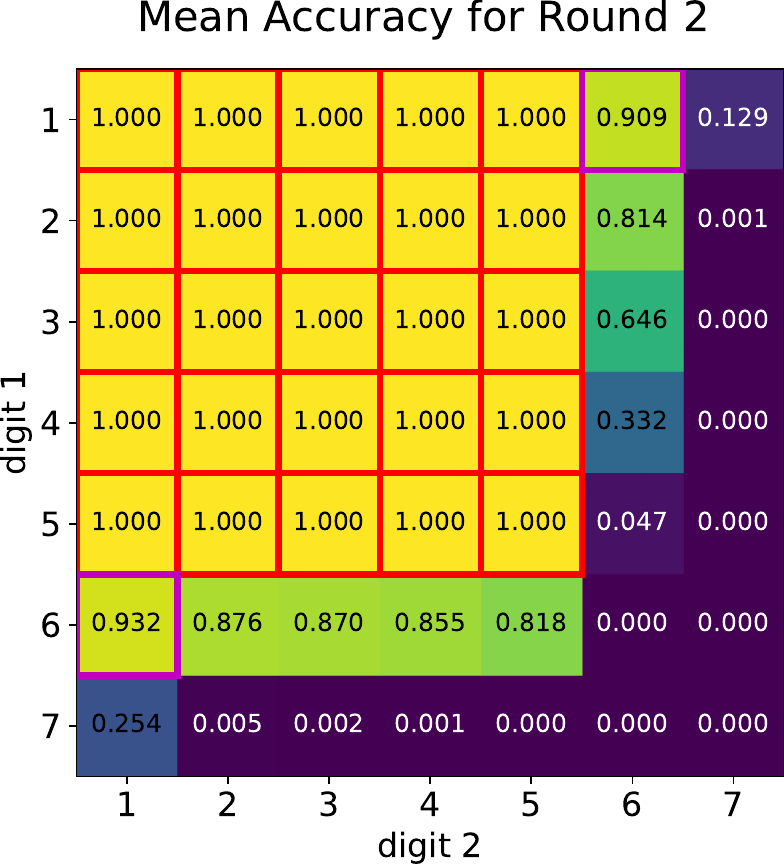}
    \includegraphics[width=0.24\linewidth]{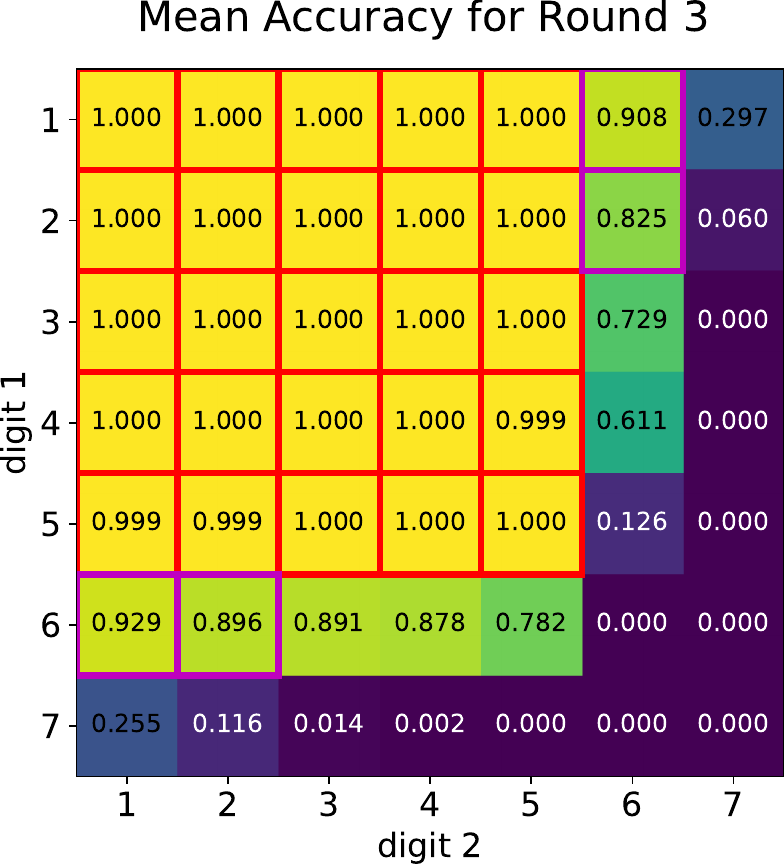}
    \includegraphics[width=0.24\linewidth]{fig/mult/vanilla/multiplication_vanilla_4_acc.pdf}
    \vspace{1mm}
    \includegraphics[width=0.24\linewidth]{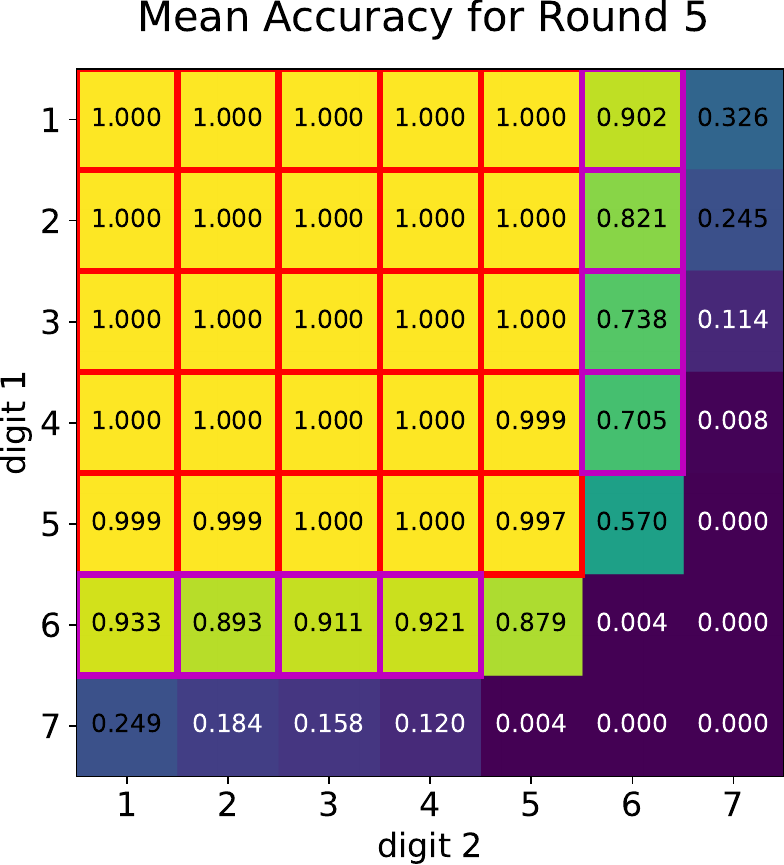}
    \includegraphics[width=0.24\linewidth]{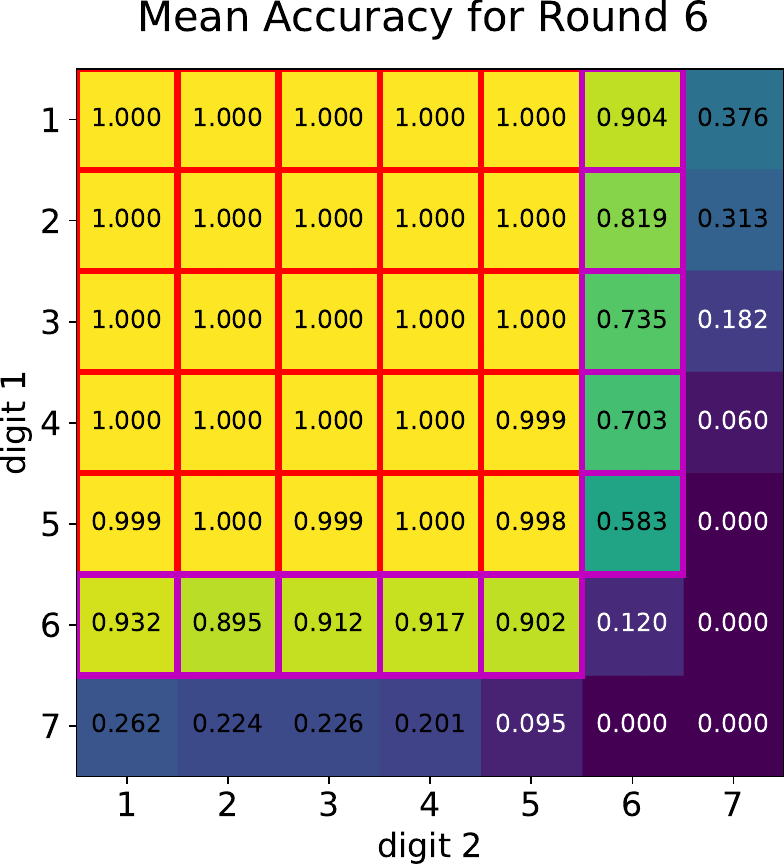}
    \includegraphics[width=0.24\linewidth]{fig/mult/vanilla/multiplication_vanilla_7_acc.pdf}  
    \end{minipage}
    }
    \caption{Results for multiplication without filtering. Each cell represents the accuracy on $n$-digit by $m$-digit multiplication. Red boxes indicate labeled in-distribution examples, while magenta boxes indicate evaluations after training on self-improved data. The model is initially trained on up to $5$-by-$5$ multiplication. Generalizing to larger multiplications is challenging without data filtering.}
    \label{fig:multiplication_vanilla_full}
\end{figure}

\begin{figure}[ht!]
    \centering
    \resizebox{0.8\linewidth}{!}{
    \begin{minipage}{\linewidth}
    \includegraphics[width=0.24\linewidth]{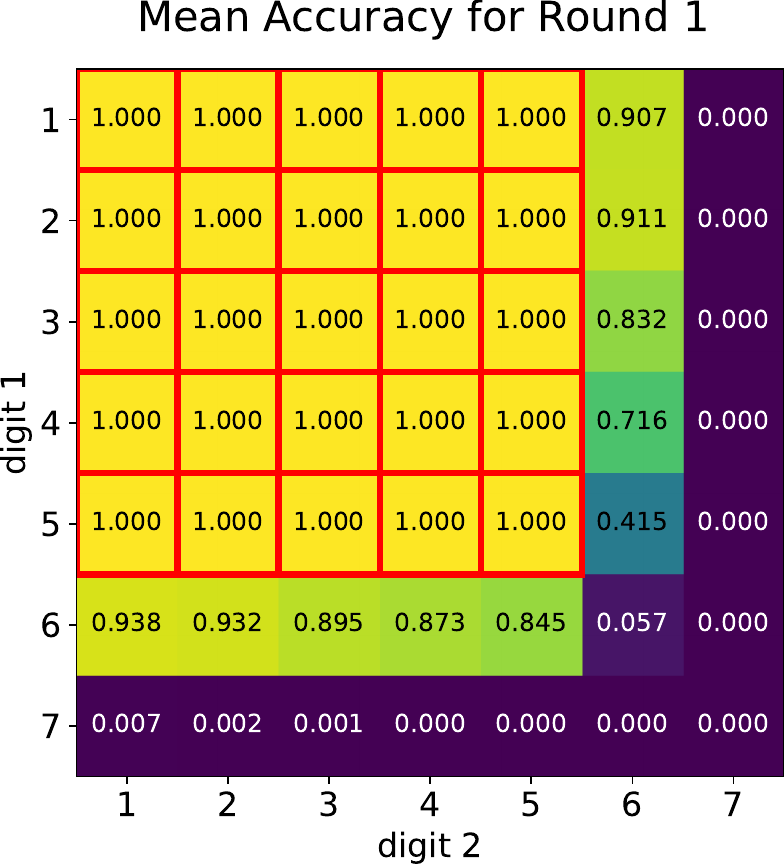}
    \includegraphics[width=0.24\linewidth]{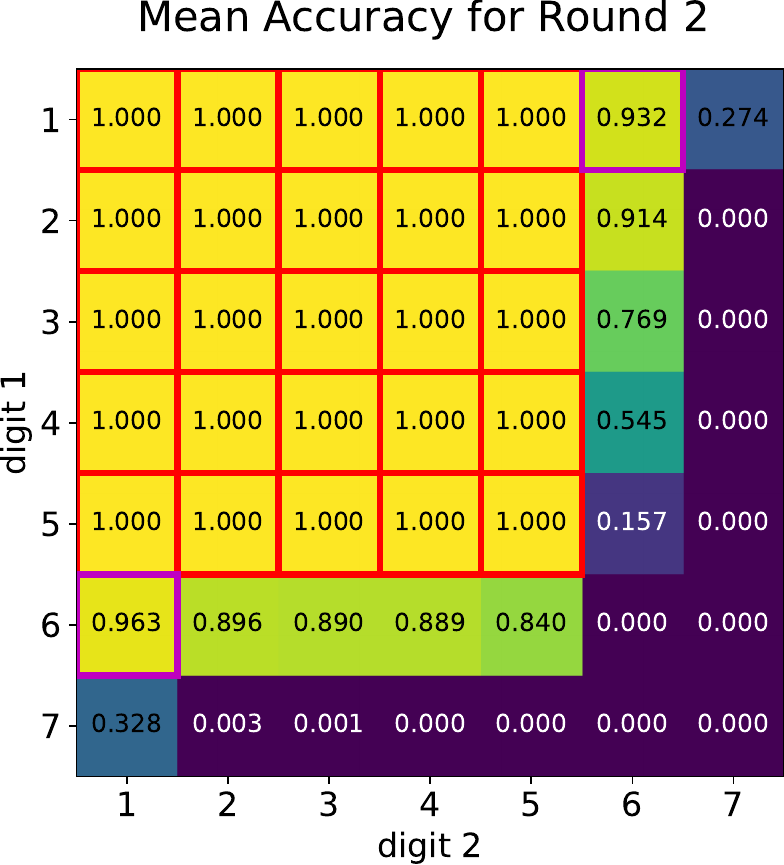}
    \includegraphics[width=0.24\linewidth]{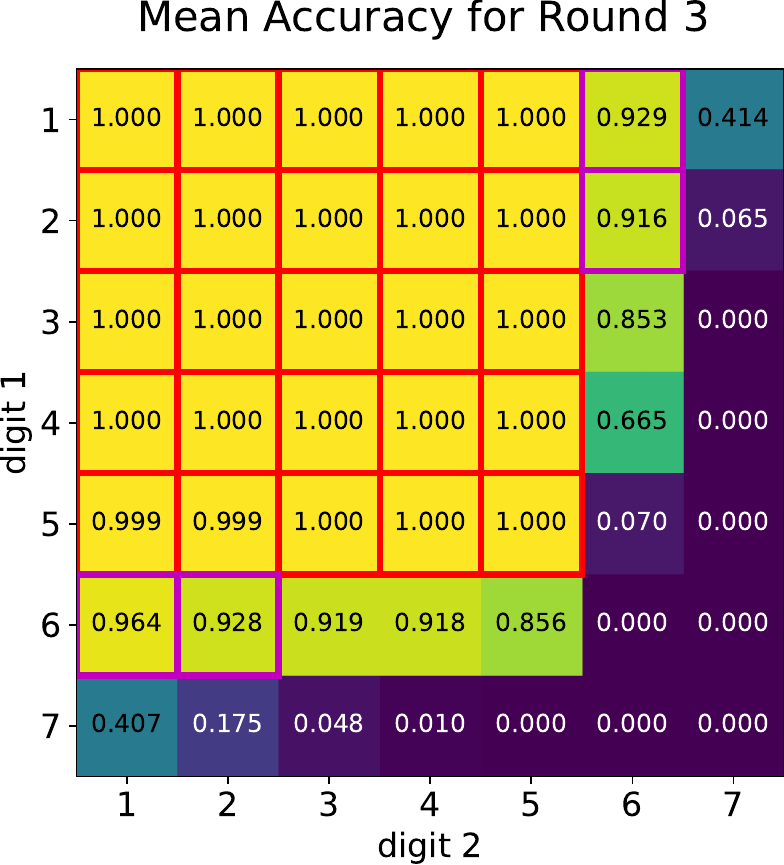}
    \includegraphics[width=0.24\linewidth]{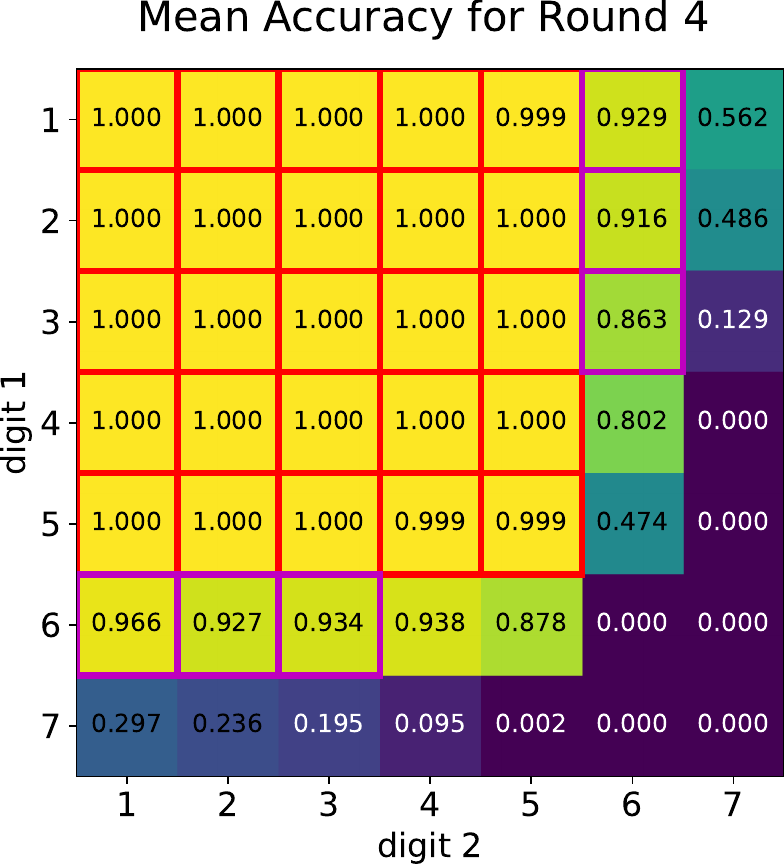}
    \includegraphics[width=0.24\linewidth]{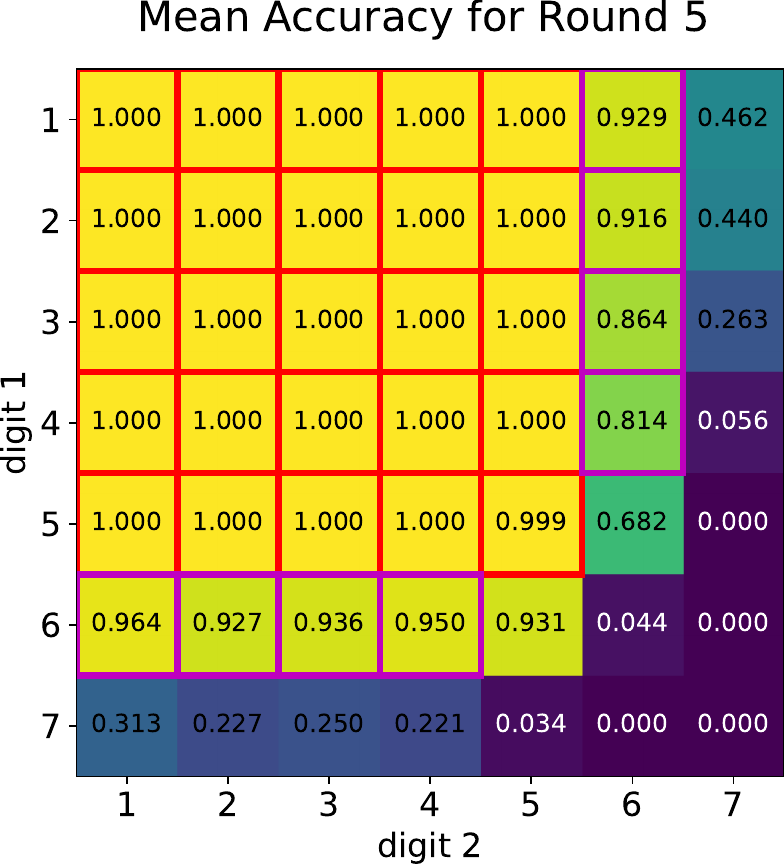}
    \includegraphics[width=0.24\linewidth]{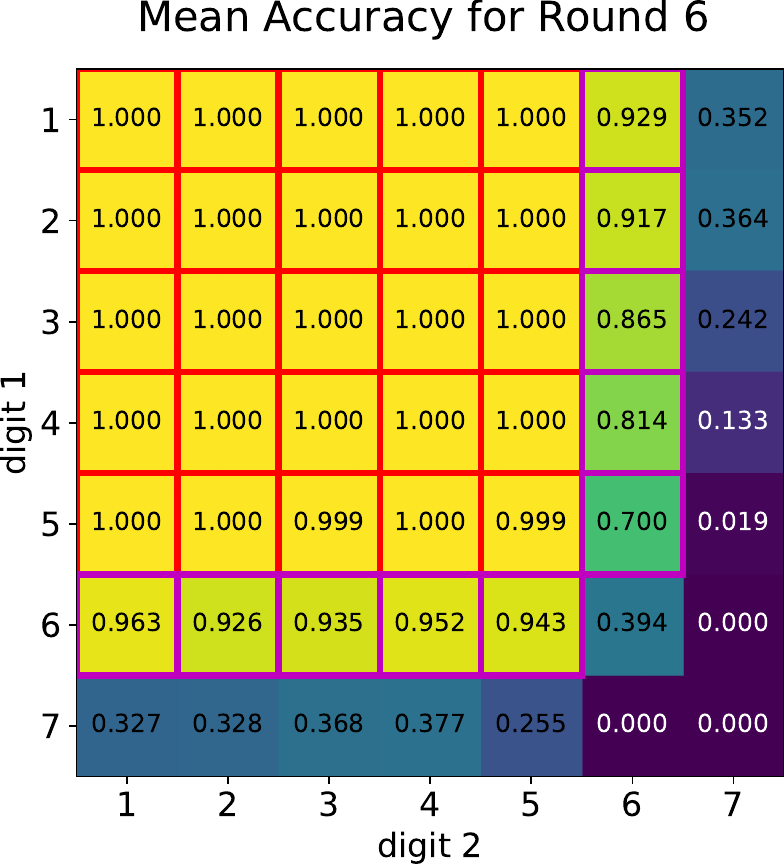}
    \includegraphics[width=0.24\linewidth]{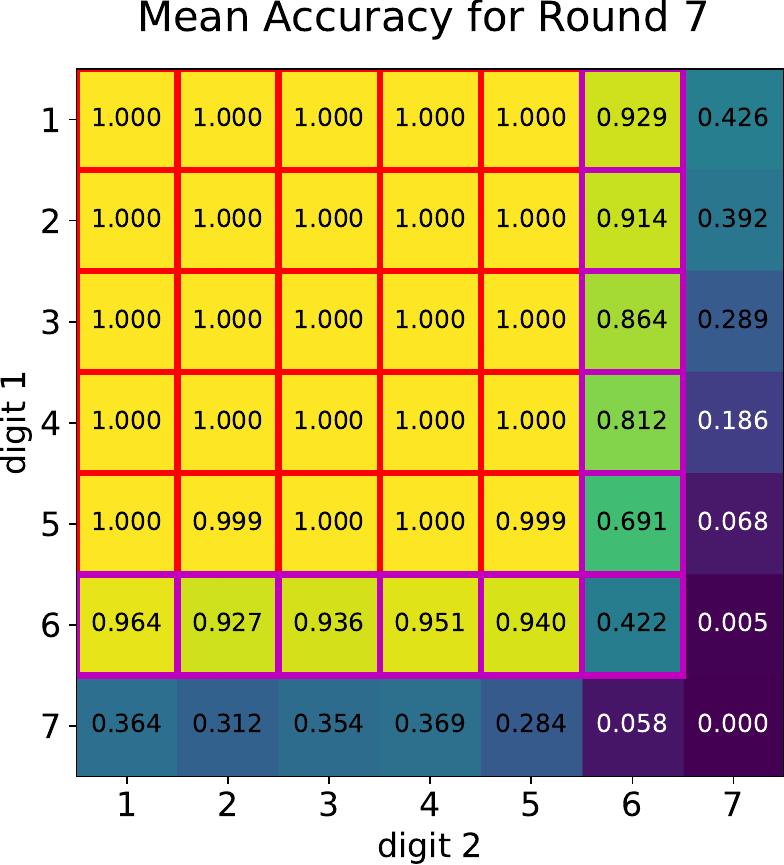}
    \includegraphics[width=0.24\linewidth]{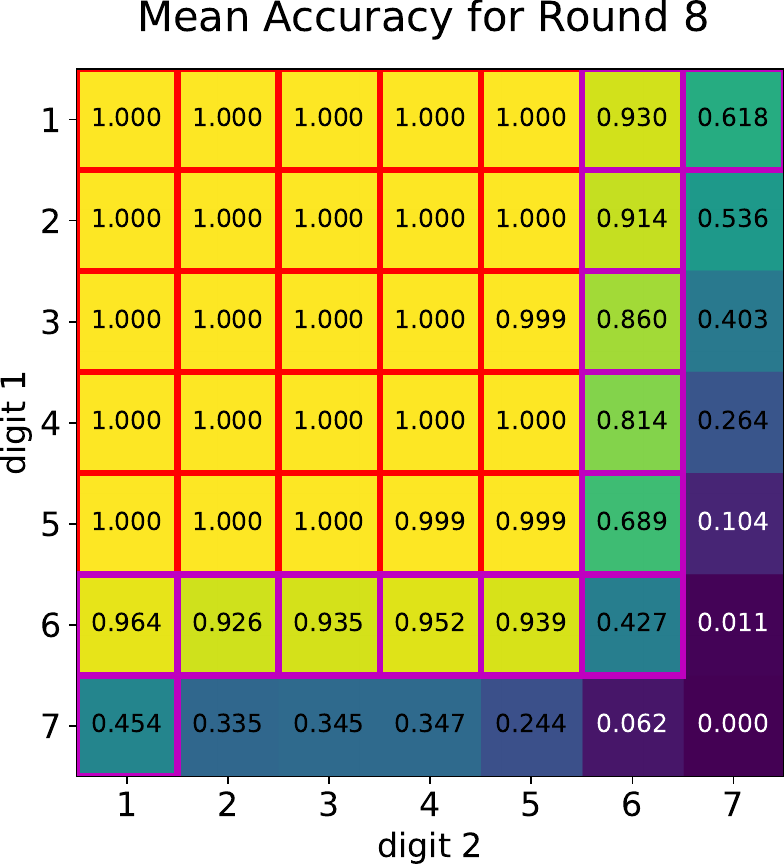}
    \includegraphics[width=0.24\linewidth]{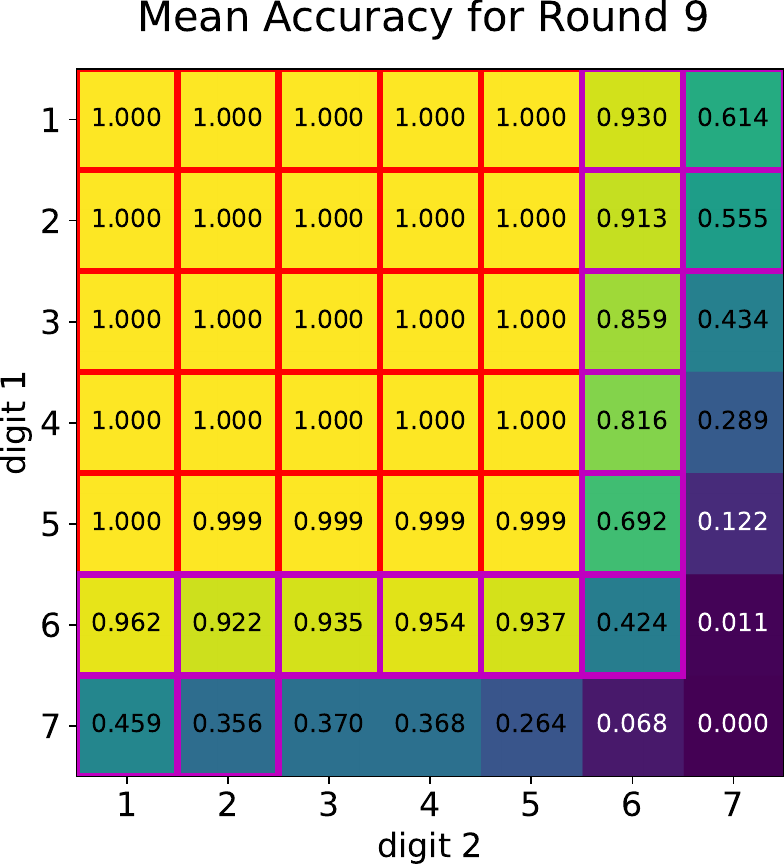}
    \includegraphics[width=0.24\linewidth]{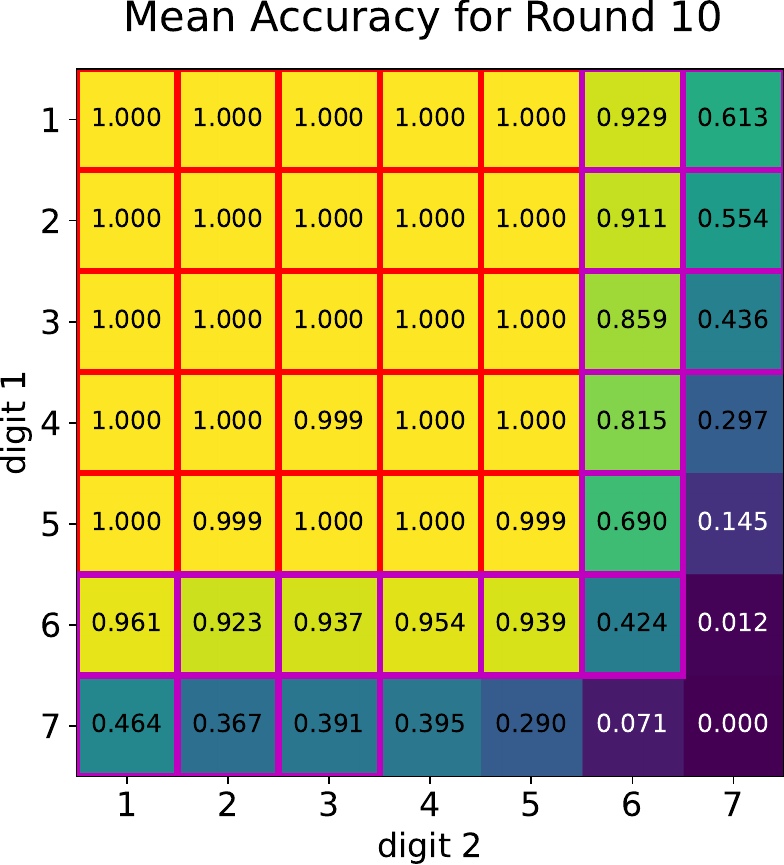}
    \includegraphics[width=0.24\linewidth]{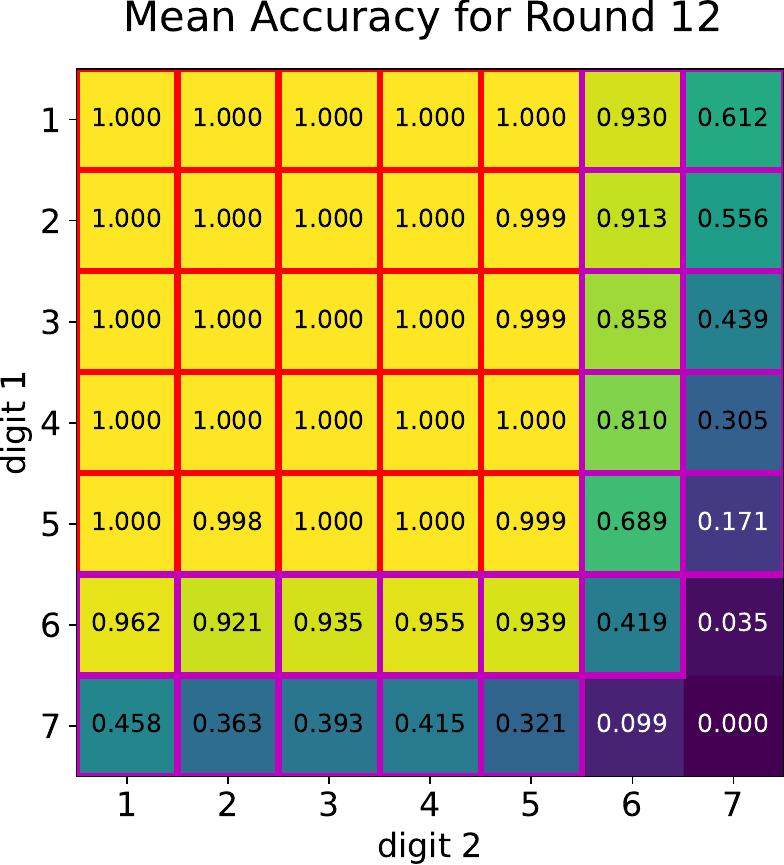}
    \includegraphics[width=0.24\linewidth]{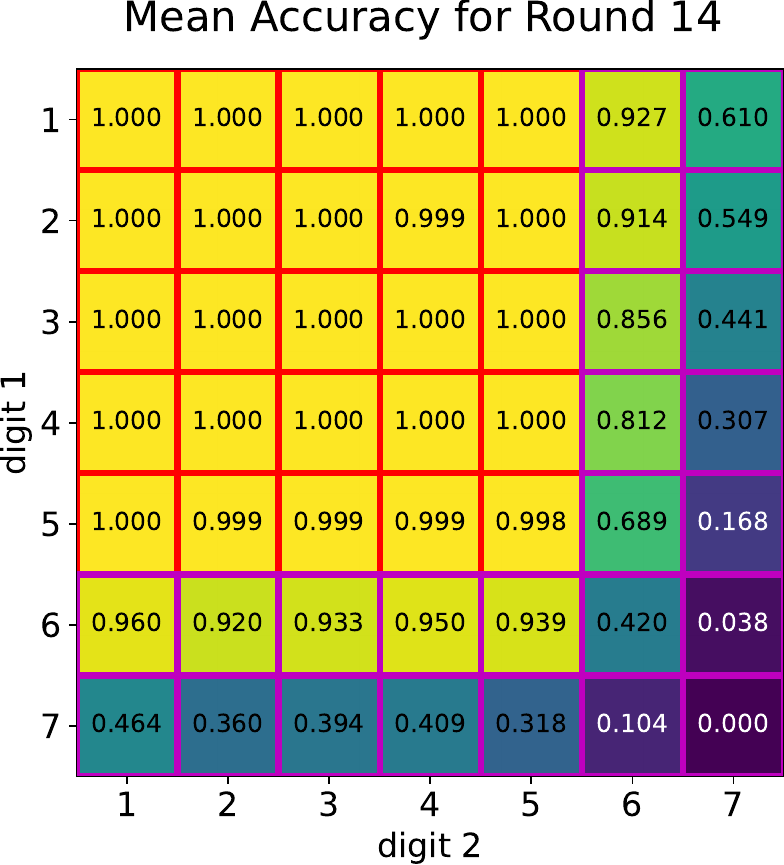}
    \end{minipage}
    }
    \caption{ Results for multiplication with length filtering with length threshold of 10.}
    \label{fig:multiplication_len_filter}
\end{figure}

\begin{figure}
    \centering
    \resizebox{0.8\linewidth}{!}{
    \begin{minipage}{\linewidth}
    \includegraphics[width=0.24\linewidth]{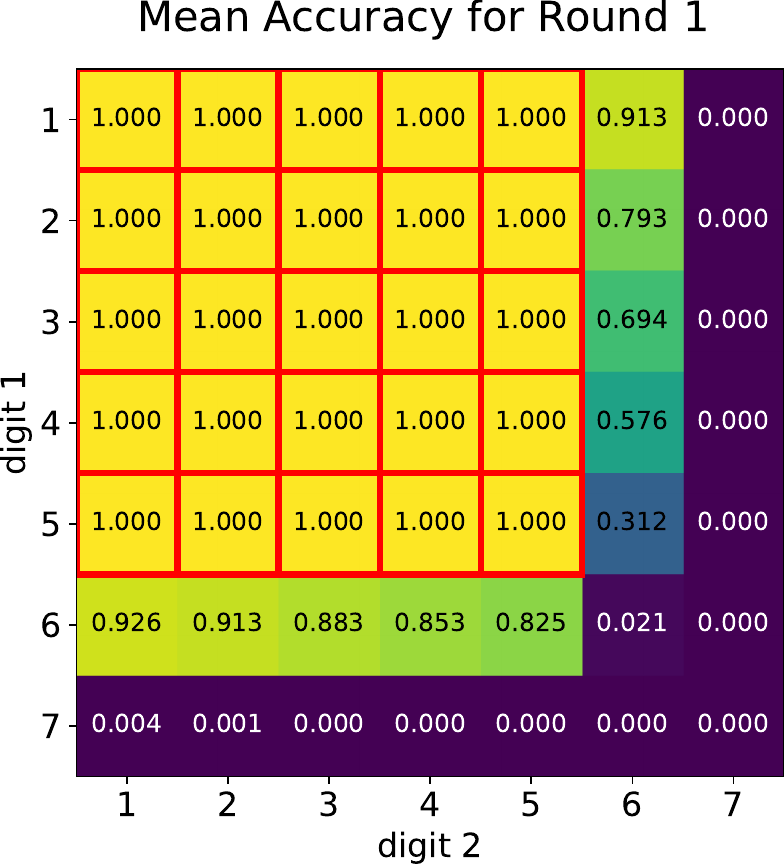}
    \includegraphics[width=0.24\linewidth]{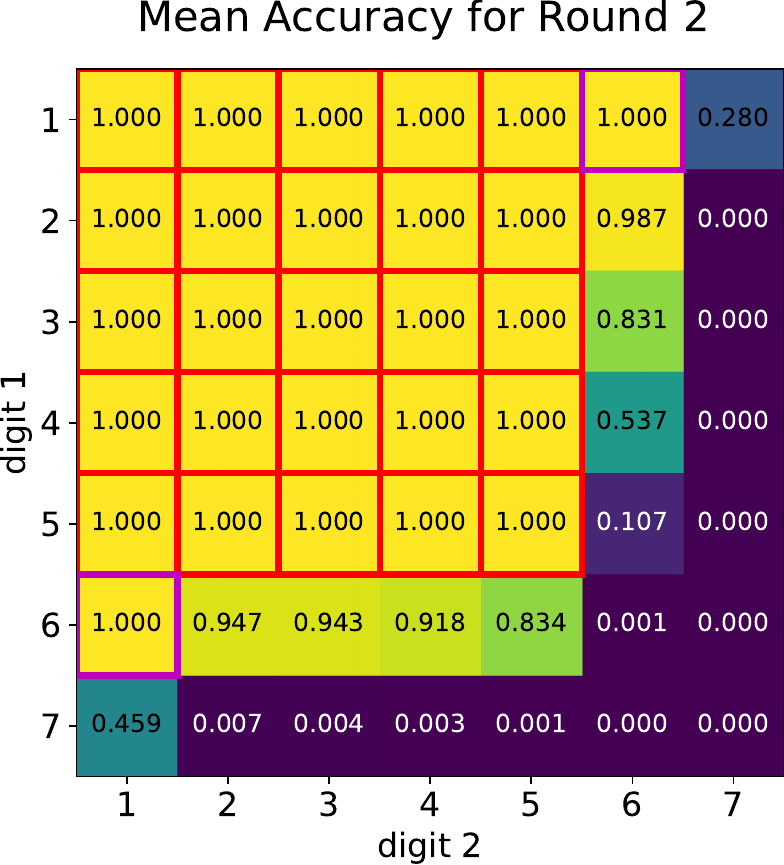}
    \includegraphics[width=0.24\linewidth]{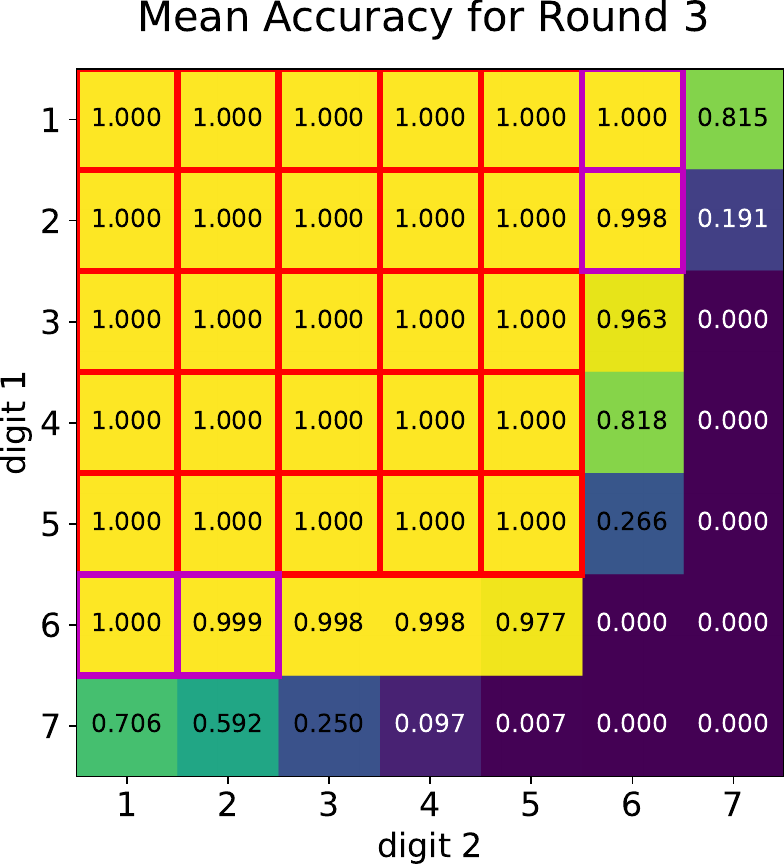}
    \includegraphics[width=0.24\linewidth]{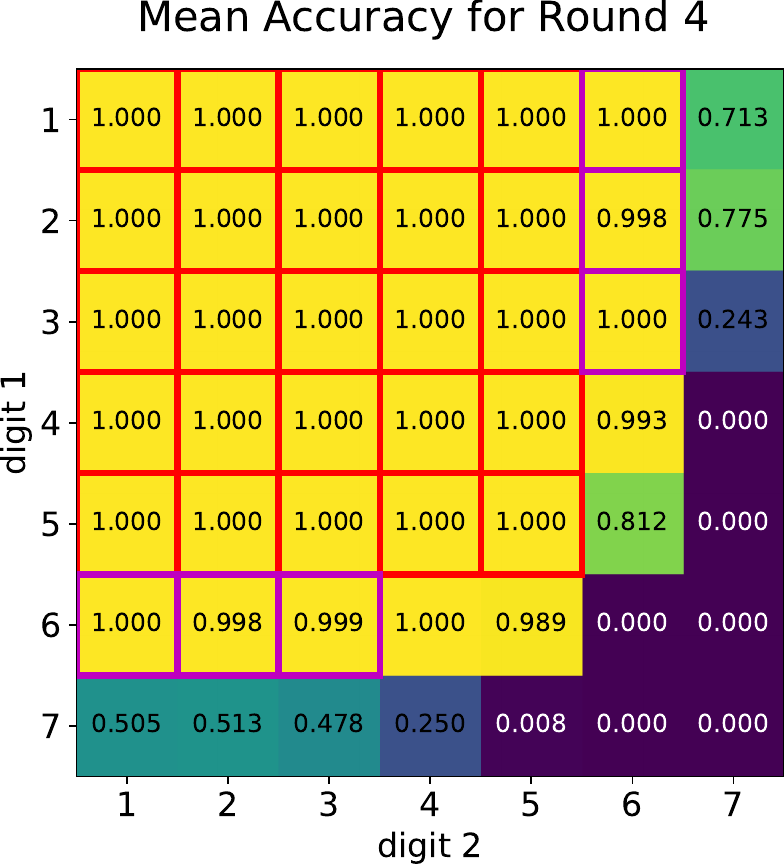}
    \includegraphics[width=0.24\linewidth]{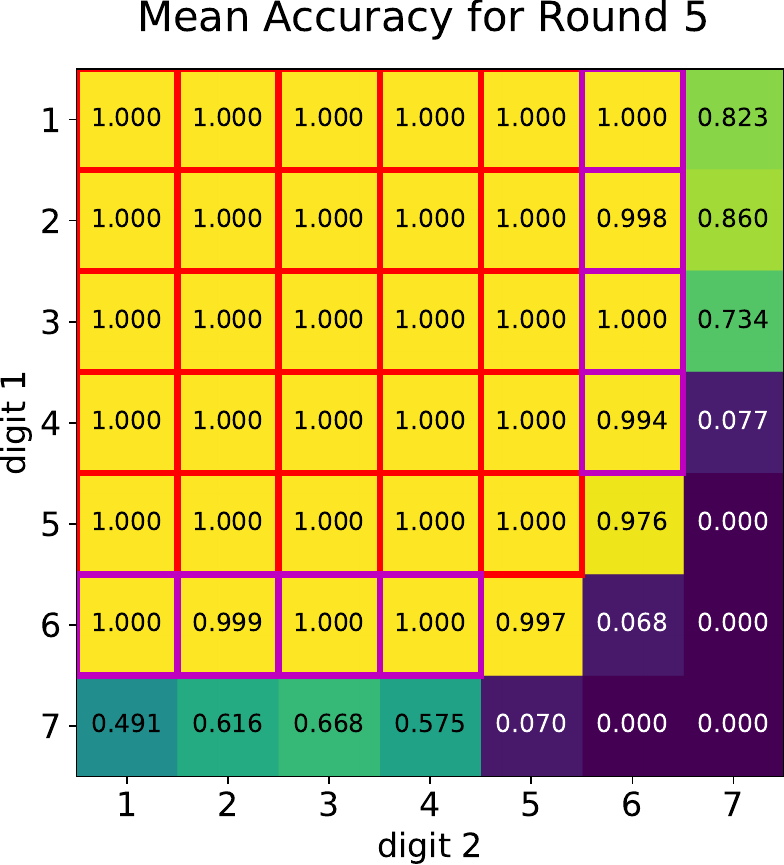}
    \includegraphics[width=0.24\linewidth]{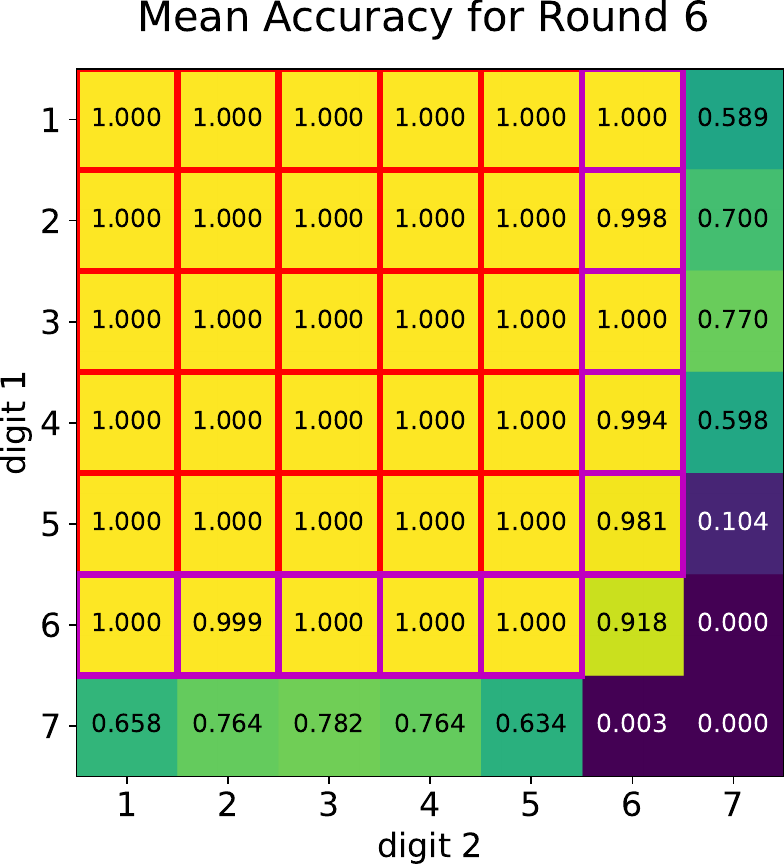}
    \includegraphics[width=0.24\linewidth]{fig/mult/majority/multiplication_mv_mult_7_acc.pdf}
    \includegraphics[width=0.24\linewidth]{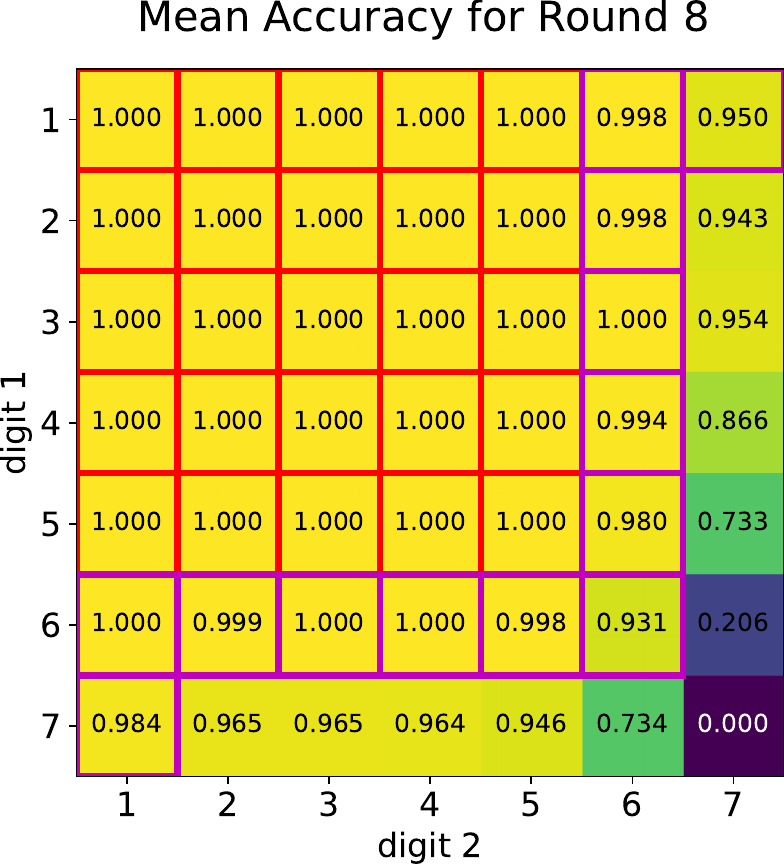}
    \vspace{1mm}
    \includegraphics[width=0.24\linewidth]{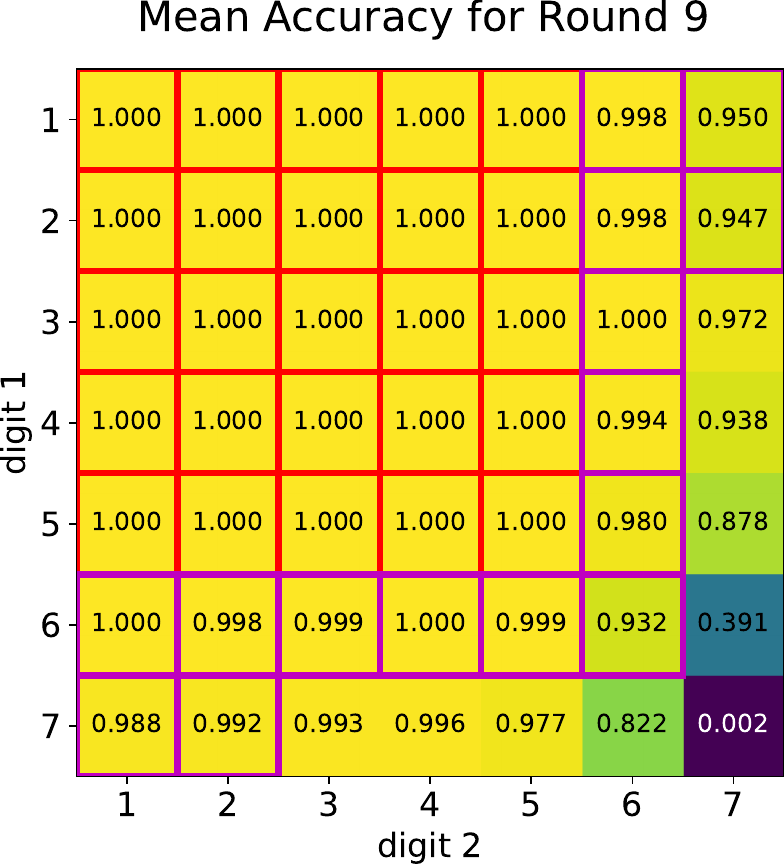}
    \includegraphics[width=0.24\linewidth]{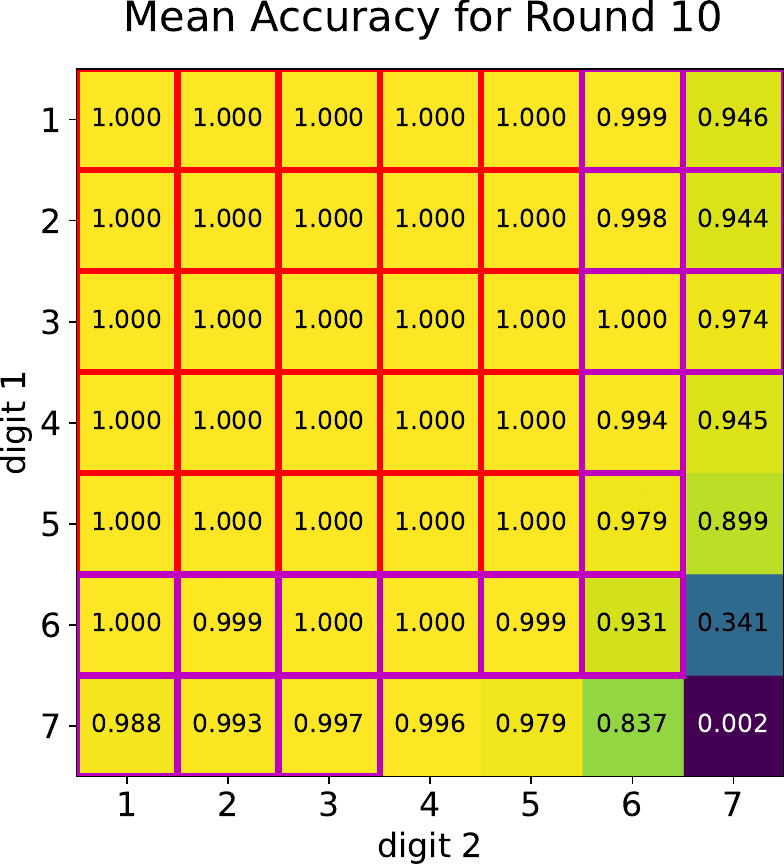}
    \includegraphics[width=0.24\linewidth]{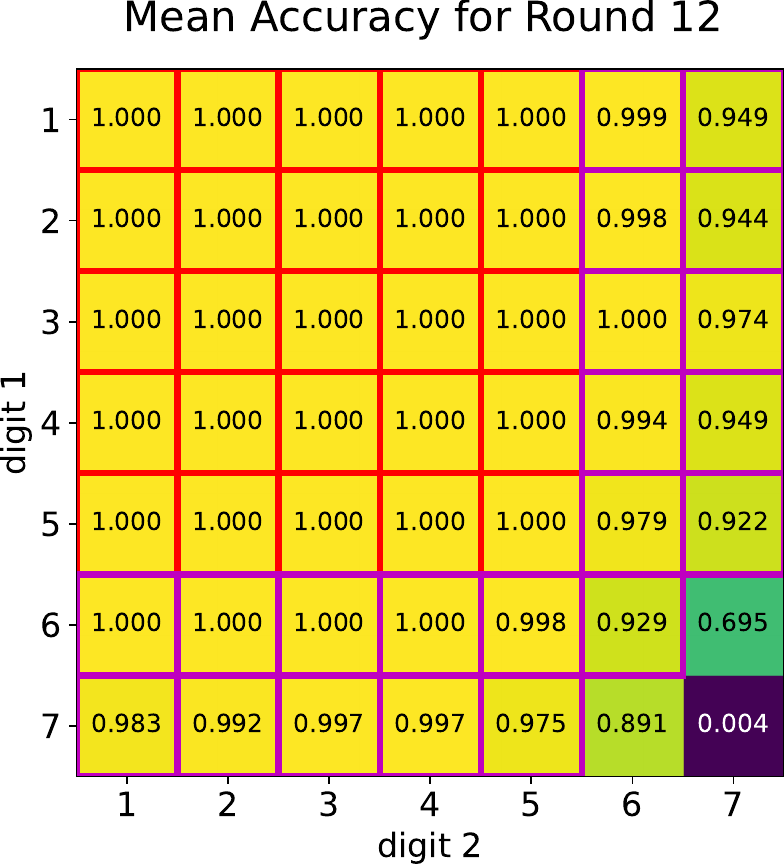}
    \includegraphics[width=0.24\linewidth]{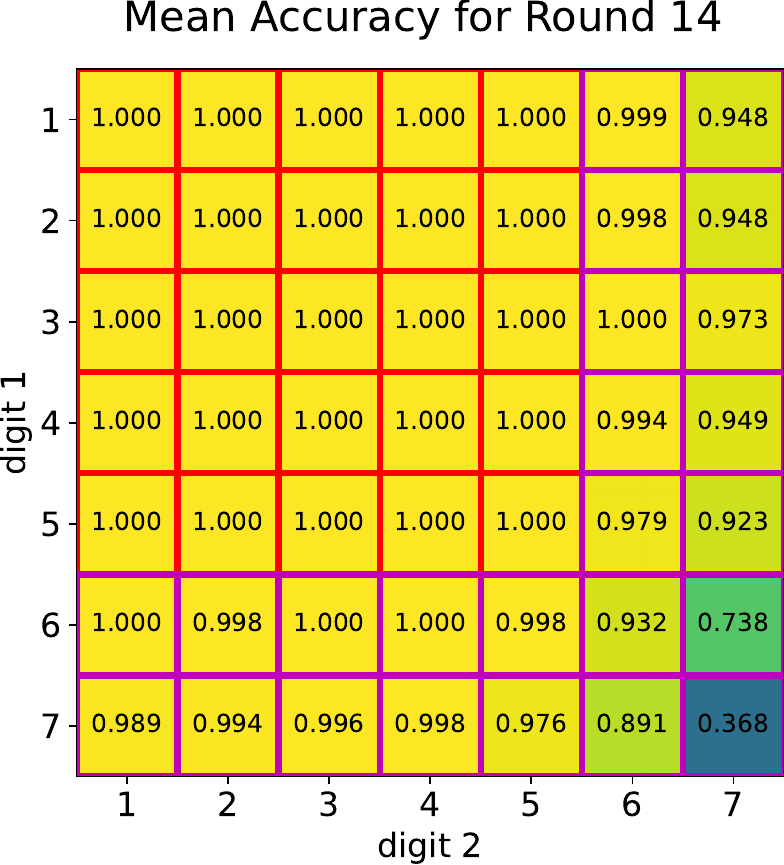}
    \end{minipage}}
    \caption{Multiplication with majority voting where filtering is based on agreement of at least 4 out of 5 models. Applying majority voting enables effective generalization from $n$-by-$n$ to $(n+1)$-by-$(n+1)$ multiplication tasks.}
    \label{fig:multiplication_mv_full}
\end{figure}

\begin{figure}
    \centering
    \resizebox{0.8\linewidth}{!}{
    \begin{minipage}{\linewidth}
    \centering
    \includegraphics[width=0.24\linewidth]{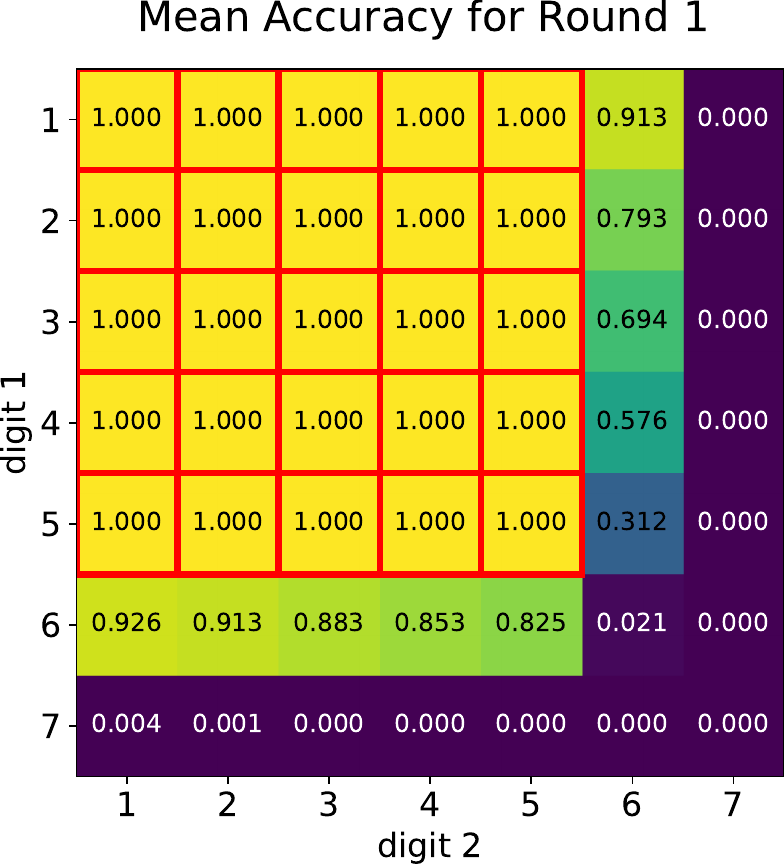}
    \includegraphics[width=0.24\linewidth]{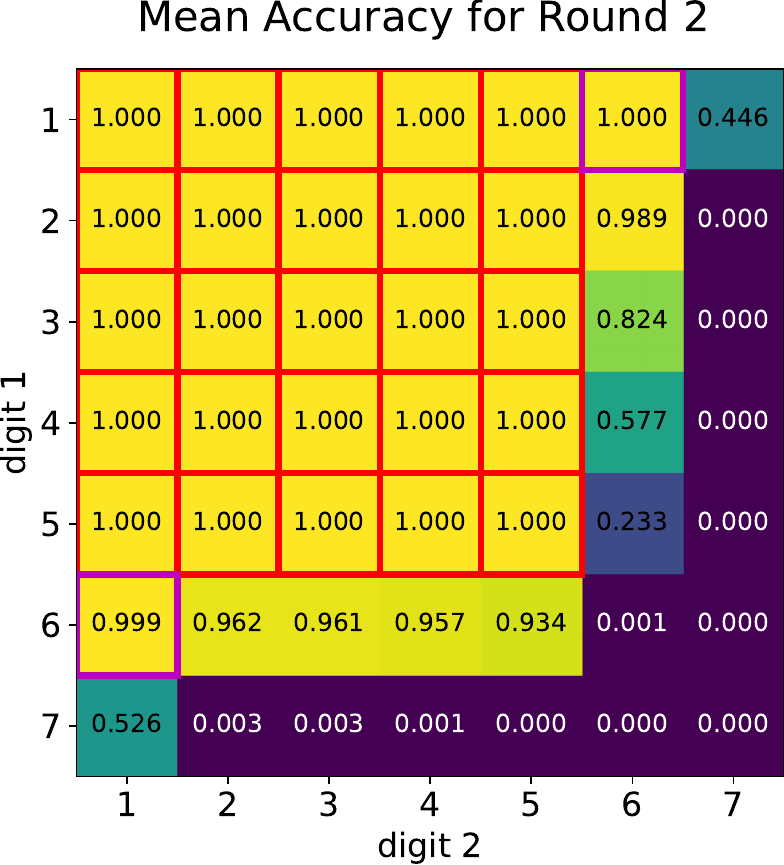}
    \includegraphics[width=0.24\linewidth]{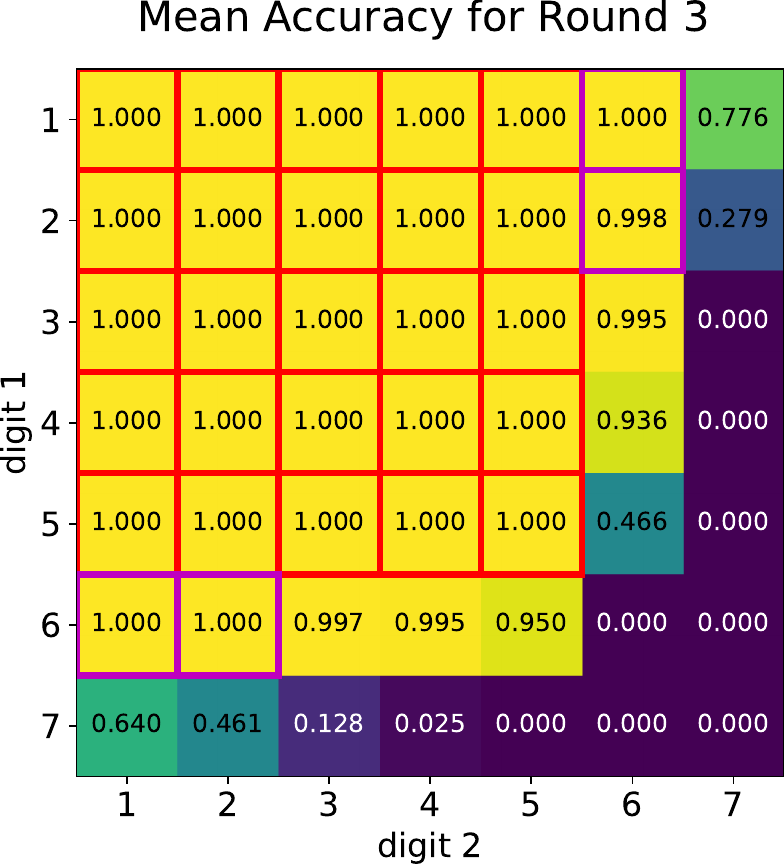}
    \includegraphics[width=0.24\linewidth]{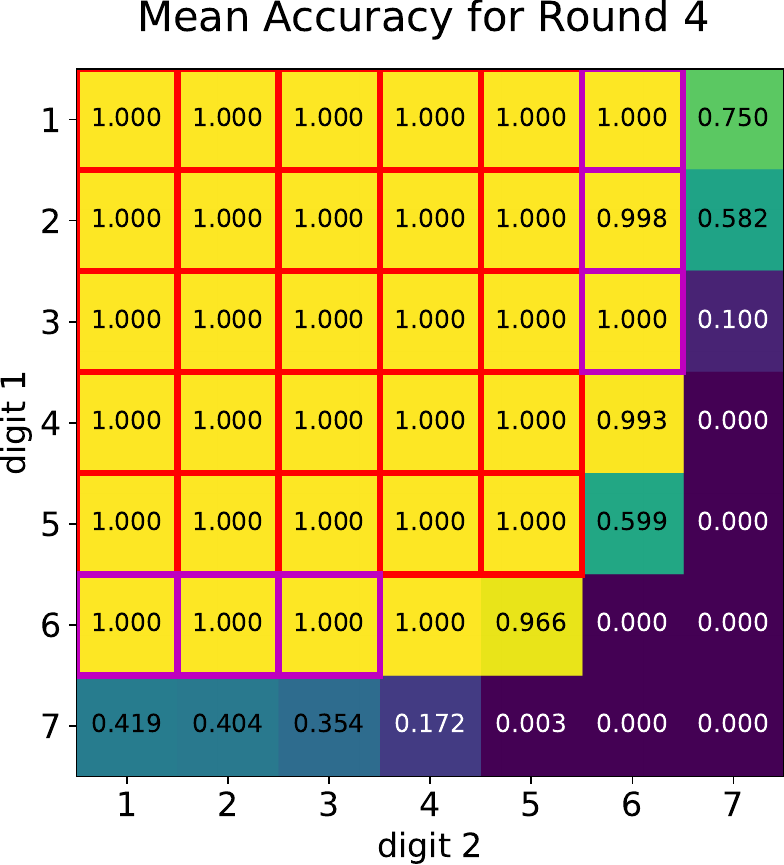}
    \includegraphics[width=0.24\linewidth]{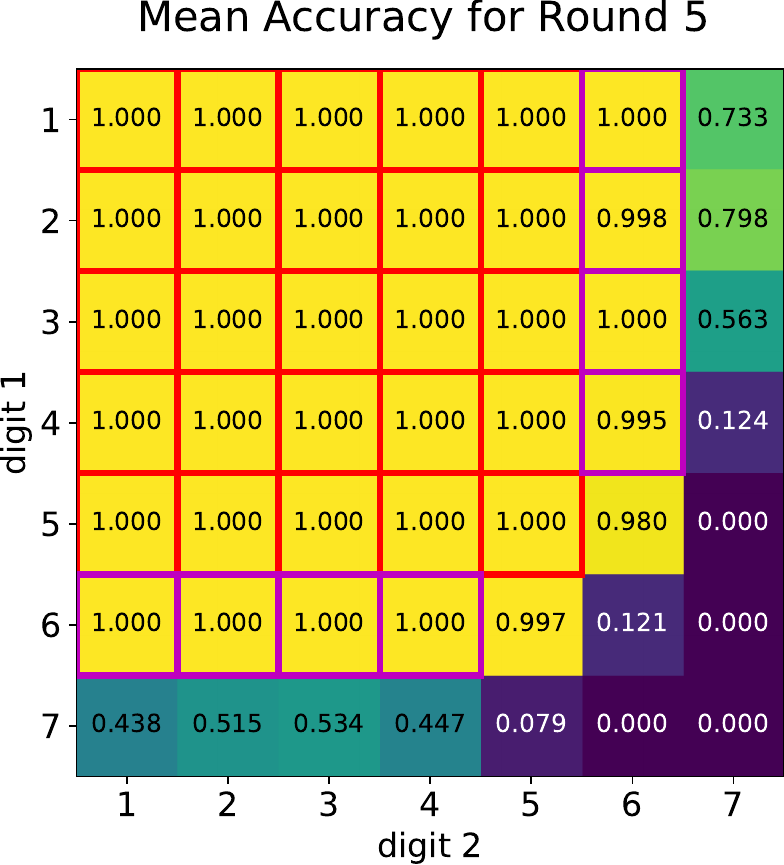}
    \includegraphics[width=0.24\linewidth]{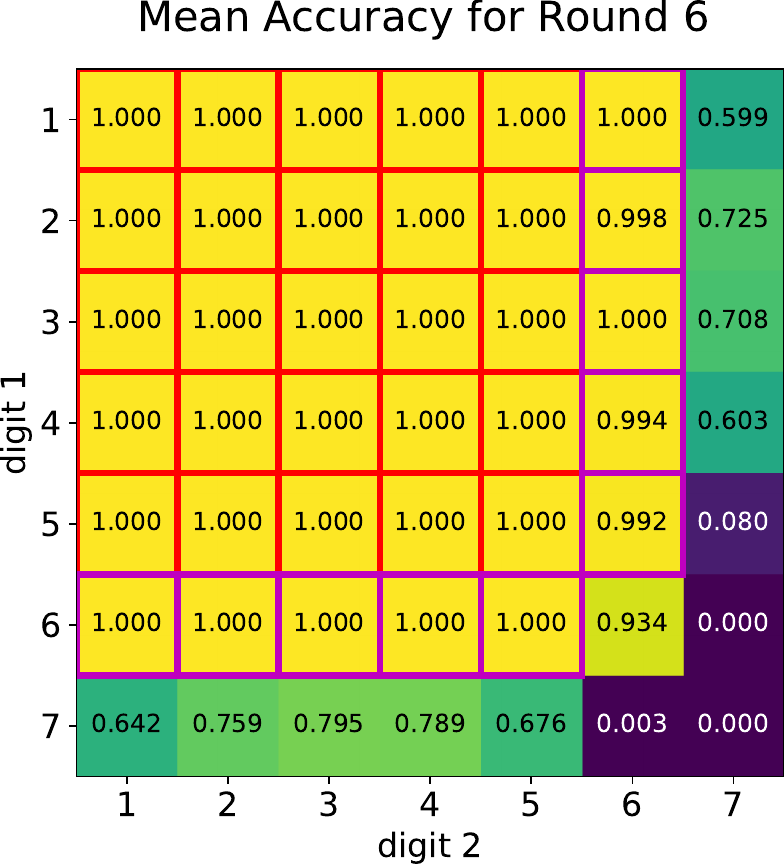}
    \includegraphics[width=0.24\linewidth]{fig/mult/majority_fixed_SI/multiplication_mv_7_acc.pdf}
    \includegraphics[width=0.24\linewidth]{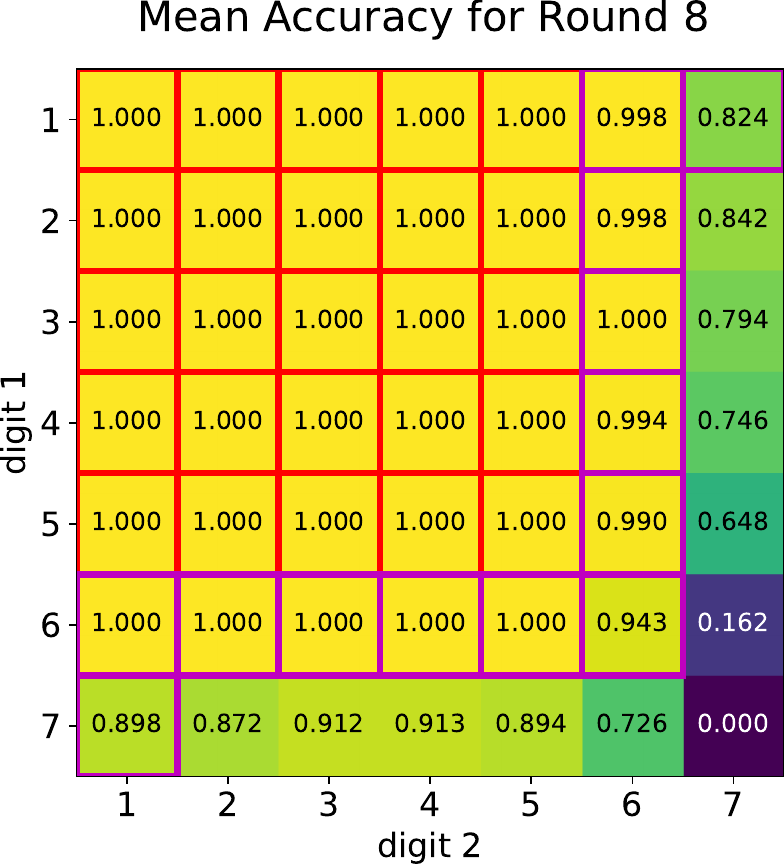}
    \includegraphics[width=0.24\linewidth]{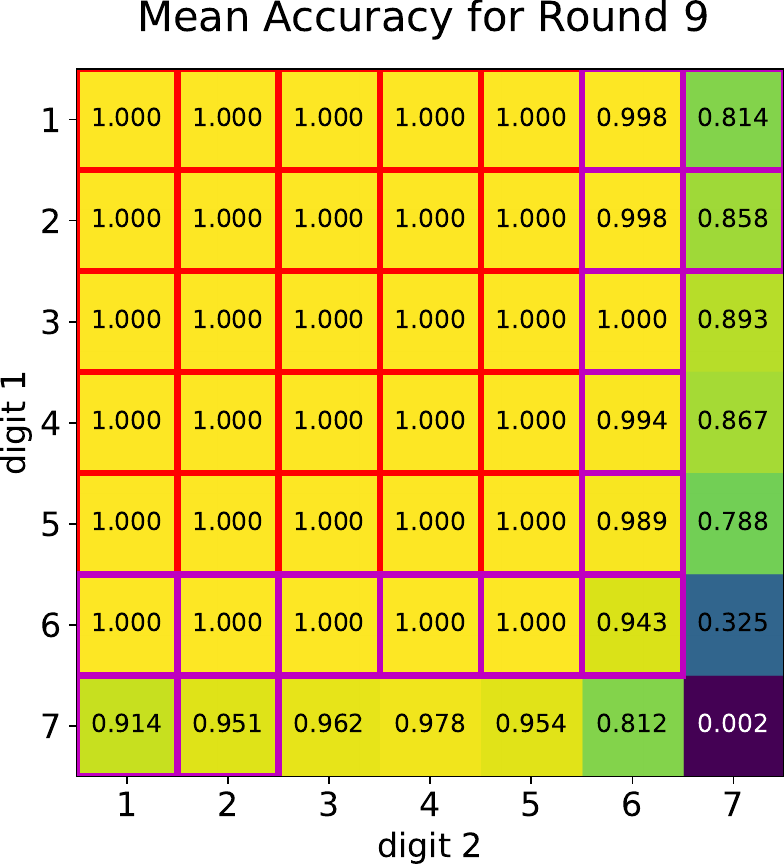}
    \includegraphics[width=0.24\linewidth]{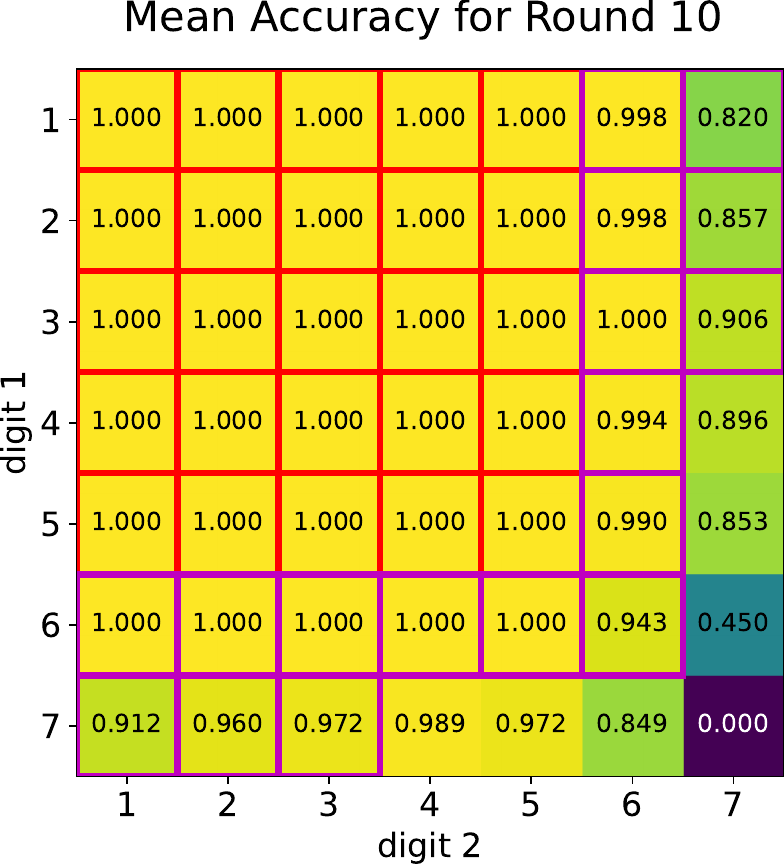}
    \includegraphics[width=0.24\linewidth]{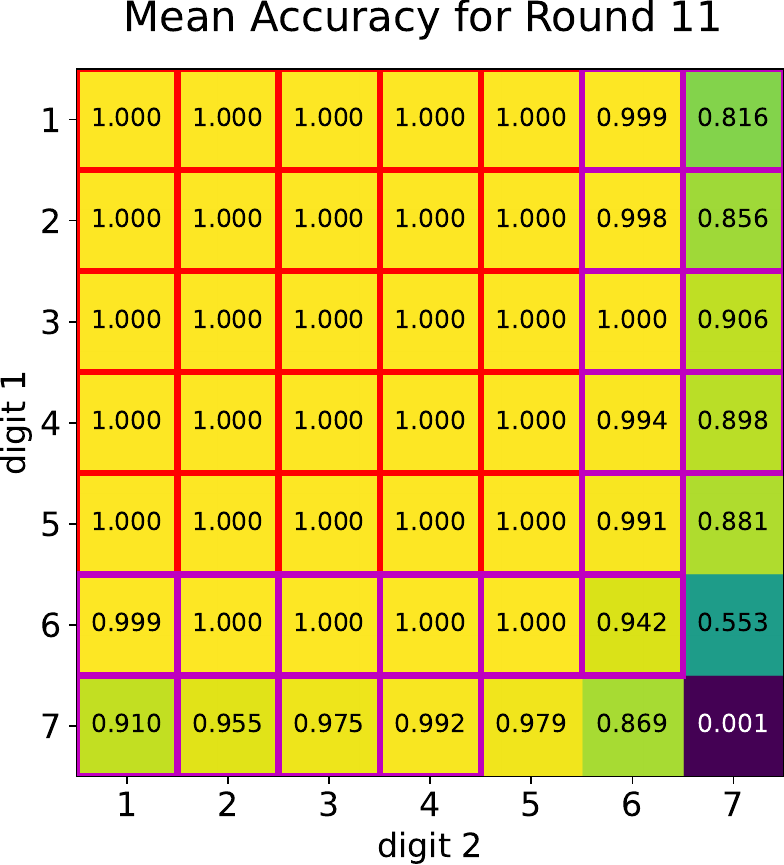}
    \includegraphics[width=0.24\linewidth]{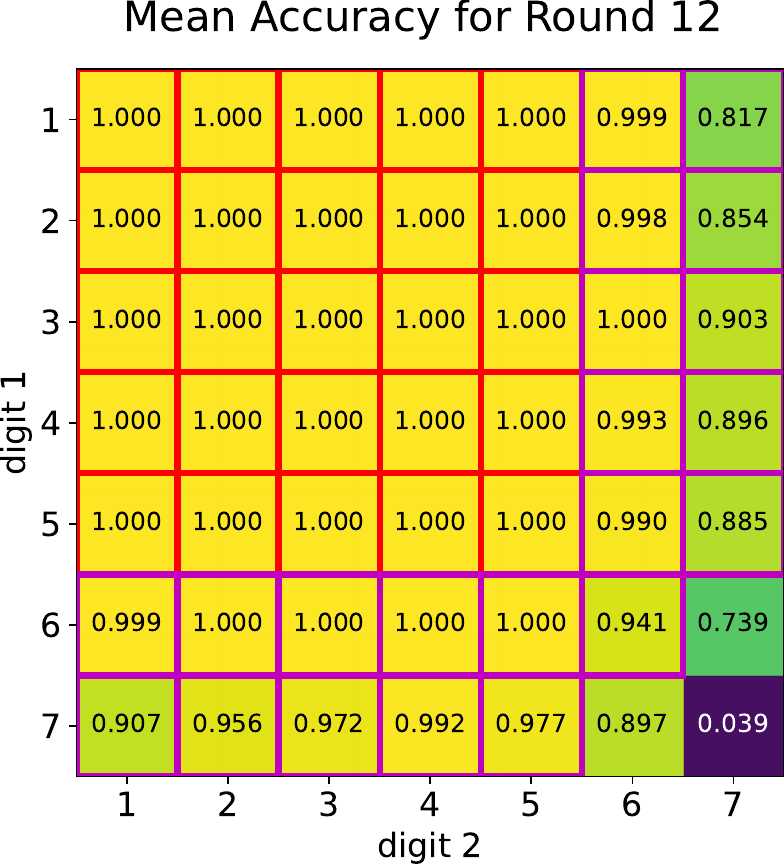}
    \end{minipage}}
    \caption{Multiplication task with majority voting with shared self-improve data (See Section~\ref{sec:mv_ablations}).}
    \label{fig:multiplication_mv_fixed_SI}
\end{figure}

\begin{figure}
    \centering
    \includegraphics[width=0.24\linewidth]{fig/mult/majority_len_n10/multiplication_majority_mult_len_filtered_n_10_1_acc.pdf}
    \includegraphics[width=0.24\linewidth]{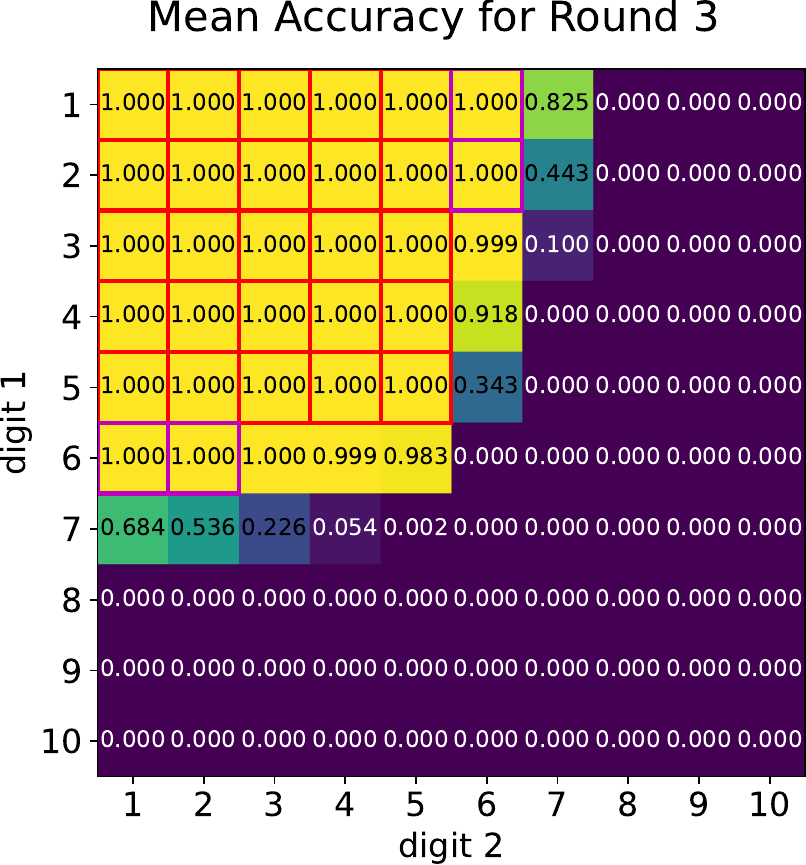}
    \includegraphics[width=0.24\linewidth]{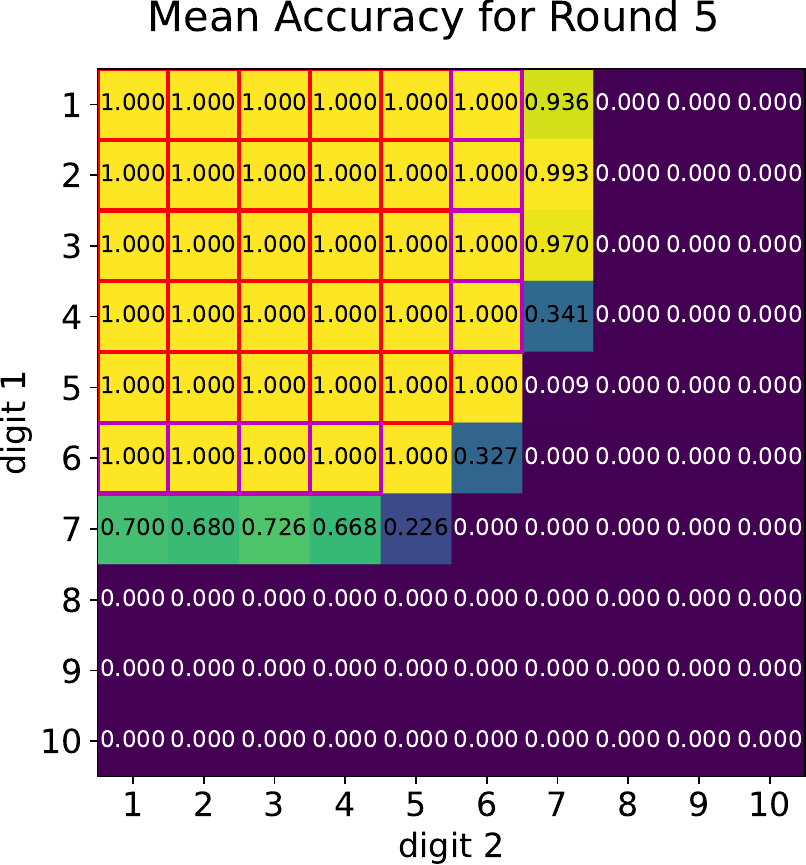}
    \includegraphics[width=0.24\linewidth]{fig/mult/majority_len_n10/multiplication_majority_mult_len_filtered_n_10_7_acc.pdf}
    \vspace{1mm}
    \includegraphics[width=0.24\linewidth]{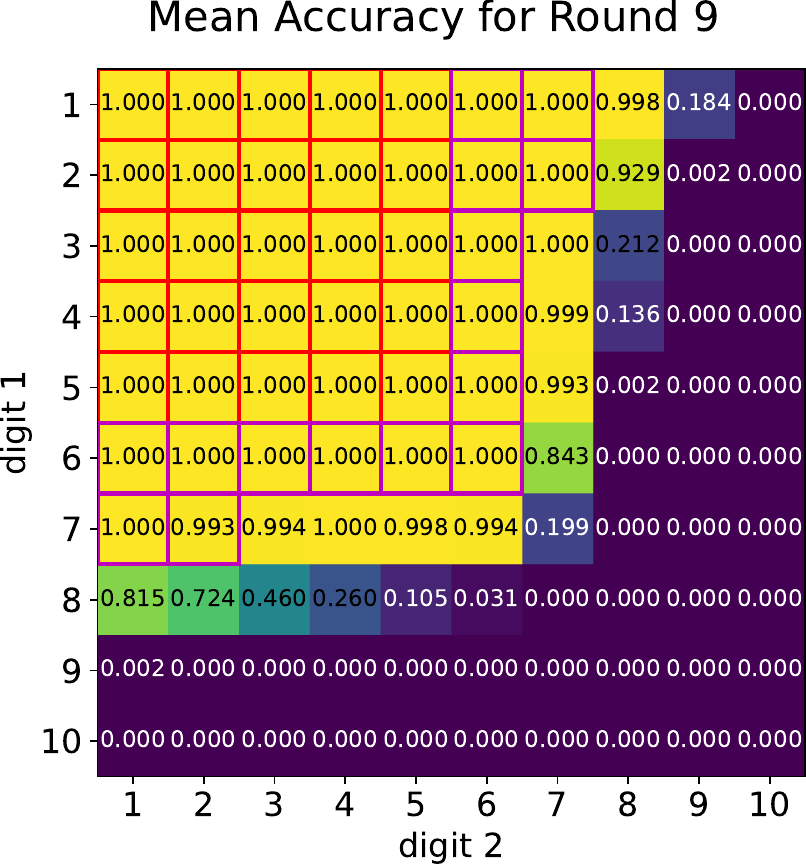}
    \includegraphics[width=0.24\linewidth]{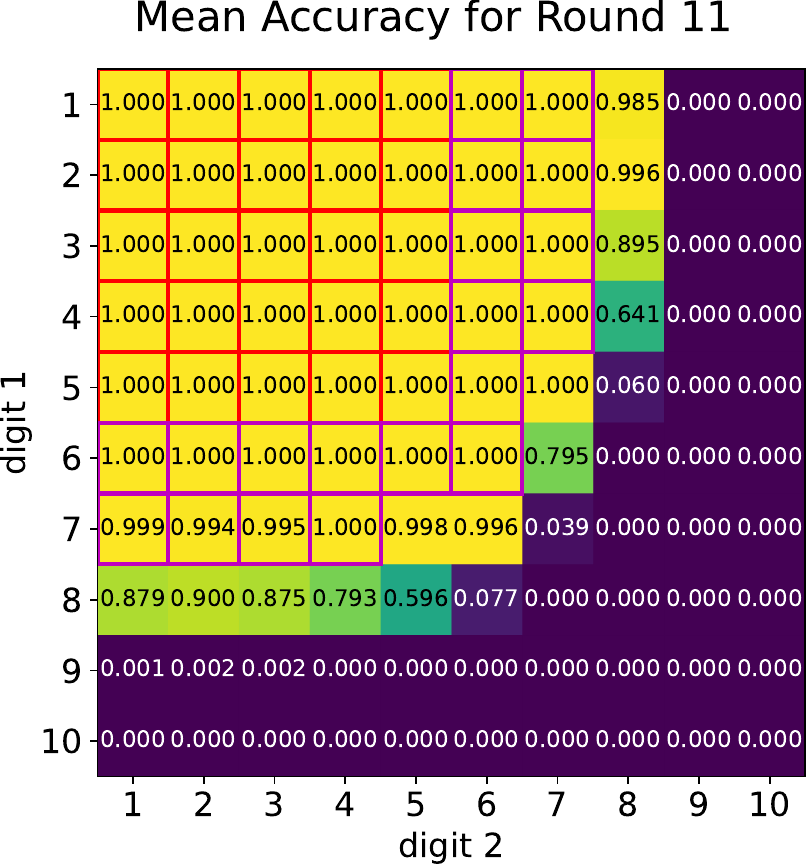}
    \includegraphics[width=0.24\linewidth]{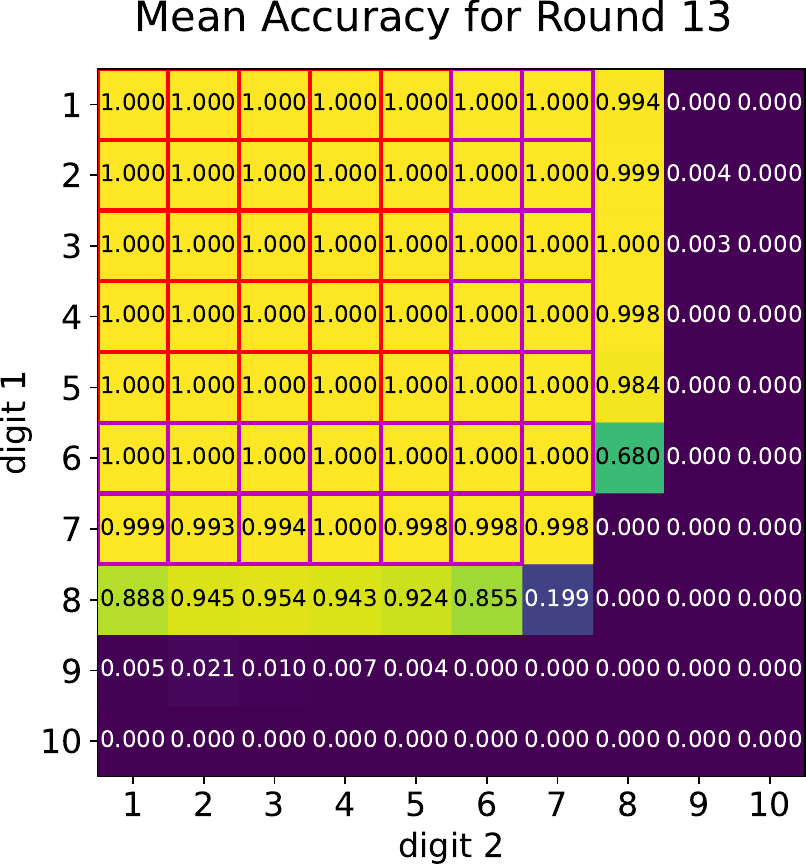}
    \includegraphics[width=0.24\linewidth]{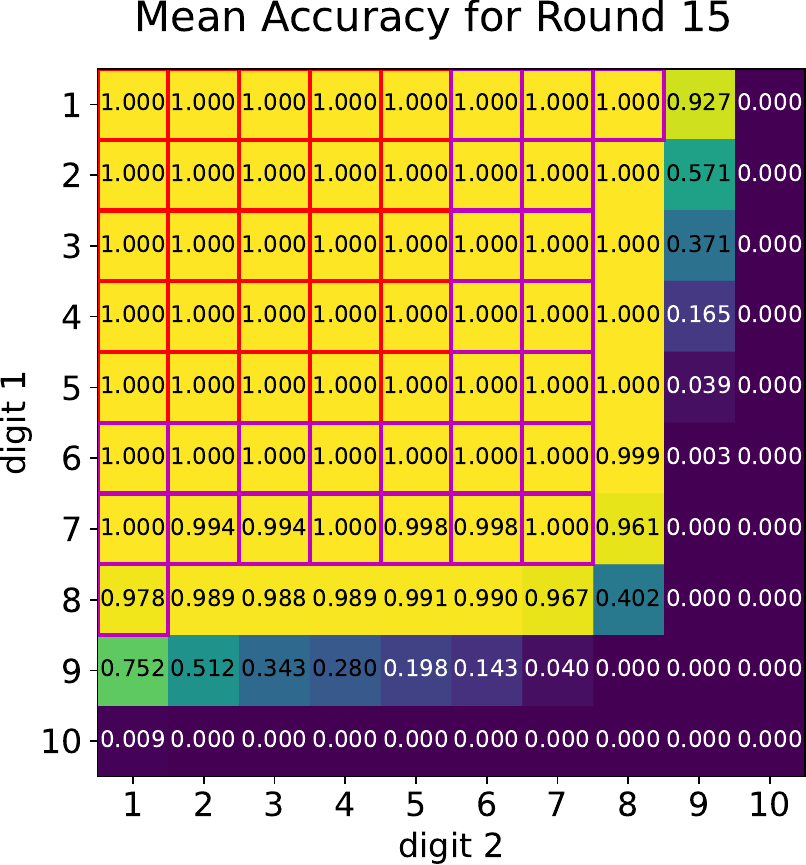}
    \vspace{1mm}
    \includegraphics[width=0.24\linewidth]{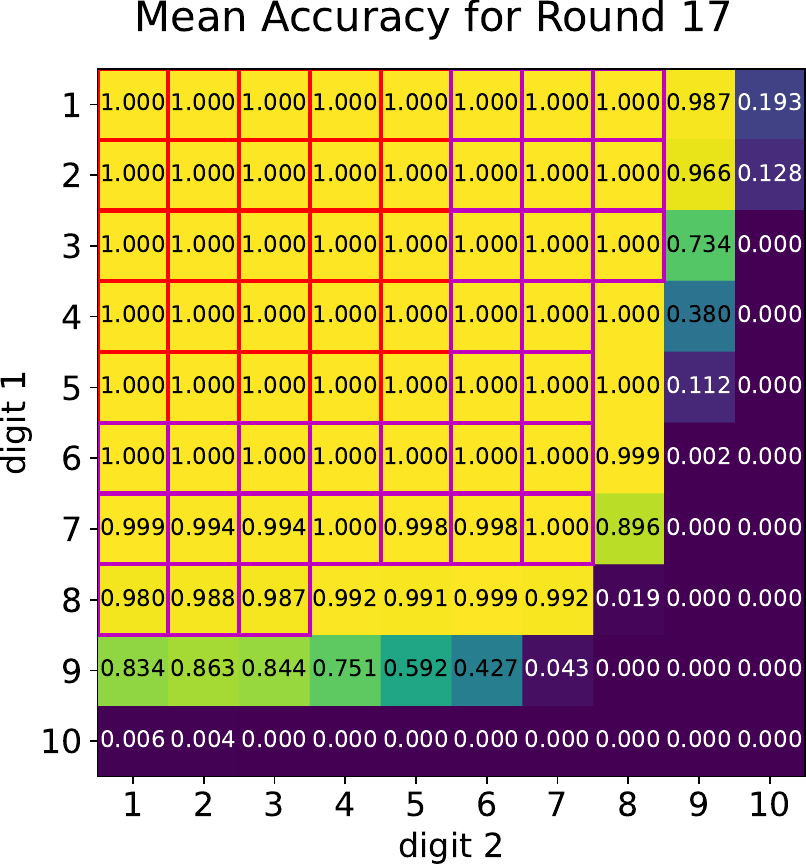}    
    \includegraphics[width=0.24\linewidth]{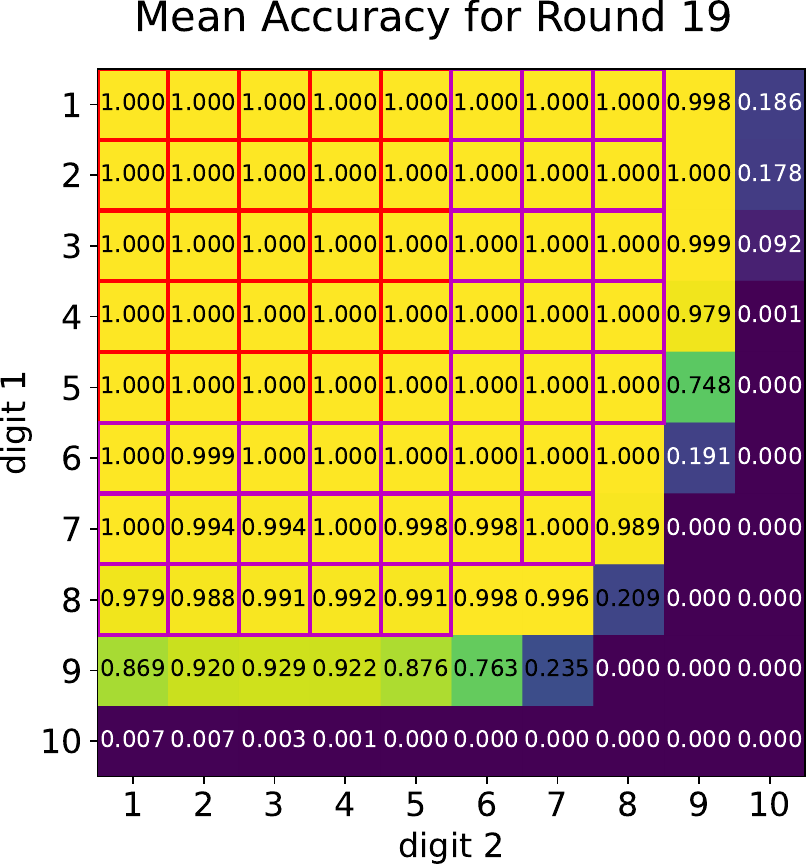}
    \includegraphics[width=0.24\linewidth]{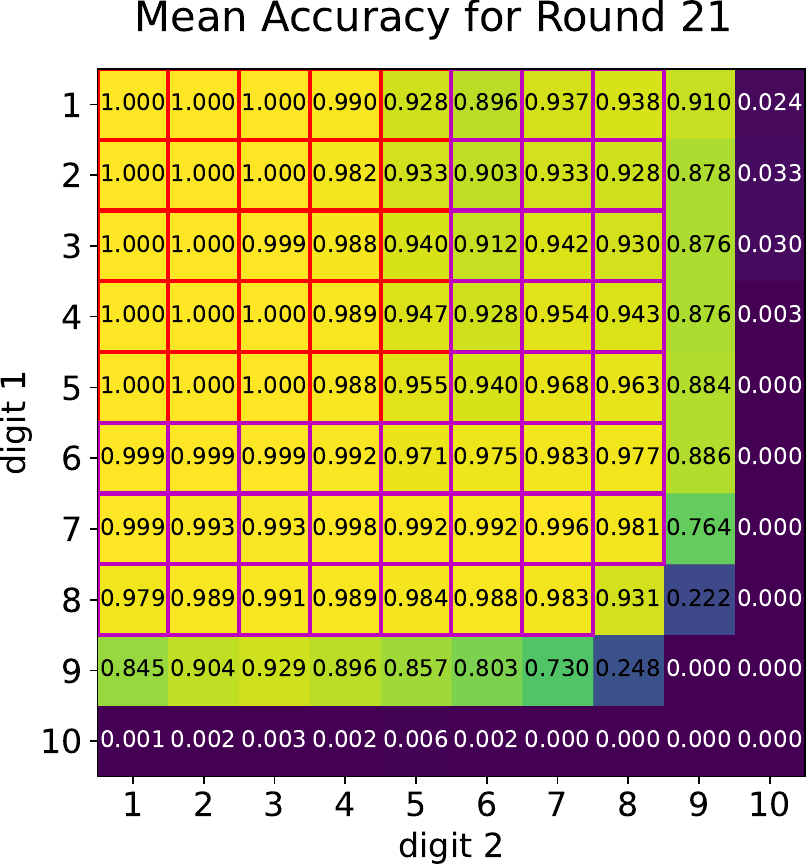}
    \includegraphics[width=0.24\linewidth]{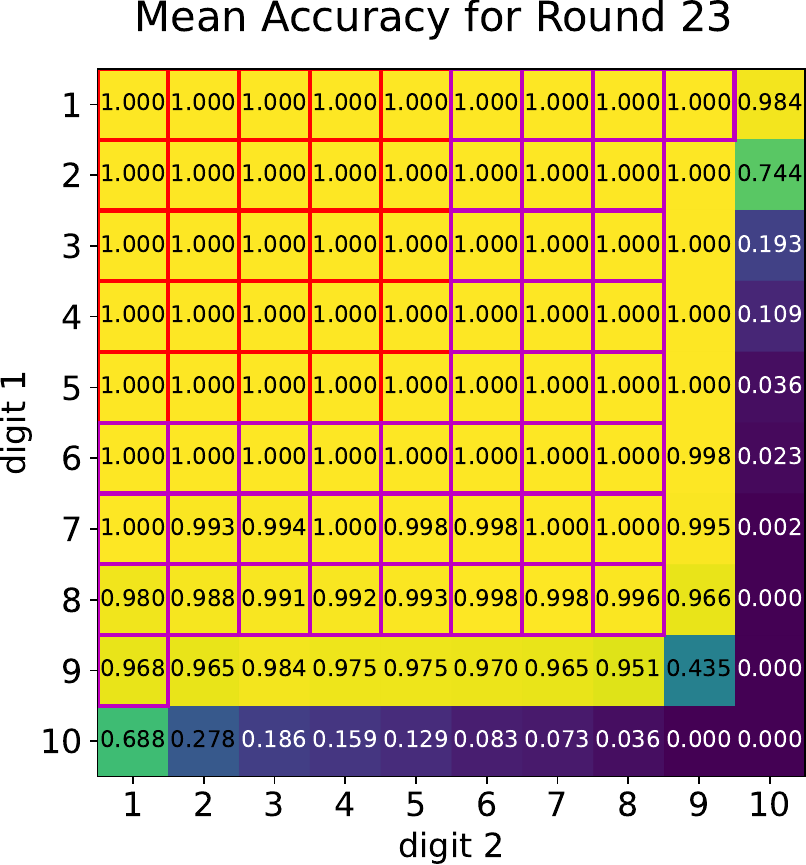}
    \vspace{1mm}
    \includegraphics[width=0.24\linewidth]{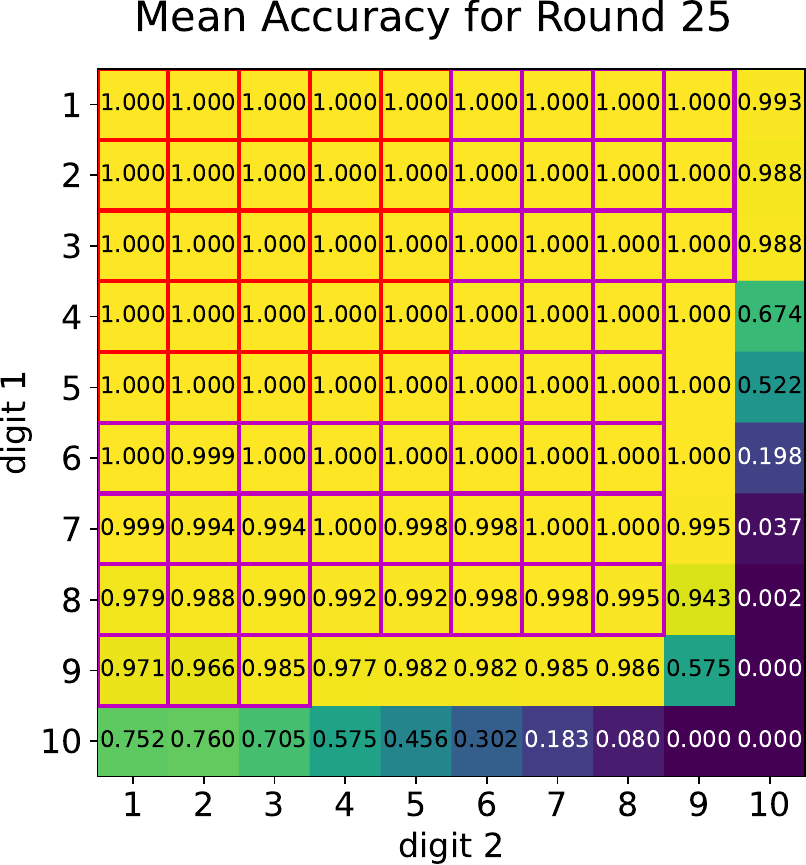}
    \includegraphics[width=0.24\linewidth]{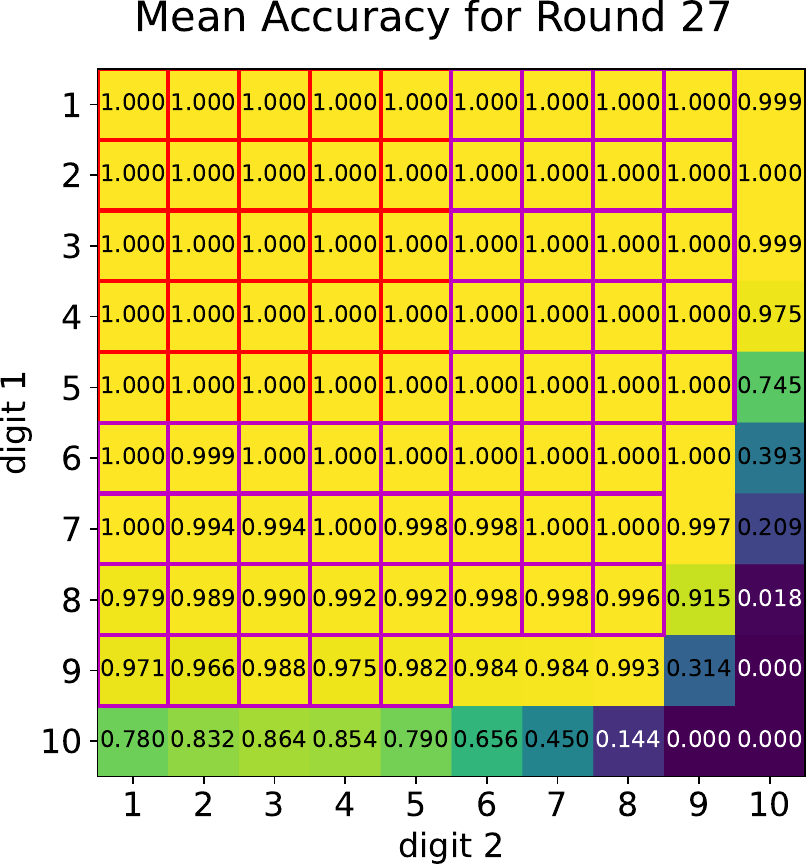}
    \includegraphics[width=0.24\linewidth]{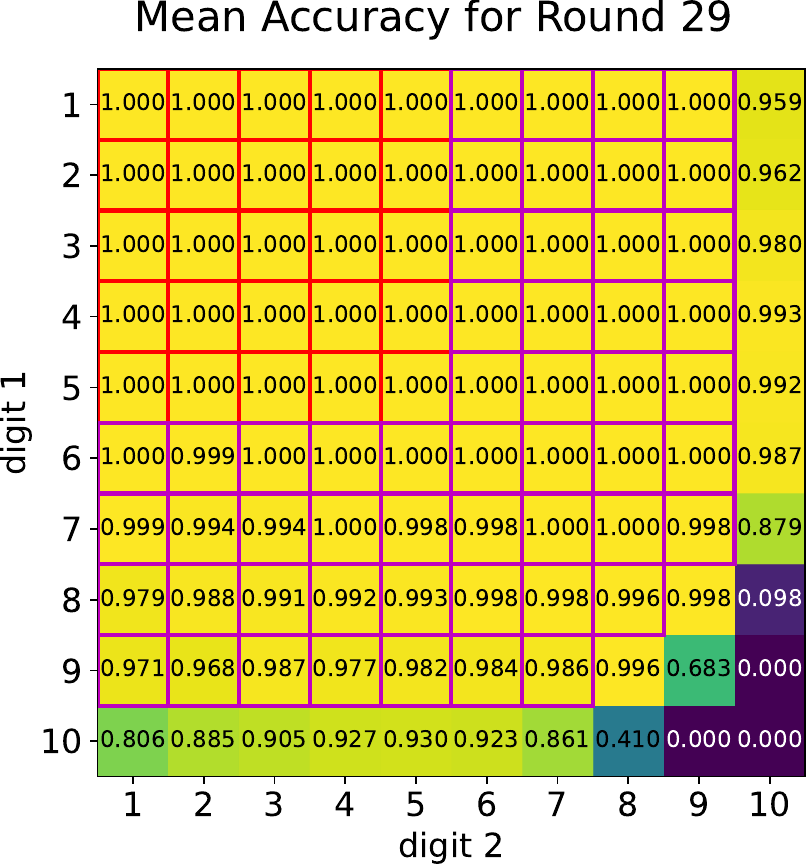}
    \includegraphics[width=0.24\linewidth]{fig/mult/majority_len_n10/multiplication_majority_mult_len_filtered_n_10_31_acc.pdf}
    \caption{Combining majority voting with length filtering. This approach achieves near-perfect length generalization up to $9 \times 9$, and potentially achieving further generalization. }
    \label{fig:multiplication_mv_len_n10_full}
\end{figure}

\begin{figure}
    \centering
    \includegraphics[width=0.24\linewidth]{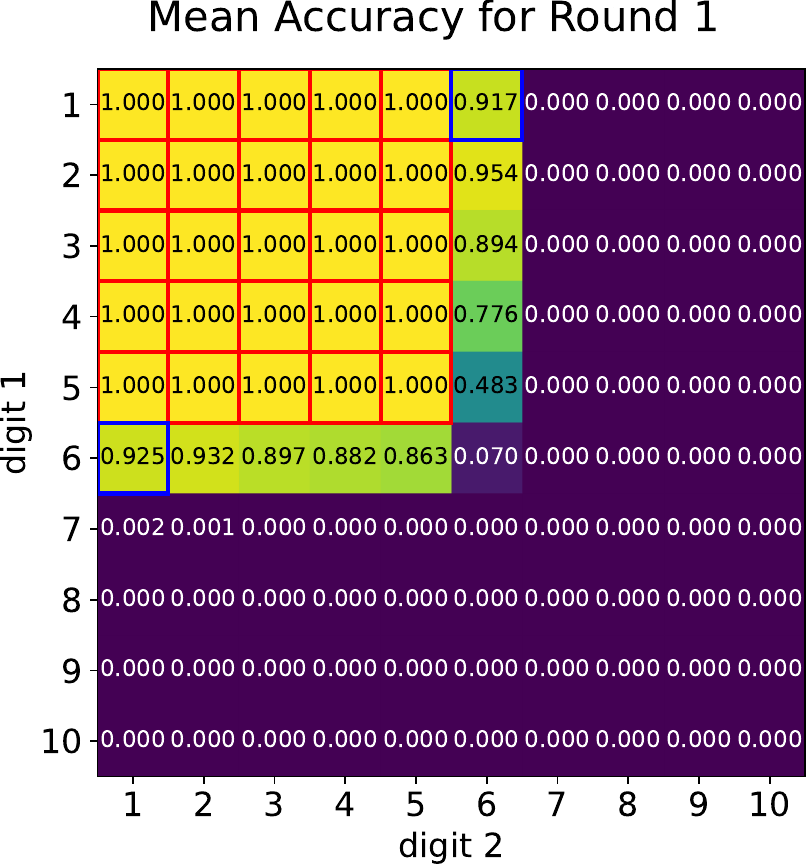}
    \includegraphics[width=0.24\linewidth]{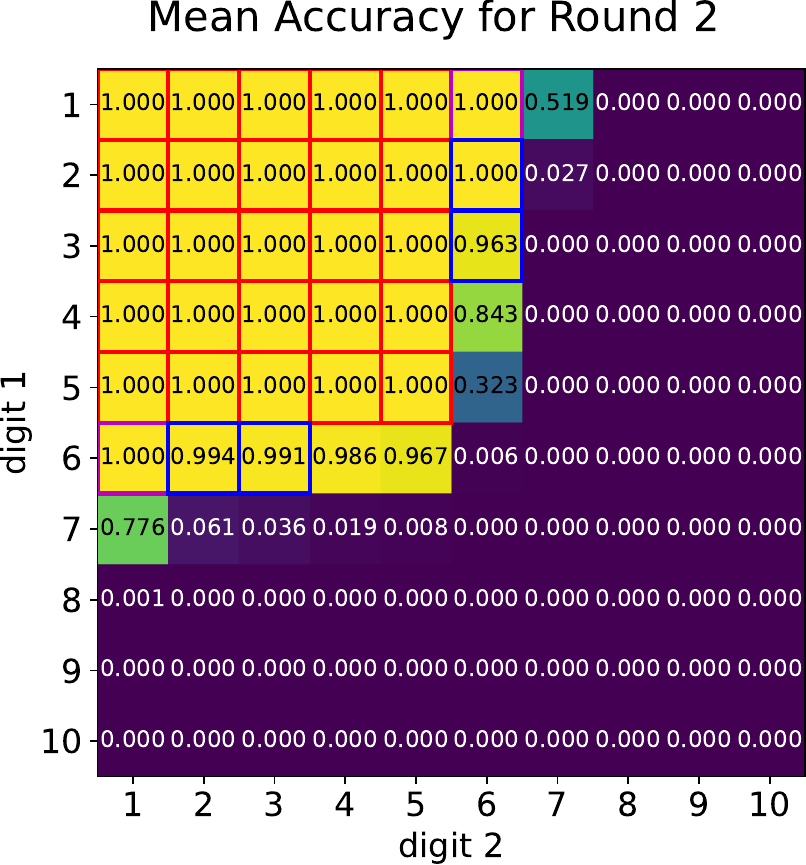}
    \includegraphics[width=0.24\linewidth]{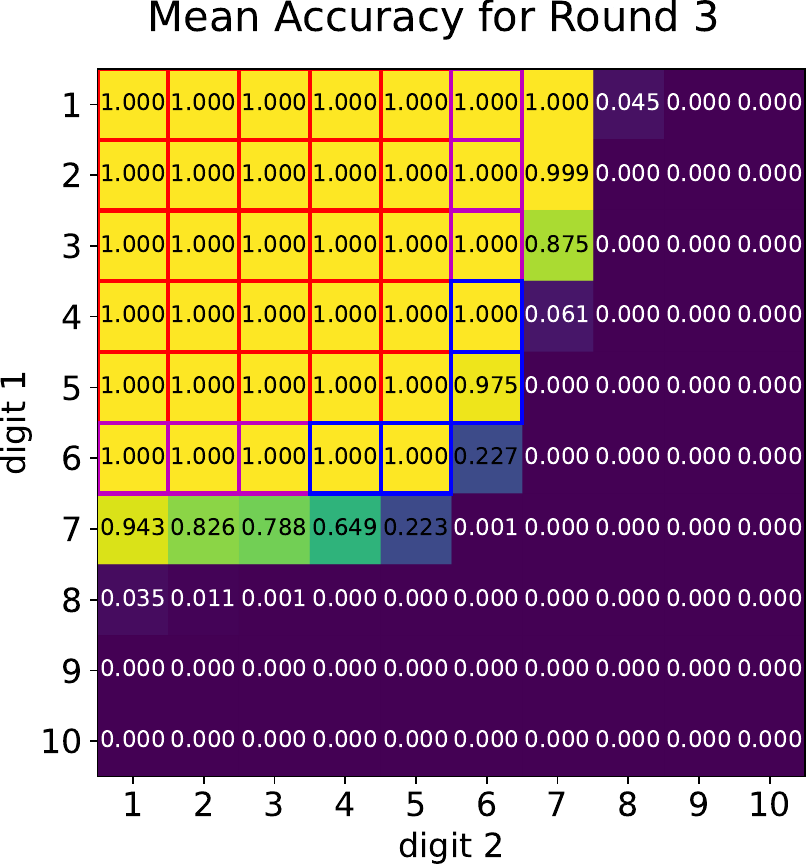}
    \includegraphics[width=0.24\linewidth]{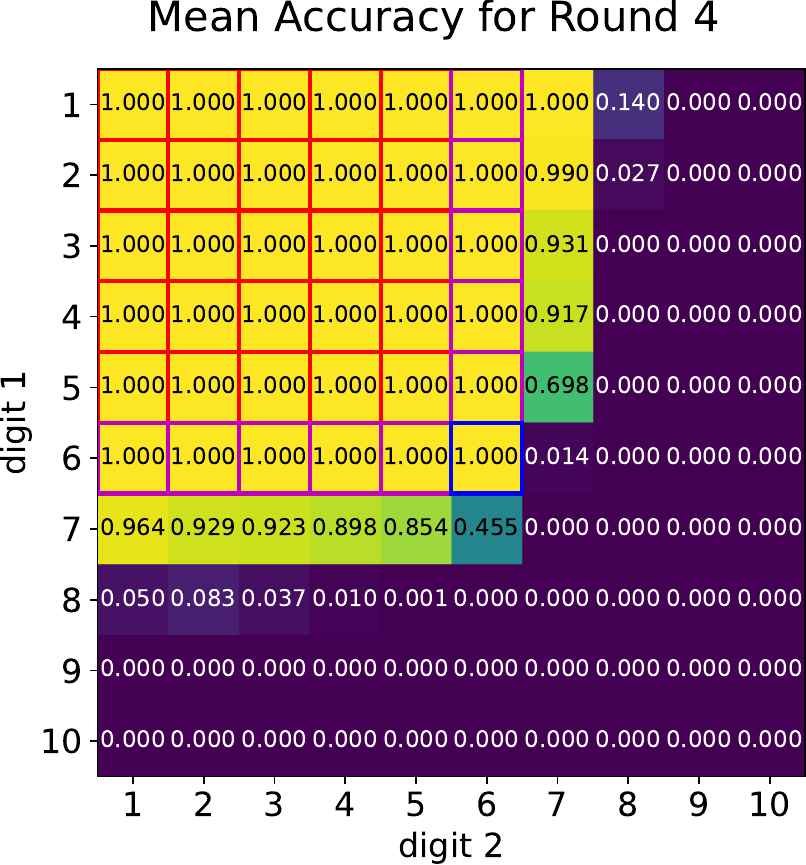}
    \includegraphics[width=0.24\linewidth]{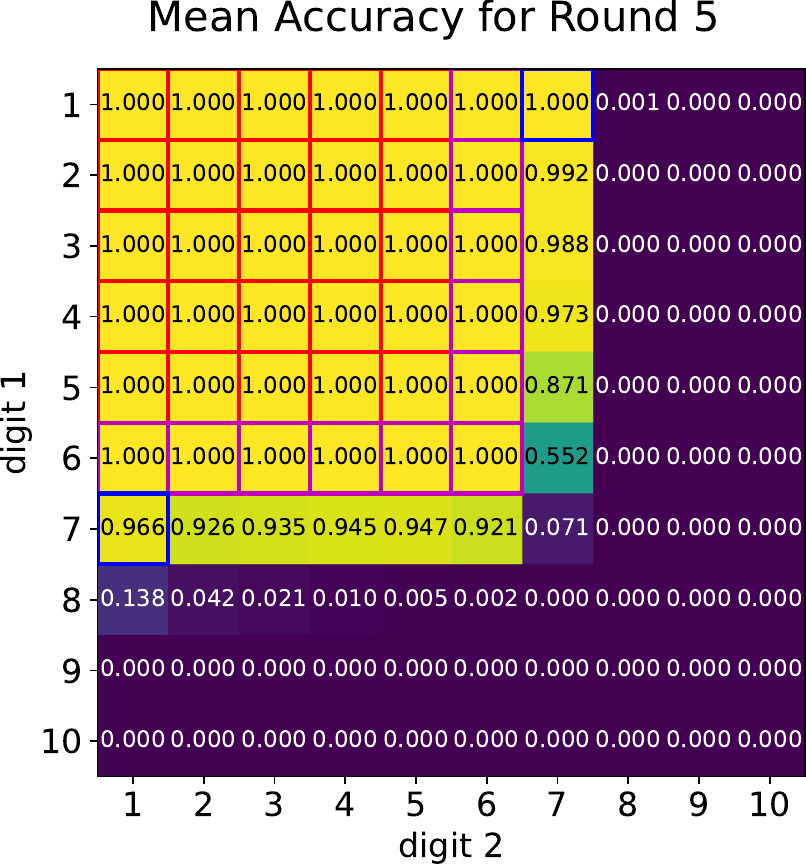}
    \includegraphics[width=0.24\linewidth]{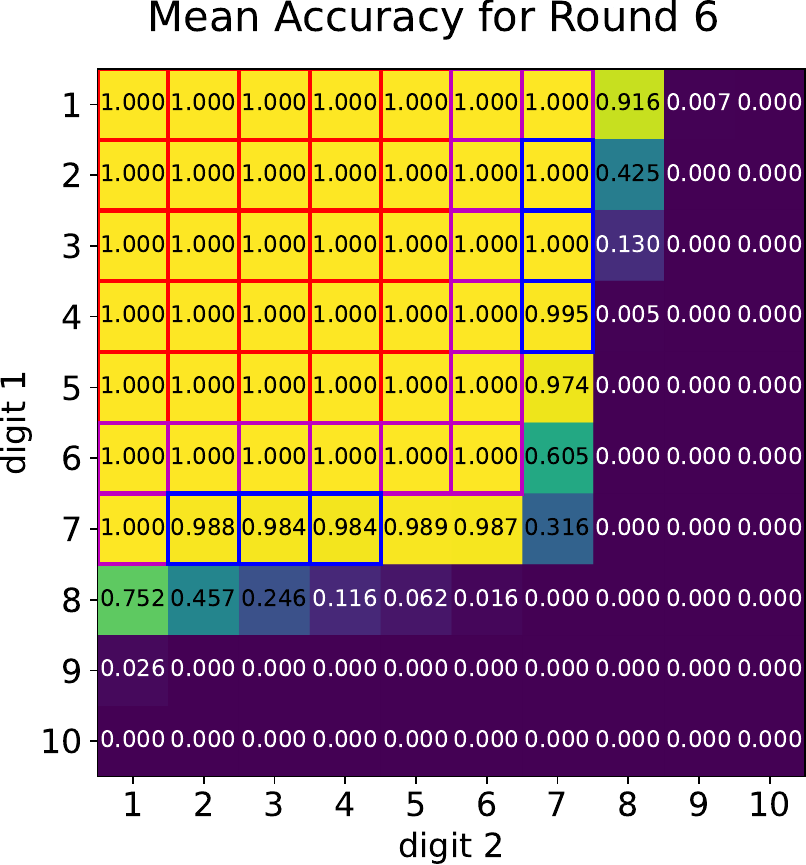}
    \includegraphics[width=0.24\linewidth]{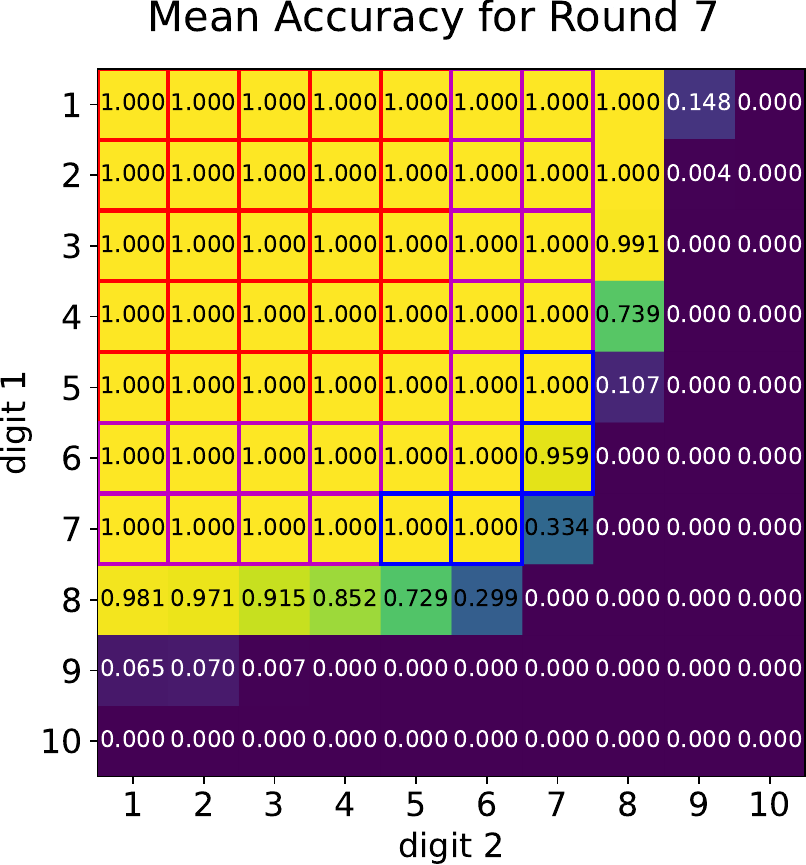}
    \includegraphics[width=0.24\linewidth]{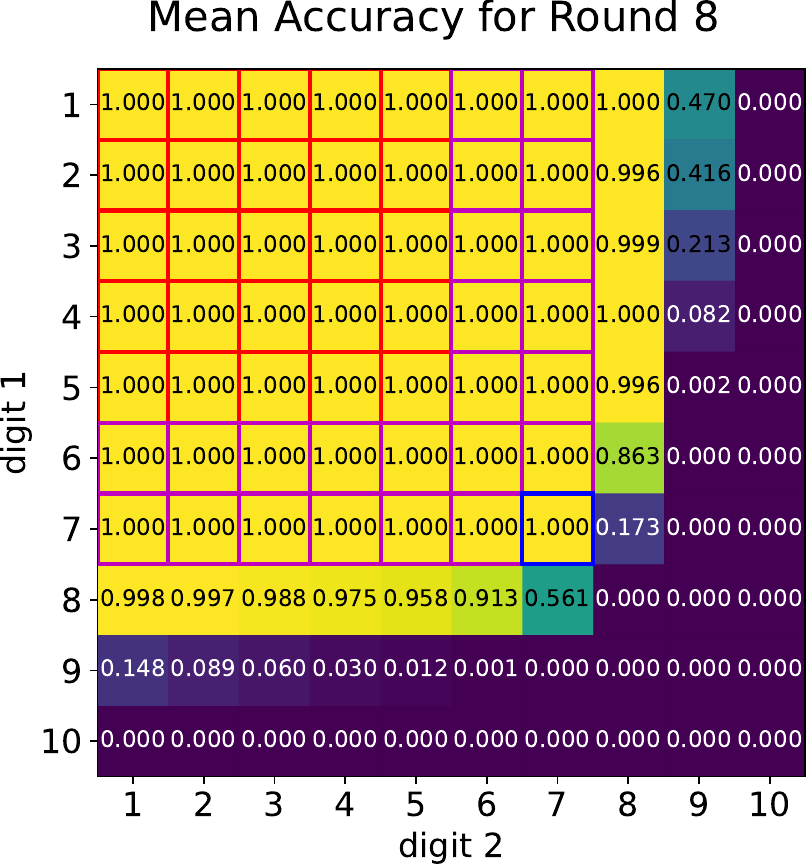}
    \includegraphics[width=0.24\linewidth]{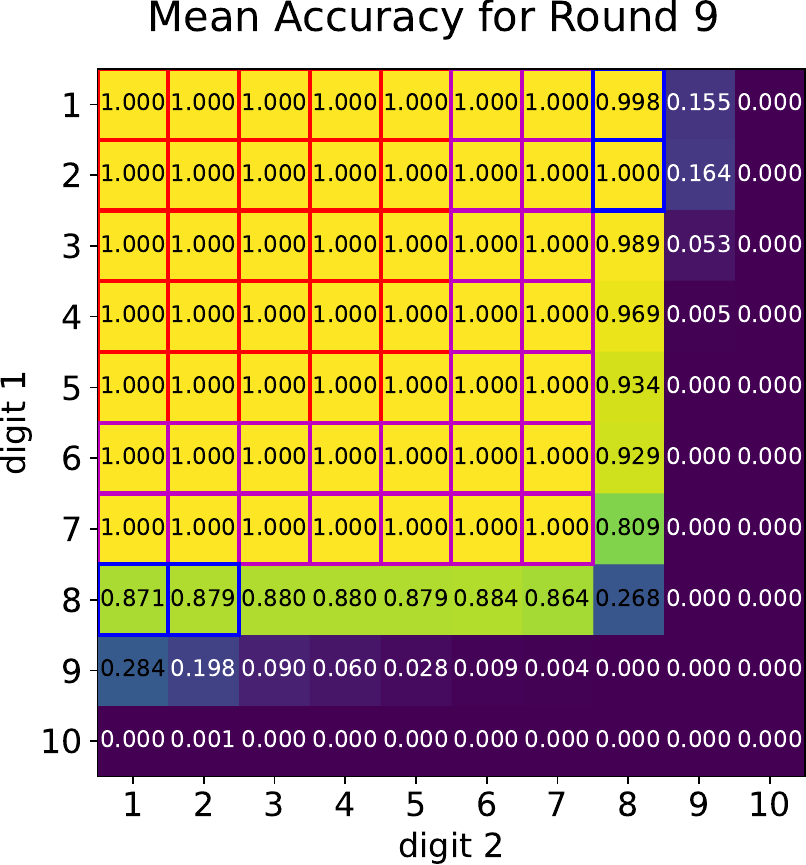}
    \includegraphics[width=0.24\linewidth]{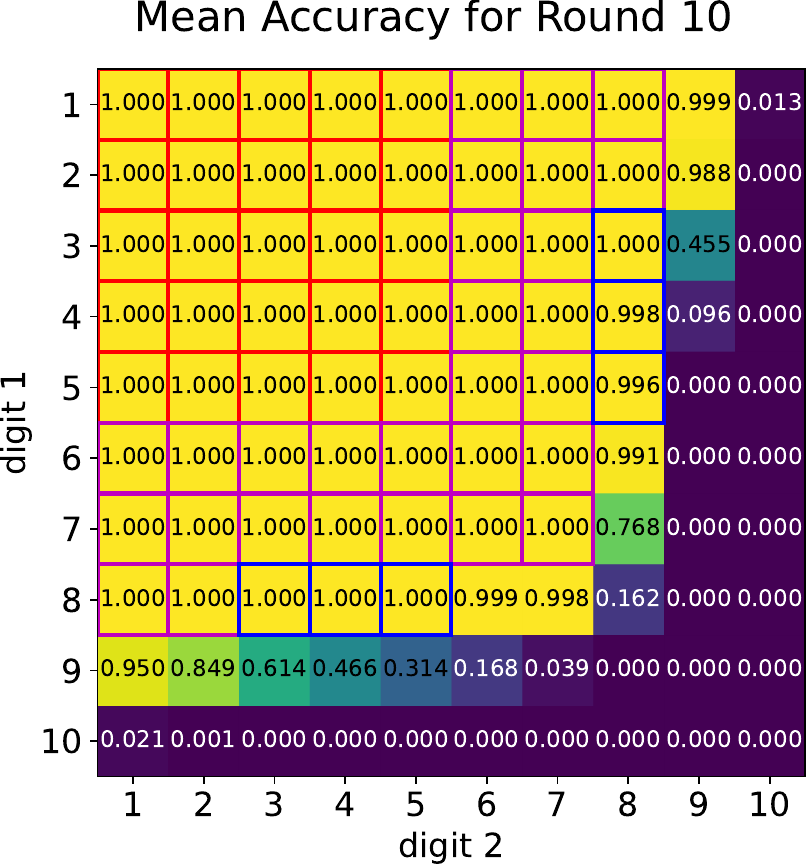}
    \includegraphics[width=0.24\linewidth]{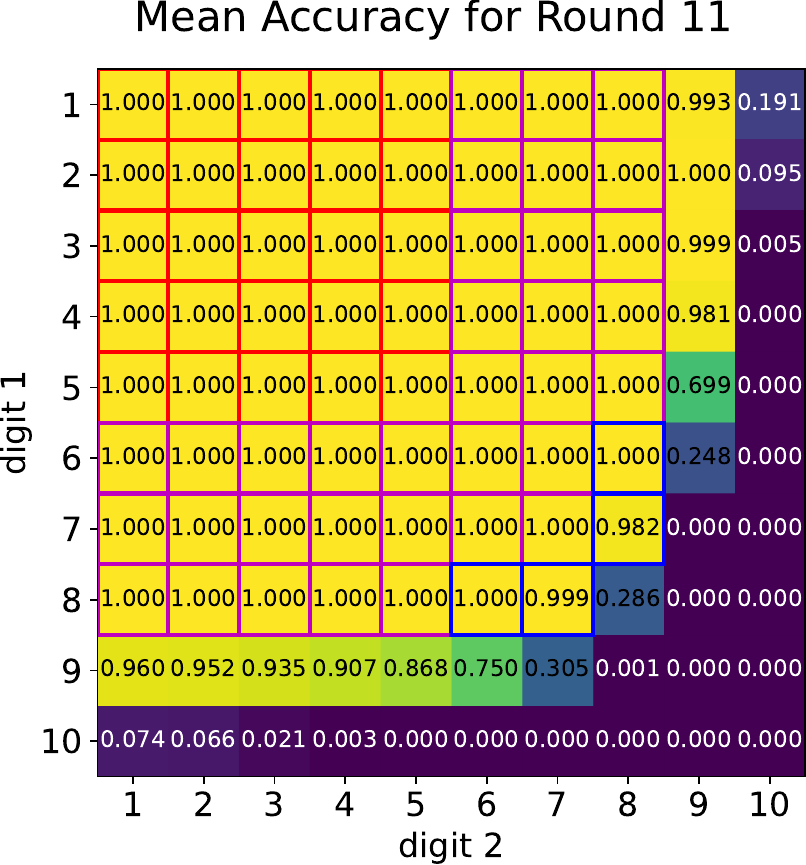}
    \includegraphics[width=0.24\linewidth]{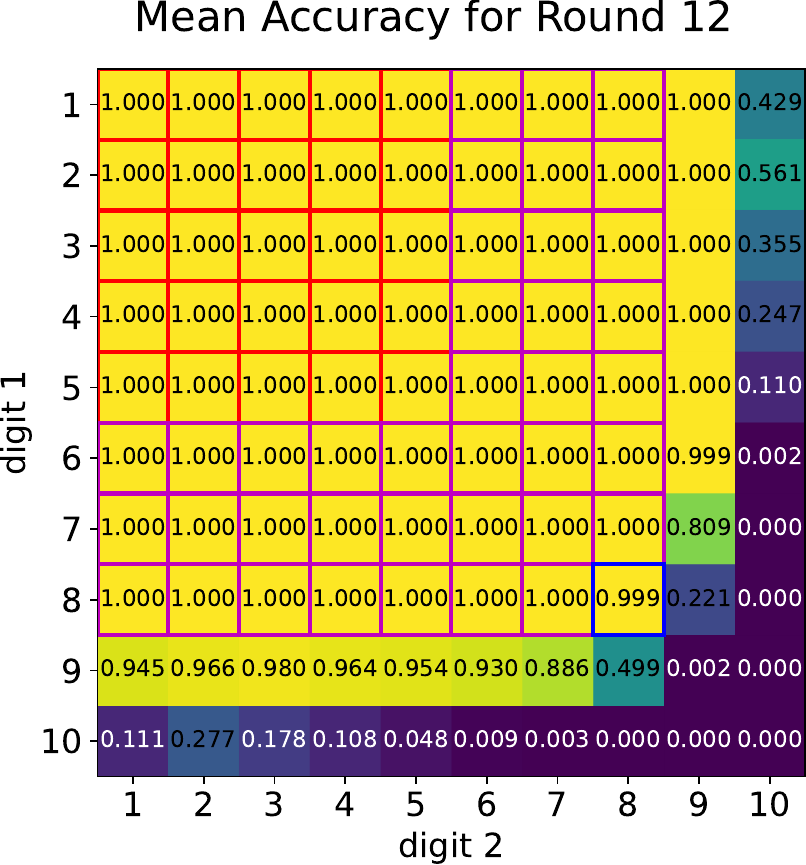}
    \includegraphics[width=0.24\linewidth]{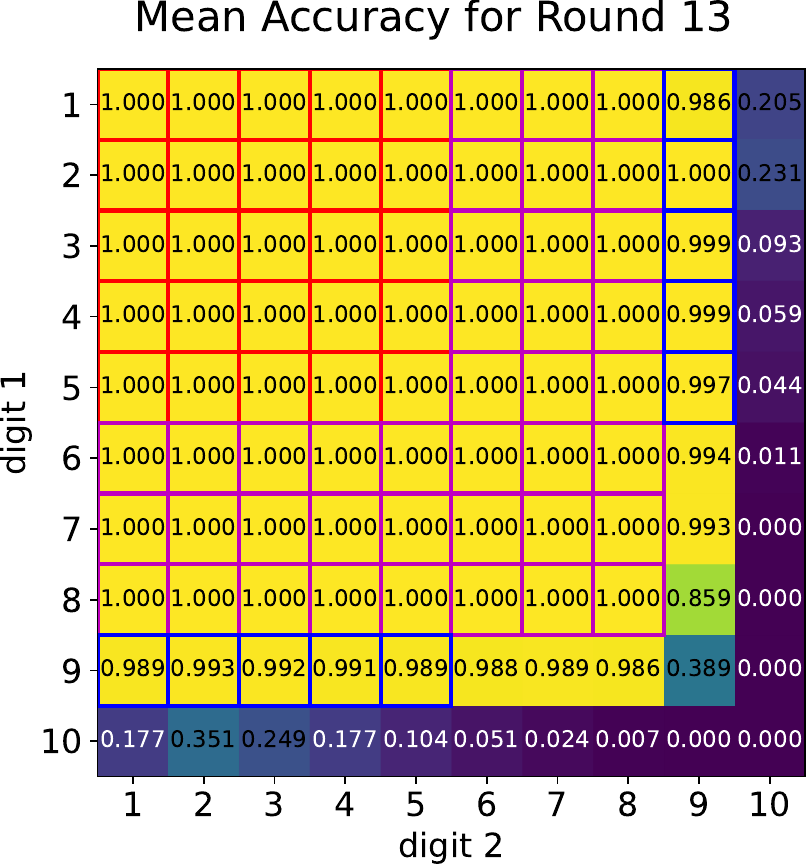}
    \includegraphics[width=0.24\linewidth]{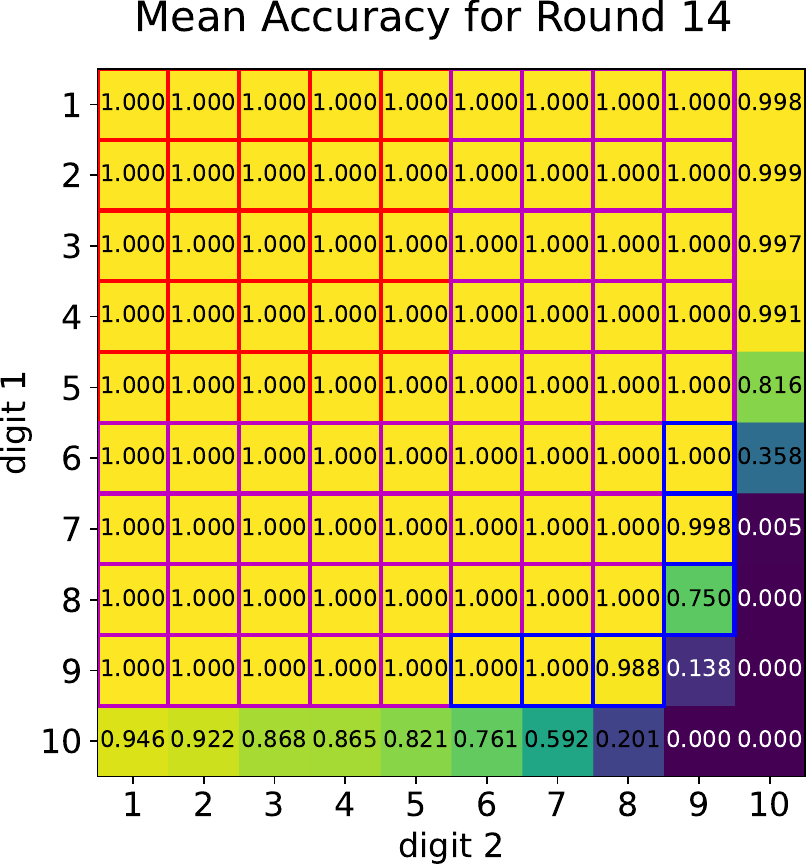}
    \includegraphics[width=0.24\linewidth]{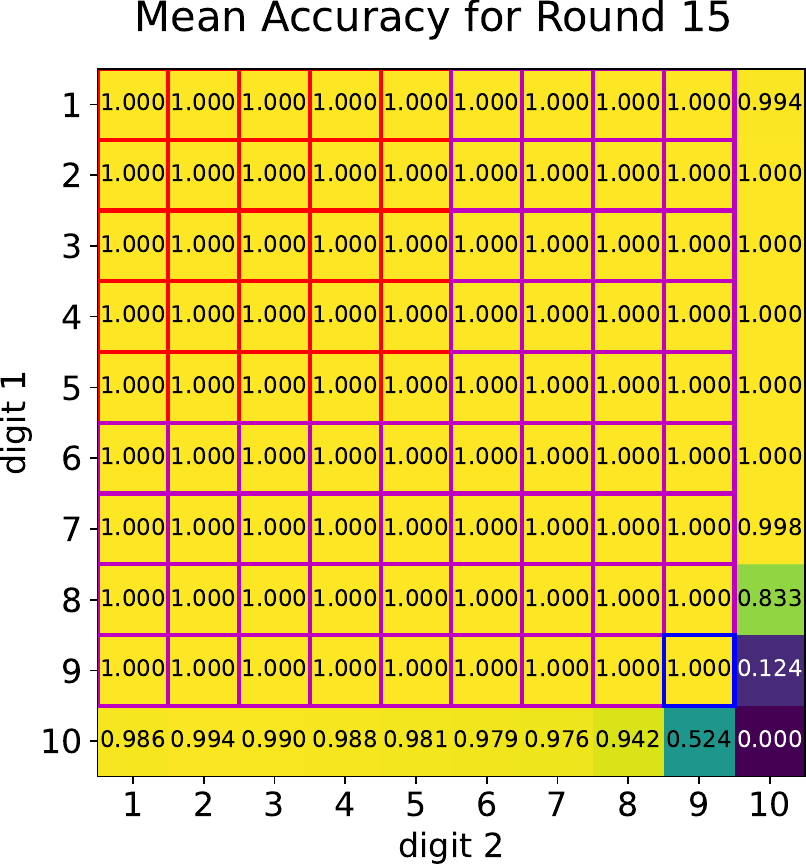}
    \includegraphics[width=0.24\linewidth]{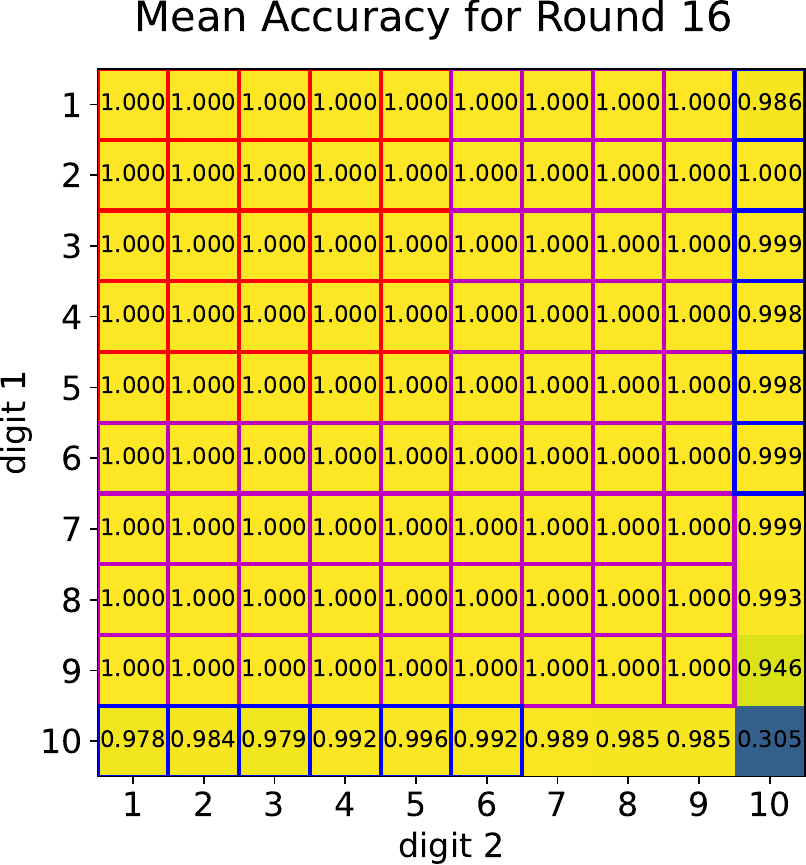}
    \includegraphics[width=0.24\linewidth]{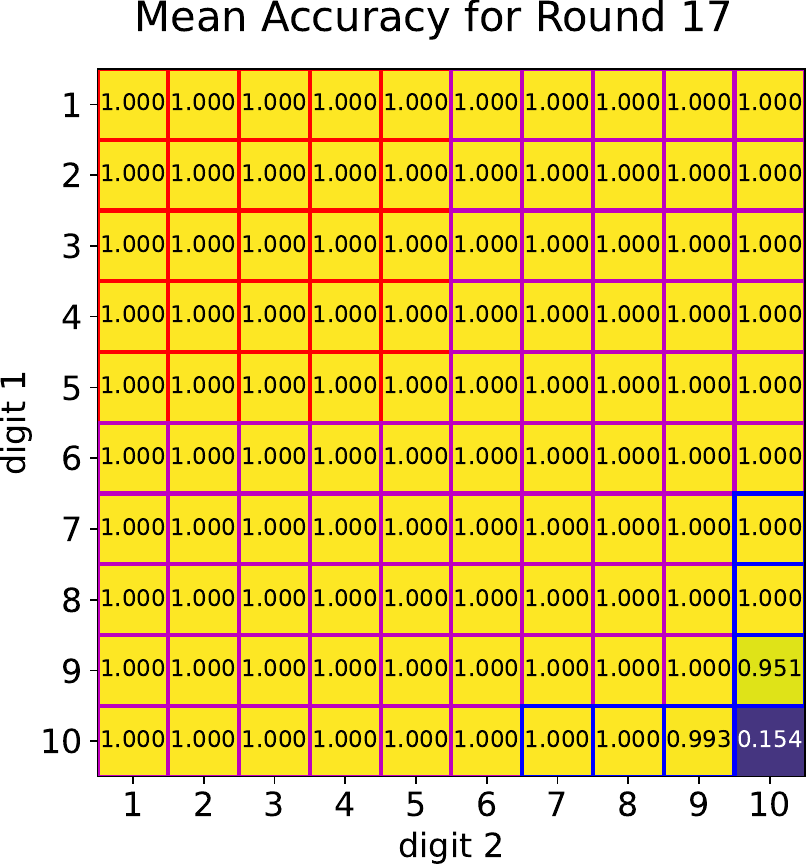}
    \includegraphics[width=0.24\linewidth]{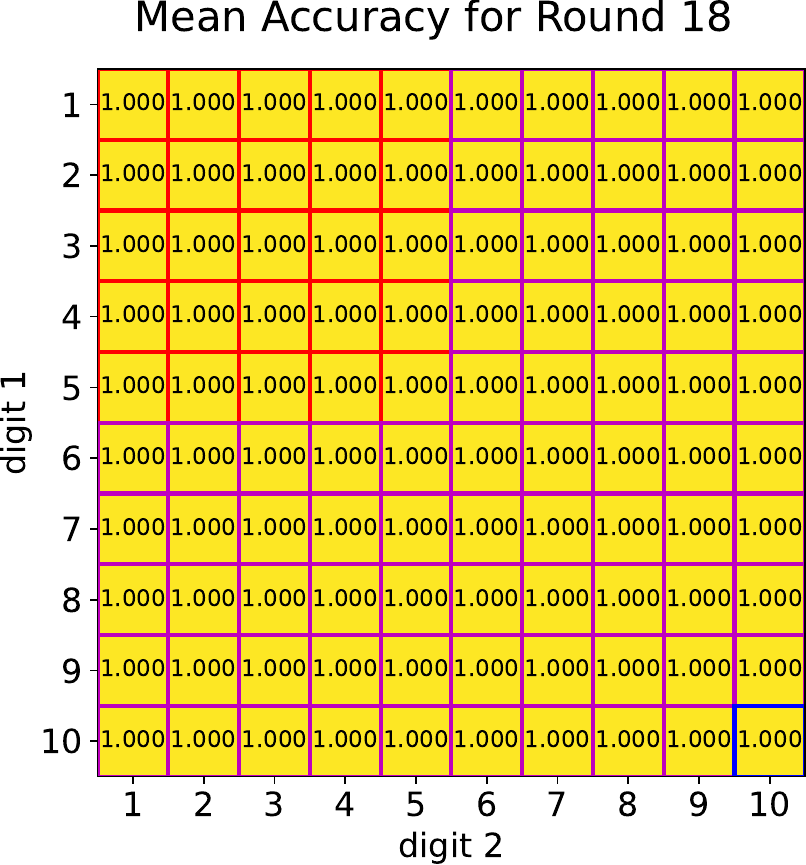}
    \includegraphics[width=0.24\linewidth]{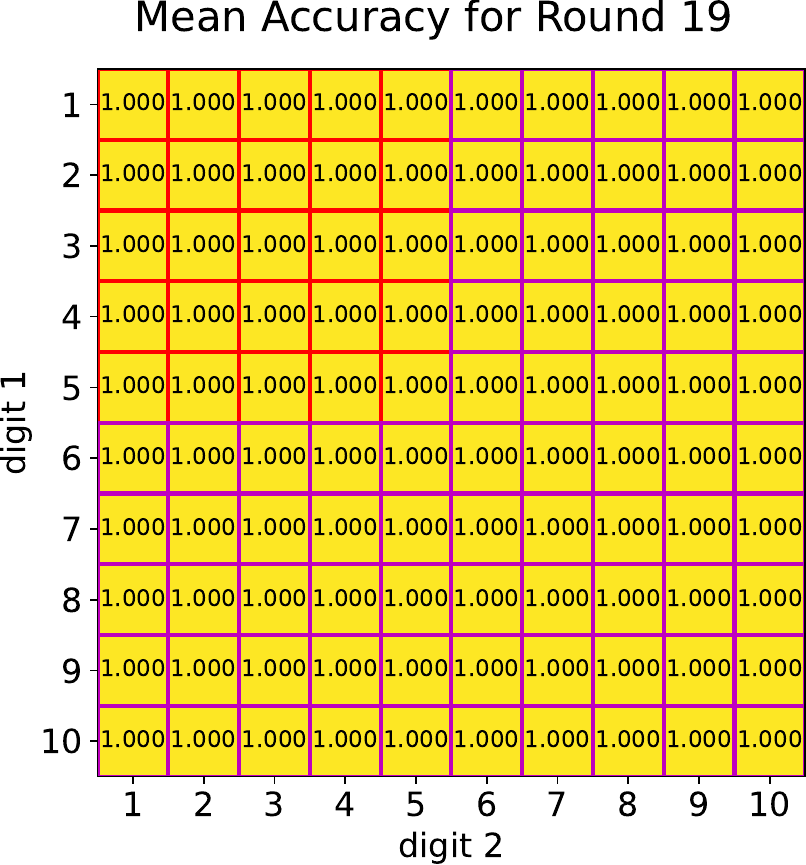}

    \caption{Accelerated multiplication. We can significantly reduce the self-improvement rounds by carefully sampling a wider range of difficulties at every round. Perfect length generalization is achieved up to 10-by-10 multiplication with 19 self-improvement rounds. }
    \label{fig:multiplication_accelerated_len_n10_full}
\end{figure}

\subsection{Results on Mazes}\label{sec:maze_full_results}
We provide additional evaluation on mazes, based on the validity of moves and correctness of end nodes. 

\begin{figure}[ht!]
    \centering
    \includegraphics[width=0.19\linewidth]{fig/maze/maze_hops/vanilla_mean_acc.pdf}
    \includegraphics[width=0.19\linewidth]{fig/maze/maze_hops/mv_mean_acc.pdf}
    \includegraphics[width=0.19\linewidth]{fig/maze/maze_hops/verifier-move-ends_mean_acc.pdf}
    \includegraphics[width=0.19\linewidth]{fig/maze/maze_hops/verifier_mean_acc.pdf}
    \includegraphics[width=0.19\linewidth]{fig/maze/maze_hops/verifier-ends_mean_acc.pdf}    
    \\
    \includegraphics[width=0.19\linewidth]{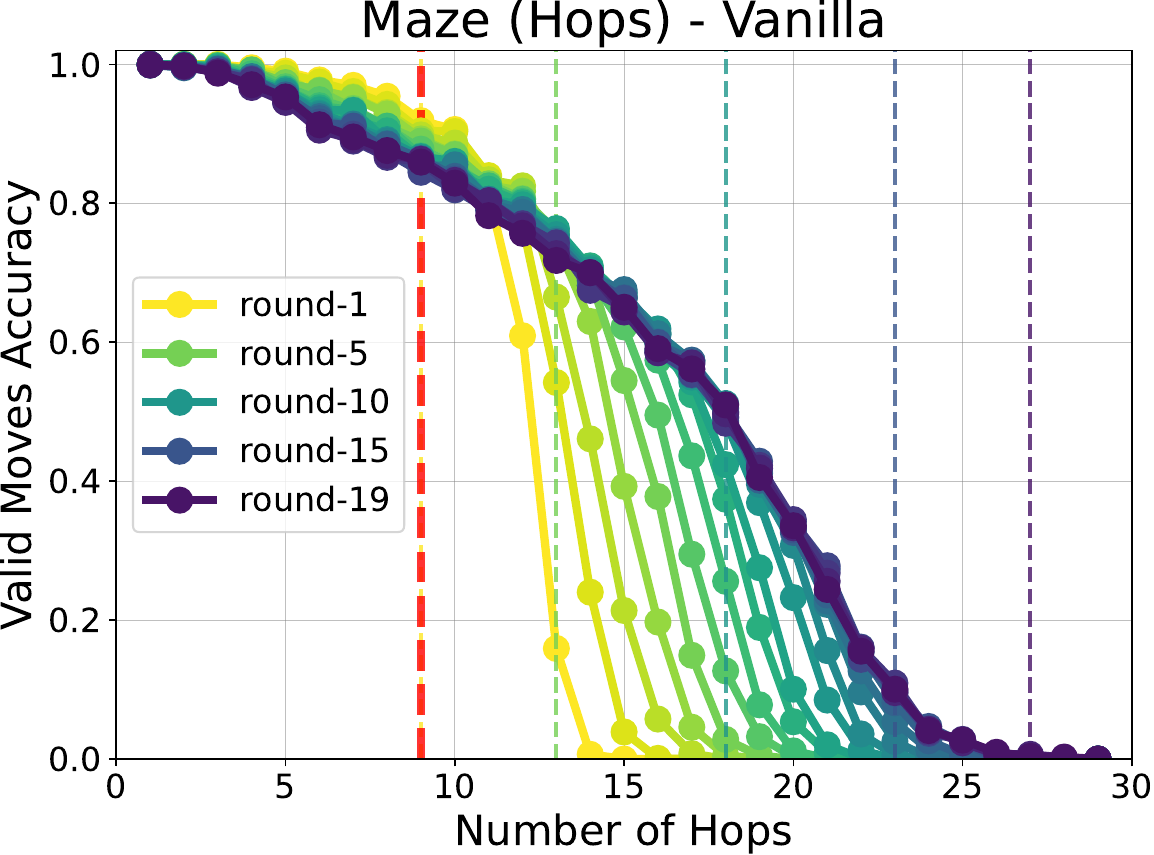}
    \includegraphics[width=0.19\linewidth]{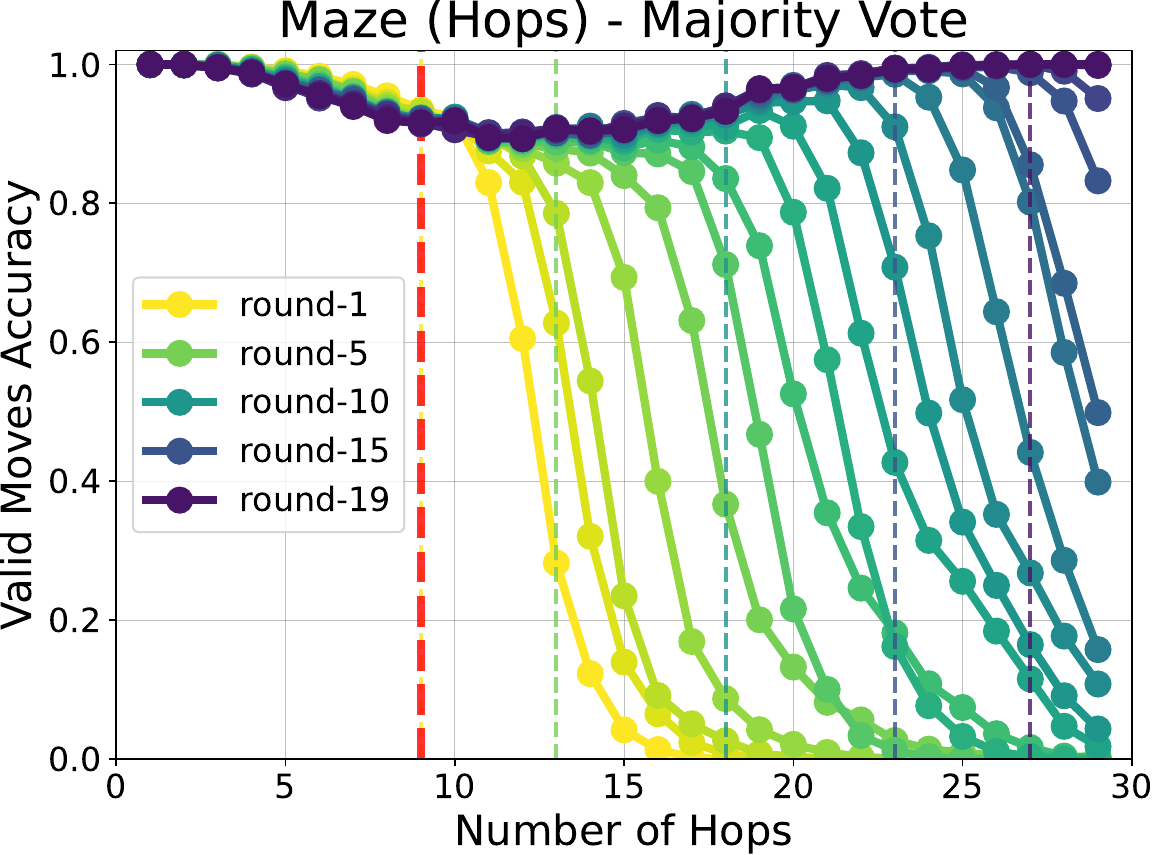}
    \includegraphics[width=0.19\linewidth]{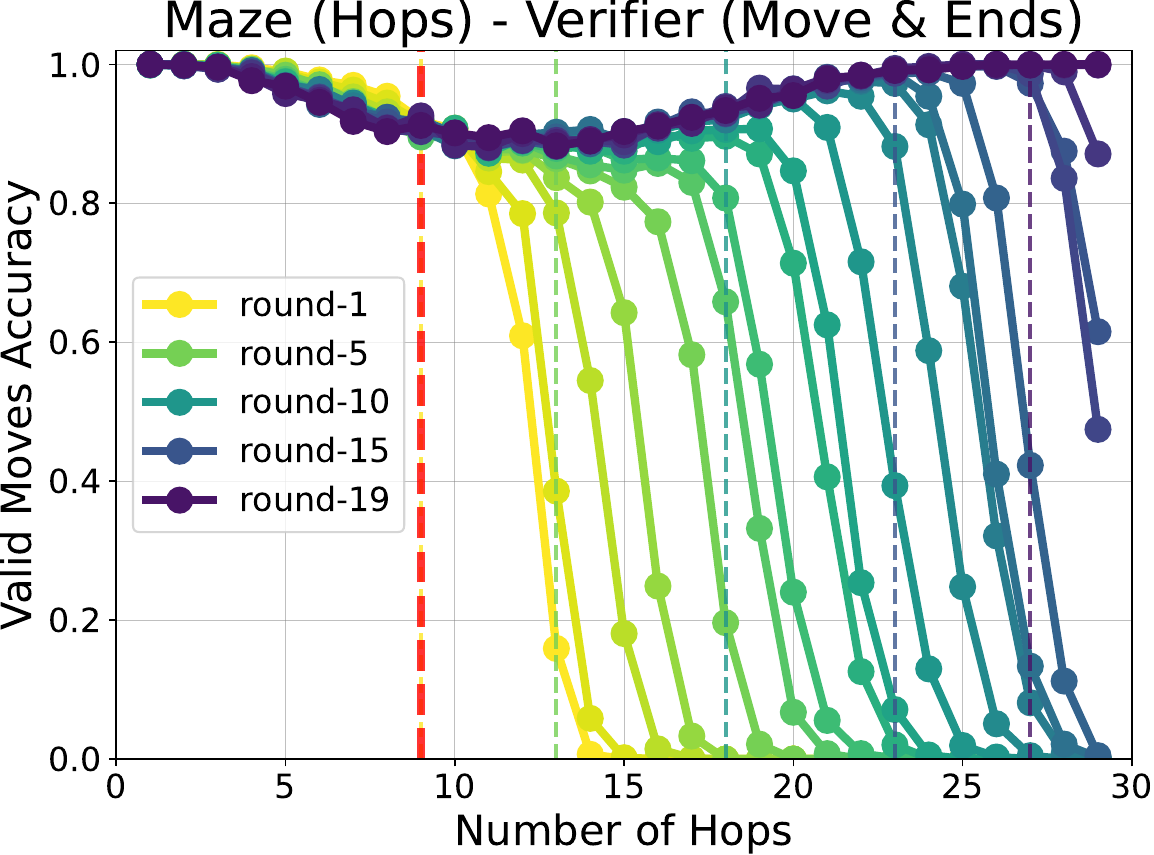}
    \includegraphics[width=0.19\linewidth]{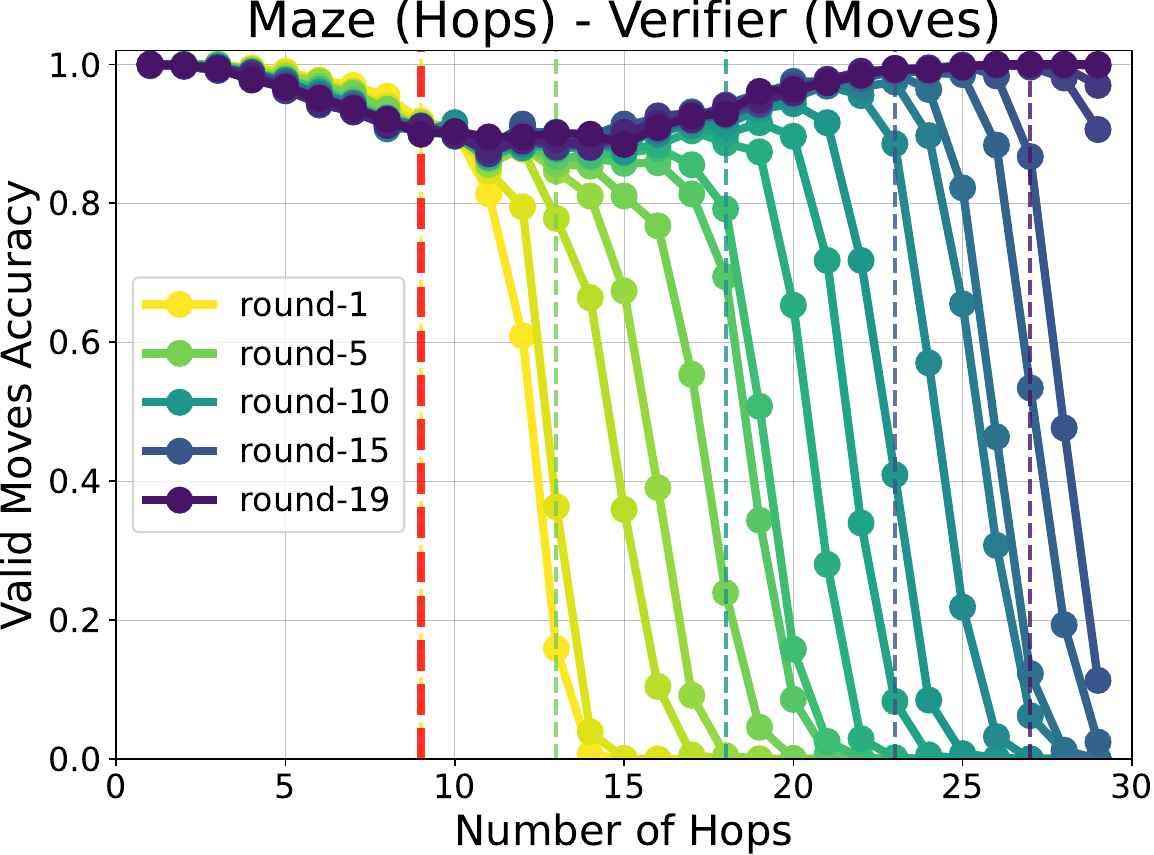}
    \includegraphics[width=0.19\linewidth]{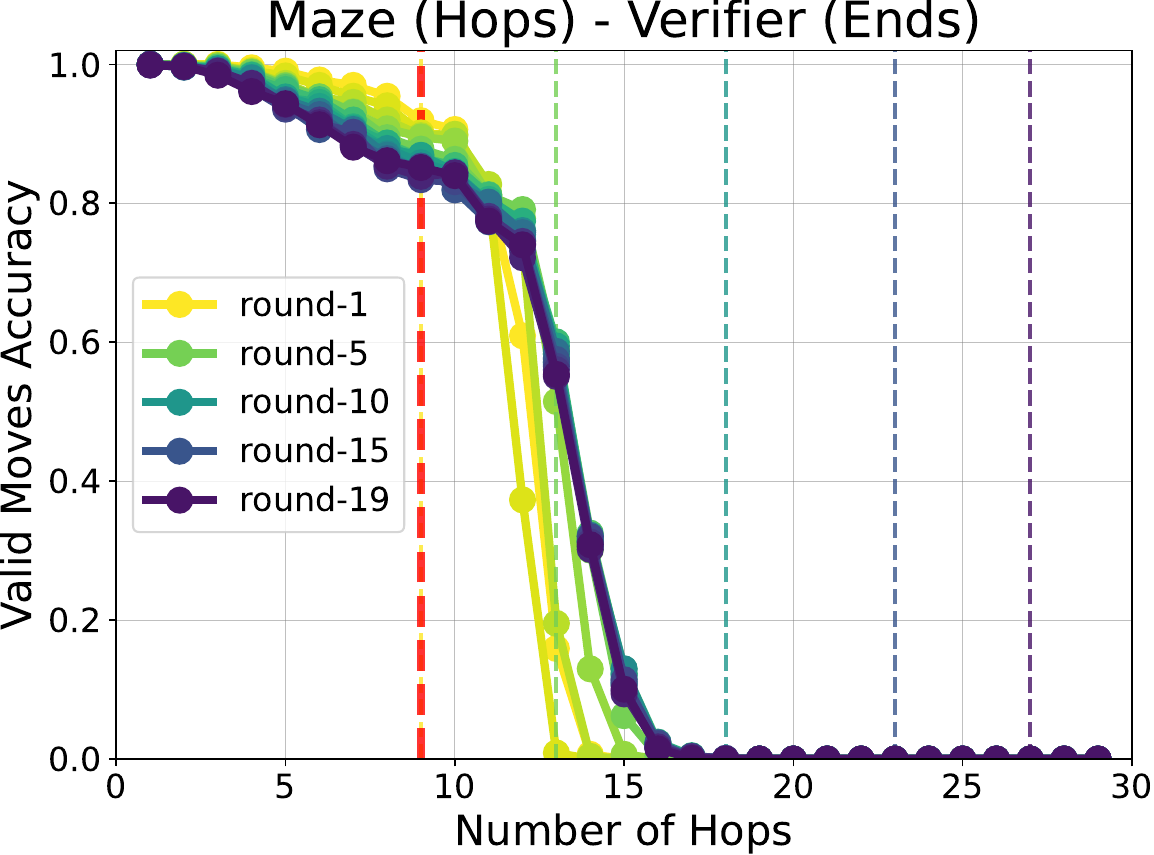}    
    \\
    \includegraphics[width=0.19\linewidth]{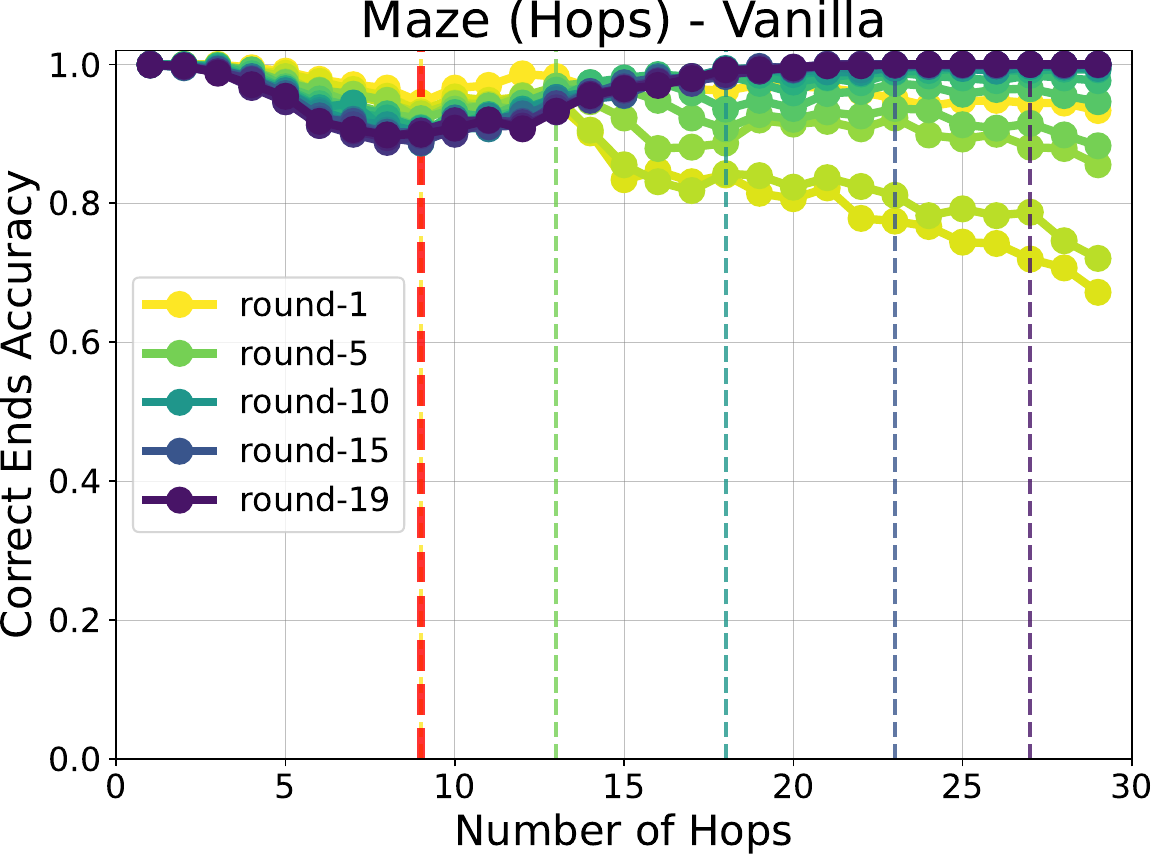}
    \includegraphics[width=0.19\linewidth]{fig/maze/maze_hops/mv_mean_acc.pdf}
    \includegraphics[width=0.19\linewidth]{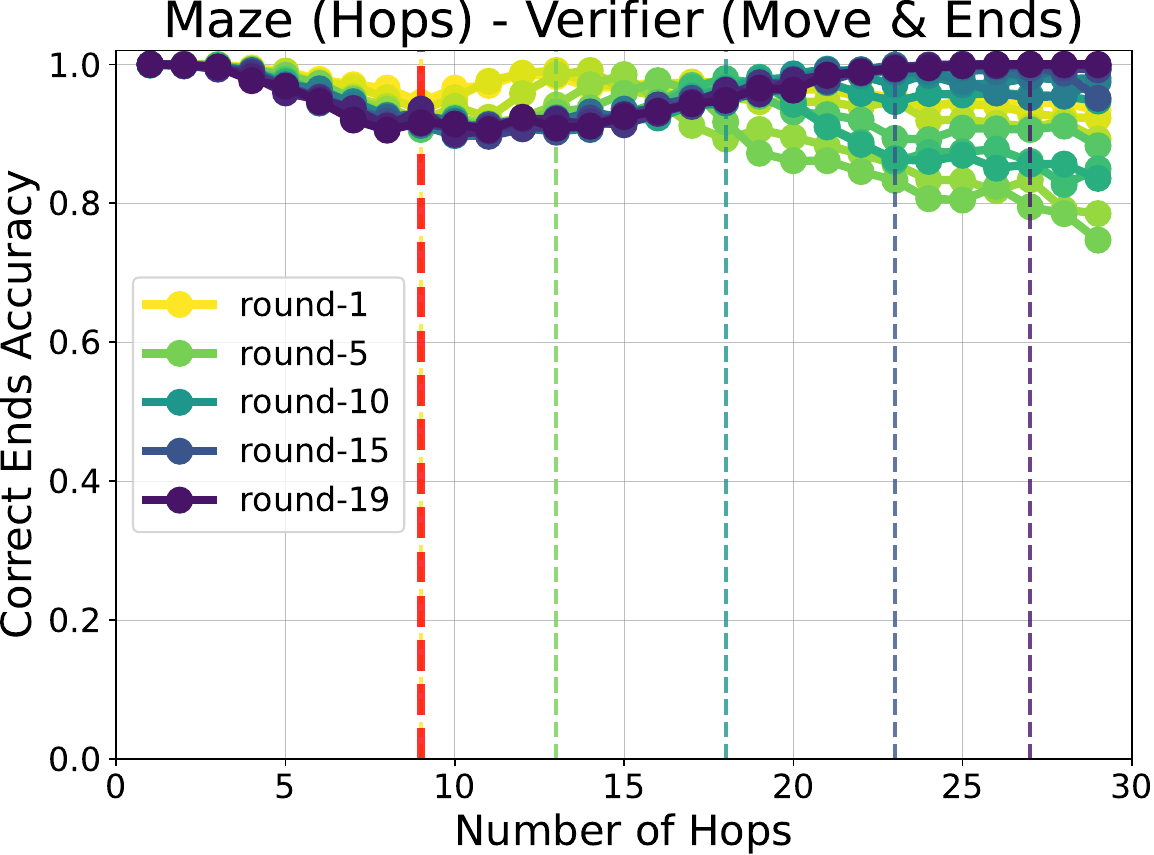}
    \includegraphics[width=0.19\linewidth]{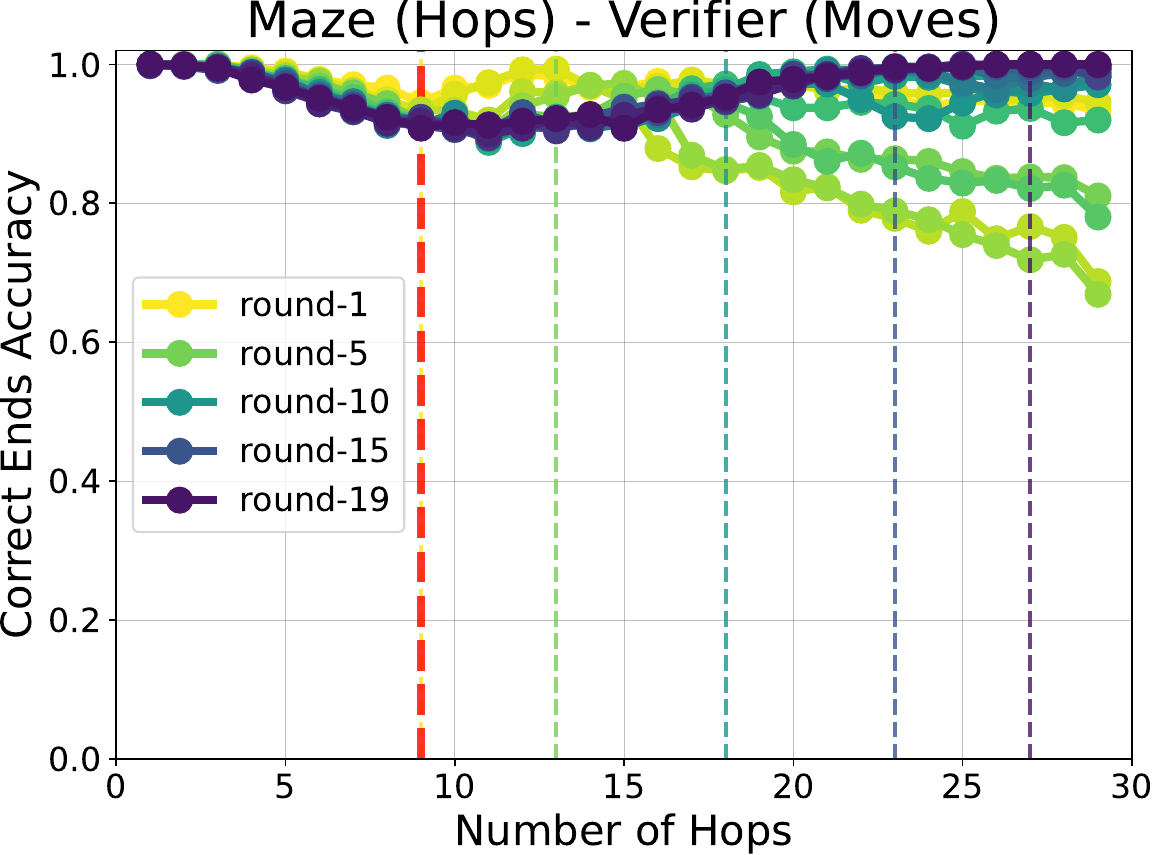}
    \includegraphics[width=0.19\linewidth]{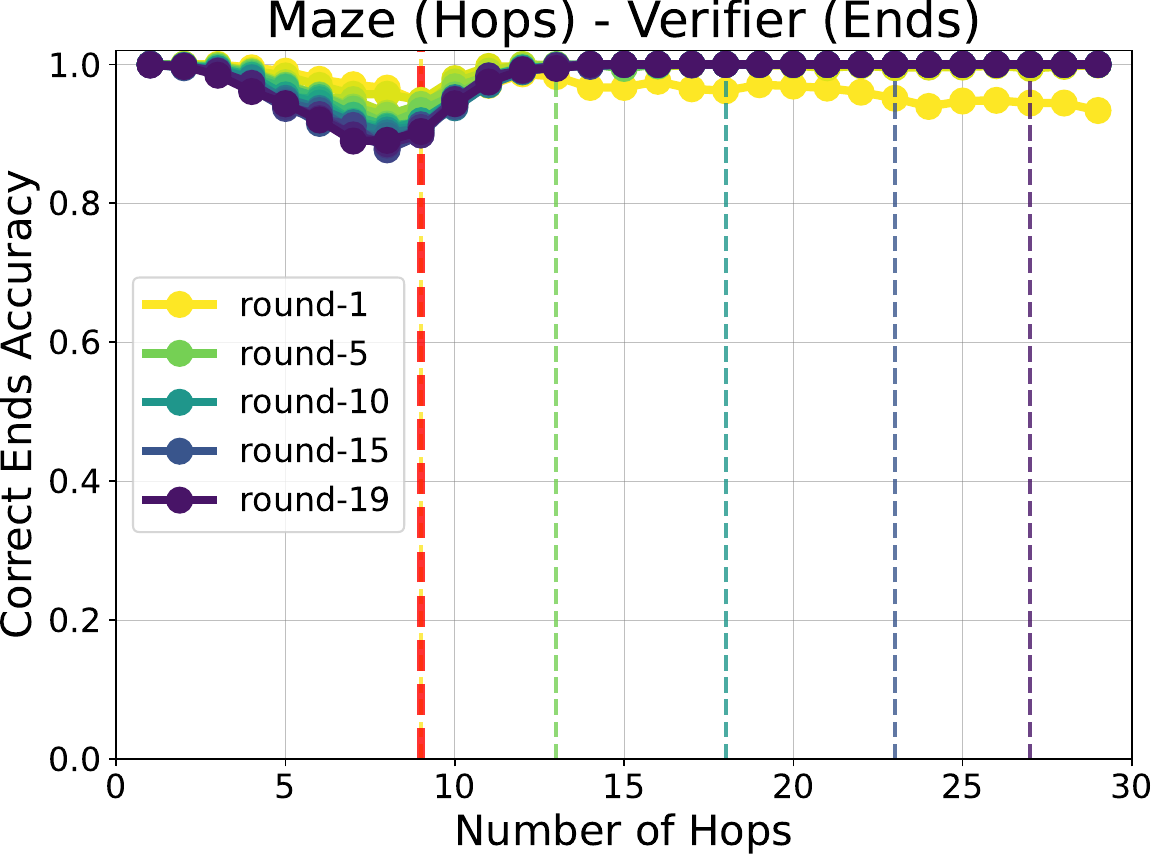}    
    \caption{ Maze solving task with increasing hops. (Top to bottom) Exact match accuracy, move validation accuracy, and end validation accuracy.  (Left to right) No data filtering, majority voting based filtering, verifier on both moves and ends, verifier on moves only, verifier on ends only.}
    \label{fig:maze_verifier_hops_full}
\end{figure}

\begin{figure}[ht!]
    \centering
    \includegraphics[width=0.19\linewidth]{fig/maze/maze_nodes/vanilla_mean_acc.pdf}
    \includegraphics[width=0.19\linewidth]{fig/maze/maze_nodes/mv_mean_acc.pdf}
    \includegraphics[width=0.19\linewidth]{fig/maze/maze_nodes/verifier-move-ends_mean_acc.pdf}
    \includegraphics[width=0.19\linewidth]{fig/maze/maze_nodes/verifier_mean_acc.pdf}
    \includegraphics[width=0.19\linewidth]{fig/maze/maze_nodes/verifier-ends_mean_acc.pdf}    
    \\
    \includegraphics[width=0.19\linewidth]{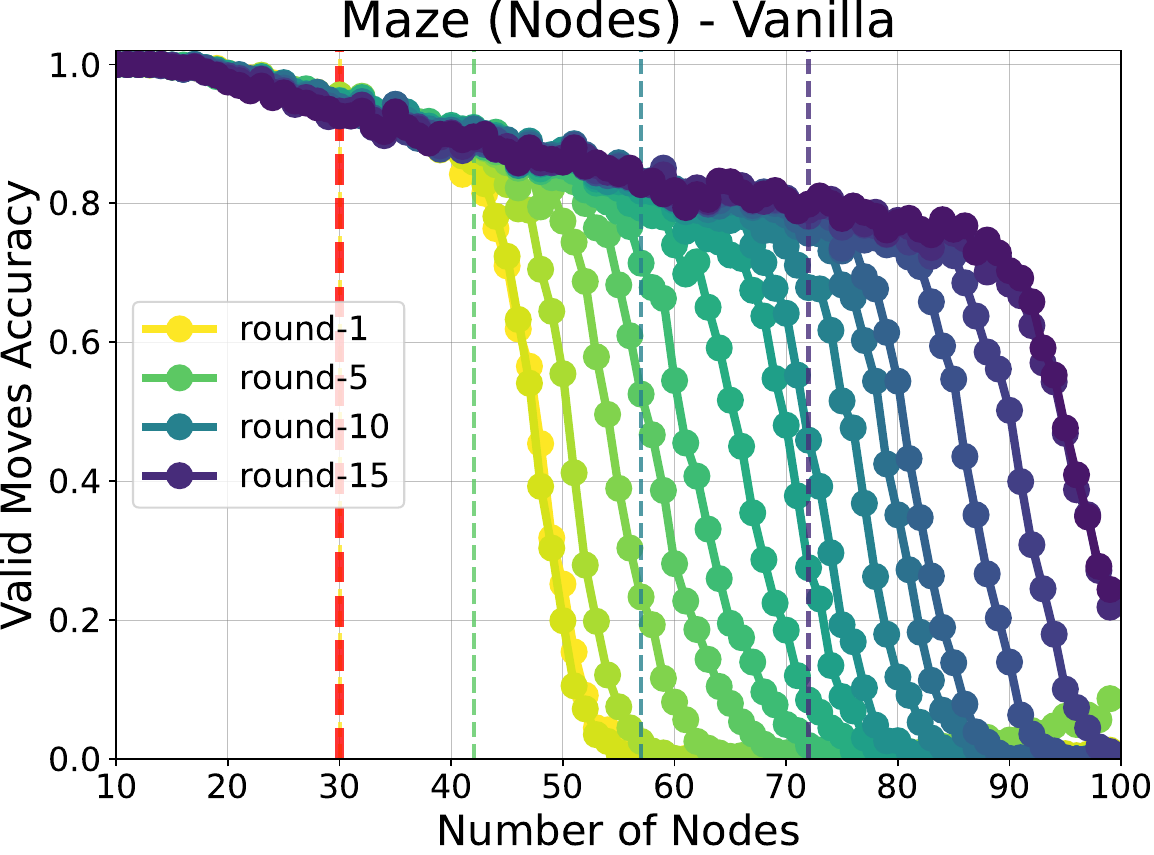}
    \includegraphics[width=0.19\linewidth]{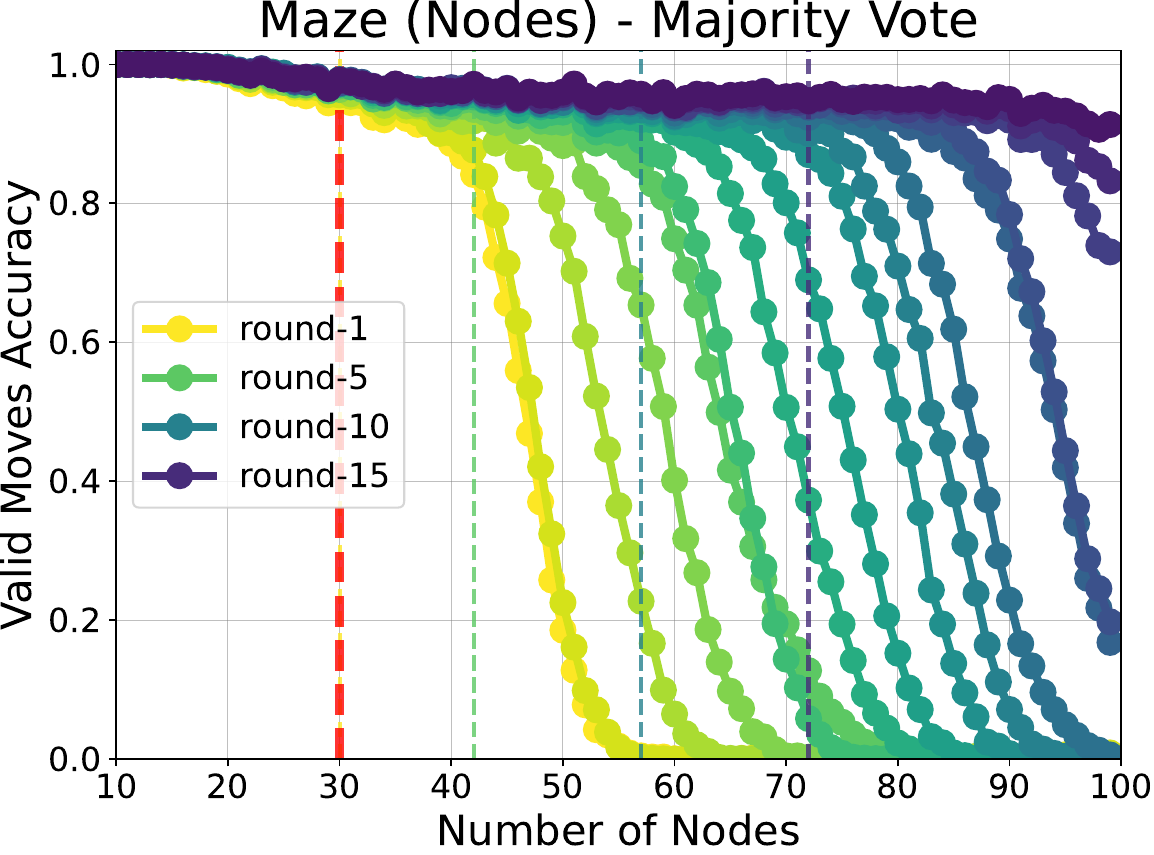}
    \includegraphics[width=0.19\linewidth]{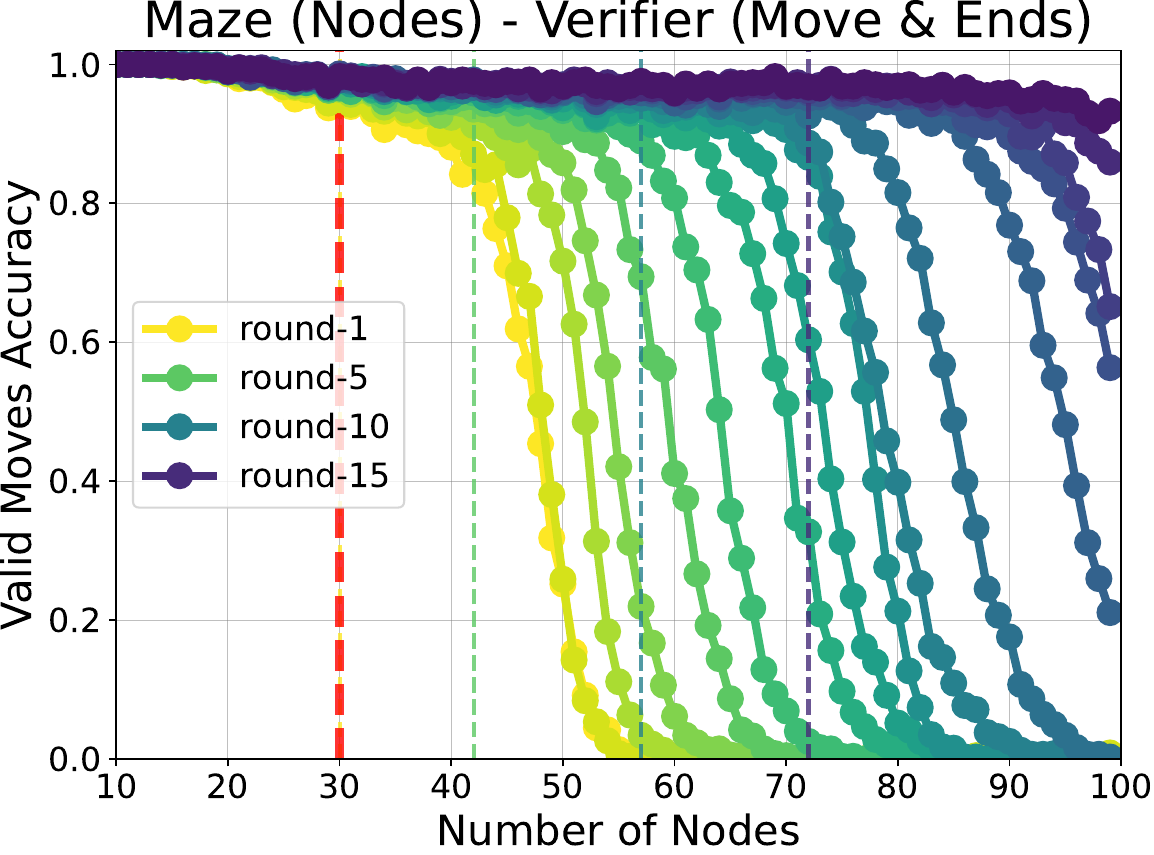}
    \includegraphics[width=0.19\linewidth]{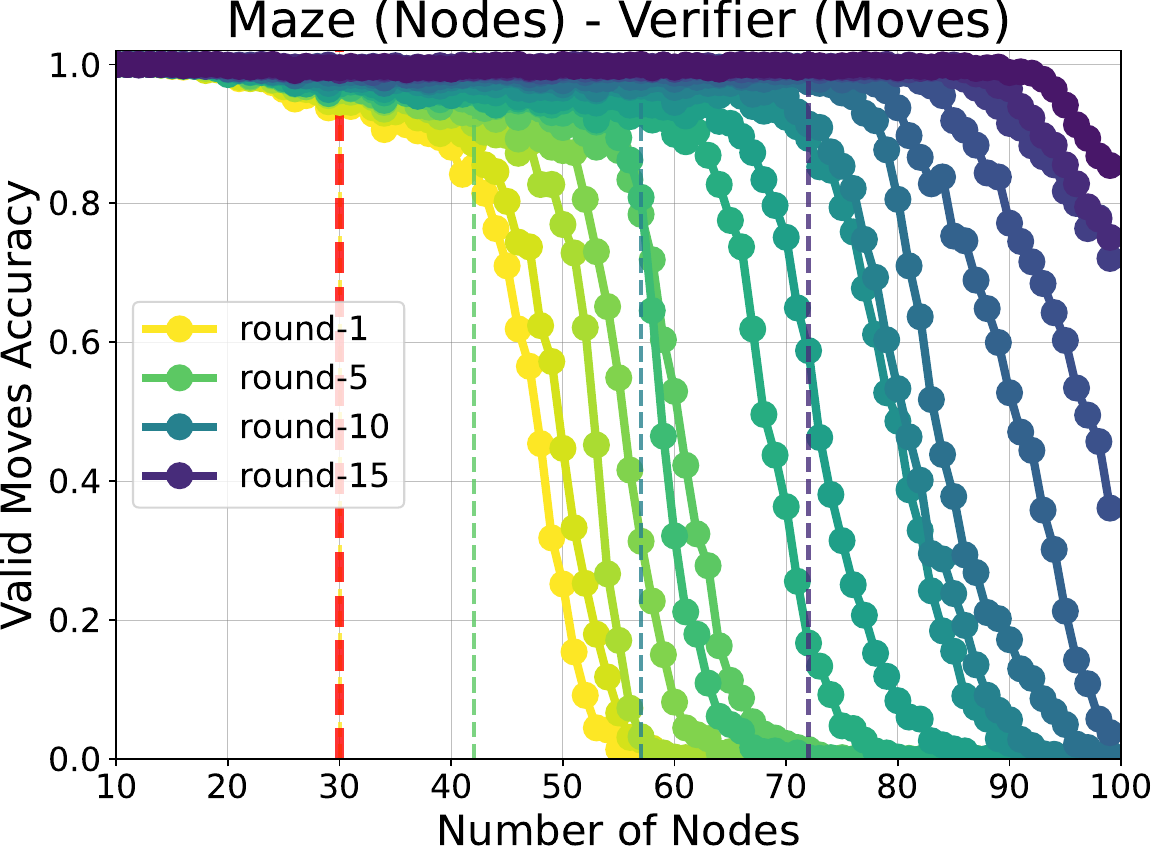}
    \includegraphics[width=0.19\linewidth]{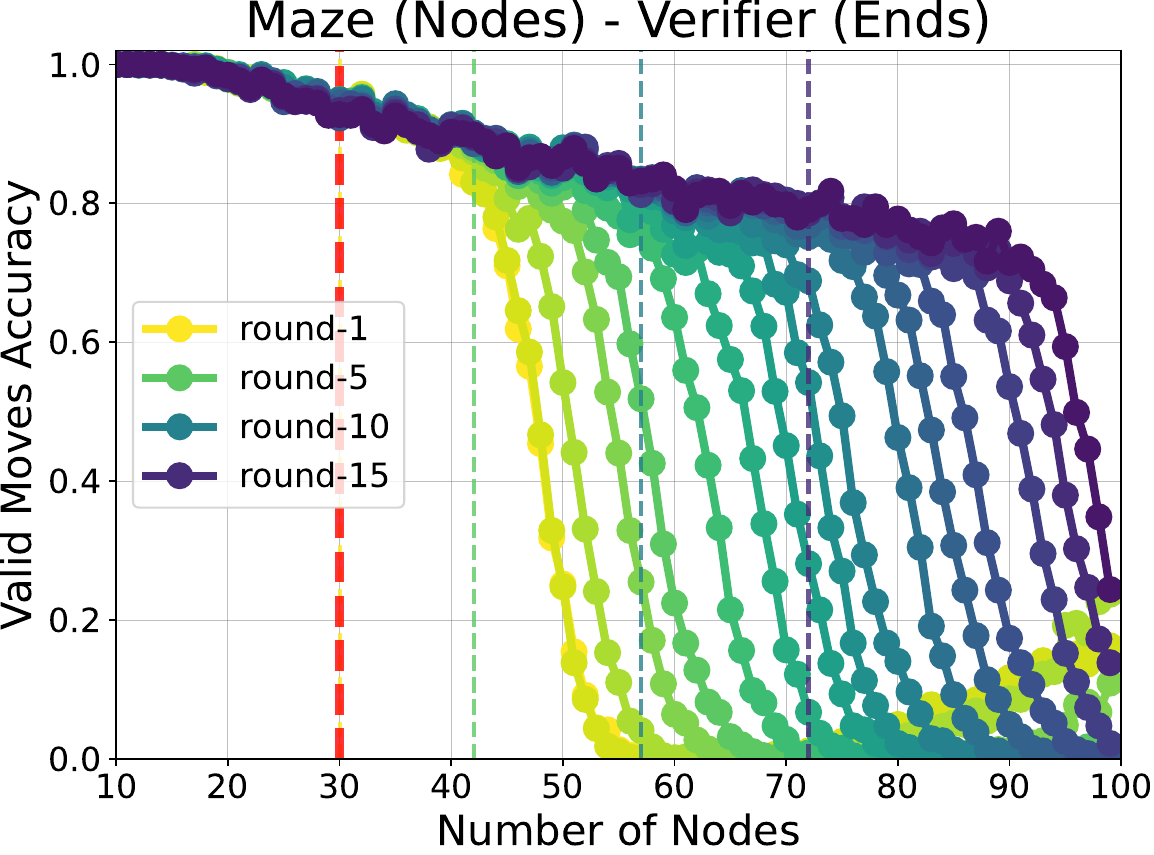}    
    \\
    \includegraphics[width=0.19\linewidth]{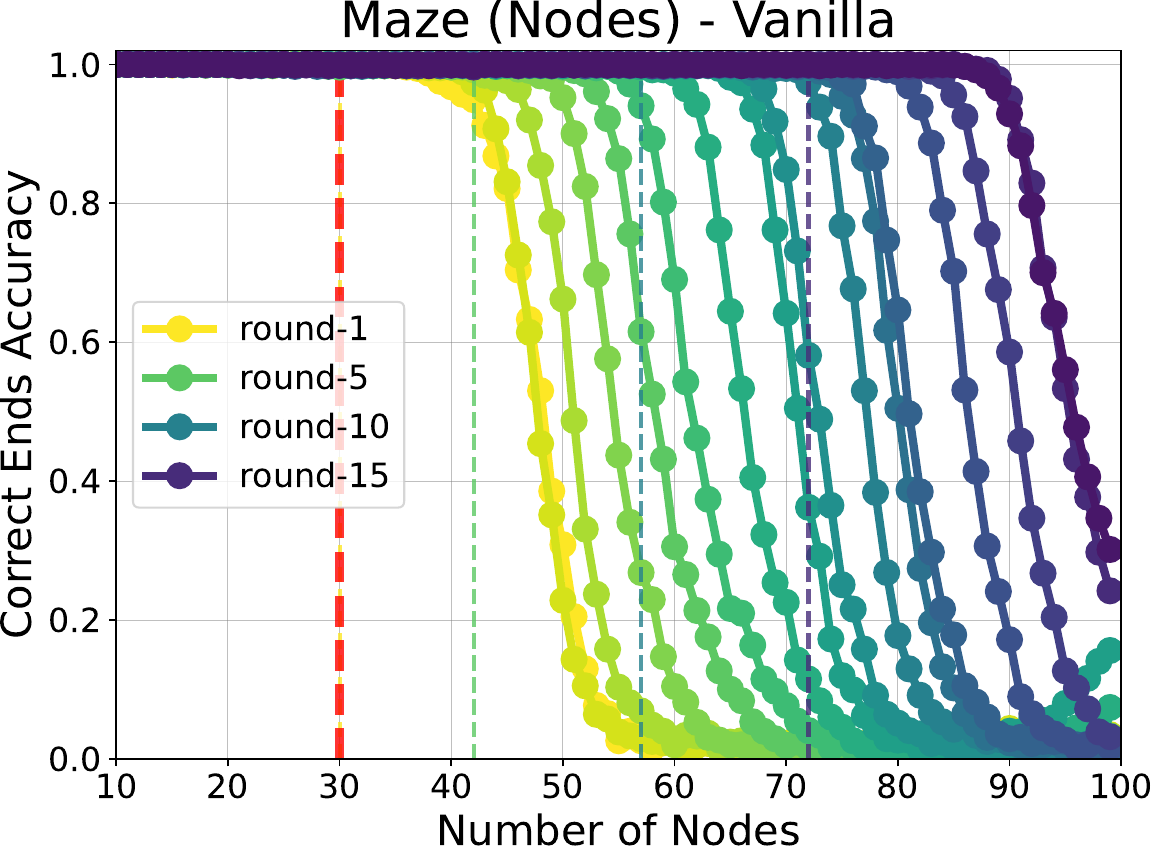}
    \includegraphics[width=0.19\linewidth]{fig/maze/maze_nodes/mv_mean_acc.pdf}
    \includegraphics[width=0.19\linewidth]{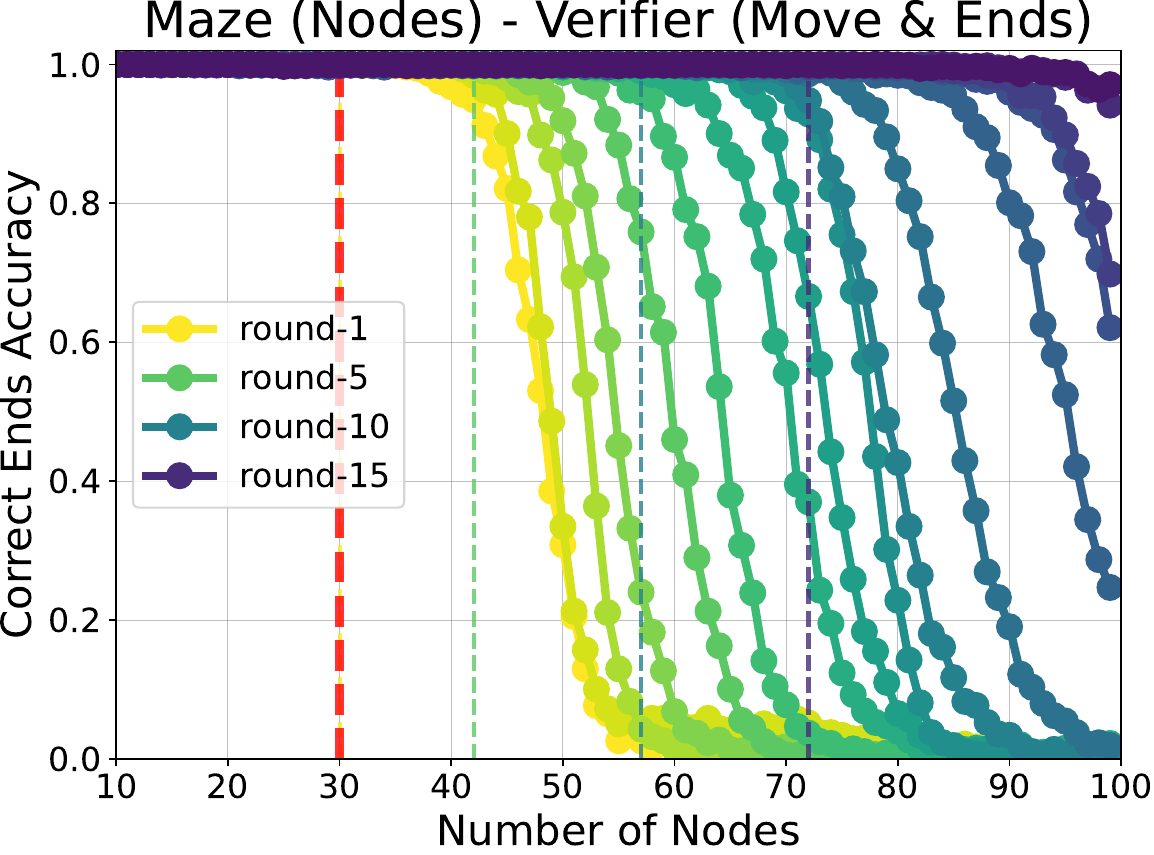}
    \includegraphics[width=0.19\linewidth]{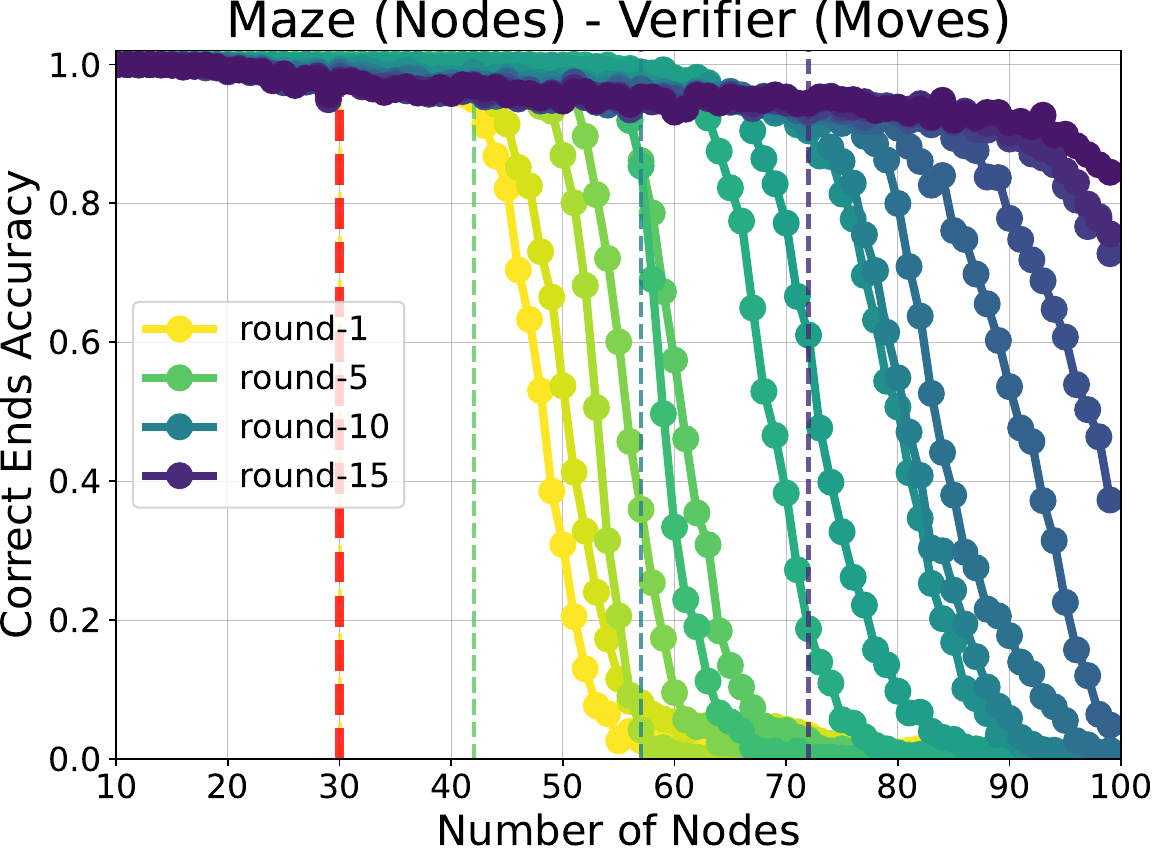}
    \includegraphics[width=0.19\linewidth]{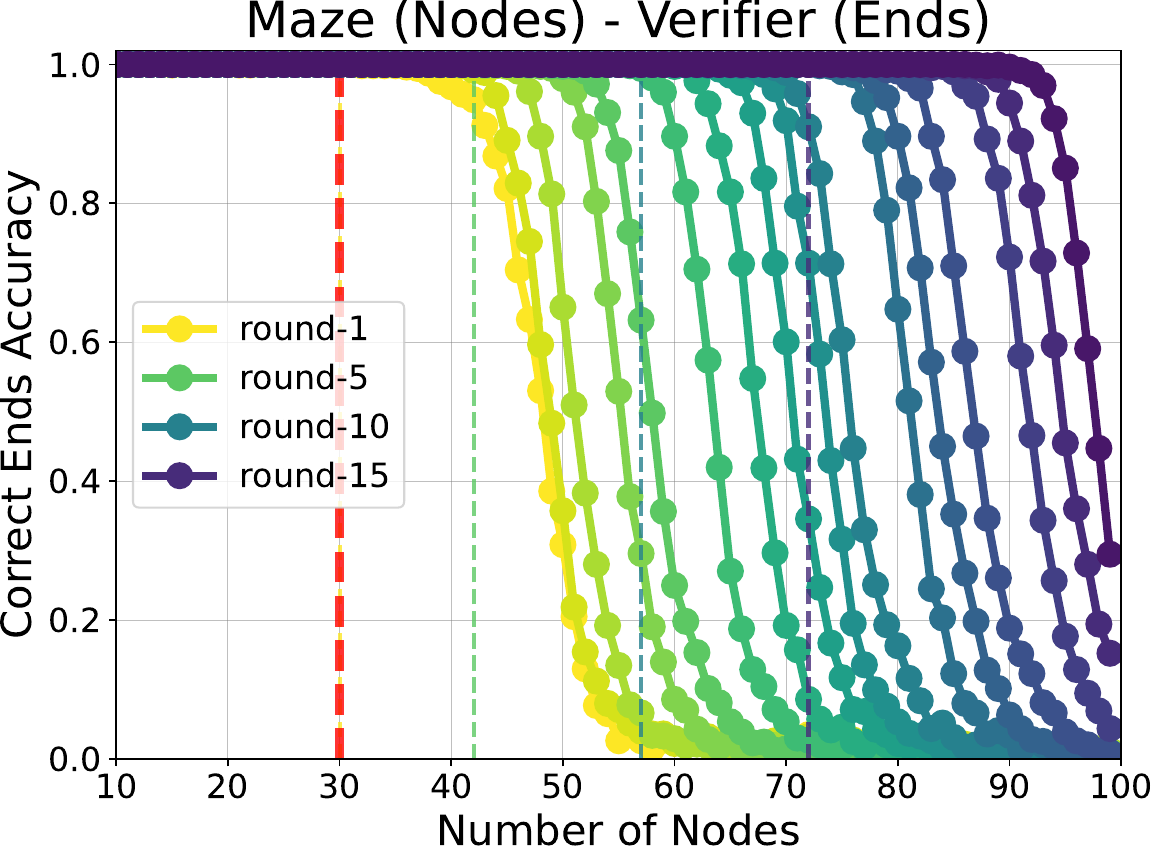}    
    \caption{ Maze solving task with increasing nodes. (Top to bottom) Exact match accuracy, move validation accuracy, and end validation accuracy.  (Left to right) No data filtering, majority voting based filtering, verifier on both moves and ends, verifier on moves only, verifier on ends only.}
    \label{fig:maze_verifier_nodes_full}
\end{figure}

\end{document}